\documentclass{article}

\usepackage[preprint]{neurips_2025}

\usepackage[utf8]{inputenc} 
\usepackage[T1]{fontenc}    
\usepackage{hyperref}       
\usepackage{url}            
\usepackage{booktabs}       
\usepackage{amsfonts}       
\usepackage{nicefrac}       
\usepackage{array}

\usepackage{microtype}      
\usepackage{xcolor}         

\usepackage{wrapfig}
\usepackage{multirow}
\usepackage{makecell}
\usepackage{enumitem}       
\usepackage{graphicx}       
\usepackage{subcaption}
\usepackage{caption}
\usepackage{amsmath}
\usepackage{mathtools}
\usepackage{rotating}
\usepackage{caption}
\usepackage{subcaption}
\usepackage{bm}
\usepackage{appendix}
\usepackage{perpage} 
\MakePerPage{footnote} 
\usepackage{threeparttable}

\usepackage{xspace}
\newcommand{\system}{{TSGym}\xspace}

\newcommand{\nmodules}{{6}\xspace}

\newcommand{\bal}{\begin{align}}
\newcommand{\ean}{\end{align}}
\newcommand{\bit}{\begin{itemize}}
\newcommand{\eit}{\end{itemize}}
\newcommand{\ben}{\begin{enumerate}}
\newcommand{\een}{\end{enumerate}}
\newcommand{\beq}{\begin{equation}}
\newcommand{\eeq}{\end{equation}}

\newcommand{\rv}[1] {\textcolor{blue}{#1}}

\definecolor{aliceblue}{rgb}{0.91796875, 0.91796875, 0.999}
\definecolor{aliceyellow}{rgb}{0.999, 0.96875, 0.91796875}
\definecolor{alicegreen}{rgb}{0.91796875, 0.95703125, 0.91796875}
\definecolor{blue}{rgb}{0,0,1}

\newcommand{\update}[1]{{\textcolor{black}{#1}}}
\newcommand{\boldres}[1]{{\textbf{\textcolor{red}{#1}}}}
\newcommand{\secondres}[1]{{\underline{\textcolor{blue}{#1}}}}
\renewcommand{\rv}[1]{{{\textcolor{black}{#1}}}}

\newenvironment{rve}{\color{black}\bfseries}{}

\title{\system: Design Choices for Deep Multivariate Time-Series Forecasting}

\author{%
  \textbf{Shuang Liang}$^{1,\ast}$,
  \textbf{Chaochuan Hou}$^{1,\ast}$,
  \textbf{Xu Yao}$^{1,\ast}$,
  \textbf{Shiping Wang}$^{2,\ast}$,\\
  \textbf{Minqi Jiang}$^{1,\ast,\dagger}$,
  \textbf{Songqiao Han}$^{1,3,\dagger}$\textbf{,}
  \textbf{Hailiang Huang}$^{1,3,\dagger}$\textbf{,}\\
  $^1$AI Lab, Shanghai University of Finance and Economics
  $^2$Ant Group\\
  $^3$MoE Key Laboratory of Interdisciplinary Research of Computation and Economics\\
  \texttt{\{liangs1104,houchaochuan,yaoxu\}@stu.sufe.edu.cn},
  \texttt{shiping.wsp@antgroup.com},\\
  \texttt{\{jiangminqi,han.songqiao,hlhuang\}@shufe.edu.cn}
}

\begin{document}

\maketitle

\begin{abstract}



Recently, deep learning has driven significant advancements in multivariate time series forecasting (MTSF) tasks. However, much of the current research in MTSF tends to evaluate models from a holistic perspective, which obscures the individual contributions and leaves critical issues unaddressed. Adhering to the current modeling paradigms, this work bridges these gaps by systematically decomposing deep MTSF methods into their core, fine-grained components like series-patching tokenization, channel-independent strategy, attention modules, or even Large Language Models and Time-series Foundation Models. Through extensive experiments and component-level analysis, our work offers more profound insights than previous benchmarks that typically discuss models as a whole.

Furthermore, we propose a novel automated solution called TSGym for MTSF tasks. Unlike traditional hyperparameter tuning, neural architecture searching or fixed model selection, TSGym performs fine-grained component selection and automated model construction, which enables the creation of more effective solutions tailored to diverse time series data, therefore enhancing model transferability across different data sources and robustness against distribution shifts. Extensive experiments indicate that TSGym significantly outperforms existing state-of-the-art MTSF \rv{and AutoML} methods. All code is publicly available on \url{https://github.com/SUFE-AILAB/TSGym}.
\end{abstract}


\section{Introduction}
\label{sec:intro}



Multivariate time series refer to time series data involving multiple interdependent variables, which are widely present in various fields such as finance~\cite{sezer2020financial}, energy~\cite{alvarez2010energy,deb2017review}, traffic~\cite{cirstea2022towards,yin2016forecasting}, and health~\cite{bui2018time,kaushik2020ai}. Among the numerous analysis tasks, multivariate time series forecasting (MTSF) attracts substantial attention from the research community due to its significant practical applications. 
Traditional approaches to MTSF are largely based on statistical methods~\cite{abraham2009statistical,zhang2003time} and machine learning techniques~\cite{hartanto2023stock,masini2023machine}. In recent years, deep learning (DL) has become the most active area of research for MTSF, driven by its ability to handle complex patterns and large-scale datasets effectively~\cite{wang2024deep}.

Early academic efforts of deep MTSF methods like RNN-type methods~\cite{yamak2019comparison} are reported to struggle with capturing long-term temporal dependencies due to their inherent limitations of gradient vanishing or exploding problem~\cite{zhou2021informer,zhou2022fedformer}. More recently, Transformer~\cite{vaswani2017attention} shows significant potential, largely due to the effectiveness of its attention mechanisms in modeling temporal correlation~\cite{vaswani2017attention,wen2022transformers}.
Consequently, attention mechanism has continuously been studied in MTSF, with a focus on adapting them to time series data, for instance, by exploiting sparsity inductive bias~\cite{li2019enhancing,zhou2021informer}, transforming time and frequency domains~\cite{zhou2022fedformer}, and fusing multi-scale series~\cite{liu2022pyraformer}. 
While simpler MLP-based structures emerged~\cite{Zeng2022AreTE} offering alternatives to the established Transformer architecture in MTSF, notable modeling strategies like series-patching and channel-independent~\cite{nie2023PatchTST}, significantly enhanced the performance of Transformer-based methods, thereby sustaining research interest in them. Building upon these developments, large time-series models including large language models (LLMs)~\cite{jin2024position,zhou2023one,jin2023time} and time series foundation models (TSFMs)~\cite{jin2023lm4ts,liu2024timer} have recently been introduced, achieving promising results and fostering new research directions for MTSF.
Alongside these advancements in model architectures, active research within the deep MTSF community also focuses on other critical topics, such as variable (channels) dependency modeling~\cite{nie2023PatchTST,liu2024itransformer,zhang2023crossformer}, series normalization methods~\cite{Liu2022NonstationaryTR,fan2023dish}, and trend-seasonal decomposition~\cite{Zeng2022AreTE,liu2023koopa}.

As the field of MTSF continues to diversify, existing studies typically address critical concerns about methodological effectiveness, either by conducting large-scale benchmarks~\cite{wang2024deep,shao2024exploring,qiu2024tfb} or performing model selection via AutoML~\cite{abdallah2022autoforecast,fischer2024autoxpcr}. However, we  identify three main challenges with these prevailing approaches:
First, \textit{the granularity of existing studies is insufficient}. Current benchmarking works evaluate or select models as a whole, which hinders a deeper understanding of the mechanisms that drive model performance. In AutoML, this lack of granularity prevents breakthroughs beyond the limits of existing models. Second, \textit{the scope of existing studies is limited}. Current benchmarking and automated selection efforts are often confined to restricted model architectures or hyperparameters, without covering a broad range of data processing methods or feature modeling techniques. Third, \textit{the range of existing studies is narrow}. Existing studies tend to cover only a subset of network architectures and often lack discussions on more diverse models, such as LLMs and TSFMs.

\begin{figure}[htbp]
    \vspace{-0.1in}
    \centering
    \includegraphics[width=0.95\columnwidth]{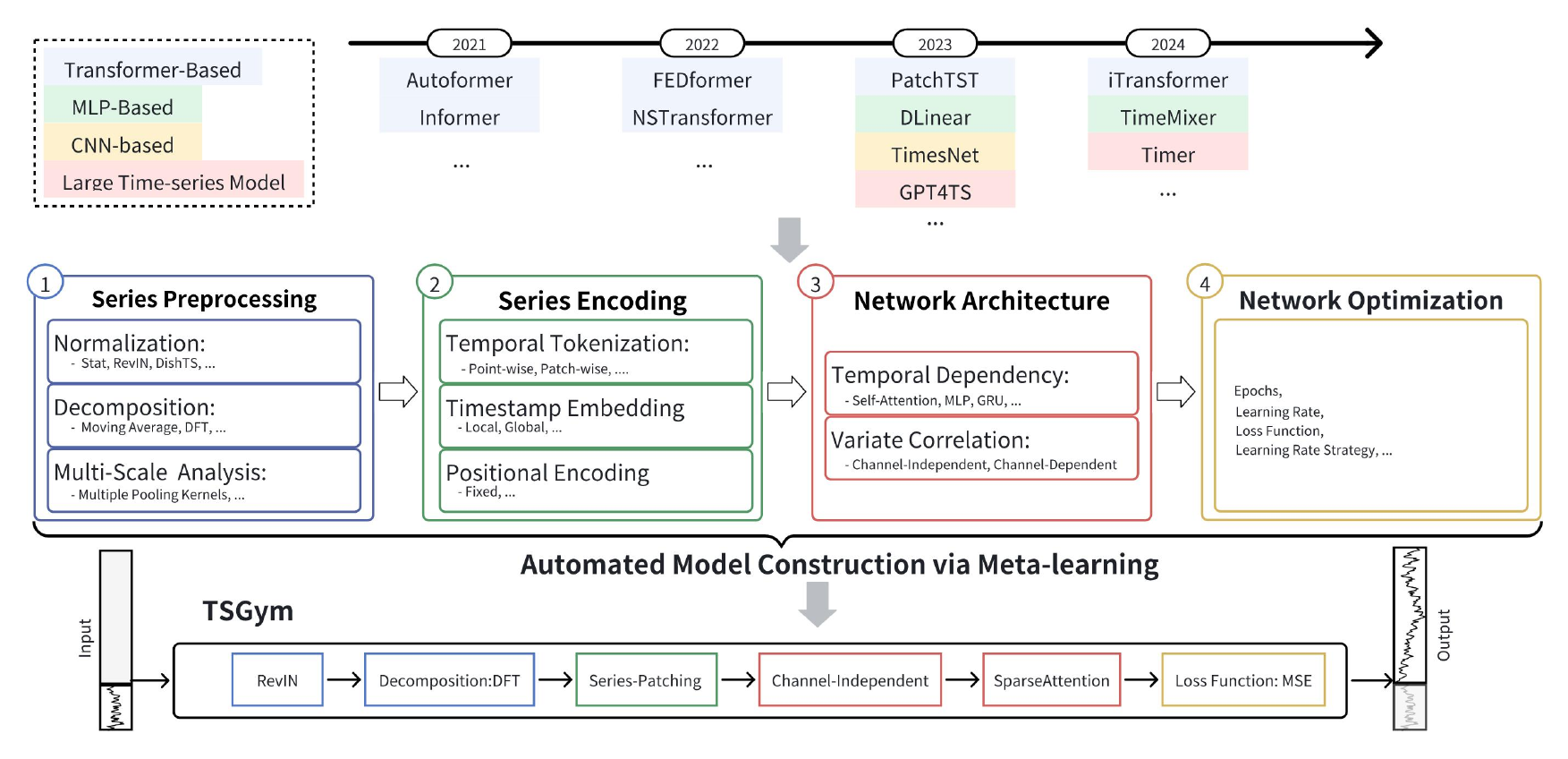}
    \caption{The Pipeline of Existing Multivariate Time Series Forecasting Methods.}
    \vspace{-0.1in}
    \label{fig:pipeline}
\end{figure}

To bridge these gaps, we propose TSGym—a framework designed for the \textbf{Large-scale Evaluation}, \textbf{Analysis}, and \textbf{Automated Model Construction} in deep MTSF tasks. Rather than viewing models as unified entities, TSGym systematically deconstructs popular deep MTSF methods by organizing them into distinct design dimensions that cover the entire time series modeling pipeline (see Fig.~\ref{fig:pipeline} and Table \ref{tab:design space}). 
Through extensive experiments, TSGym conducts fine-grained, isolated evaluations of core components, thereby identifying key design dimensions/choices and valuable insights from the vast MTSF methods. \rv{Leveraging a large-scale benchmark, TSGym systematically validates or refutes a series of prevailing claims within the MTSF community, including comparisons between Transformer and MLP architectures, as well as the adaptability of channel-independent approaches.}
Moreover, TSGym proposes the first component-level model construction in MTSF tasks, which effectively overcomes limitations in the previous automation methods by enabling more flexible and customized model designs tailored to data characteristics. Extensive experimental results indicate that the proposed TSGym generally outperforms existing SOTA methods. We summarize the key contributions of TSGym as follows:

\textbf{Component-level evaluation of MTSF methods}.
We propose TSGym, the first large-scale benchmark that systematically decouples deep MTSF methods. By evaluating $16$ design dimensions across $10$ benchmark datasets, TSGym \rv{elucidates contested issues in the current community} and offers key insights to inform future development for MTSF.

\textbf{Automated MTSF model construction}.
Leveraging meta-learning, TSGym develops models that outperform current SOTA methods, offering the MTSF community an effective, automated, and data-adaptive solution for model design.

\textbf{Discussion on emerging large time-series models}.
TSGym broadens current MTSF scope by applying systematic evaluation and automated combination not only to well-established models like MLP and Transformer, but also to novel large time-series models like LLMs and TSFMs.

\section{Related Work}
\label{sec:related}
\vspace{-0.1in}
\subsection{Deep Learning-based MTSF}

MTSF evolves from traditional statistical methods like ARIMA and Gaussian processes 
to modern deep learning approaches. Recurrent Neural Networks (RNNs) introduce memory mechanisms for sequential data but struggle with long-term dependencies. Temporal Convolutional Networks (TCNs) improve this by capturing multi-scale patterns, though their fixed window sizes limit global context. Transformers, using self-attention, enable long-range forecasting but introduce high computational complexity, leading to efficient variants like sparse attention \cite{wu2021autoformer} and patch-based models \cite{nie2023PatchTST}. Multilayer Perceptrons (MLPs) regain attention as simple yet effective models \cite{zeng2023dlinear}, with numerous variants offering competitive performance \cite{chen2023tsmixer, yi2023frets, das2023TiDE, liu2023koopa}. Leveraging NLP foundation models, LLM adaptation approaches use frozen backbones and prompt engineering \cite{jin2024timellm, zhou2023one} or fine-tuning \cite{chang2023llm4ts} to transfer pretrained knowledge. Simultaneously, pure TSFMs trained on large datasets achieve zero-shot generalization \cite{liu2024timer, goswami2024moment}, though constrained by Transformers' complexity. Our TSGym framework modularizes six core backbones—RNNs, CNNs, Transformers, MLPs, LLMs, and TSFMs—offering flexible, hybrid integration based on temporal dependencies and resource needs.

In recent advancements in MTSF, we summarize the design paradigm through a unified pipeline  (Fig.~\ref{fig:pipeline}), consisting of four stages: \textit{Series Preprocessing}$\rightarrow$\textit{Series Encoding}$\rightarrow$\textit{Network Architecture}$\rightarrow$\textit{Network Optimization}. Additionally, several specialized modules are proposed to enhance predictive accuracy by addressing non-stationarity, multi-scale dependencies, and inter-variable interactions. We categorize these developments into 6 specialized modules: 

\textbf{(1) Normalization} methods like RevIN \cite{kim2021RevIN} adjust non-stationary data, improving robustness against distribution shifts. \textbf{(2) Decomposition methods}, such as Autoformer \cite{wu2021autoformer}’s trend-seasonality separation, isolate non-stationary components, making the data more predictable by separating trends from seasonality. \textbf{(3) Multi-scale analysis} extracts temporal patterns across granularities, as in TimeMixer \cite{wang2024timemixer}, capturing both high-frequency fluctuations and low-frequency trends through hierarchical resolution modeling. \textbf{(4) Temporal tokenization techniques} like PatchTST\cite{nie2023PatchTST}’s subseries-level embedding  represent time series hierarchically, improving the capture of complex temporal semantics. \textbf{(5) Temporal dependency} modeling through architectures like Transformers leverages self-attention to capture long-range dependencies, effectively modeling both short- and long-term relationships. \textbf{(6) Variate correlation learning}, exemplified by DUET \cite{qiu2025DUET}, models inter-variable dependencies using frequency-domain metric learning, improving predictions by capturing interactions across variables.

To provide a more detailed categorization and comprehensive technical specifications, please refer to Appx.~\ref{appx:modules}. Due to the extensive focus and continuous evolution of these modules in MTSF research, TSGym strives to decouple and modularize these key modules, exploring their real contributions and enabling more flexible model structure selection and configuration.

\vspace{-0.1in}

\subsection{Benchmarks for Time Series Forecasting}
Recent time series forecasting benchmark studies~\cite{wang2024deep,shao2024exploring,qiu2024tfb,liutimer} have conducted large-scale experiments across a diverse range of datasets. However, most of these works treat current models as monolithic entities. 
TSlib\footnote{https://github.com/thuml/Time-Series-Library}~\cite{wang2024deep}, one of the most popular repositories for time series analysis, provides a comprehensive survey and evaluates recent time series models across various time series analysis tasks. 
From the perspective of time series characteristics, BasicTS~\cite{shao2024exploring} analyzes model architectures and the strategy of treating channels (or variables) independently. 
With a more extensive experimental setup, TFB~\cite{qiu2024tfb} additionally includes machine learning and statistical forecasting methods, and covers datasets from a broader range of domains. More recently, 
OpenLTM\footnote{https://github.com/thuml/OpenLTM}~\cite{liutimer} provides a system to evaluate Time Series Foundation Models as well as Large language Models for time series methods. 
Although some surveys and benchmarks analyze the fine-grained components of time series models, their scope is often limited. Wen et al.~\cite{wen2020time} discuss various time series data augmentation techniques and evaluate their effectiveness. Another survey~\cite{wen2022transformers} systematically reviews fine-grained components within Transformer-based architectures, but lacks broader coverage of model structures and experimental evaluations. 

\rv{These limitations prevent the aforementioned studies from comprehensively and meticulously evaluating the entire MTSF pipeline, spanning from sequence preprocessing to model parameter optimization.}
To the best of our knowledge, TSGym is the first benchmark that not only provides component-level fine-grained analysis, but also conducts large-scale empirical evaluations.

\subsection{AutoML for Time Series Forecasting}
\vspace{-0.05cm}

Current automated approaches for DL-based MTSF can be categorized into ensemble-based~\cite{shchur2023autogluon} and meta-learning-based~\cite{abdallah2022autoforecast,fischer2024autoxpcr} methods. The former fits and integrates various models from a predefined pool with ensemble techniques, which inevitably incurs substantial computational cost. The latter leverages meta-features to characterize datasets and selects optimal models for the given datasets. However, both approaches operate at the model level and struggle to surpass the performance ceiling of existing methods. AutoCTS++~\cite{wu2024autocts++} achieves automated selection by searching over model architectures and hyperparameters, but its search space is limited in scope. In contrast, TSGym is the first framework to support automated selection over a wide range of fine-grained components for MTSF, extending beyond narrow model structures, hyperparameters, and data processing strategies.

A closely related work is our previous effort, ADGym\cite{jiang2023adgym}, which is designed for tabular anomaly detection with model decomposition. Differently, TSGym deals with multivariate time series data, which presents more complex data processing design choices, such as series sampling, series normalization, and series decomposition. \rv{Second,} TSGym considers finer-grained model structures, such as various attention variants in Transformers, and broader network types, including LLMs and TSFMs. \rv{Third, TSGym explores the value of Optuna\cite{akiba2019optuna}, a Bayesian-optimization-driven intelligent search framework, which attains a superior design space at markedly lower cost and thus enhances the efficacy of TSGym.} It is worth mentioning that the success of TSGym validates the universality of the model decomposition framework, marking an innovation and progression distinct from ADGym. 
Further details on the differences between two works can be found in Appx.\ref{appx:tsgym_vs_adgym}.

\vspace{-0.1in}

\section{\system: \textit{Benchmarking} and \textit{Automating} Design Choices in Deep MTSF}
\label{sec: setting}
\vspace{-0.1in}
\subsection{Problem Definition for MTSF}
\vspace{-0.1in}

In this paper, we focus on the common MTSF settings for time series data containing $C$ variates. Given historical data $\mathbf{\chi}=\left\{\boldsymbol{x}_{1}^{t},\ldots, \boldsymbol{x}_{C}^{t}\right\}^{L}_{t=1}$, where $L$ is the look-back sequence length and $\boldsymbol{x}_{i}^{t}$ is the $i$-th variate, the forecasting task is to predict $T$-step future sequence $\mathbf{\hat{\chi}}=\left\{\boldsymbol{\hat{x}}_{1}^{t},\ldots, \boldsymbol{x}_{C}^{t}\right\}^{L+T}_{t=L+1}$. 
To avoid error accumulation ($T>1$), we directly predict all future steps, following~\cite{zhou2021informer}.

\vspace{-0.1in}
\subsection{Large Benchmarking towards Design Choices of Deep MTSF}
\label{subsec:design_choices}
\vspace{-0.1in}
Considering the above numerous methods \rv{proposed} for MTSF tasks, the foremost priority involves decoupling the current state-of-the-art (SOTA) methods and further conducting large-scale benchmark to identify the core components that really drive the improvements in time-series forecasting. 

Following the taxonomy of the previous study \cite{wen2022transformers, zeng2023dlinear}, we decouple existing SOTA methods according to the standard process of MTSF modeling, while significantly expanding the diversity of the modeling pipeline. Based on the flow direction from the input to the output sequence, the \textbf{Pipeline} of TSGym includes: \textit{Series Preprocessing}$\rightarrow$\textit{Series Encoding}$\rightarrow$\textit{Network Architecture}$\rightarrow$\textit{Network Optimization}, as is demonstrated in Fig.~\ref{fig:pipeline}. Moreover, we structure each pipeline step according to distinct \textbf{Design Dimensions}, where a DL-based time-series forecasting model can be instantiated by specified \textbf{Design Choices}, as is shown in Table~\ref{tab:design space}.

\begin{table}[!t]
\footnotesize
  \centering
  \caption{
  \system supports comprehensive design choices for deep time-series forecasting methods.}
  \resizebox{0.95\textwidth}{!}{
    \begin{tabular}{l|ll}
    \toprule
    \multicolumn{1}{l}{\textbf{Pipeline}} & \textbf{Design Dimensions} & \textbf{Design Choices} \\
    \midrule
    \multirow{3}{*}{$\downarrow$Series Preprocessing} 
     &    Series Normalization   & [None, Stat, RevIN, DishTS] \\
     &    Series Decomposition   & [None, MA, MoEMA, DFT] \\
     &    Series Sampling/Mixing   & [False, True] \\
    \hline
    \multirow{2}{*}{$\downarrow$Series Encoding} 
    &   Channel Independent   & [False, True] \\
    &  Sequence Length    & [48, 96, 192, 512] \\
    &   Series Embedding    & [Inverted Encoding, Positional Encoding, Series Patching] \\
    \hline
    \multirow{7}{*}{$\downarrow$Network Architecture} 
    & Network Type & [MLP, RNN, Transformer, LLM, TSFM] \\
    & Series Attention & [Null, SelfAttn, AutoCorr, SparseAttn, FrequencyAttn, DestationaryAttn] \\
    & Feature Attention & [Null, SelfAttn, SparseAttn, FrequencyAttn] \\
          & $d\_model$ & [64, 256] \\
          & $d\_ff$ & [256, 1024] \\
          & Encoder Layers & [2, 3] \\
    \hline
    \multirow{4}{*}{$\downarrow$Network Optimization} 
    & Epochs & [10, 20, 50] \\
          & Loss Function & [MSE, MAE, HUBER] \\
          & Learning Rate & [1e-3, 1e-4] \\
          & Learning Rate Strategy & [Null, Type1] \\
    \bottomrule
    \end{tabular}%
    }
  \label{tab:design space}%
\end{table}


Through the proposed design dimensions and choices, TSGym provides detailed description of time-series modeling pipeline, disentangling key elements within mainstream time-series forecasting methods and facilitating component-level comparison/automated construction. For example, TSGym includes multi-scale mixing module proposed in TimeMixer~\cite{wang2024timemixer}, Inverted Encoding method proposed in iTransformer~\cite{liu2024itransformer}, Channel-independent strategy and Series-Patching encoding used in PatchTST~\cite{nie2023PatchTST}, various attention mechanism discussed in \cite{wen2022transformers}, and also LLM and TSFM network type choices that are often integrated without fully considering their interactions with other design dimensions. Detailed descriptions of all design choices are provided in Appx.~\ref{appx:design_choices_details}.



\subsection{Automated construction MTSF models via \system}
\label{subsec:TSGym_overview}
\textbf{Overview}. Differing from traditional methods that focus on selecting an off-the-shelf model, TSGym aims to customize models given the downstream MTSF tasks and data descriptions. Given a pre-defined conflict-free model set $\mathcal{M}=\{M_1,...,M_m\}$, each model $M_{i}$ is instantiated by the design choice combinations illustrated in Table~\ref{tab:design space}. TSGym learns the mapping function from these automatically combined models to their associated forecasting performance on the training datasets, and generalize to the test dataset(s) to select the best model based on predicted results.


%


\textbf{Meta-learning for automated MTSF model construction}. 
Formally speaking, TSGym propose $k$ design dimensions $\mathcal{DD}=\{DD_1,...,DD_k\}$ for comprehensively describing each step of aforementioned pipeline in deep learning time-series modeling. Each design dimension $DD_{i}$ represents a set containing elements of different design choices $DC$. By taking the Cartesian product of the sets $\mathcal{DD}$ corresponding to different design dimensions, we obtain the pool of all valid model combinations
$\mathcal{M}=DD_1 \times DD_2 \times \cdots \times DD_k = \{(DC_1, DC_2, \ldots, DC_k) \mid DC_i \in DD_i, i = 1, 2, \ldots, k\}$. Considering the potentially large number of combinations and the computational cost, we randomly sampled $\mathcal{M}$ to $\mathcal{M}_s$, where $M_i=(DC_1=RevIN, DC_1=DFT, ..., DC_k=Type1) \in \mathcal{M}_s$, for example, which means $M_i$ instantiates RevIN method to normalize input series, then decompose it to the seasonal and trend term. Subsequently, following the Series Encoding and Network Architecture constructing pipeline (as illustrated in Table~\ref{tab:design space}), finally the Type1, i.e., a step decay learning rate strategy is employed to adjust the learning rate for updating the model parameters.

Suppose we have $n$ training datasets $\bm{\mathcal{D}}_{\text{train}}=\{\mathcal{D}_1,\ldots,\mathcal{D}_n\}$ and the number of sampled model combinations (i.e., the size of the set $\mathcal{M}_s$) is $m$, TSGym conducts extensive experiments on $n$ historical training datasets to evaluate and further collect the forecasting performance of $m$ model combinations. TSGym then acquire the MSE performance matrix $\bm{P}\in \mathbb{R}^{n \times m}$, where $\bm{P}_{i,j}$ corresponds to the $j$-th auto-constructed MTSF model's performance on the $i$-th training dataset. Since the difficulty of prediction tasks varies across training datasets, leading to significant differences in the numerical range of performance metrics. Directly using these metrics (e.g., MSE) as training targets of a meta-predictor may result in overfitting on more difficult dataset(s). Therefore, we convert the performance metrics of $\mathcal{M}_s$ into their corresponding normalized ranking, where $\bm{R}_{i,j}=rank(P_{i,j})/m \in [0,1]$ and smaller values indicate better performance on the corresponding dataset.

Distinguished from previous model selection approaches~\cite{abdallah2022autoforecast, abdallah2025evaluation}, TSGym decouples more recently MTSF methods (including MLP-Mixer-type, Transformer-based, LLM and TSFM models), and supports fine-grained model construction at the component level, rather than being constrained to a fixed, limited set of existing models, which enables significantly greater flexibility and effectiveness. Specifically, TSGym follows the idea of meta-learning to construct a meta-predictor that learns the mapping function $f(\cdot)$ from training dataset $\mathcal{D}_i$ and model combination $M_j$, to the performance rankings $\bm{R}_{i,j}$, as is shown in Eq.~\ref{eq:meta-predictor}. We leverage meta-features $\mathbf{E}_{i}^{meta}$ \rv{from the training set of each dataset}, which capture multiple aspects such as statistical, temporal, spectral, fractal features, and distribution shift metrics to fully describe the complex data characteristics of time series datasets. Learnable continuous embeddings $\mathbf{E}_{j}^{comp}$ are used to represent different model combinations and are updated through the gradient backpropagation of the meta-predictor. \rv{At test time, only the meta-features extracted from the training split of the target dataset are required to identify the top-performing component, eliminating any redundant experimentation on the test set.}

\vspace{-0.1in}
\begin{equation}
    \color{black}
    f(\mathcal{D}_i, M_j)=\bm{R}_{i,j}, 
    f \;: \underbracket{\mathbf{E}_{i}^{meta}}_{\text{meta features}}, 
    \underbracket{\mathbf{E}_{j}^{comp}}_{\text{component embed.}}
    \mapsto \bm{R}_{i,j} \;\;,\;\;  
    i\in \{1,\ldots,n\}, \; j\in \{1,\ldots,m\}
    \label{eq:meta-predictor}
\end{equation} 
We used a simple two-layer MLP as the meta-predictor and trained it through a regression problem, thereby transferring the learned mapping to new test datasets.
For a newcoming dataset (i.e., test dataset $\mathbf{X}_{\text{test}}$), we acquire the predicted relative ranking of different components using the trained $f(\cdot)$, and select top-$1$ ($k$) to construct MTSF model(s). Note this procedure is zero-shot without needing any neural network training on $\mathbf{X}_{\text{test}}$ but only extracting meta-features and pipeline embeddings. We show the effectiveness of the meta-predictor in \S \ref{exp:meta-predictor}.

\vspace{-0.1in}

\section{Experiments}
\label{sec: exp}
\subsection{Experiment Settings}

\textbf{Datasets}. 
Following most prior works~\cite{wu2021autoformer,wu2023timesnet,jin2024timellm}, we adopt 9 datasets as experimental data for MTSF tasks, ETT (4 subsets), Traffic, Electricity, Weather, Exchange, ILI. And we utilize the M4 dataset for short-term forecasting tasks. The forecast horizon \( L \) for long-term forecasting is \( \{96, 192, 336, 720\} \), while for the ILI dataset, it is  \( \{24, 36, 48, 60\} \). For short-term forecasting, the forecast horizons are \( \{6, 8, 13, 14, 18, 48\} \). More details can be seen in Appx. \ref{appx:data}.


\textbf{Baseline}. We present a comprehensive set of baseline comparison experiments \rv{including MTSF and AutoML methods,} to demonstrate the superior performance of the pipelines automatically constructed by TSGym. Due to space limitations, the baseline methods presented in this section include the latest clustering-based approach DUET~\cite{qiu2025DUET}, time series mixing methods (TimeMixer~\cite{wang2024timemixer}), MLP-based methods (MICN~\cite{wang2023micn}), RNN-based methods (SegRNN~\cite{lin2023segrnn}), CNN-based methods (TimesNet~\cite{wu2023timesnet}), and Transformer-based methods (PatchTST~\cite{nie2023PatchTST}, Crossformer~\cite{zhang2023crossformer}, Autoformer~\cite{wu2021autoformer}). We present experiments based on the complete baseline in the Appx. \ref{appx:complete_results}.

\textbf{Evaluation Metrics}. We follow the experimental setup of most prior works, using Mean Squared Error (MSE) and Mean Absolute Error (MAE) as evaluation metrics for long-term forecasting tasks, and using Symmetric Mean Absolute Percentage Error (SMAPE), Mean Absolute Scaled Error (MASE), and Overall Weighted Average (OWA) as metrics for short-term forecasting tasks. The mathematical formulas for these evaluation metrics are provided in the Appx. \ref{appx:metrics}.

\textbf{Meta-predictor in \system}. 
The meta-predictor is instantiated as a two-layer MLP and trained for 100 epochs with early stopping. The training process utilizes the Adam optimizer \cite{kingma2014adam} with a learning rate of 0.001 and batch size of 512. See details in Appx. \ref{appx:meta-predictors}.

\subsection{Large Evaluation on Time-Series Design Choices}
\vspace{-0.1in}
\label{exp:benchmark}

In this work, we perform large evaluations on the decoupled pipelines according to the standard procedure of MTSF methods. 
Such analysis is often overlooked in previous studies, and we investigate each design dimension of decoupled pipelines by fixing its corresponding design choice (e.g., Self Attention), and randomly sampling other dimensional design choices to construct MTSF pipelines.

\begin{figure}[t!]
     \centering
     \begin{subfigure}[t]{0.28\textwidth}
         \centering
         \includegraphics[width=\textwidth]{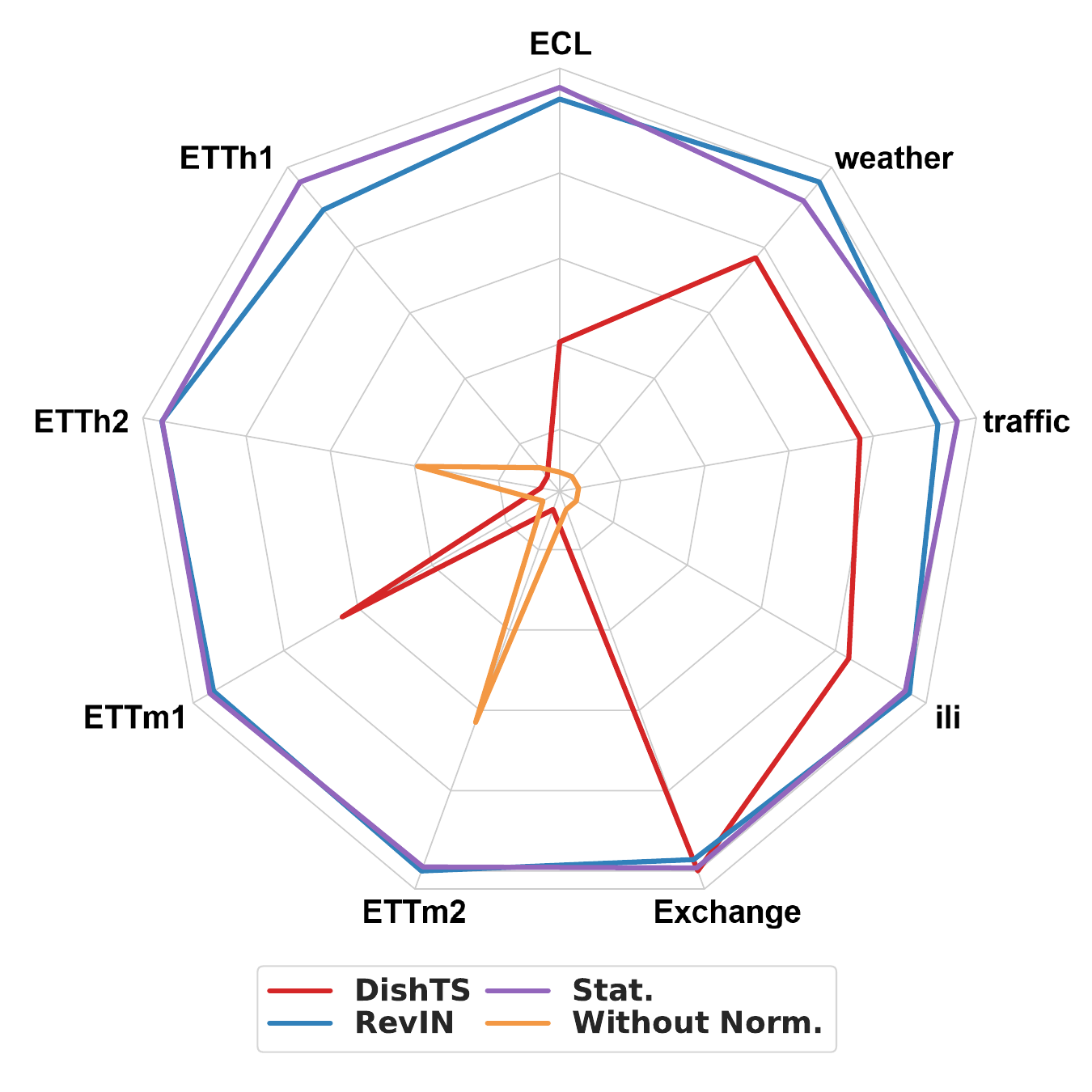}
         \caption{Series Normalization}
         \label{fig:exp-Normalization}
     \end{subfigure}
     \hspace{10pt}
     \begin{subfigure}[t]{0.28\textwidth}
         \centering
         \includegraphics[width=\textwidth]{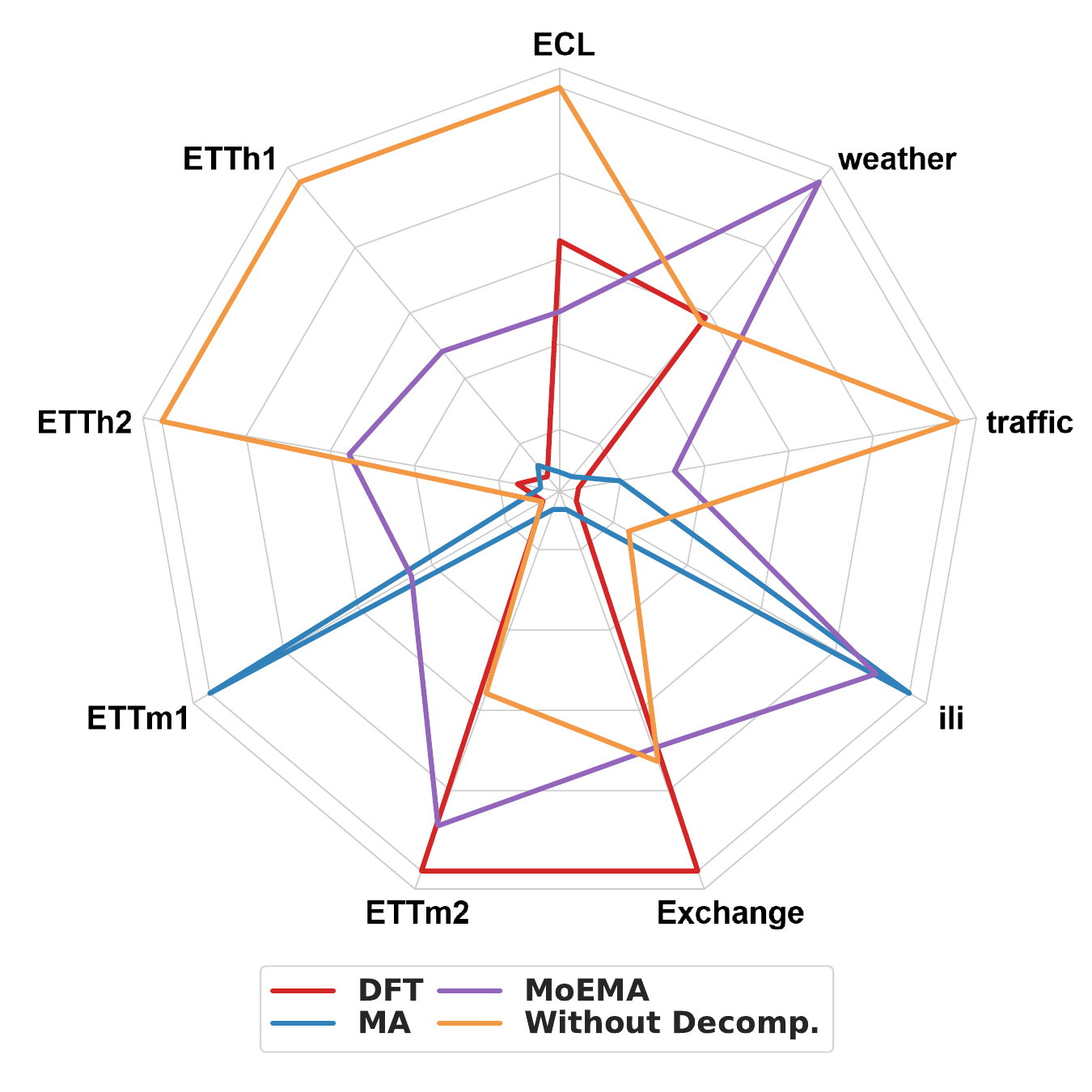}
         \caption{Series Decomposition}
         \label{fig:exp-Decomposition}
     \end{subfigure}
     \hspace{10pt}
    \begin{subfigure}[t]{0.28\textwidth}
         \centering
         \includegraphics[width=\textwidth]{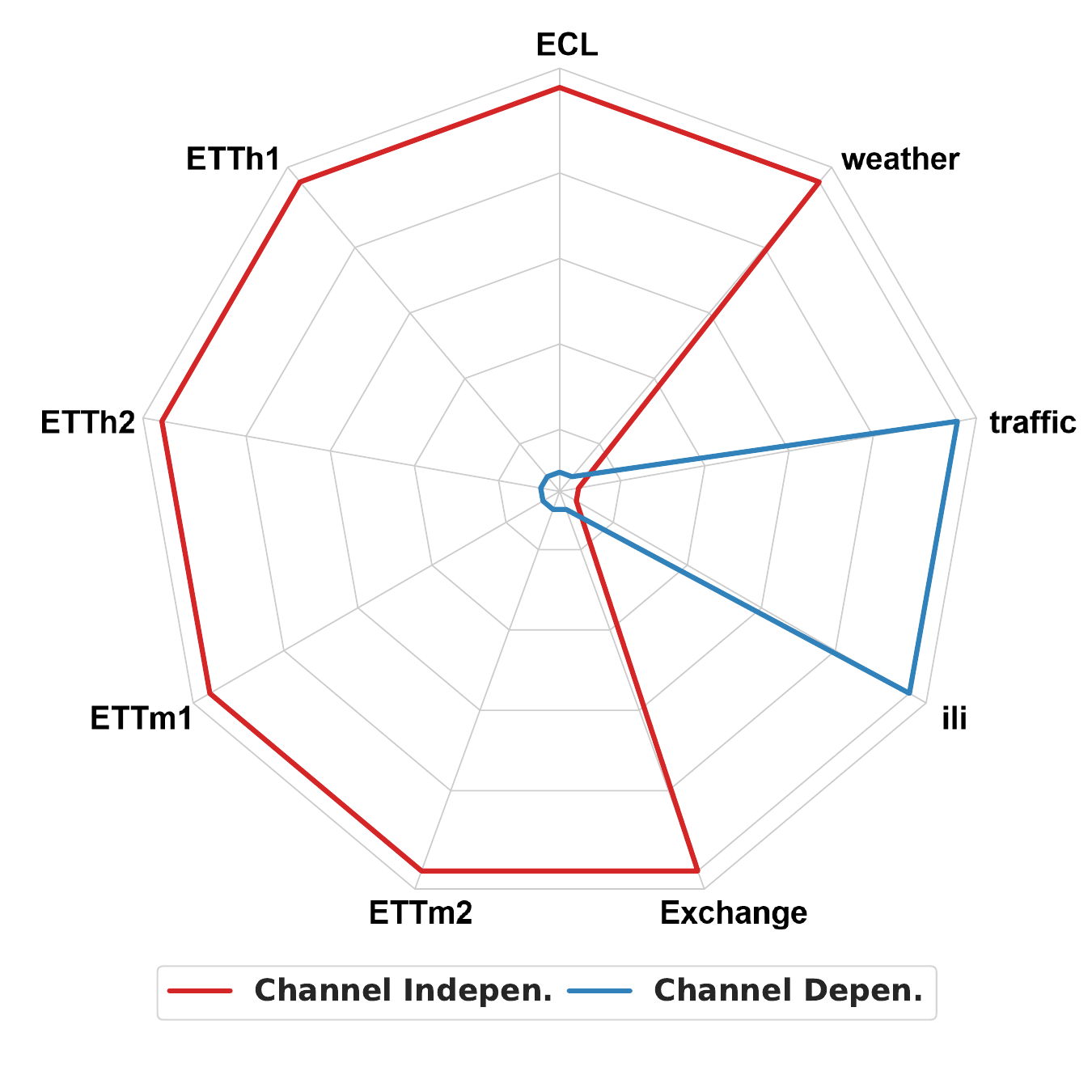}
         \caption{Channel Independent}
         \label{fig:exp-CI}
     \end{subfigure}
     \begin{subfigure}[t]{0.28\textwidth}
         \centering
         \includegraphics[width=\textwidth]{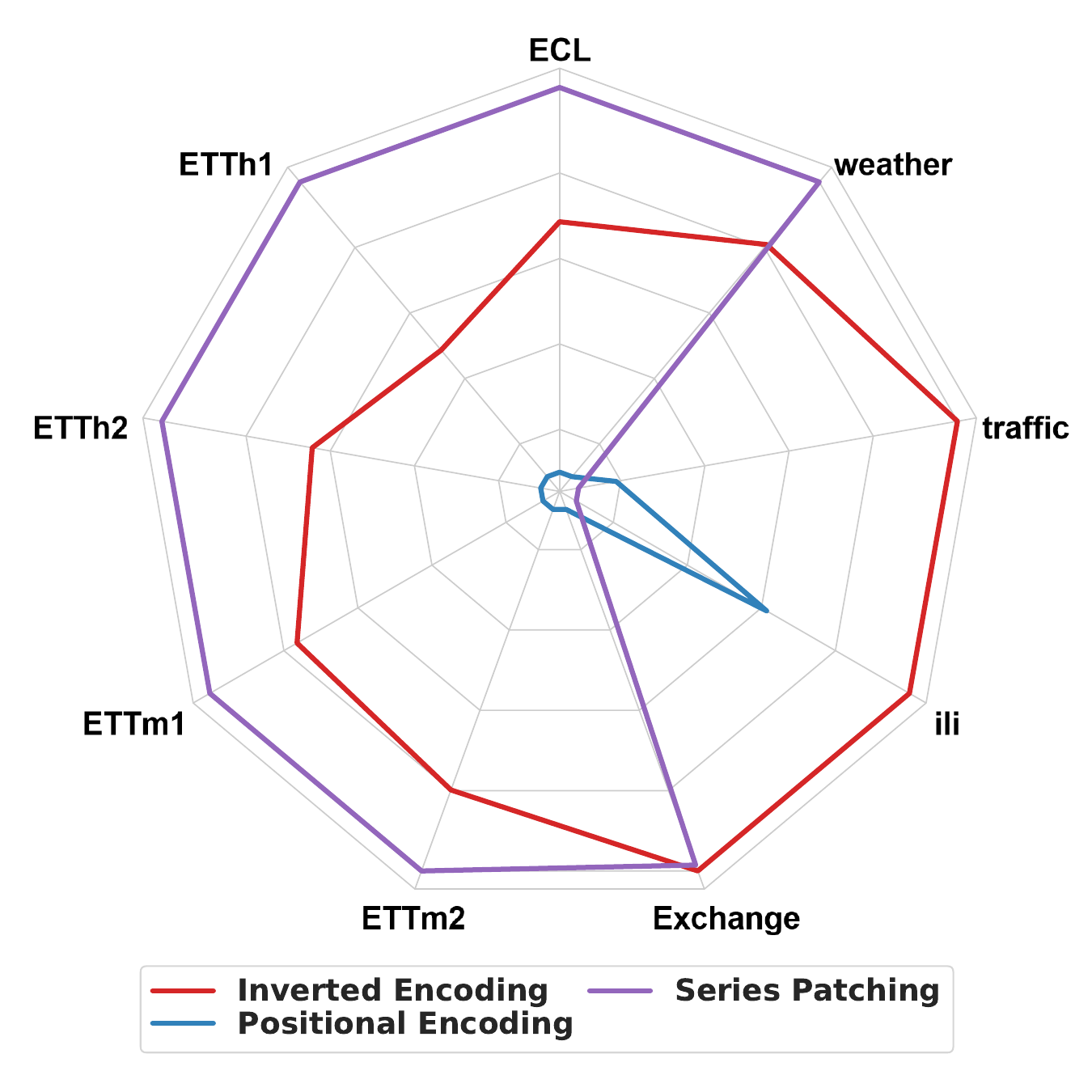}
         \caption{Series Tokenization}
         \label{fig:exp-Tokenization}
     \end{subfigure}
     \hspace{10pt}
     \begin{subfigure}[t]{0.28\textwidth}
         \centering
         \includegraphics[width=\textwidth]{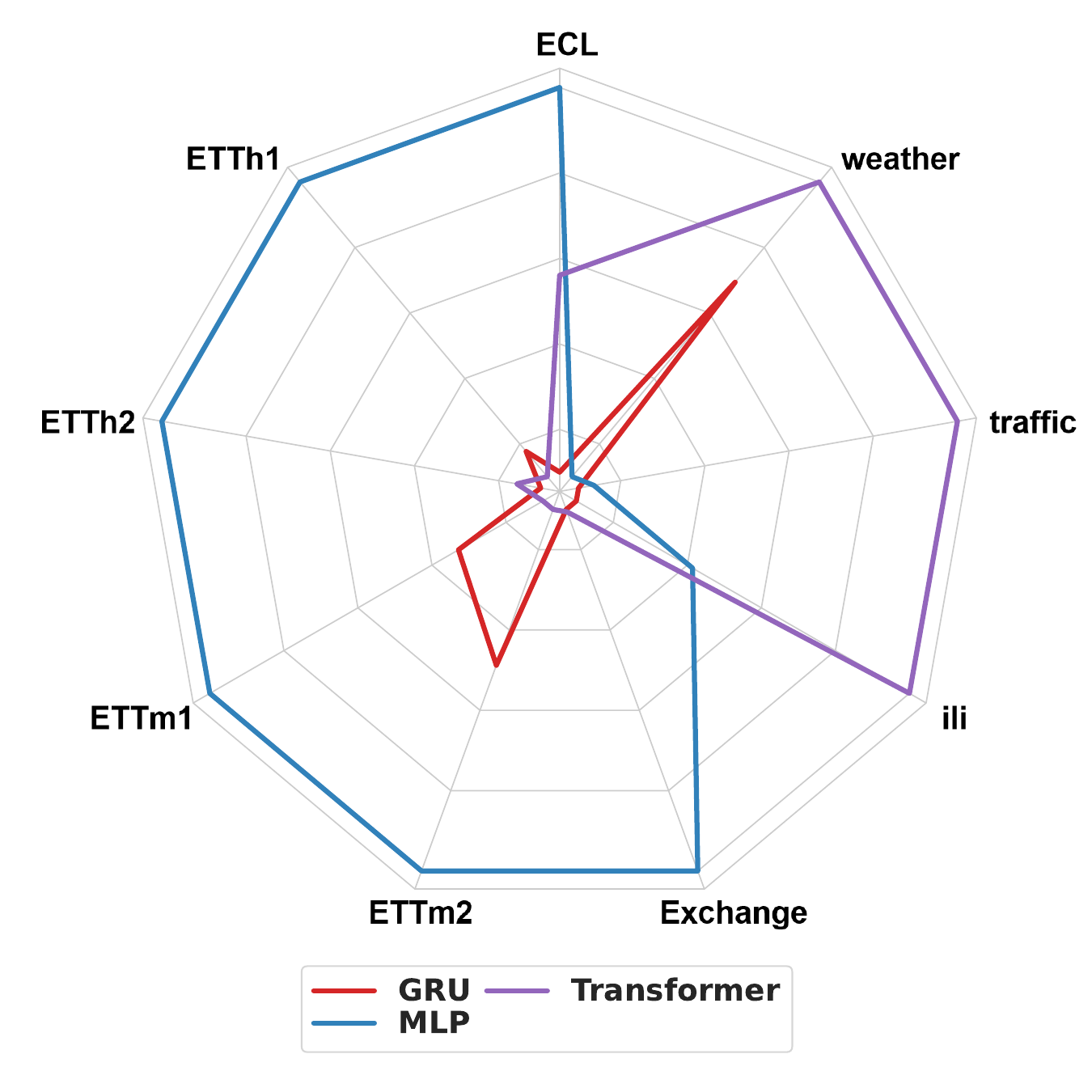}
         \caption{Network Backbone}
         \label{fig:exp-backbone}
     \end{subfigure}
     \hspace{10pt}
     \begin{subfigure}[t]{0.28\textwidth}
         \centering
         \includegraphics[width=\textwidth]{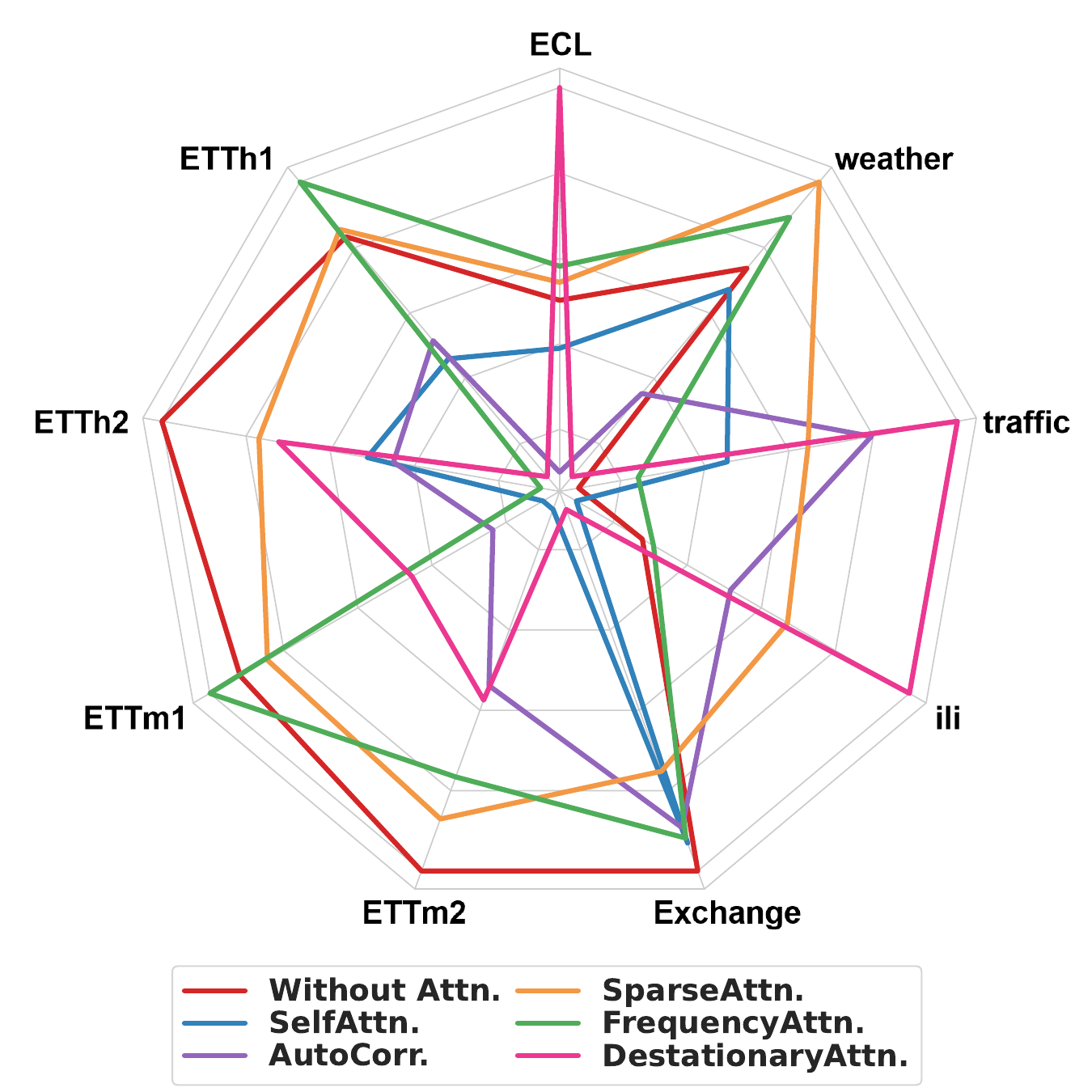}
         \caption{Series Attention}
         \label{fig:exp-attention}
     \end{subfigure}
     \caption{Overall performance across key design dimensions. The results (MSE) are based on the top 25th percentile across all forecasting horizons.}
     \vspace{-0.1in}
     \label{fig:exp-rada}
\end{figure}

In the following sections, we provide systematic conclusions based on long-term MTSF experimental results, addressing several gaps in the current MTSF research community. Specifically, different design choices are compared and demonstrated using a spider chart, where each vertex represents a dataset. Design choices that are closer to the vertices exhibit superior performance in their respective design dimensions. We analyze these components based on the different design dimensions, namely \textbf{\textit{Series Preprocessing}}, \textbf{\textit{Series Encoding}} and \textbf{\textit{Network Construction}}. Finally, we evaluated the performance on four datasets under the fixed model architectures of LLMs or TSFMs, and discussed the effects of various design choices in the subsection on \textbf{\textit{Large Time Series Models}}. \rv{Figure~\ref{fig:exp-rada} presents the average results of key components across diverse datasets and forecast horizons.} More detailed information, including the complete design choices and their performance on short-term time series forecasting tasks, is provided in the Appx. \ref{appx:complete_results}. \rv{Based on the comprehensive experiments results, we conduct an in-depth investigation of the components, formulate the seven most contentious claims in the research community, and clarify them with our benchmark; the results are presented in Appx.  \ref{appx:deep_analysis}.}


\textbf{\textit{Series Preprocessing}}. Based on the systematic evaluation of Series Preprocessing strategies in Fig.~\ref{fig:exp-Normalization} and Fig.~\ref{fig:exp-Decomposition}, several key insights emerge. 
Series Normalization (Fig.~\ref{fig:exp-Normalization}) proves universally effective, with RevIN and Stationary achieving the lowest MSE (top 25th percentile) across diverse datasets, establishing them as essential baselines for stabilizing non-stationary dynamics. 
In contrast, Series Decomposition (Fig.~\ref{fig:exp-Decomposition}) demonstrates varying effectiveness depending on the dataset. For example, decomposition methods improve performance on datasets like ETTm1 when using MA, while others, such as ETTh1, ECL, ETTh2, and traffic datasets, perform better without decomposition. 

\textbf{\textit{Series Encoding}}.
Several clear observations emerge from the Series Encoding stage, as illustrated in Fig.\ref{fig:exp-CI} and Fig.\ref{fig:exp-Tokenization}.
First, as shown in panel Fig.~\ref{fig:exp-CI}, channel-independent methods consistently outperform channel-dependent ones, with the exception of traffic and ILI, highlighting the advantage of modeling each variable separately.
Regarding tokenization strategies Fig.~\ref{fig:exp-Tokenization}, both patch-wise (sequence patching) and series-wise (inverted encoding) approaches outperform the traditional point-wise encoding. Among them, patch-wise encoding shows strong performance across most datasets, whereas inverted encoding proves particularly effective on the traffic dataset.


\textbf{\textit{Network Construction}}. 
As is shown in Fig.~\ref{fig:exp-backbone}, we find that complex network architectures like the Transformer are not always necessary, which only performs better than MLP on weather, traffic and ILI datasets. This result is reasonable since more sophisticated DL backbone like Transformer and its variants often require specific hyperparameter combinations and tailored architectural designs to achieve satisfactory performance. We indicate that these observed results further emphasize the importance of automated model construction tailored to specific data characteristics. Indeed, we show in \S~\ref{exp:meta-predictor} that including the Transformer architecture in the set of design choices would enhance the performance of constructed model via TSGym.
Fig.~\ref{fig:exp-attention} suggests that employing Attention mechanisms for modeling dependencies of input sequence does not offer significant advantages, which aligns with recent findings~\cite{Zeng2022AreTE}. Additionally, the spider chart shows no significant performance differences were observed among different variants of attention mechanisms.

\begin{figure}[t!]
     \centering
     \begin{subfigure}[t]{0.3\textwidth}
         \centering
         \includegraphics[width=\textwidth]{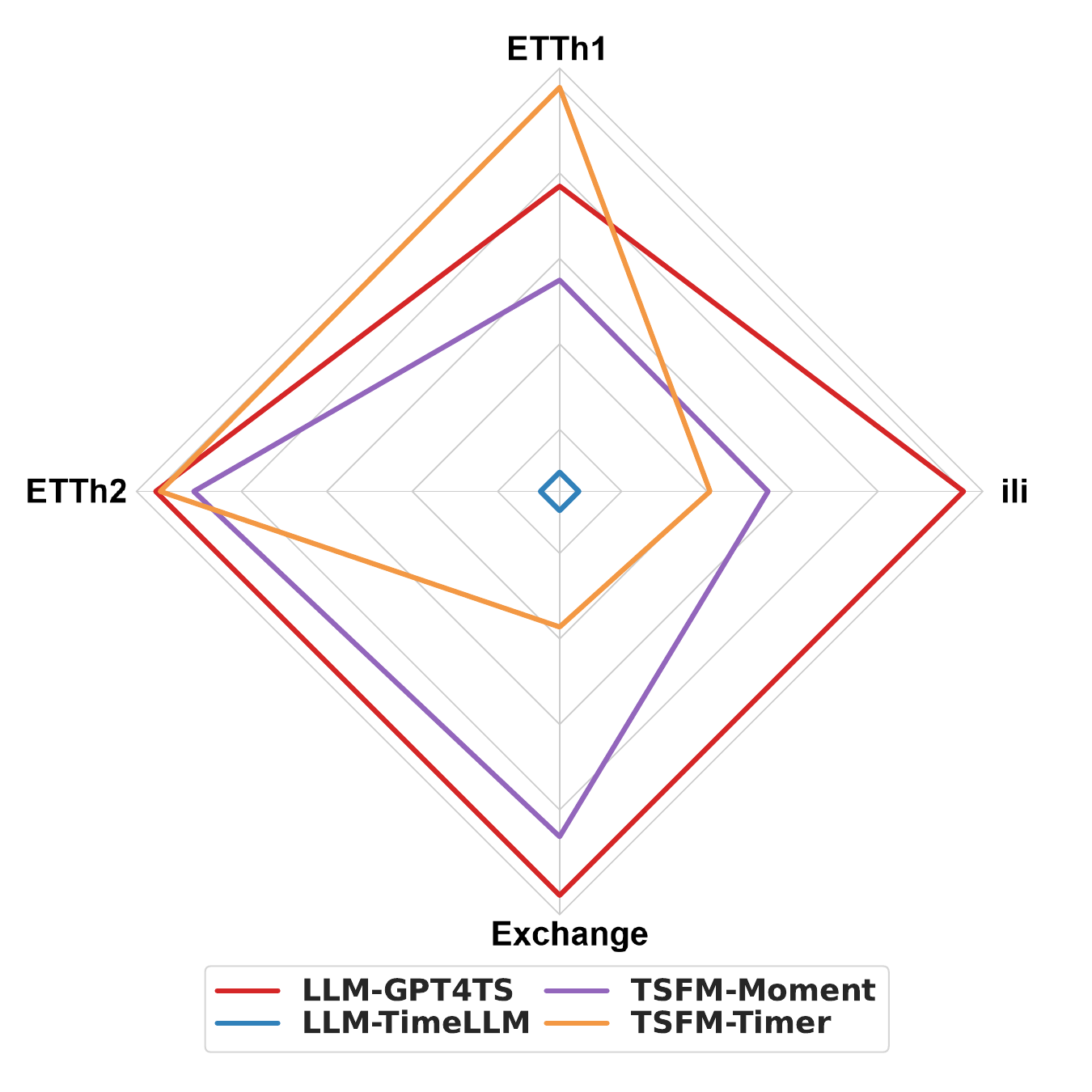}
         \caption{Network Backbone}
         \label{fig:exp-llmbackbone}
     \end{subfigure}
     \hspace{10pt}
     \begin{subfigure}[t]{0.3\textwidth}
         \centering
         \includegraphics[width=\textwidth]{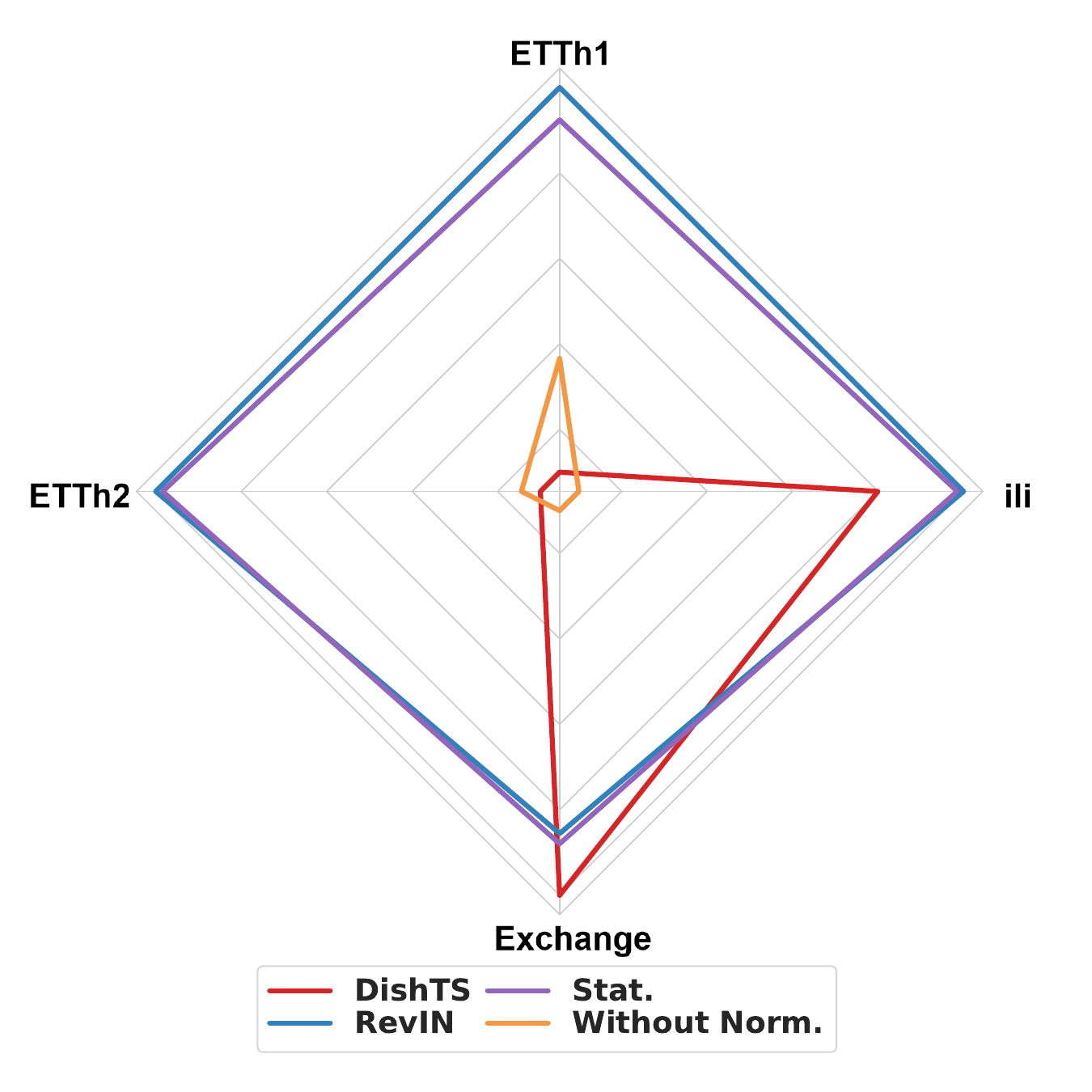}
         \caption{Series Normalization}
         \label{fig:exp-llmnorm}
     \end{subfigure}
     \hspace{10pt}
     \begin{subfigure}[t]{0.3\textwidth}
         \centering
         \includegraphics[width=\textwidth]{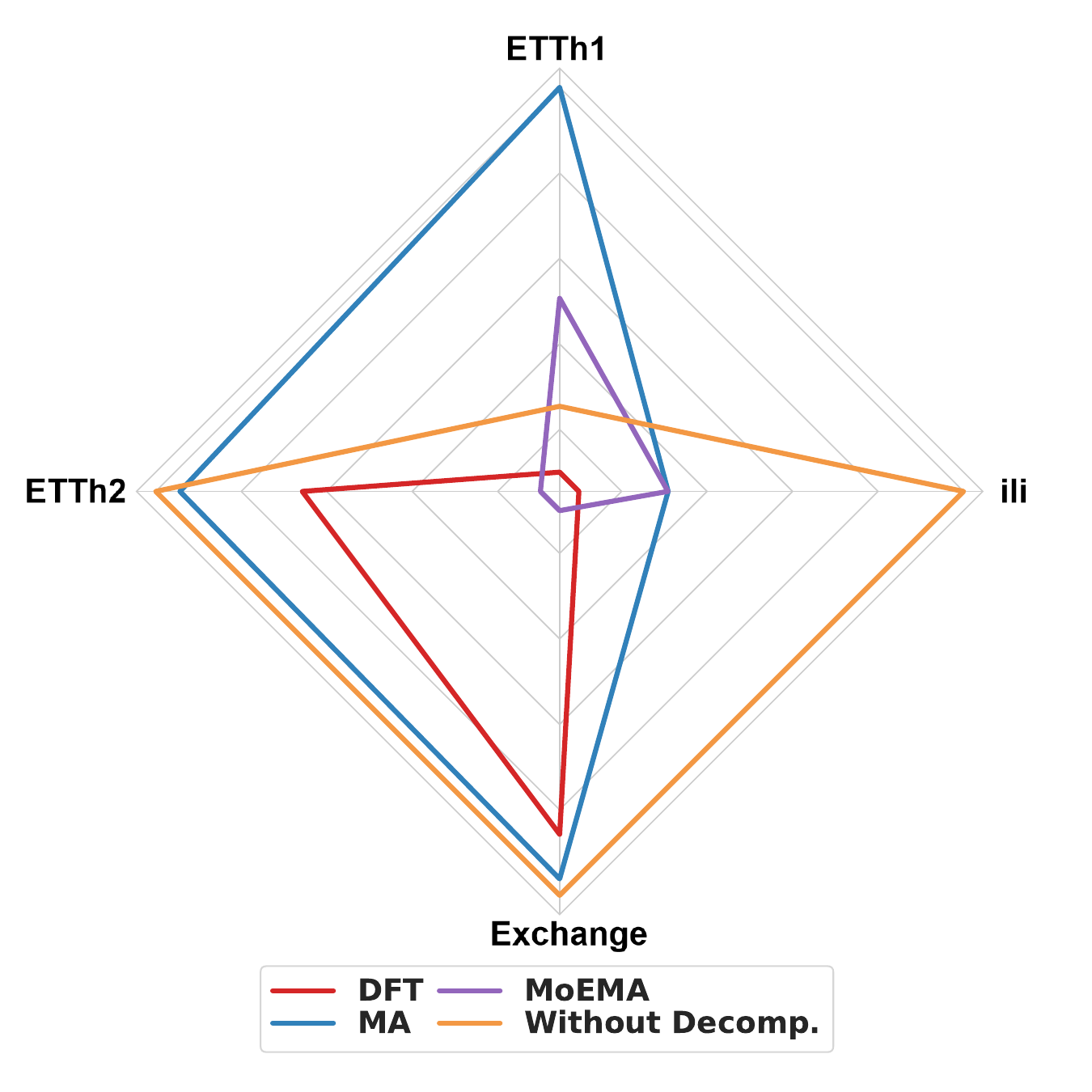}
         \caption{Series Decomposition}
         \label{fig:exp-llmdecomp}
     \end{subfigure}
     \caption{Overall performance of different design choices across 3 design dimensions (Network Backbone \ref{fig:exp-llmbackbone}, Series Normalization \ref{fig:exp-Normalization}, Series Decomposition \ref{fig:exp-Decomposition}) when using LLMs or TSFMs. The results (MSE) are based on the top 25th percentile across all forecasting horizons.}
     \label{fig:exp-largemodels}
\end{figure}

\textbf{\textit{Large Time Series Models}}. 
We choose three design dimensions crucial to large time-series models for discussion, as is shown in Fig.\ref{fig:exp-largemodels}. The most surprising finding pertains to the model backbones. We observe that GPT4TS demonstrates stable and competitive performance across most datasets. In contrast, Time-LLM, which is also based on LLMs, exhibits opposite results. A highly plausible explanation is that time series embeddings processed through different methods struggle to align consistently with the embedding space of word representations, which is an important part in Time-LLM, resulting in suboptimal performance of Time-LLM under diverse experimental configurations.

\subsection{Automatic Component Construction via \system}
\label{exp:meta-predictor}
Extensive experimental results discussed above indicate that in deep time series modeling, most design choices are determined by data characteristics, meaning one-size-fits-all approaches are seldom effective. This, in turn, emphasizes the necessity of automated model construction.

In this subsection, we compare the MTSF pipeline selected by TSGym with existing SOTA methods. Through large-scale experiments, we found that TSGym outperforms existing SOTA models in both long- and short-term MTSF tasks. Regarding algorithm efficiency, our experiments demonstrate that even when limited to a search pool of lightweight model structures, such as MLP and RNNs, TSGym can still achieve competitive results. We analyze the effectiveness of the pipelines automatically constructed by TSGym through five key questions as follows. Additional details, such as the results based on more metrics and more complex meta-features, can be found in the Appx.\ref{appx:complete_results}.

\textbf{Question 1: Is the model constructed by meta-predictor better than existing SOTA methods?}
Comprehensive forecasting results are presented in Table \ref{tab:tsgym_vs_sota_short} and \ref{tab:tsgym_vs_sota}, where the best performances are highlighted in \textcolor{red}{\textbf{red}} and the second-best results are \underline{\textcolor{blue}{underlined}}. Compared with other state-of-the-art forecasters, TSGym demonstrates superior capability, especially in handling high-dimensional time series data. Its consistent top-ranking performance across multiple datasets underlines its robustness and effectiveness for complex multivariate forecasting tasks. 

\begin{table}[h!]
\small
  \centering
  \vspace{-0.2in}
  \caption{\update{Short-term forecasting task on M4. The results are averaged from several datasets under different sample intervals. See Table in Appendix for the full results.}}
    \resizebox{1.0\textwidth}{!}{
    \begin{tabular}{c|c|c|c|c|c|c|c|c|c}
    \toprule
    \textbf{Models} & \textbf{TSGym (ours)} & \textbf{TimeMixer} & \textbf{MICN} & \textbf{TimesNet} & \textbf{PatchTST} & \textbf{DLinear} & \textbf{Crossformer} & \textbf{Autoformer} & \textbf{SegRNN} \\

\midrule
OWA & \textcolor{red}{\textbf{0.872}} & \underline{\textcolor{blue}{0.884}} & 0.984 & 0.907 & 0.965 & 0.922 & 8.856 & 1.273 & 1.007 \\
SMAPE & \underline{\textcolor{blue}{12.013}} & \textcolor{red}{\textbf{11.985}} & 13.025  & 12.199& 12.848 & 12.511 & >30 & 16.392 & 13.509 \\
MASE & \textcolor{red}{\textbf{1.575}} & \underline{\textcolor{blue}{1.615}} & 1.839 & 1.662 & 1.738 & 1.693 & >10 & 2.317 & 1.823 \\
    \bottomrule
    \end{tabular}}%
  \label{tab:tsgym_vs_sota_short}%
\end{table}%
\begin{table}[h!]
  \centering
  \caption{\update{Long-term forecasting task.} The past sequence length is set as 36 for ILI and 96 for the others. All the results are averaged from 4 different prediction lengths, that is $\{24,36,48,60\}$ for ILI and $\{96,192,336,720\}$ for the others. See Table in Appendix for the full results.}
    \resizebox{1\textwidth}{!}{
    \begin{tabular}{c|r@{\hspace{3pt}}r|r@{\hspace{3pt}}r|r@{\hspace{3pt}}r|r@{\hspace{3pt}}r|r@{\hspace{3pt}}r|r@{\hspace{3pt}}r|r@{\hspace{3pt}}r|r@{\hspace{3pt}}r|r@{\hspace{3pt}}r|r@{\hspace{3pt}}r}
    \toprule
    
    \multicolumn{1}{c}{\multirow{2}{*}{Models}} & 
    \multicolumn{2}{c}{\rotatebox{0}{\scalebox{0.76}{\textbf{TSGym}}}} &
    \multicolumn{2}{c}{\rotatebox{0}{\scalebox{0.76}{DUET}}} &
    \multicolumn{2}{c}{\rotatebox{0}{\scalebox{0.76}{TimeMixer}}}  &
    \multicolumn{2}{c}{\rotatebox{0}{\scalebox{0.76}{MICN}}} & 
    \multicolumn{2}{c}{\rotatebox{0}{\scalebox{0.76}{TimesNet}}} &
    \multicolumn{2}{c}{\rotatebox{0}{\scalebox{0.76}{PatchTST}}}  &
    \multicolumn{2}{c}{\rotatebox{0}{\scalebox{0.76}{DLinear}}}&
    \multicolumn{2}{c}{\rotatebox{0}{\scalebox{0.76}{Crossformer}}} & 
    \multicolumn{2}{c}{\rotatebox{0}{\scalebox{0.76}{Autoformer}}} &
    \multicolumn{2}{c}{\rotatebox{0}{\scalebox{0.76}{SegRNN}}} \\
    \multicolumn{1}{c}{} & \multicolumn{2}{c}{\scalebox{0.76}{(\textbf{Ours})}} & 
    \multicolumn{2}{c}{\scalebox{0.76}{\citep{qiu2025DUET}}} &
    \multicolumn{2}{c}{\scalebox{0.76}{\citep{wang2024timemixer}}}  &
    \multicolumn{2}{c}{\scalebox{0.76}{\citep{wang2023micn}}} & 
    \multicolumn{2}{c}{\scalebox{0.76}{\citep{wu2023timesnet}}}  & 
    \multicolumn{2}{c}{\scalebox{0.76}{\citep{nie2023PatchTST}}}  & 
    \multicolumn{2}{c}{\scalebox{0.76}{\citep{zeng2023dlinear}}}  & 
    \multicolumn{2}{c}{\scalebox{0.76}{\citep{zhang2023crossformer}}} & 
    \multicolumn{2}{c}{\scalebox{0.76}{\citep{lin2023segrnn}}}  &  
    \multicolumn{2}{c}{\scalebox{0.76}{\citep{wu2021autoformer}}} \\
    \cmidrule(lr){2-3}\cmidrule(lr){4-5}\cmidrule(lr){6-7}\cmidrule(lr){8-9}\cmidrule(lr){10-11}\cmidrule(lr){12-13}\cmidrule(lr){14-15}\cmidrule(lr){16-17}\cmidrule(lr){18-19}\cmidrule(lr){20-21}
    \multicolumn{1}{c}{Metric} & \scalebox{0.76}{MSE} & \scalebox{0.76}{MAE} & \scalebox{0.76}{MSE} & \scalebox{0.76}{MAE} & \scalebox{0.76}{MSE} & \scalebox{0.76}{MAE} & \scalebox{0.76}{MSE} & \scalebox{0.76}{MAE} & \scalebox{0.76}{MSE} & \scalebox{0.76}{MAE} & \scalebox{0.76}{MSE} & \scalebox{0.76}{MAE} & \scalebox{0.76}{MSE} & \scalebox{0.76}{MAE} & \scalebox{0.76}{MSE} & \scalebox{0.76}{MAE} & \scalebox{0.76}{MSE} & \scalebox{0.76}{MAE} & \scalebox{0.76}{MSE} & \scalebox{0.76}{MAE} \\

\midrule
\textbf{ETTm1} & \textcolor{red}{\textbf{0.357}} & \textcolor{red}{\textbf{0.383}} & 0.407 & 0.409 & \underline{\textcolor{blue}{0.384}} & \underline{\textcolor{blue}{0.399}} & 0.402 & 0.429 & 0.432 & 0.430 & 0.390 & 0.404 & 0.404 & 0.407 & 0.501 & 0.501 & 0.532 & 0.496 & 0.388 & 0.404 \\
\textbf{ETTm2} & \textcolor{red}{\textbf{0.261}} & \textcolor{red}{\textbf{0.319}} & 0.296 & 0.338 & 0.277 & 0.325 & 0.342 & 0.391 & 0.296 & 0.334 & 0.288 & 0.334 & 0.349 & 0.399 & 1.487 & 0.789 & 0.330 & 0.368 & \underline{\textcolor{blue}{0.273}} & \underline{\textcolor{blue}{0.322}} \\
\textbf{ETTh1} & \underline{\textcolor{blue}{0.426}} & 0.440 & 0.433 & \underline{\textcolor{blue}{0.437}} & 0.448 & 0.438 & 0.589 & 0.537 & 0.474 & 0.464 & 0.454 & 0.449 & 0.465 & 0.461 & 0.544 & 0.520 & 0.492 & 0.485 & \textcolor{red}{\textbf{0.422}} & \textcolor{red}{\textbf{0.429}} \\
\textbf{ETTh2} & \textcolor{red}{\textbf{0.358}} & \textcolor{red}{\textbf{0.400}} & 0.380 & \underline{\textcolor{blue}{0.403}} & 0.383 & 0.406 & 0.585 & 0.530 & 0.415 & 0.424 & 0.385 & 0.409 & 0.566 & 0.520 & 1.552 & 0.908 & 0.446 & 0.460 & \underline{\textcolor{blue}{0.374}} & 0.405 \\
\textbf{ECL} & \textcolor{red}{\textbf{0.170}} & \underline{\textcolor{blue}{0.265}} & \underline{\textcolor{blue}{0.179}} & \textcolor{red}{\textbf{0.262}} & 0.185 & 0.273 & 0.186 & 0.297 & 0.219 & 0.314 & 0.209 & 0.298 & 0.225 & 0.319 & 0.193 & 0.289 & 0.234 & 0.340 & 0.216 & 0.302 \\
\textbf{Traffic} & \textcolor{red}{\textbf{0.435}} & \textcolor{red}{\textbf{0.313}} & 0.797 & 0.427 & \underline{\textcolor{blue}{0.496}} & \underline{\textcolor{blue}{0.313}} & 0.544 & 0.320 & 0.645 & 0.348 & 0.497 & 0.321 & 0.673 & 0.419 & 1.458 & 0.782 & 0.637 & 0.397 & 0.807 & 0.411 \\
\textbf{Weather} & \textcolor{red}{\textbf{0.229}} & \textcolor{red}{\textbf{0.268}} & 0.252 & 0.277 & \underline{\textcolor{blue}{0.244}} & \underline{\textcolor{blue}{0.274}} & 0.264 & 0.316 & 0.261 & 0.287 & 0.256 & 0.279 & 0.265 & 0.317 & 0.253 & 0.312 & 0.339 & 0.379 & 0.251 & 0.298 \\
\textbf{Exchange} & 0.410 & 0.431 & \textcolor{red}{\textbf{0.322}} & \textcolor{red}{\textbf{0.384}} & 0.359 & \underline{\textcolor{blue}{0.402}} & \underline{\textcolor{blue}{0.346}} & 0.422 & 0.405 & 0.437 & 0.381 & 0.412 & 0.346 & 0.414 & 0.904 & 0.695 & 0.506 & 0.500 & 0.408 & 0.423 \\
\textbf{ILI} & 2.233 & 1.015 & 2.640 & 1.018 & 4.502 & 1.557 & 2.938 & 1.178 & \textcolor{red}{\textbf{2.140}} & \underline{\textcolor{blue}{0.907}} & \underline{\textcolor{blue}{2.160}} & \textcolor{red}{\textbf{0.901}} & 4.367 & 1.540 & 4.311 & 1.396 & 3.156 & 1.207 & 4.305 & 1.397 \\

\midrule
\textbf{{$1^{\text{st}}$ Count}} & 
\multicolumn{2}{c}{\boldres{11}} & 
\multicolumn{2}{c}{\underline{\textcolor{blue}{3}}} &
\multicolumn{2}{c}{0} & 
\multicolumn{2}{c}{0}  & 
\multicolumn{2}{c}{1}  & 
\multicolumn{2}{c}{1}  & 
\multicolumn{2}{c}{0}  & 
\multicolumn{2}{c}{0}  & 
\multicolumn{2}{c}{0}  & 
\multicolumn{2}{c}{2} 
\\
\bottomrule
\end{tabular}}%
  \label{tab:tsgym_vs_sota}%
\end{table}%

Specifically, TSGym achieves the lowest MSE and MAE on a total of 11 occasions, reflecting its strong generalization ability over both medium and long forecasting horizons. While some baseline models, like DUET, PatchTST and SegRNN, occasionally show competitive results on certain datasets. As for short-term forecasting tasks, both TSGym and \rv{TimeMixer} demonstrate competitive performance, with TSGym outperforming on most evaluation metrics.

\textbf{Question 2: Is TSGym with lightweight architecture better than existing SOTA methods?}

In the previous section, we compared TSGym using the full component pool with SOTA and found that TSGym outperforms SOTA on 66.7\% of the datasets evaluated. In this ablation experiment Table \ref{tab:tsgym_ablation}, we specifically compare the \texttt{-Transformer} configuration of TSGym with DUET. 
Remarkably, even after removing Transformer-related components from the TSGym component pool and retaining only the more computationally efficient MLP- and RNN-based models, TSGym still outperforms DUET on the majority of datasets. This demonstrates the robustness and efficiency of TSGym’s architecture and highlights the strong predictive power of the simplified MLP-based design.

\textbf{Question 3: Does the training strategies bring significant improvement for TSGym?}

Following Table~\ref{tab:tsgym_ablation}, we find that the introduction of the resampling method enhances TSGym's meta-predictor performance across 2 datasets, improving both robustness and accuracy. The \texttt{+AllPL} configuration, which trains on datasets with varying prediction lengths and transfers this knowledge to a test set with a single prediction length, further improves generalization, with the best performance observed on the ETTh1 dataset. Additionally, removing the Transformer component (\texttt{-Transformer}) leads to performance gains on certain datasets, suggesting that a simplified MLP- or RNN-based architecture can be more effective in specific scenarios. These results highlight the flexibility of TSGym’s design and the potential benefits of customizing the component pool to suit dataset characteristics.

\begin{table}[h!]
  \centering
  \begin{minipage}{0.25\textwidth}
  \vspace{-0.1in}
  \caption{Ablation study evaluates the removal of Transformer-based components and different training strategies, with results averaged over all prediction lengths, and the final row shows how often TSGym variants outperform DUET.}
  \label{tab:tsgym_ablation}
  \end{minipage}%
  \hspace*{0.1cm}
  \begin{minipage}{0.92\textwidth}
  \resizebox{0.8\textwidth}{!}{
    \begin{tabular}{c|rr|rr|rr|rr|rr}
    \toprule
          \multicolumn{1}{c}{Models} & \multicolumn{2}{c|}{\textbf{TSGym}} & \multicolumn{2}{c|}{\textbf{-Transformer}} & \multicolumn{2}{c|}{\textbf{+Resample}} & \multicolumn{2}{c|}{\textbf{+AllPL}} & \multicolumn{2}{c}{\textbf{DUET}}\\
    \midrule
          \multicolumn{1}{c}{Metric}& \multicolumn{1}{c}{\textbf{MSE}} & \multicolumn{1}{c|}{\textbf{MAE}} & \multicolumn{1}{c}{\textbf{MSE}} & \multicolumn{1}{c|}{\textbf{MAE}} & \multicolumn{1}{c}{\textbf{MSE}} & \multicolumn{1}{c|}{\textbf{MAE}} & \multicolumn{1}{c}{\textbf{MSE}} & \multicolumn{1}{c|}{\textbf{MAE}} & \multicolumn{1}{c}{\textbf{MSE}} & \multicolumn{1}{c}{\textbf{MAE}}\\

\midrule
\textbf{ETTm1} & \textcolor{red}{\textbf{0.357}} & \textcolor{red}{\textbf{0.383}} & 0.363 & 0.388 & \underline{\textcolor{blue}{0.361}} & \underline{\textcolor{blue}{0.384}} & 0.367 & 0.391 & 0.407 & 0.409 \\
\textbf{ETTm2} & \underline{\textcolor{blue}{0.261}} & \textcolor{red}{\textbf{0.319}} & 0.274 & 0.330 & \textcolor{red}{\textbf{0.260}} & \underline{\textcolor{blue}{0.320}} & 0.265 & 0.320 & 0.296 & 0.338 \\
\textbf{ETTh1} & \underline{\textcolor{blue}{0.426}} & 0.440 & 0.441 & 0.450 & 0.432 & 0.440 & \textcolor{red}{\textbf{0.424}} & \textcolor{red}{\textbf{0.434}} & 0.433 & \underline{\textcolor{blue}{0.437}} \\
\textbf{ETTh2} & \underline{\textcolor{blue}{0.358}} & \underline{\textcolor{blue}{0.400}} & 0.358 & 0.401 & \textcolor{red}{\textbf{0.357}} & \textcolor{red}{\textbf{0.398}} & 0.361 & 0.404 & 0.380 & 0.403 \\
\textbf{ECL} & \textcolor{red}{\textbf{0.170}} & \underline{\textcolor{blue}{0.265}} & 0.174 & 0.273 & \underline{\textcolor{blue}{0.170}} & 0.265 & 0.185 & 0.277 & 0.179 & \textcolor{red}{\textbf{0.262}} \\
\textbf{Traffic} & 0.435 & 0.313 & \textcolor{red}{\textbf{0.415}} & \textcolor{red}{\textbf{0.293}} & \underline{\textcolor{blue}{0.429}} & 0.307 & 0.429 & \underline{\textcolor{blue}{0.306}} & 0.797 & 0.427 \\
\textbf{Weather} & \underline{\textcolor{blue}{0.229}} & 0.268 & \textcolor{red}{\textbf{0.226}} & \textcolor{red}{\textbf{0.265}} & 0.229 & \underline{\textcolor{blue}{0.267}} & 0.234 & 0.271 & 0.252 & 0.277 \\
\textbf{Exchange} & 0.410 & 0.431 & 0.409 & 0.435 & 0.399 & 0.430 & \underline{\textcolor{blue}{0.377}} & \underline{\textcolor{blue}{0.411}} & \textcolor{red}{\textbf{0.322}} & \textcolor{red}{\textbf{0.384}} \\
\textbf{ILI} & \underline{\textcolor{blue}{2.233}} & \textcolor{red}{\textbf{1.015}} & 2.437 & 1.053 & 2.698 & 1.114 & \textcolor{red}{\textbf{2.173}} & 1.020 & 2.640 & \underline{\textcolor{blue}{1.018}} \\
\midrule
\textbf{Better Count} & \multicolumn{2}{c|}{\textbf{14/18}} & \multicolumn{2}{c|}{\textbf{12/18}} & \multicolumn{2}{c|}{\textbf{12/18}}  & \multicolumn{2}{c|}{\textbf{12/18}} \\
    \bottomrule
    \end{tabular}}%
    \end{minipage}%
  
\end{table}%

\textbf{Question 4: Does large time-series models bring significant improvement for TSGym?}
\begin{wraptable}{r}{0.5\textwidth}
  \centering
  \vspace{-0.15in}
  \caption{Ablation study of TSGym incorporating LLM and TSFM in 4 datasets. The average results of all prediction lengths are listed here.}
  \resizebox{0.5\textwidth}{!}{
    \begin{tabular}{c|rr|rr|rr}
    \toprule
          \multicolumn{1}{c}{Models} & \multicolumn{2}{c|}{\textbf{TSGym}} & \multicolumn{2}{c|}{\textbf{+LLM}} & \multicolumn{2}{c}{\textbf{+TSFM}} \\
    \midrule
          \multicolumn{1}{c}{Metric}& \multicolumn{1}{c}{\textbf{MSE}} & \multicolumn{1}{c|}{\textbf{MAE}} & \multicolumn{1}{c}{\textbf{MSE}} & \multicolumn{1}{c|}{\textbf{MAE}} & \multicolumn{1}{c}{\textbf{MSE}} & \multicolumn{1}{c}{\textbf{MAE}} \\
    \midrule

\textbf{ETTh1} & \underline{\textcolor{blue}{0.442}} & 0.440 & 0.442 & \underline{\textcolor{blue}{0.438}} & \textcolor{red}{\textbf{0.432}} & \textcolor{red}{\textbf{0.437}} \\
\textbf{ETTh2} & 0.411 & 0.429 & \textcolor{red}{\textbf{0.364}} & \textcolor{red}{\textbf{0.402}} & \underline{\textcolor{blue}{0.376}} & \underline{\textcolor{blue}{0.409}} \\
\textbf{Exchange} & 0.673 & 0.522 & \underline{\textcolor{blue}{0.503}} & \underline{\textcolor{blue}{0.466}} & \textcolor{red}{\textbf{0.374}} & \textcolor{red}{\textbf{0.411}} \\
\textbf{ILI} & \underline{\textcolor{blue}{2.682}} & \underline{\textcolor{blue}{1.112}} & \textcolor{red}{\textbf{2.470}} & \textcolor{red}{\textbf{1.063}} & 2.860 & 1.163 \\

    \bottomrule
    \end{tabular}}%
    \vspace{-0.15in}
  \label{tab:ablationllm}%
\end{wraptable}

Table \ref{tab:ablationllm} evaluates the impact of incorporating LLM and TSFM into the base TSGym framework. It is evident that both the introduction of LLM and TSFM consistently improve forecasting accuracy compared to the baseline TSGym configuration. This demonstrates the effectiveness of leveraging advanced model architectures and fusion strategies to further enhance the predictive performance of TSGym, particularly on complex multivariate time series datasets.

\rv{\textbf{Question 5: Does smarter sampling strategy bring improvement for TSGym?}}
\begin{table}[htbp]
  \centering
  \caption{\rv{TSGym Performance Comparison Across Sampling Strategies and DUET. The average results of all prediction lengths are listed here.}}
    \begin{tabular}{ccc|cc|cc}
    \toprule
    \textbf{Models} & \multicolumn{2}{c|}{\textbf{TSGym (Optuna)}} & \multicolumn{2}{c|}{\textbf{TSGym}} & \multicolumn{2}{c}{\textbf{DUET}} \\
    \midrule
    \textbf{Metric} & \textbf{MSE} & \textbf{MAE} & \textbf{MSE} & \textbf{MAE} & \textbf{MSE} & \textbf{MAE} \\
    \midrule
    \textbf{ETTm1} & \textcolor{red}{\textbf{0.350}} & \textcolor{red}{\textbf{0.373}} & 0.357 & 0.383 & 0.407 & 0.409 \\
    \textbf{ETTm2} & \textcolor{red}{\textbf{0.252}} & \textcolor{red}{\textbf{0.310}} & \underline{\textcolor{blue}{0.261}} & \underline{\textcolor{blue}{0.319}} & 0.296 & 0.338 \\
    \textbf{ETTh1} & \underline{\textcolor{blue}{0.429}} & \underline{\textcolor{blue}{0.440}} & \textcolor{red}{\textbf{0.426}} & \underline{\textcolor{blue}{0.440}} & 0.433 & \textcolor{red}{\textbf{0.437}} \\
    \textbf{ETTh2} & \textcolor{red}{\textbf{0.354}} & \textcolor{red}{\textbf{0.392}} & \underline{\textcolor{blue}{0.358}} & \underline{\textcolor{blue}{0.400}} & 0.380 & 0.403 \\
    \textbf{ECL} & \textcolor{red}{\textbf{0.166}} & \textcolor{red}{\textbf{0.262}} & \underline{\textcolor{blue}{0.170}} & \underline{\textcolor{blue}{0.265}} & 0.179 & \textcolor{red}{\textbf{0.262}} \\
    \textbf{Traffic} & \underline{\textcolor{blue}{0.439}} & \textcolor{red}{\textbf{0.295}} & \textcolor{red}{\textbf{0.435}} & \underline{\textcolor{blue}{0.313}} & 0.797 & 0.427 \\
    \textbf{Weather} & \textcolor{red}{\textbf{0.229}} & \textcolor{red}{\textbf{0.258}} & \textcolor{red}{\textbf{0.229}} & \underline{\textcolor{blue}{0.268}} & \underline{\textcolor{blue}{0.252}} & 0.277 \\
    \textbf{Exchange} & 0.457 & 0.446 & \underline{\textcolor{blue}{0.410}} & \underline{\textcolor{blue}{0.431}} & \textcolor{red}{\textbf{0.322}} & \textcolor{red}{\textbf{0.384}} \\
    \textbf{ILI} & 3.194 & 1.234 & \textcolor{red}{\textbf{2.233}} & \textcolor{red}{\textbf{1.015}} & \underline{\textcolor{blue}{2.640}} & \underline{\textcolor{blue}{1.018}} \\
    \midrule
    \textbf{1st Count} & \multicolumn{2}{c|}{\textcolor{red}{\textbf{11}}} & \multicolumn{2}{c|}{\underline{\textcolor{blue}{5}}} & \multicolumn{2}{c}{4} \\
    \bottomrule
    \end{tabular}
  \label{tab:tsgym_optuna}%
\end{table}%

\rv{To further improve efficiency and reduce meta-training costs, we adopted Optuna, a smarter sampling strategy based on Bayesian optimization. Starting with 50 random trials as a cold start, Optuna then guided 50 additional trials using prior search knowledge, significantly enhancing search efficiency and identifying better model configurations with fewer evaluations. See Appx.\ref{appx:optuna} for details. Table~\ref{tab:tsgym_optuna} shows that the meta-learner trained on Optuna-selected samples achieves comparable or better performance than the one trained on random search. The only exception is the ILI dataset, where Optuna used a much smaller trial budget; performance is expected to improve with more trials.}




\section{Conclusions, Limitations, and Future Directions}
\label{sec: conclu}
To advance beyond holistic evaluations in multivariate time-series forecasting (MTSF), this paper introduced \system, a novel framework centered on fine-grained component analysis and the automated construction of specialized forecasting models. By systematically decomposing MTSF pipelines into design dimensions and choices informed by recent studies, \system uncovers crucial insights into component-level forecasting performance and leverages meta-learning method for the automated construction of customized models. Extensive experimental results indicate that the MTSF models constructed by the 
proposed \system significantly outperform current MTSF SOTA solutions—demonstrating the advantage of adaptively customizing models according to distinct data characteristics. Our results show that \system is highly effective, even without exhaustively covering all SOTA components, and \system is made publicly available to benefit the MTSF community.

Future efforts will focus on expanding TSGym's range of forecasting techniques with emerging techniques and refining its meta-learning capabilities by incorporating multi-objective optimization to balance predictive performance against computational costs, especially for large time-series models, while also broadening its applicability across diverse time series analysis tasks.

\clearpage
\newpage

\bibliographystyle{abbrv}
\small
\bibliography{ref}


\newpage

\clearpage
\appendix
\setcounter{table}{0}
\setcounter{figure}{0}

\renewcommand{\thetable}{\Alph{section}\arabic{table}}
\renewcommand{\thefigure}{\Alph{section}\arabic{figure}}

\section*{Appendix}
For further details, we provide more information in the Appendix, including the evaluated $10$ datasets (\S \ref{appx:data}), key modules (\S \ref{appx:modules}), compared baselines (\S \ref{appx:baseline}), metrics mathematical formula (\S \ref{appx:metrics}), system configuration (\S \ref{appx:system_config}), ADGym comparison analysis (\S \ref{appx:tsgym_vs_adgym}), the details of proposed \system (\S \ref{appx:adgym_details}), and additional experimental results (\S \ref{appx:complete_results}). 

\section{Dataset List}\label{appx:data}
We conduct extensive evaluations on nine standard long-term forecasting benchmarks - four ETT variants (ETTh1, ETTh2, ETTm1, ETTm2), Electricity (abbreviated as ECL), Traffic, Weather, Exchange, and ILI, complemented by the M4 dataset for short-term forecasting tasks, with complete dataset specifications provided in Table~\ref{tab:datasets}.

\begin{table}[!h]
\centering
\footnotesize
	\caption{Data description of the 10 datasets included in \system.} 
\footnotesize
\resizebox{1\textwidth}{!}{
    \begin{tabular}{@{}c l l l r r l @{}}
        \toprule
        Task & Dataset      & Domain      & Frequency & Lengths & Dim  & Description\\ \midrule

        \multirow{9}{*}{LTF} & ETTh1        & Electricity & 1 hour     & 14,400      & 7         & Power transformer 1, comprising seven indicators such as oil temperature and useful load\\
                            & ETTh2        & Electricity & 1 hour    & 14,400      & 7         & Power transformer 2, comprising seven indicators such as oil temperature and useful load\\
                            & ETTm1        & Electricity & 15 mins   & 57,600      & 7         & Power transformer 1, comprising seven indicators such as oil temperature and useful load\\
                            & ETTm2        & Electricity & 15 mins   & 57,600      & 7         & Power transformer 2, comprising seven indicators such as oil temperature and useful load\\
                            & ECL  & Electricity & 1 hour    & 26,304      & 321       & Electricity records the electricity consumption in kWh every 1 hour from 2012 to 2014\\
                            & Traffic      & Traffic     & 1 hour    & 17,544      & 862       & Road occupancy rates measured by 862 sensors on San Francisco Bay area freeways\\
                            & Weather      & Environment & 10 mins   & 52,696      & 21        & Recorded
                            every for the whole year 2020, which contains 21 meteorological indicators\\
                            & Exchange & Economic    & 1 day      & 7,588       & 8         & ExchangeRate collects the daily exchange rates of eight countries\\
                            & ILI          & Health      & 1 week     & 966        & 7         & Recorded indicators of patients data from Centers for Disease Control and Prevention\\

        \midrule
        \multirow{6}{*}{STF} & \multirow{6}{*}{M4}    & \multirow{6}{2cm}{Demographic, Finance, Industry, Macro, Micro and Other} & Yearly   & \multirow{6}{*}{19-9933} & \multirow{6}{*}{100000}    & \multirow{6}{10cm}{M4 competition dataset containing 100,000 unaligned time series with varying lengths and time periods} \\
                            &  &  & Quarterly    &   &  & \\
                            &  &  & Monthly    &   &  & \\
                            &  &  & Weakly    &   &   & \\
                            &  &  & Daily    &   &   & \\
                            &  &  & Hourly    &   &   & \\             
        \bottomrule
        
    \end{tabular}
    
}
\label{tab:datasets} 
\end{table}

 
\section{Key Modules}\label{appx:modules} 
Modern deep learning for MTSF utilizes several specialized modules to tackle non-stationarity, multi-scale dependencies, and inter-variable interactions. In this section, we analyze the design and efficacy of prevalent specialized modules adopted in state-of-the-art models (Fig.~\ref{fig:pipeline}).

\textbf{Normalization modules} address temporal distribution shifts through adaptive statistical alignment. While z-score normalization employs fixed moments, modern techniques enhance adaptability: RevIN \cite{kim2021RevIN} introduces learnable affine transforms with reversible normalization/denormalization; Dish-TS \cite{fan2023dish} decouples inter-/intra-series distribution coefficients; Non-Stationary Transformer \cite{liu2022nonstationary} integrates statistical moments into attention via de-stationary mechanisms. These methods balance stationarized modeling with inherent non-stationary dynamics.

\textbf{Decomposition methods}, standard in time series analysis, break down series into components like trend and seasonality to improve predictability and handle distribution shifts.
\textbf{(1) Time-domain decomposition} utilizes moving average operations to isolate slowly-varying trends from high-frequency fluctuations that represent seasonality (e.g., DLinear \cite{zeng2023dlinear}, Autoformer, FEDformer). 
\textbf{(2) Frequency-domain decomposition} partitions series via Discrete Fourier Transform (DFT), assigning low-frequency spectra to trends and high-frequency bands to seasonality, which is applied in the Koopa \cite{liu2023koopa} model.

\textbf{Multi-Scale modeling} addresses the inherent temporal hierarchy in time series data, where patterns manifest differently across various granularities (e.g., minute-level fluctuations vs. daily trends). 
Pyraformer \cite{liu2022pyraformer} integrates multi-convolution kernels via pyramidal attention to establish hierarchical temporal dependencies. FEDformer \cite{zhou2022fedformer} employs mixed experts to combine trend components from multiple pooling kernels with varying receptive fields, where larger kernels capture macro patterns while smaller ones preserve local details. TimeMixer \cite{wang2024timemixer} extends this paradigm through bidirectional mixing operations - upward propagation refines fine-scale seasonal features while downward aggregation consolidates coarse-scale trends. FiLM 
 \cite{zhou2022film} dynamically adjusts temporal resolutions through learnable lookback windows, enabling adaptive focus on relevant historical contexts across scales. Crossformer \cite{zhang2023crossformer} implements flexible patchsize configurations, where multi-granular patches independently model short-term fluctuations and long-term cycles through dimension-aware processing. 

\textbf{Temporal Tokenization strategies}, originating from Transformers \cite{wang2024deep, liu2024itransformer} and now extended to RNNs \cite{lin2023segrnn}, vary by temporal representation granularity:
\textbf{(1) Point-wise} methods (e.g., Informer \cite{zhou2021informer}, Pyraformer \cite{liu2022pyraformer}) process individual timestamps as tokens. They offer temporal precision but face quadratic complexity, requiring attention sparsification that may hinder long-range dependency capture.
\textbf{(2) Patch-wise} strategies (e.g., PatchTST \cite{nie2023PatchTST}) aggregate local temporal segments into patches. Pathformer \cite{chen2024pathformer} similarly employs patch-based processing via adaptive multi-scale pathways.
\textbf{(3) Series-wise} approaches (e.g., iTransformer \cite{liu2024itransformer}) construct global variate representations, enabling cross-variate modeling but risking temporal misalignment. TimeXer \cite{wang2024timexer} uses hybrid tokenization: patch-level for endogenous variables and series-level for exogenous, bridged by a learnable global token.

\textbf{Temporal Dependency Modeling} captures dynamic inter-step dependencies through diverse architectural mechanisms, balancing local interactions and global patterns. Recurrent state transitions (e.g., LSTM) model sequential memory via gated memory cells; temporal convolutions (e.g., TCN \cite{bai2018tcn}) construct multi-scale receptive fields using dilated kernels; attention mechanisms (e.g., Transformers) enable direct pairwise interactions across arbitrary time steps. Efficiency-driven innovations include sparse attention (Informer \cite{zhou2021informer}), periodicity-based aggregation (Autoformer \cite{wu2021autoformer}), and state-space hybrids (Mamba \cite{gu2024mamba}), achieving tractable long-range dependency modeling while preserving temporal fidelity.

\textbf{Variate Correlation}, fundamental to modeling critical correlations in multivariate time series forecasting (MTSF), operates through two primary paradigms \cite{qiu2025DUET}:
\textbf{(1) Channel-Independent (CI) Strategy}: Processes channels independently with shared parameters (e.g., PatchTST \cite{nie2023PatchTST}), ensuring robustness and efficiency but ignoring multivariate dependencies, limiting use with strong inter-channel interactions \cite{qiu2025comprehensive}.

\textbf{(2) Channel-Dependent (CD) Strategy}: Integrates channel information via methods like channel-wise self-attention (iTransformer \cite{liu2024itransformer}) or MLP-based mixing (TSMixer \cite{chen2023tsmixer}). This allows explicit dependency modeling but risks overfitting and struggles with noise in high dimensions.


\section{Compared Baselines}\label{appx:baseline}
We systematically compare state-of-the-art forecasting models using the \nmodules architectural modules introduced in Section \ref{appx:modules}. Table \ref{tab:Baseline} presents the configuration of each baseline in terms of these modules. The "Notes" column provides concise annotations of each model’s key methodological features, allowing for quick identification of the technical differentiators among the baselines.
\begin{table}[!h]
\centering
\footnotesize
\caption{Component Configurations of 27 Baseline Models} 
\label{tab:Baseline} 
\footnotesize
\makebox[\linewidth]{%
\resizebox{1\textwidth}{!}{
\renewcommand{\arraystretch}{1.1} 
\setlength{\tabcolsep}{2pt} 
    \begin{tabular}{@{}c l m{1cm}<{\centering} m{1.7cm}<{\centering} m{1cm}<{\centering} m{1.7cm}<{\centering} m{2cm}<{\centering} m{1cm}<{\centering} m{7cm}@{}}
        \toprule
        \textbf{Backbone} & \textbf{Method} & \textbf{Normali\-zation} & \textbf{Decom\-position} & \textbf{Multi\-Scale} & \textbf{Token\-izations}& \textbf{Temporal Dependency} & \textbf{Variate Correlation} & \multicolumn{1}{c}{\textbf{Notes}} \\ 
        \midrule 
\multirow{3}{*}{\centering \textbf{RNN}}         & SegRNN\cite{lin2023segrnn}        & SubLast       &               &            & Patch-wise             & GRU                                               & CI                  & Reduces iterations via patch-wise processing and parallel multi-step forecasting.                                                                                 \\
            & Mamba\cite{gu2024mamba}         & Stat          &               &            & Point-wise             & Selective State Space Model                       & CD                  & Efficient model selectively propagating information without attention or MLP blocks.              
            \\
            \midrule
            
\multirow{5}{*}{\centering \textbf{CNN}}         & SCINet\cite{liu2022scinet}        & Stat          &               & TRUE       & Point-wise             & Conv1d                                            & CD                  & Recursively downsamples, convolves, and interacts with data to capture complex temporal dynamics.                                                                 \\
            & MICN\cite{wang2023micn}          &               & MA            & TRUE       & Point-wise             & Conv1d                                            & CD                  & Combines local features and global correlations using multi-scale convolutions with linear complexity.                                                            \\
            & TimesNet\cite{wu2023timesnet}      & Stat          &               & TRUE       & Point-wise             & Conv2d                                            & CD                  & Transforms 1D time series into 2D tensors to capture multi-periodicity and temporal variations.                                                                   \\
            \midrule
            
\multirow{14}{*}{\centering \textbf{MLP}}         & FiLM\cite{zhou2022film}          & RevIN         &               & TRUE       & Point-wise             & Legendre Projection Unit                          & CD                  & Preserves historical info and reduces noise with Legendre and Fourier projections.                                            \\
            & LightTS\cite{zhang2022LightTS}       &               &               &            & Patch-wise             & MLP                                               & CD                  & Lightweight MLP model for multivariate forecasting, using continuous and interval sampling for efficiency.                                                        \\
            & DLinear\cite{zeng2023dlinear}       &               & MA            &            & Point-wise             & MLP                                               & CI/CD               & Decomposes series into trend and seasonal components, then applies linear layers for improved forecasting.                                                   \\
            & Koopa\cite{liu2023koopa}         & Stat          & DFT           &            & Point-wise, Patch-wise & MLP                                               & CD                  & Uses Koopman theory to model non-stationary dynamics, handling time-variant and time-invariant components.                                                        \\
            & TSMixer\cite{chen2023tsmixer}       &               &               &            & Point-wise             & MLP                                               & CD                  & Simple MLP model efficiently captures both time and feature dependencies for forecasting.                                                                         \\
            & FreTS\cite{yi2023frets}         &               &               &            & Point-wise             & Frequency-domain MLP                              & CI/CD               & Uses frequency-domain MLPs to capture global dependencies and focus on key frequency components.                                                                  \\
            & TiDE\cite{das2023TiDE}          & Stat          &               &            & Point-wise             & MLP                                               & CI                  & Fast MLP-based model for long-term forecasting, handling covariates and non-linear dependencies.                                                                  \\
            & TimeMixer\cite{wang2024timemixer}     & RevIN         & MA            & TRUE       & Point-wise             & MLP                                               & CI/CD               & Fully MLP-based model, disentangles and mixes multiscale temporal patterns.                                                   \\
            \midrule
            
\multirow{28}{*}{\centering \textbf{Transformer}}  & Reformer\cite{Kitaev2020Reformer}      &               &               &            & Point-wise             & LSHSelf\-Attention                                  & CD                  & Memory-efficient Transformer with locality-sensitive hashing for faster training on long sequences.                                                               \\
            & Informer\cite{zhou2021informer}      &               &               &            & Point-wise             & ProbSparse-Attention                              & CD                  & Efficient Transformer with ProbSparse-Attention and a generative decoder for faster long-sequence forecasting.                                                    \\
            & TFT\cite{lim2021tft}           & Stat          &               &            & Point-wise             & Self-Attention                                    & CD                  & High-performance, interpretable multi-horizon forecasting model combining recurrent layers for local processing and attention layers for long-term dependencies.  \\
            & Autoformer\cite{wu2021autoformer}    &               & MA            &            & Point-wise             & Auto-Correlation                                  & CD                  & Uses Auto-Correlation and decomposition for accurate long-term predictions.                                                                                       \\
            & PyraFormer\cite{liu2022pyraformer}    &               &               & TRUE       & Point-wise             & Pyramid-Attention                                 & CD                  & Captures temporal dependencies at multiple resolutions with constant signal path length.                                              \\
            & NS\-Transformer\cite{liu2022nonstationary} & Stat          &               &            & Point-wise             & De-stationary Attention                           & CD                  & Restores non-stationary information through de-stationary attention for improved forecasting.                                                                     \\
            & ETSformer\cite{woo2022etsformer}     &               & DFT           &            & Point-wise             & Exponential\-Smoothing\-Attention & CD                  & Integrates exponential smoothing and frequency attention for accuracy, efficiency, and interpretability.                                                 \\
            & FEDformer\cite{zhou2022fedformer}     &               & MA            & TRUE       & Point-wise             & AutoCorrelation                                   & CD                  & Combines seasonal-trend decomposition with frequency-enhanced Transformer for efficient forecasting.                                                              \\
            & Crossformer\cite{zhang2023crossformer}   &               &               & TRUE       & Patch-wise             & TwoStage\-Attention                                 & CD                  & Captures both temporal and cross-variable dependencies with two-stage attention.                                                                                  \\
            & PatchTST\cite{nie2023PatchTST}      & Stat          &               &            & Patch-wise             & FullAttention                                     & CI                  & Segments time series into patches and uses channel-independent embeddings.                                           \\
            & iTransformer\cite{liu2024itransformer}  & Stat          &               &            & Series-wise            & FullAttention                                     & CD                  & Redefines token embedding to treat time points as series-wise tokens for better multivariate modeling.                                                            \\
            & TimeXer\cite{wang2024timexer}       & Stat          &               &            & Series-wise            & FullAttention                                     & CD                  & Enhances forecasting by incorporating exogenous variables via patch-wise and variate-wise attention.                                                              \\
            & PAttn\cite{tan2024pattn}         & Stat          &               &            & Patch-wise             & FullAttention                                     & CI                  & Similar to PatchTST, uses attention-based patching for efficient forecasting without large language models.                                                       \\
            & DUET\cite{qiu2025DUET}           & RevIN         & MA            &            & Point-wise             & FullAttention                                     & CI/CD               & Enhances multivariate forecasting by using Mixture of Experts (MOE) for temporal clustering and a frequency-domain similarity mask matrix for channel clustering.  \\

        \bottomrule
    \end{tabular}
}
}	
\end{table}

\section{Metrics Mathematical Formula}\label{appx:metrics}
The metrics used in this paper can be calculated as follows\cite{wu2023timesnet}:
\begin{align*} \label{equ:metrics}
    \text{MSE} &= \frac{1}{H} \sum_{i=1}^H (\mathbf{X}_{i} - \widehat{\mathbf{X}}_{i})^2,
    &
    \text{MAE} &= \frac{1}{H} \sum_{i=1}^H |\mathbf{X}_{i} - \widehat{\mathbf{X}}_{i}|, \\
    \text{SMAPE} &= \frac{200}{H} \sum_{i=1}^H \frac{|\mathbf{X}_{i} - \widehat{\mathbf{X}}_{i}|}{|\mathbf{X}_{i}| + |\widehat{\mathbf{X}}_{i}|},
    &
    \text{MAPE} &= \frac{100}{H} \sum_{i=1}^H \frac{|\mathbf{X}_{i} - \widehat{\mathbf{X}}_{i}|}{|\mathbf{X}_{i}|}, \\
    \text{MASE} &= \frac{1}{H} \sum_{i=1}^H \frac{|\mathbf{X}_{i} - \widehat{\mathbf{X}}_{i}|}{\frac{1}{H-m}\sum_{j=m+1}^{H}|\mathbf{X}_j - \mathbf{X}_{j-m}|},
    &
    \text{OWA} &= \frac{1}{2} \left[ \frac{\text{SMAPE}}{\text{SMAPE}_{\textrm{Naïve2}}}  + \frac{\text{MASE}}{\text{MASE}_{\textrm{Naïve2}}}  \right],
\end{align*}
where $m$ is the periodicity of the data. $\mathbf{X},\widehat{\mathbf{X}}\in\mathbb{R}^{H\times C}$ are the ground truth and prediction results of the future with $H$ time points and $C$ dimensions. $\mathbf{X}_{i}$ means the $i$-th future time point.

\section{System Configuration}\label{appx:system_config}
We conducted all experiments in the same experimental environment, which includes four NVIDIA A100 GPUs with 80GB and eight 40GB of memory. We saved overall experimental time by running experiments in parallel.

\section{Compared with ADGym}\label{appx:tsgym_vs_adgym}

\begin{table}[!ht]
    \centering
    \caption{Compared with ADGym, TSGym covers a broader and more in-depth design space, as well as a more structured and extensive automated selection experiment.}
    \resizebox{1\textwidth}{!}{
    \begin{tabular}{lll}
    \hline
        ~ & ADGym & TSGym \\ \hline
        Design Dimensions & 13 & 16 \\ 
        Design Space Size & 195,9552 & 796,2624 \\ 
        Model Architectures & MLP,AE,ResNet,FTTransformer & MLP,RNN, Transformers, LLM, TSFM \\ 
        Max of Data Samples & 3000 & 57,600 \\ 
        Meta Feature Dimensions & 200 & 1404 \\ 
        Baseline Methods & 7 & 27 \\ \hline
    \end{tabular}
    }
    \label{tab:adgym_vs_tsgym}%
\end{table}
Compared with ADGym~\cite{jiang2023adgym}, TSGym exhibits the following differences and advantages:

\textbf{(1) Broader model structure design choices}. ADGym includes only MLP, autoencoder (AE), ResNet, and Transformer architectures, while TSGym provides an in-depth decoupling of different attention mechanisms within Transformers and incorporates two pre-trained large models: LLMs and TSFM.
\textbf{(2) More diverse data processing design choices}. ADGym focuses solely on data augmentation and two normalization methods, whereas TSGym encompasses series sampling, series normalization, series decomposition, as well as various series encoding options.
\textbf{(3) More complex meta-features}. The meta-features in ADGym include statistical metrics for tabular datasets, while TSGym considers multiple sequence characteristics across different channels in multivariate time series, such as distribution drift, sequence autocorrelation, and more.
\textbf{(4) More standardized automated selection experiments}. Due to time constraints, ADGym limits the sample size to fewer than 3000 samples, whereas TSGym imposes no such restriction, providing a larger-scale experimental design that leads to more solid experimental conclusions.

In summary, compared with ADGym, TSGym makes \textbf{significant progress and development in both components benchmarking and automated selection.}
More details can be seen in table \ref{tab:adgym_vs_tsgym}.

\section{Details of \system} \label{appx:adgym_details}
In this section, we introduce detailed descriptions of the design choices, extracted meta-features and the trained meta-predictors.

\subsection{More Details of Design Choices in TSGym.}
\label{appx:design_choices_details}

\begin{figure}[htbp]
  \centering
    \includegraphics[width=0.8\textwidth]{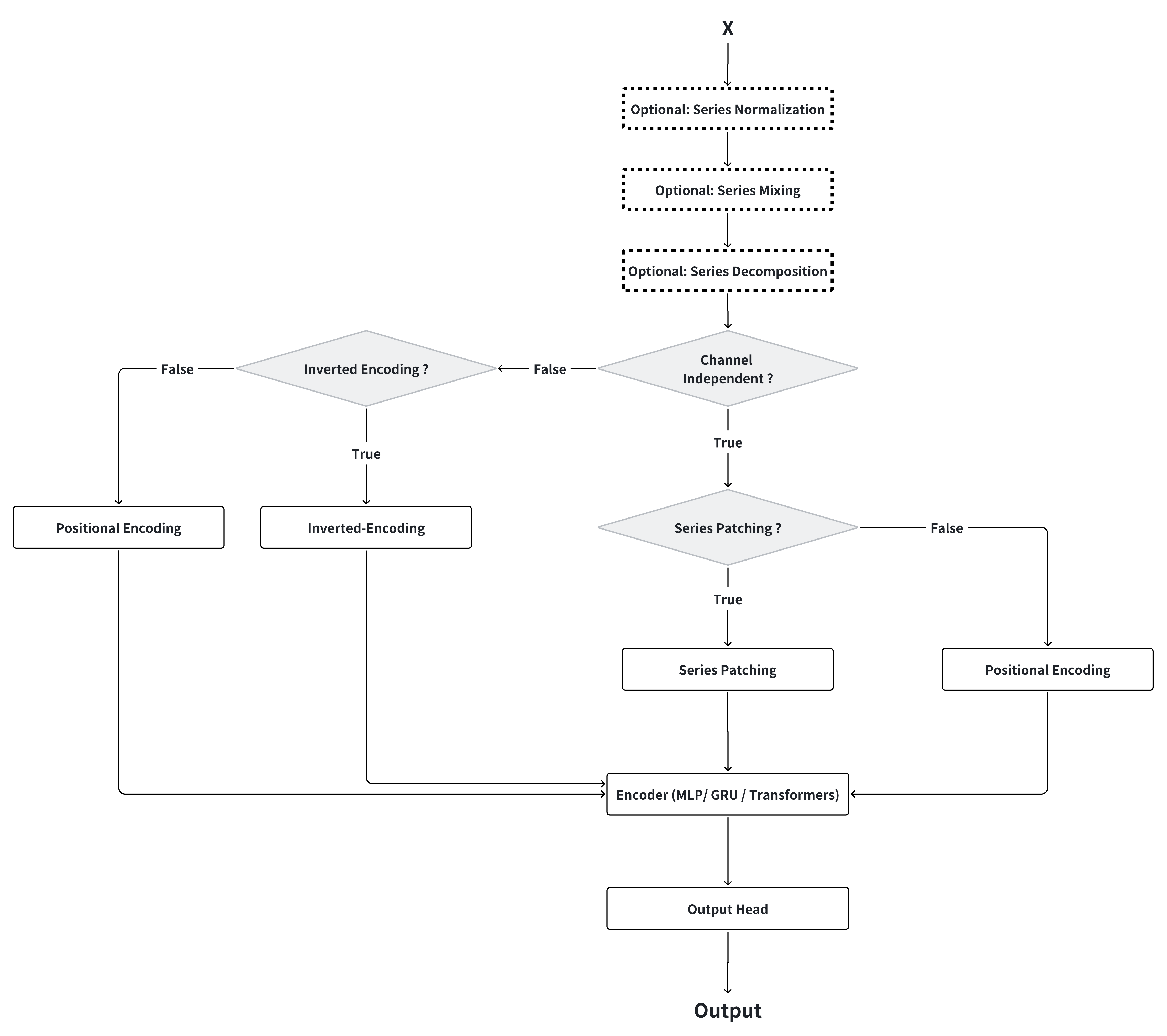}
  \caption{\rv{TSGym pipeline framework. This diagram shows the component-based design of TSGym. The final TSGym structure is formed by combining different component options.}}
  \label{fig:tsgym_pipeline}
\end{figure}

We selected competitive components from the key modules of existing state-of-the-art (SOTA) works as our design choices. In Appx. \ref{appx:modules}, we introduced the underlying principles of these components according to different design dimensions. While most of the individual components have demonstrated their effectiveness through ablation studies in their respective original papers, the interactions and synergies among them when combined have never been systematically explored. Notably, when assembling complete pipelines from different design choices, we automatically exclude incompatible combinations, such as pairing MLP-based architectures with diverse series attention modules.
\rv{To further enhance transparency, we have also created a new, detailed diagram Fig~\ref{fig:tsgym_pipeline} that illustrates the workflow step-by-step through the pipeline. }

\subsection{Meta-features and Meta-predictors}
\label{appx:meta-details}



\begin{figure}[htbp]
  \centering
  \begin{subfigure}[t]{0.4\textwidth}
    \includegraphics[width=\textwidth]{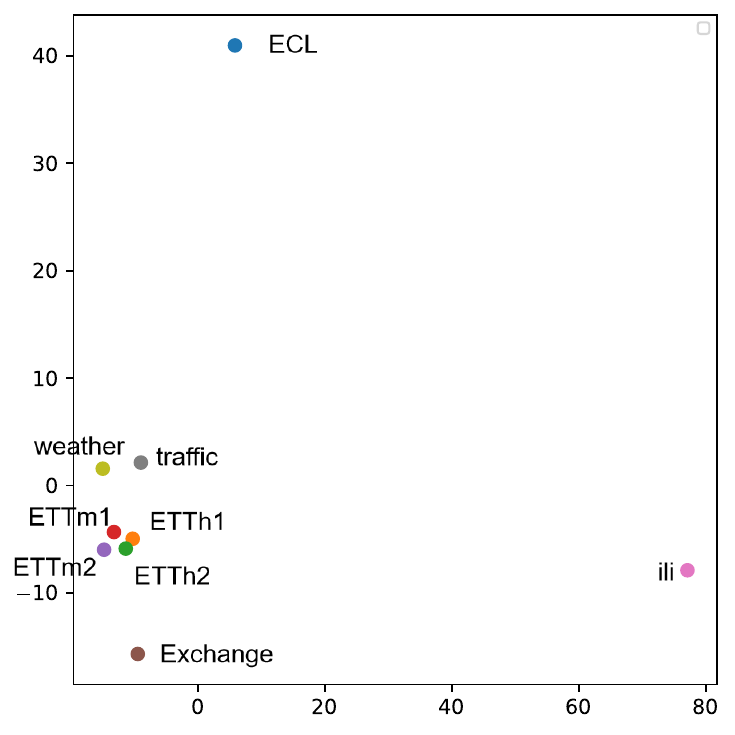}
    \caption{PCA projection of meta-features for 9 long-term forecasting datasets}
    \label{fig:longpca}
  \end{subfigure}
  \hspace{0.05\textwidth}
  \begin{subfigure}[t]{0.4\textwidth}
    \includegraphics[width=\textwidth]{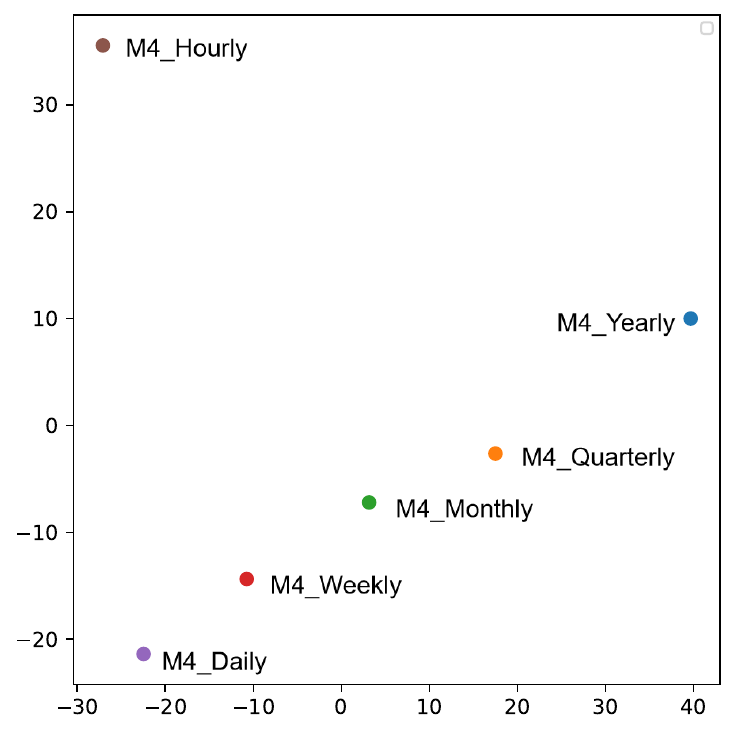}
    \caption{PCA projection of meta-features for 6 short-term forecasting datasets}
    \label{fig:shortpca}
  \end{subfigure}
  \caption{Distributions of meta-features after PCA dimensionality reduction, comparing datasets for long-term and short-term time series forecasting tasks.}
  \label{fig:meta_features_pca}
\end{figure}

\textbf{Details and the selected list of meta-features}.
\label{appx:meta-features}
All meta-features in this paper integrate two complementary perspectives: (1) static characteristics extracted via TSFEL \cite{qiu2024tfb} spanning temporal, statistical, spectral, and fractal domains, and (2) dynamic behavioral metrics from TFB \cite{barandas2020tsfel} to quantify temporal distribution shifts. In Section \ref{exp:benchmark}, we present the results of the meta-predictor trained on meta-features derived from static characteristics, which corresponds to the default setting in TSGym. Furthermore, in Fig. \ref{fig:meta_features_pca}, we visualize the dimension-reduced meta-features across different datasets. In Table~\ref{tab:appx_diverse_mf}, we report the performance of the meta-predictor under various meta-feature configurations.
The following categorizes these features with their analytical purposes (see Tables \ref{tab:meta_features-Temporal}--\ref{tab:meta_features-Fractal} for implementation details):
\begin{itemize}[leftmargin=10pt]
\item \textbf{Temporal features} (Table \ref{tab:meta_features-Temporal}): Characterize sequential dynamics through trend detection, entropy analysis, and change-point statistics, preserving sensitivity to temporal ordering.
\item \textbf{Statistical features} (Table \ref{tab:meta_features-Statistical}): Capture distribution properties via central tendency (mean/median), dispersion (variance/IQR), and shape descriptors (skewness/kurtosis), invariant to observation order.
\item \textbf{Spectral features} (Table \ref{tab:meta_features-Spectral}): Decompose signals into frequency components using Fourier/wavelet transforms, identifying dominant periodicities and hidden oscillations.
\item \textbf{Fractal features} (Table \ref{tab:meta_features-Fractal}): Quantify multiscale complexity through fractal dimensions and Hurst exponents, reflecting self-similarity patterns across temporal resolutions.
\item \textbf{Shifting Metric}: To complement static features, this TFB-derived metric measures temporal distribution drift via KL-divergence between adjacent windows. Values approaching 1 indicate severe shifts caused by external perturbations or systemic transitions, providing a diagnostic tool for non-stationary dynamics.
\end{itemize}

\begin{table*}[ht]
\footnotesize
\centering
	\caption{\textbf{Temporal} Meta-feature Specifications} 
\resizebox{1\textwidth}{!}{
\renewcommand{\arraystretch}{1.3} 
\setlength{\tabcolsep}{4pt} 
	\begin{tabular}{p{3cm}  p{6cm} p{10cm}  } 
		\toprule
		Feature & Description & Functionality \\
		\midrule
        Absolute Energy & Computes the absolute energy of the signal. & Measures the total energy of the signal, often used to understand signal power and activity levels. \\ 
        Area Under the Curve & Computes the area under the curve of the signal computed with the trapezoid rule. & Provides a measure of the overall signal amplitude or ""energy"" over time. \\ 
        Autocorrelation & Calculates the first 1/e crossing of the autocorrelation function (ACF). & Measures the correlation of the signal with its own past values, useful for identifying repeating patterns. \\ 
        Average Power & Computes the average power of the signal. & Averages the squared values of the signal, capturing its power over time. \\ 
        Centroid & Computes the centroid along the time axis. & Indicates the ""center"" or ""balance point"" of the signal in time, providing insight into its distribution. \\ 
        Signal Distance & Computes signal traveled distance. & Measures the total path length covered by the signal over time, capturing the extent of signal fluctuations. \\ 
        Negative Turning & Computes number of negative turning points of the signal. & Counts the number of times the signal changes direction from positive to negative. \\ 
        Neighbourhood Peaks & Computes the number of peaks from a defined neighbourhood of the signal. & Identifies the number of peak points within a specified window, useful for pattern detection. \\ 
        Peak-to-Peak Distance & Computes the peak to peak distance. & Measures the time interval between successive peaks, indicating the period of oscillations. \\ 
        Positive Turning & Computes number of positive turning points of the signal. & Counts the number of times the signal changes direction from negative to positive. \\ 
        Root Mean Square & Computes root mean square of the signal. & Calculates the square root of the average squared values of the signal, often used as a measure of signal strength. \\ 
        Slope & Computes the slope of the signal. & Measures the rate of change in the signal's amplitude over time, indicating trends or shifts. \\ 
        Sum of Absolute Differences & Computes sum of absolute differences of the signal. & Measures the total variation in the signal by summing the absolute differences between consecutive values. \\ 
        Zero-Crossing Rate & Computes Zero-crossing rate of the signal. & Counts how many times the signal crosses the zero axis, indicating its frequency and periodicity. \\ 
        \bottomrule
	\end{tabular}}
	\label{tab:meta_features-Temporal} 
\end{table*}

\begin{table*}[ht]
\footnotesize
\centering
	\caption{\textbf{Statistical} Meta-feature Specifications} 
\resizebox{1\textwidth}{!}{
\renewcommand{\arraystretch}{1.3} 
\setlength{\tabcolsep}{4pt} 
	\begin{tabular}{p{3cm}  p{6cm} p{10cm}  } 
		\toprule
		Feature & Description & Functionality \\
		\midrule
        Maximum Value & Computes the maximum value of the signal. & Identifies the highest amplitude or peak value in the signal, useful for determining extreme values. \\ 
        Mean Value & Computes mean value of the signal. & Calculates the average value of the signal, providing insight into its central tendency. \\ 
        Median & Computes the median of the signal. & Finds the middle value of the signal when sorted, offering robustness to outliers. \\ 
        Minimum Value & Computes the minimum value of the signal. & Identifies the lowest amplitude or trough value in the signal, useful for detecting minima. \\ 
        Standard Deviation & Computes standard deviation (std) of the signal. & Measures the variation or spread of the signal values, indicating how much the signal deviates from the mean. \\ 
        Variance & Computes variance of the signal. & Quantifies the spread of signal values, related to the square of the standard deviation. \\ 
        Empirical Cumulative Distribution Function & Computes the values of ECDF along the time axis. & Provides a cumulative distribution function, representing the probability distribution of the signal values. \\ 
        ECDF Percentile & Computes the percentile value of the ECDF. & Extracts specific percentiles from the cumulative distribution, useful for understanding the signal's quantiles. \\ 
        ECDF Percentile Count & Computes the cumulative sum of samples that are less than the percentile. & Measures the number of samples falling below a given percentile, providing distribution insights. \\ 
        ECDF Slope & Computes the slope of the ECDF between two percentiles. & Measures the steepness or rate of change in the cumulative distribution, indicating distribution sharpness. \\ 
        Histogram Mode & Compute the mode of a histogram using a given number of bins. & Finds the most frequent value in the signal's histogram, representing the peak of the signal's distribution. \\ 
        Interquartile Range & Computes interquartile range of the signal. & Measures the range between the 25th and 75th percentiles, indicating the spread of the central 50\% of the signal values. \\ 
        Kurtosis & Computes kurtosis of the signal. & Measures the ""tailedness"" of the signal distribution, indicating the presence of outliers or extreme values. \\ 
        Mean Absolute Deviation & Computes mean absolute deviation of the signal. & Measures the average deviation of the signal values from the mean, providing an indication of signal variability. \\ 
        Mean Absolute Difference & Computes mean absolute differences of the signal. & Calculates the average of absolute differences between successive signal values, reflecting the signal's smoothness. \\ 
        Mean Difference & Computes mean of differences of the signal. & Computes the average of the first-order differences, used to measure overall signal change. \\ 
        Median Absolute Deviation & Computes median absolute deviation of the signal. & Measures the spread of the signal values around the median, offering a robust measure of variability. \\ 
        Median Absolute Difference & Computes median absolute differences of the signal. & Similar to mean absolute difference but based on the median, used to assess signal smoothness. \\ 
        Median Difference & Computes median of differences of the signal. & Calculates the median of first-order differences, providing insights into signal trend stability. \\ 
        Skewness & Computes skewness of the signal. & Measures the asymmetry of the signal's distribution, indicating whether it is skewed towards higher or lower values. \\ 
        \bottomrule
	\end{tabular}}
	\label{tab:meta_features-Statistical} 
\end{table*}

\begin{table*}[ht]
\footnotesize
\centering
	\caption{\textbf{Spectral} Meta-feature Specifications} 
\resizebox{1\textwidth}{!}{
\renewcommand{\arraystretch}{1.3} 
\setlength{\tabcolsep}{4pt} 
	\begin{tabular}{p{3cm}  p{6cm} p{10cm}  } 
		\toprule
		Feature & Description & Functionality \\
		\midrule
        Entropy & Computes the entropy of the signal using Shannon Entropy. & Quantifies the uncertainty or randomness in the signal, offering insights into its complexity. \\ 
        Fundamental Frequency & Computes the fundamental frequency of the signal. & Identifies the primary frequency at which the signal oscillates, crucial for detecting periodic behaviors. \\ 
        Human Range Energy & Computes the human range energy ratio. & Measures the energy in the human audible range, useful for identifying signals relevant to human hearing. \\ 
        Linear Prediction Cepstral Coefficients & Computes the linear prediction cepstral coefficients. & Extracts features related to the signal's frequency components, commonly used in speech and audio processing. \\ 
        Maximum Frequency & Computes maximum frequency of the signal. & Identifies the highest frequency component of the signal, providing insight into its frequency range. \\ 
        Maximum Power Spectrum & Computes maximum power spectrum density of the signal. & Measures the peak value in the power spectral density, identifying dominant frequencies in the signal. \\ 
        Median Frequency & Computes median frequency of the signal. & Identifies the frequency that divides the signal's power spectrum into two equal halves. \\ 
        Mel-Frequency Cepstral Coefficients & Computes the MEL cepstral coefficients. & Used to extract features representing the spectral characteristics of the signal, primarily used in speech analysis. \\ 
        Multiscale Entropy & Computes the Multiscale entropy (MSE) of the signal, that performs entropy analysis over multiple scales. & Quantifies the signal's complexity at different scales, useful for detecting non-linear temporal behaviors. \\ 
        Power Bandwidth & Computes power spectrum density bandwidth of the signal. & Measures the width of the frequency band where the majority of the signal's power is concentrated. \\ 
        Spectral Centroid & Barycenter of the spectrum. & Identifies the ""center"" of the signal's frequency spectrum, used in sound and audio analysis. \\ 
        Spectral Decrease & Represents the amount of decreasing of the spectra amplitude. & Measures how rapidly the spectral amplitude decreases across frequency, useful for identifying spectral roll-off. \\ 
        Spectral Distance & Computes the signal spectral distance. & Quantifies the difference between the signal's spectrum and a reference, helpful in pattern recognition. \\ 
        Spectral Entropy & Computes the spectral entropy of the signal based on Fourier transform. & Measures the randomness or complexity in the frequency domain of the signal. \\ 
        Spectral Kurtosis & Measures the flatness of a distribution around its mean value. & Quantifies the tail heaviness of the signal's frequency distribution, identifying outliers or abnormal distributions. \\ 
        Spectral Positive Turning & Computes number of positive turning points of the fft magnitude signal. & Counts the points where the signal's Fourier transform changes direction from negative to positive. \\ 
        Spectral Roll-Off & Computes the spectral roll-off of the signal. & Measures the frequency below which a specified percentage of the total spectral energy is contained. \\ 
        Spectral Roll-On & Computes the spectral roll-on of the signal. & Similar to roll-off but identifies the frequency above which a specified amount of energy is concentrated. \\ 
        Spectral Skewness & Measures the asymmetry of a distribution around its mean value. & Measures the skew in the signal’s frequency distribution, highlighting the presence of spectral biases. \\ 
        Spectral Slope & Computes the spectral slope. & Quantifies the slope of the power spectral density, often used to distinguish between harmonic and non-harmonic signals. \\ 
        Spectral Spread & Measures the spread of the spectrum around its mean value. & Measures the dispersion or spread of the signal's spectral energy. \\ 
        Spectral Variation & Computes the amount of variation of the spectrum along time. & Quantifies how much the frequency content of the signal changes over time. \\ 
        Spectrogram Mean Coefficients & Calculates the average power spectral density (PSD) for each frequency throughout the entire signal. & Averages the power spectral density across all time intervals, capturing the signal's overall spectral energy distribution. \\ 
        Wavelet Absolute Mean & Computes CWT absolute mean value of each wavelet scale. & Measures the average wavelet transform magnitude across scales, useful for detecting changes in signal frequency. \\ 
        Wavelet Energy & Computes CWT energy of each wavelet scale. & Quantifies the energy at each wavelet scale, reflecting the signal's energy distribution across frequencies. \\ 
        Wavelet Entropy & Computes CWT entropy of the signal. & Measures the complexity or unpredictability of the signal at different wavelet scales. \\ 
        Wavelet Standard Deviation & Computes CWT std value of each wavelet scale. & Measures the variation or spread of the wavelet transform across different scales. \\ 
        Wavelet Variance & Computes CWT variance value of each wavelet scale. & Quantifies the dispersion of the signal at different wavelet scales. \\ 
        \bottomrule
	\end{tabular}}
	\label{tab:meta_features-Spectral} 
\end{table*}

\begin{table*}[ht]
\footnotesize
\centering
	\caption{\textbf{Fractal} Meta-feature Specifications} 
\resizebox{1\textwidth}{!}{
\renewcommand{\arraystretch}{1.3} 
\setlength{\tabcolsep}{4pt} 
	\begin{tabular}{p{3cm}  p{6cm} p{10cm}  } 
		\toprule
		Feature & Description & Functionality \\
		\midrule
        Detrended Fluctuation Analysis & Computes the Detrended Fluctuation Analysis (DFA) of the signal. & Measures long-range correlations and self-similarity in the signal, used for identifying fractal behavior. \\ 
        Higuchi Fractal Dimension & Computes the fractal dimension of a signal using Higuchi's method (HFD). & Measures the complexity of the signal's pattern by calculating its fractal dimension. \\ 
        Hurst Exponent & Computes the Hurst exponent of the signal through the Rescaled range (R/S) analysis. & Measures the long-term memory or persistence in the signal, useful for identifying trends and randomness. \\ 
        Lempel-Ziv Complexity & Computes the Lempel-Ziv's (LZ) complexity index, normalized by the signal's length. & Quantifies the randomness or predictability of the signal based on its compressibility. \\ 
        Maximum Fractal Length & Computes the Maximum Fractal Length (MFL) of the signal. & Measures the fractal dimension at the smallest scale of the signal, reflecting its intricate pattern complexity. \\ 
        Petrosian Fractal Dimension & Computes the Petrosian Fractal Dimension of a signal. & Measures the signal's fractal dimension based on its variation across different scales. \\
        \bottomrule
	\end{tabular}}
	\label{tab:meta_features-Fractal} 
\end{table*}


\textbf{Details of the trained meta-predictors}.
\label{appx:meta-predictors}
For each design choice, we first use the LabelEncoder class from scikit-learn to convert it into a numerical class index. This index is then fed into an $nn.Embedding$ layer within our model to obtain a dense vector representation. These learned embeddings, along with other meta-features, subsequently form the input to the meta-predictor.
The meta-predictor is optimized using Pearson loss to learn the relative performance ranks of different design choices, thereby emphasizing the linear correlation between predicted and actual rankings.

Moreover, we experimented with different training strategies to guide the meta-predictor in selecting the top-$1$ design pipelines. 

(1) \textbf{+Resample}: Constraining the number of combinations from different datasets to be equal when training the meta-predictor. 

(2) \textbf{+AllPL}: Training on datasets with varying prediction lengths and transfers this knowledge to a test set with a single prediction length. 

(3) We train the meta-predictor using diverse meta features, including those generated by segmenting the datasets based on timestamps (\textbf{Sub}), those combining information from different time periods (\textbf{Whole}), and those designed to capture distributional shifts (\textbf{Delta}). The symbol \textbf{"+"} denotes the concatenation of multiple meta features.

We report the results of \textbf{+Resample} and \textbf{+AllPL} in Table \ref{tab:tsgym_ablation}, and the results of diverse meta-features in Table~\ref{tab:appx_diverse_mf}.

\clearpage

\section{Additional Experimental Results}
\label{appx:complete_results}

\subsection{Comprehensive Results of TSGym Against State-of-the-Art Methods}
Due to space limitations in the main text, here we provide complete experimental comparisons for both long-term and short-term forecasting tasks. Table \ref{tab:TSGym_vs_Sota_fullAvg} details the full long-term forecasting performance across all prediction horizons, while Table \ref{tab:TSGym_vs_Sota_short_full} presents the comprehensive short-term forecasting results. Following standard benchmarking conventions, we highlight top-performing methods in \textcolor{red}{\textbf{red}} and second-best results with \underline{\textcolor{blue}{underlined}} formatting. These extensive evaluations consistently validate TSGym's competitive performance across diverse temporal prediction scenarios.
In addition, we investigate the impact of different meta-feature configurations through controlled ablation studies. As demonstrated in Table~\ref{tab:appx_diverse_mf}, no individual meta-feature configuration exhibits consistent superiority across all datasets. 
\begin{table}
\centering

\makebox[\textwidth][c]{  
\begin{minipage}{1\textwidth}  

\caption{\update{Full results for the long-term forecasting task.} All the results are averaged from 4 different prediction lengths, that is $\{24,36,48,60\}$ for ILI and $\{96,192,336,720\}$ for the others.}
\label{tab:TSGym_vs_Sota_fullAvg}
\centering
\resizebox{1\textwidth}{!}{
\setlength{\tabcolsep}{3pt} 
\begin{tabular}{c|rr|rr|rr|rr|rr|rr|rr|rr|rr|rr|rr|rr|rr} 
\toprule
\multicolumn{1}{c}{\multirow{2}{*}{Models}} & \multicolumn{2}{c}{\textbf{TSGym}}                                & \multicolumn{2}{c}{DUET}                                          & \multicolumn{2}{c}{TimeMixer}           & \multicolumn{2}{c}{MICN}                & \multicolumn{2}{c}{TimesNet}                                      & \multicolumn{2}{c}{PatchTST}            & \multicolumn{2}{c}{DLinear} & \multicolumn{2}{c}{Crossformer} & \multicolumn{2}{c}{Autoformer} & \multicolumn{2}{c}{SegRNN}                                        & \multicolumn{2}{c}{Mamba} & \multicolumn{2}{c}{iTransformer}        & \multicolumn{2}{c}{TimeXer}                                        \\

\multicolumn{1}{c}{}                        & 
\multicolumn{2}{c}{(\textbf{Ours})}                               & \multicolumn{2}{c}{\cite{qiu2025DUET}} &
    \multicolumn{2}{c}{\cite{wang2024timemixer}}  &
    \multicolumn{2}{c}{\cite{wang2023micn}} & 
    \multicolumn{2}{c}{\cite{wu2023timesnet}}  & 
    \multicolumn{2}{c}{\cite{nie2023PatchTST}}  & 
    \multicolumn{2}{c}{\cite{zeng2023dlinear}}  & 
    \multicolumn{2}{c}{\cite{zhang2023crossformer}} &
    \multicolumn{2}{c}{\cite{wu2021autoformer}} &
    \multicolumn{2}{c}{\cite{lin2023segrnn}}  &
     \multicolumn{2}{c}{\cite{gu2024mamba}} & 
     \multicolumn{2}{c}{\cite{liu2024itransformer}}
     & 
     \multicolumn{2}{c}{\cite{wang2024timexer}}                                              \\ 
\cmidrule(l){2-27}
\multicolumn{1}{c}{Metric}                  & MSE                             & MAE                             & MSE                             & MAE                             & MSE   & MAE                             & MSE                             & MAE   & MSE                             & MAE                             & MSE   & MAE                             & MSE   & MAE                 & MSE   & MAE                     & MSE   & MAE                    & MSE                             & MAE                             & MSE   & MAE               & MSE                             & MAE   & MSE                             & MAE                              \\ 
\midrule
\textbf{ETTm1}                              & \textcolor{red}{\textbf{0.360}} & \textcolor{red}{\textbf{0.384}} & 0.407                           & 0.409                           & 0.384 & 0.399                           & 0.402                           & 0.429 & 0.432                           & 0.430                           & 0.390 & 0.404                           & 0.404 & 0.407               & 0.501 & 0.501                   & 0.532 & 0.496                  & 0.388                           & 0.404                           & 0.501 & 0.466             & 0.414                           & 0.415 & 0.386                           & 0.400                            \\
\textbf{ETTm2}                              & \underline{\textcolor{blue}{0.265}} & \textcolor{red}{\textbf{0.322}} & 0.296                           & 0.338                           & 0.277 & 0.325                           & 0.342                           & 0.391 & 0.296                           & 0.334                           & 0.288 & 0.334                           & 0.349 & 0.399               & 1.487 & 0.789                   & 0.330 & 0.368                  & 0.273                           & \underline{\textcolor{blue}{0.322}} & 0.356 & 0.370             & 0.290                           & 0.332 & 0.279                           & 0.325                            \\
\textbf{ETTh1}                              & \underline{\textcolor{blue}{0.425}} & \underline{\textcolor{blue}{0.434}} & 0.433                           & 0.437                           & 0.448 & 0.438                           & 0.589                           & 0.537 & 0.474                           & 0.464                           & 0.454 & 0.449                           & 0.465 & 0.461               & 0.544 & 0.520                   & 0.492 & 0.485                  & \textcolor{red}{\textbf{0.422}} & \textcolor{red}{\textbf{0.429}} & 0.544 & 0.504             & 0.462                           & 0.452 & 0.446                           & 0.443                            \\
\textbf{ETTh2}                              & \textcolor{red}{\textbf{0.371}} & 0.406                           & 0.380                           & \underline{\textcolor{blue}{0.403}} & 0.383 & 0.406                           & 0.585                           & 0.530 & 0.415                           & 0.424                           & 0.385 & 0.409                           & 0.566 & 0.520               & 1.552 & 0.908                   & 0.446 & 0.460                  & 0.374                           & 0.405                           & 0.465 & 0.448             & 0.382                           & 0.406 & \underline{\textcolor{blue}{0.372}} & \textcolor{red}{\textbf{0.399}}  \\
\textbf{ECL}                                & \textcolor{red}{\textbf{0.179}} & 0.275                           & \underline{\textcolor{blue}{0.179}} & \textcolor{red}{\textbf{0.262}} & 0.185 & \underline{\textcolor{blue}{0.273}} & 0.186                           & 0.297 & 0.219                           & 0.314                           & 0.209 & 0.298                           & 0.225 & 0.319               & 0.193 & 0.289                   & 0.234 & 0.340                  & 0.216                           & 0.302                           & 0.209 & 0.312             & 0.190                           & 0.277 & 0.191                           & 0.286                            \\
\textbf{Traffic}                            & \textcolor{red}{\textbf{0.434}} & \textcolor{red}{\textbf{0.310}} & 0.797                           & 0.427                           & 0.496 & \underline{\textcolor{blue}{0.313}} & 0.544                           & 0.320 & 0.645                           & 0.348                           & 0.497 & 0.321                           & 0.673 & 0.419               & 1.458 & 0.782                   & 0.637 & 0.397                  & 0.807                           & 0.411                           & 0.679 & 0.380             & \underline{\textcolor{blue}{0.474}} & 0.318 & 0.509                           & 0.333                            \\
\textbf{Weather}                            & \textcolor{red}{\textbf{0.229}} & \textcolor{red}{\textbf{0.267}} & 0.252                           & 0.277                           & 0.244 & 0.274                           & 0.264                           & 0.316 & 0.261                           & 0.287                           & 0.256 & 0.279                           & 0.265 & 0.317               & 0.253 & 0.312                   & 0.339 & 0.379                  & 0.251                           & 0.298                           & 0.291 & 0.315             & 0.259                           & 0.280 & 0.243                           & 0.273                            \\
\textbf{Exchange}                           & 0.392                           & 0.418                           & \textcolor{red}{\textbf{0.322}} & \textcolor{red}{\textbf{0.384}} & 0.359 & \underline{\textcolor{blue}{0.402}} & \underline{\textcolor{blue}{0.346}} & 0.422 & 0.405                           & 0.437                           & 0.381 & 0.412                           & 0.346 & 0.414               & 0.904 & 0.695                   & 0.506 & 0.500                  & 0.408                           & 0.423                           & 0.714 & 0.562             & 0.369                           & 0.410 & 0.410                           & 0.424                            \\
\textbf{ILI}                                & 2.345                           & 1.053                           & 2.640                           & 1.018                           & 4.502 & 1.557                           & 2.938                           & 1.178 & \underline{\textcolor{blue}{2.140}} & \underline{\textcolor{blue}{0.907}} & 2.160 & \textcolor{red}{\textbf{0.901}} & 4.367 & 1.540               & 4.311 & 1.396                   & 3.156 & 1.207                  & 4.305                           & 1.397                           & 3.729 & 1.335             & 2.305                           & 0.974 & 2.633                           & 1.034                            \\ 
\midrule
\textbf{$1^{\text{st}}$ Count}              & \multicolumn{2}{c}{\textcolor{red}{9} }                                            & \multicolumn{2}{c}{\underline{\textcolor{blue}{3}}}                   & \multicolumn{2}{c}{0}                   & \multicolumn{2}{c}{0}                   & \multicolumn{2}{c}{0}                                             & \multicolumn{2}{c}{1}                   & \multicolumn{2}{c}{0}       & \multicolumn{2}{c}{0}           & \multicolumn{2}{c}{0}          & \multicolumn{2}{c}{2}                                             & \multicolumn{2}{c}{0}     & \multicolumn{2}{c}{0}                   & \multicolumn{2}{c}{1}                                              \\
\bottomrule
\toprule
\multicolumn{1}{c}{\multirow{2}{*}{Models}} & \multicolumn{2}{c}{PAttn} & \multicolumn{2}{c}{Koopa}                                         & \multicolumn{2}{c}{TSMixer} & \multicolumn{2}{c}{FreTS} & \multicolumn{2}{c}{Pyraformer} & \multicolumn{2}{c}{Nonstationary} & \multicolumn{2}{c}{ETSformer} & \multicolumn{2}{c}{FEDformer} & \multicolumn{2}{c}{SCINet} & \multicolumn{2}{c}{LightTS} & \multicolumn{2}{c}{Informer} & \multicolumn{2}{c}{Transformer} & \multicolumn{2}{c}{Reformer}  \\
    
    \multicolumn{1}{c}{}&
    \multicolumn{2}{c}{\cite{tan2024pattn}}            & \multicolumn{2}{c}{\cite{liu2023koopa}}& 
     \multicolumn{2}{c}{\cite{chen2023tsmixer}}           & \multicolumn{2}{c}{\cite{yi2023frets}}             & \multicolumn{2}{c}{\cite{liu2022pyraformer}}            & \multicolumn{2}{c}{\cite{Liu2022NonstationaryTR}}          & \multicolumn{2}{c}{\cite{woo2022etsformer}} &
     \multicolumn{2}{c}{\cite{zhou2022fedformer}}      & 
     \multicolumn{2}{c}{\cite{liu2022scinet}}   
     &
     \multicolumn{2}{c}{\cite{zhang2022LightTS}}  & 
     \multicolumn{2}{c}{\cite{zhou2021informer}}  & 
     \multicolumn{2}{c}{\cite{vaswani2017attention}}  & 
     \multicolumn{2}{c}{\cite{Kitaev2020Reformer}} \\ 
\cmidrule(l){2-27}
\multicolumn{1}{c}{Metric}                  & MSE   & MAE               & MSE                             & MAE                             & MSE   & MAE                 & MSE   & MAE               & MSE   & MAE                    & MSE   & MAE                       & MSE   & MAE                   & MSE   & MAE                   & MSE   & MAE                & MSE   & MAE                 & MSE   & MAE                  & MSE   & MAE                     & MSE   & MAE                   \\ 
\midrule
\textbf{ETTm1}                              & 0.384 & 0.399             & \underline{\textcolor{blue}{0.367}} & \underline{\textcolor{blue}{0.396}} & 0.527 & 0.512               & 0.409 & 0.417             & 0.695 & 0.593                  & 0.509 & 0.467                     & 0.636 & 0.592                 & 0.438 & 0.450                 & 0.409 & 0.412              & 0.438 & 0.445               & 0.969 & 0.736                & 0.836 & 0.678                   & 0.998 & 0.723                 \\
\textbf{ETTm2}                              & 0.291 & 0.336             & \textcolor{red}{\textbf{0.264}} & 0.327                           & 1.030 & 0.750               & 0.336 & 0.378             & 1.565 & 0.876                  & 0.412 & 0.398                     & 1.381 & 0.807                 & 0.301 & 0.348                 & 0.294 & 0.335              & 0.432 & 0.448               & 1.504 & 0.878                & 1.454 & 0.851                   & 1.856 & 0.996                 \\
\textbf{ETTh1}                              & 0.468 & 0.454             & 0.472                           & 0.471                           & 0.615 & 0.579               & 0.476 & 0.464             & 0.814 & 0.692                  & 0.610 & 0.543                     & 0.750 & 0.651                 & 0.448 & 0.461                 & 0.520 & 0.488              & 0.530 & 0.505               & 1.057 & 0.798                & 0.930 & 0.768                   & 0.973 & 0.739                 \\
\textbf{ETTh2}                              & 0.386 & 0.412             & 0.388                           & 0.423                           & 2.160 & 1.220               & 0.548 & 0.514             & 3.776 & 1.557                  & 0.552 & 0.505                     & 0.572 & 0.534                 & 0.427 & 0.446                 & 0.428 & 0.440              & 0.633 & 0.551               & 4.535 & 1.745                & 2.976 & 1.369                   & 2.487 & 1.238                 \\
\textbf{ECL}                                & 0.205 & 0.286             & 0.219                           & 0.319                           & 0.229 & 0.337               & 0.209 & 0.296             & 0.295 & 0.387                  & 0.194 & 0.296                     & 0.275 & 0.370                 & 0.225 & 0.336                 & 0.220 & 0.323              & 0.243 & 0.344               & 0.369 & 0.444                & 0.273 & 0.367                   & 0.324 & 0.404                 \\
\textbf{Traffic}                            & 0.513 & 0.328             & 0.595                           & 0.413                           & 0.599 & 0.403               & 0.597 & 0.377             & 0.697 & 0.391                  & 0.642 & 0.351                     & 1.035 & 0.584                 & 0.615 & 0.379                 & 0.654 & 0.419              & 0.656 & 0.428               & 0.830 & 0.464                & 0.708 & 0.384                   & 0.694 & 0.380                 \\
\textbf{Weather}                            & 0.257 & 0.280             & \underline{\textcolor{blue}{0.230}} & \underline{\textcolor{blue}{0.271}} & 0.242 & 0.301               & 0.255 & 0.299             & 0.284 & 0.349                  & 0.289 & 0.312                     & 0.365 & 0.424                 & 0.315 & 0.369                 & 0.256 & 0.283              & 0.245 & 0.295               & 0.572 & 0.523                & 0.599 & 0.531                   & 0.475 & 0.472                 \\
\textbf{Exchange}                           & 0.365 & 0.407             & 0.610                           & 0.516                           & 0.487 & 0.546               & 0.442 & 0.453             & 1.183 & 0.855                  & 0.557 & 0.490                     & 0.361 & 0.416                 & 0.520 & 0.502                 & 0.374 & 0.418              & 0.486 & 0.493               & 1.548 & 0.997                & 1.379 & 0.921                   & 1.612 & 1.044                 \\
\textbf{ILI}                                & 2.359 & 0.975             & \textcolor{red}{\textbf{2.064}} & 0.912                           & 5.617 & 1.680               & 3.447 & 1.279             & 4.691 & 1.442                  & 2.592 & 1.012                     & 4.046 & 1.419                 & 3.088 & 1.214                 & 6.505 & 1.853              & 7.078 & 1.975               & 5.035 & 1.539                & 4.682 & 1.448                   & 4.211 & 1.350                 \\ 
\midrule
\textbf{$1^{\text{st}}$ Count}              & \multicolumn{2}{c}{0}     & \multicolumn{2}{c}{2}                                             & \multicolumn{2}{c}{0}       & \multicolumn{2}{c}{0}     & \multicolumn{2}{c}{0}          & \multicolumn{2}{c}{0}             & \multicolumn{2}{c}{0}         & \multicolumn{2}{c}{0}         & \multicolumn{2}{c}{0}      & \multicolumn{2}{c}{0}       & \multicolumn{2}{c}{0}        & \multicolumn{2}{c}{0}           & \multicolumn{2}{c}{0}         \\
\bottomrule
\end{tabular}
}
\end{minipage}
}
\end{table}

\label{tab:TSGym_vs_Sota_fullAvg}

\begin{sidewaystable}[htbp]
  \centering
  \caption{Full results for the short-term forecasting task in the M4 dataset. $\ast$. in the Transformers indicates the name of $\ast$former.}
    \label{tab:addlabel}%
    \resizebox{1\textwidth}{!}{
    \renewcommand{\arraystretch}{1.2} 
\setlength{\tabcolsep}{3pt} 
    \begin{tabular}{c|c|ccccccccccccccccccccccccc}
    \toprule
    \multicolumn{2}{c}{\multirow{2}[1]{*}{Models}} & TSGym & TimeMixer & MICN  & TimesNet & PatchTST & DLinear & Cross. & Auto. & SegRNN & Mamba & iTrans. & TimeXer & PAttn & TSMixer & FreTS & Pyra. & ETS. & FED. & SCINet & LightTS & In. & Trans. & Re. & TiDE  & FiLM \\
    \multicolumn{2}{c}{}                         & \multicolumn{1}{c}{(Ours)}       &
\cite{wang2024timemixer} &   
\cite{wang2023micn}     &  
\cite{wu2023timesnet}                                 &    
\cite{nie2023PatchTST}      &     
\cite{zeng2023dlinear}                      &    
\cite{zhang2023crossformer}     &    
\cite{wu2021autoformer}   &          
\cite{lin2023segrnn}                        &   
\cite{gu2024mamba}  &              
\cite{liu2024itransformer}               &   
\cite{wang2024timexer}     &     
\cite{tan2024pattn} &  \cite{chen2023tsmixer}       &    \cite{yi2023frets}    &   \cite{liu2022pyraformer}     &      \cite{woo2022etsformer}                    &    \cite{zhou2022fedformer}    &    \cite{liu2022scinet}    &                         \cite{zhang2022LightTS}         &   \cite{zhou2021informer}     &      \cite{vaswani2017attention}   &   \cite{Kitaev2020Reformer}     &    \cite{das2023TiDE}    &    \cite{zhou2022film} \\
    \midrule
    \multirow{3}[2]{*}{\begin{sideways}Yearly\end{sideways}} & OWA   & \textcolor[rgb]{ 1,  0,  0}{\textbf{0.779 }} & 0.789  & 0.873  & \textcolor[rgb]{ 0,  0,  1}{0.784 } & 0.798  & 0.843  & 4.407  & 1.019  & 0.858  & 0.790  & 0.807  & 0.797  & 0.823  & 0.798  & 0.800  & 0.883  & 0.982  & 0.806  & 0.801  & 0.795  & 0.887  & 4.406  & 0.829  & 0.807  & 0.806  \\
          & sMAPE & \textcolor[rgb]{ 1,  0,  0}{\textbf{13.286 }} & 13.467  & 14.586  & \textcolor[rgb]{ 0,  0,  1}{13.344 } & 13.606  & 14.402  & 71.464  & 17.294  & 14.323  & 13.388  & 13.681  & 13.551  & 13.893  & 13.539  & 13.579  & 14.967  & 16.105  & 13.631  & 13.573  & 13.516  & 15.086  & 69.405  & 14.064  & 14.006  & 13.988  \\
          & MASE  & \textcolor[rgb]{ 1,  0,  0}{\textbf{2.960 }} & 3.000  & 3.392  & \textcolor[rgb]{ 0,  0,  1}{2.983 } & 3.033  & 3.198  & 17.649  & 3.897  & 3.335  & 3.021  & 3.090  & 3.043  & 3.162  & 3.051  & 3.056  & 3.377  & 3.888  & 3.088  & 3.065  & 3.030  & 3.382  & 18.144  & 3.166  & 3.011  & 3.007  \\
    \midrule
    \multirow{3}[2]{*}{\begin{sideways}Quarterly\end{sideways}} & OWA   & 0.891  & 0.911  & 1.025  & \textcolor[rgb]{ 1,  0,  0}{\textbf{0.886 }} & 0.975  & 0.928  & 8.208  & 1.290  & 1.009  & 0.912  & 1.050  & 0.942  & 0.897  & 0.903  & 0.908  & 1.002  & 1.312  & 0.940  & 0.922  & \textcolor[rgb]{ 0,  0,  1}{0.886 } & 1.248  & 8.190  & 1.007  & 0.959  & 0.960  \\
          & sMAPE & 10.232  & 10.313  & 11.427  & \textcolor[rgb]{ 1,  0,  0}{\textbf{10.063 }} & 10.975  & 10.500  & 74.297  & 14.085  & 11.193  & 10.320  & 11.752  & 10.536  & 10.203  & 10.218  & 10.329  & 11.259  & 13.600  & 10.655  & 10.426  & \textcolor[rgb]{ 0,  0,  1}{10.166 } & 13.644  & 73.944  & 11.324  & 10.719  & 10.742  \\
          & MASE  & \textcolor[rgb]{ 0,  0,  1}{1.169 } & 1.215  & 1.388  & 1.176  & 1.306  & 1.238  & 13.260  & 1.784  & 1.374  & 1.217  & 1.416  & 1.272  & 1.189  & 1.204  & 1.205  & 1.345  & 1.906  & 1.252  & 1.229  & \textcolor[rgb]{ 1,  0,  0}{\textbf{1.163 }} & 1.723  & 13.256  & 1.352  & 1.295  & 1.296  \\
    \midrule
    \multirow{3}[2]{*}{\begin{sideways}Monthly\end{sideways}} & OWA   & \textcolor[rgb]{ 1,  0,  0}{\textbf{0.863 }} & 0.895  & 0.986  & 0.939  & 1.020  & 0.936  & 7.637  & 1.369  & 1.074  & 0.931  & 0.982  & 0.944  & 0.951  & 0.898  & 0.915  & 1.043  & 1.281  & 1.001  & 0.924  & \textcolor[rgb]{ 0,  0,  1}{0.884 } & 1.151  & 7.668  & 1.448  & 0.942  & 0.942  \\
          & sMAPE & \textcolor[rgb]{ 1,  0,  0}{\textbf{12.570 }} & 12.823  & 13.798  & 13.314  & 14.156  & 13.384  & 68.873  & 18.132  & 15.052  & 13.152  & 13.737  & 13.254  & 13.421  & 12.865  & 13.059  & 14.666  & 15.449  & 14.112  & 13.146  & \textcolor[rgb]{ 0,  0,  1}{12.717 } & 15.806  & 69.992  & 18.782  & 13.381  & 13.352  \\
          & MASE  & \textcolor[rgb]{ 1,  0,  0}{\textbf{0.909 }} & 0.959  & 1.080  & 1.015  & 1.126  & 1.004  & 11.165  & 1.576  & 1.175  & 1.010  & 1.076  & 1.030  & 1.034  & 0.962  & 0.984  & 1.138  & 1.586  & 1.090  & 0.996  & \textcolor[rgb]{ 0,  0,  1}{0.943 } & 1.283  & 11.149  & 1.694  & 1.017  & 1.019  \\
    \midrule
    \multirow{3}[2]{*}{\begin{sideways}Weekly\end{sideways}} & OWA   & \textcolor[rgb]{ 1,  0,  0}{\textbf{0.983 }} & 1.266  & 1.500  & 1.310  & 1.035  & 1.461  & 28.636  & 1.575  & \textcolor[rgb]{ 0,  0,  1}{0.998 } & 1.449  & 1.424  & 1.179  & 1.124  & 1.525  & 1.286  & 1.313  & 1.024  & 1.094  & 1.438  & 1.340  & 1.537  & 28.093  & 1.473  & 1.451  & 1.280  \\
          & sMAPE & 9.467  & 11.555  & 11.790  & 11.569  & 9.546  & 11.805  & 198.371  & 12.727  & \textcolor[rgb]{ 1,  0,  0}{\textbf{9.149 }} & 12.495  & 12.050  & 10.455  & 10.326  & 12.364  & 11.394  & 11.742  & \textcolor[rgb]{ 0,  0,  1}{9.363 } & 9.635  & 12.757  & 12.060  & 11.967  & 191.424  & 11.522  & 12.425  & 11.539  \\
          & MASE  & \textcolor[rgb]{ 1,  0,  0}{\textbf{2.593 }} & 3.529  & 4.760  & 3.770  & 2.857  & 4.534  & 98.925  & 4.892  & \textcolor[rgb]{ 0,  0,  1}{2.771 } & 4.262  & 4.254  & 3.379  & 3.111  & 4.723  & 3.688  & 3.732  & 2.848  & 3.155  & 4.122  & 3.789  & 4.911  & 98.015  & 4.690  & 4.295  & 3.609  \\
    \midrule
    \multirow{3}[2]{*}{\begin{sideways}Daily\end{sideways}} & OWA   & \textcolor[rgb]{ 1,  0,  0}{\textbf{0.982 }} & 1.040  & 1.245  & 1.079  & 1.018  & 1.090  & 48.627  & 1.418  & \textcolor[rgb]{ 0,  0,  1}{0.988 } & 1.130  & 1.191  & 1.047  & 1.011  & 1.230  & 1.088  & 1.111  & 1.061  & 0.998  & 1.082  & 1.086  & 1.235  & 29.620  & 1.496  & 1.106  & 1.082  \\
          & sMAPE & \textcolor[rgb]{ 1,  0,  0}{\textbf{3.005 }} & 3.162  & 3.786  & 3.294  & 3.103  & 3.319  & 179.226  & 4.248  & \textcolor[rgb]{ 0,  0,  1}{3.009 } & 3.417  & 3.607  & 3.198  & 3.089  & 3.677  & 3.304  & 3.398  & 3.216  & 3.060  & 3.288  & 3.313  & 3.727  & 99.709  & 4.521  & 3.342  & 3.267  \\
          & MASE  & \textcolor[rgb]{ 1,  0,  0}{\textbf{3.205 }} & 3.418  & 4.089  & 3.530  & 3.332  & 3.577  & 125.892  & 4.722  & \textcolor[rgb]{ 0,  0,  1}{3.237 } & 3.732  & 3.928  & 3.423  & 3.303  & 4.109  & 3.580  & 3.626  & 3.497  & 3.247  & 3.552  & 3.557  & 4.086  & 86.873  & 4.941  & 3.653  & 3.580  \\
    \midrule
    \multirow{3}[2]{*}{\begin{sideways}Hourly\end{sideways}} & OWA   & \textcolor[rgb]{ 0,  0,  1}{0.902 } & 1.372  & 1.526  & 3.171  & 2.625  & 1.040  & 11.691  & 1.623  & 1.704  & \textcolor[rgb]{ 1,  0,  0}{\textbf{0.730 }} & 1.214  & 1.636  & 1.683  & 1.243  & 1.393  & 2.315  & 1.201  & 1.126  & 1.372  & 1.183  & 3.166  & 6.498  & 3.204  & 1.231  & 1.445  \\
          & sMAPE & 18.203  & 19.994  & 24.631  & 34.626  & 29.980  & \textcolor[rgb]{ 0,  0,  1}{17.260 } & 128.419  & 25.407  & 34.523  & \textcolor[rgb]{ 1,  0,  0}{\textbf{14.944 }} & 19.573  & 21.244  & 23.431  & 19.809  & 20.805  & 26.112  & 19.751  & 18.858  & 24.196  & 21.382  & 34.038  & 99.324  & 34.755  & 20.828  & 21.088  \\
          & MASE  & \textcolor[rgb]{ 0,  0,  1}{1.947 } & 3.966  & 4.100  & 10.680  & 8.667  & 2.732  & 39.269  & 4.465  & 3.663  & \textcolor[rgb]{ 1,  0,  0}{\textbf{1.549 }} & 3.266  & 5.067  & 5.009  & 3.372  & 3.964  & 7.685  & 3.178  & 2.937  & 3.420  & 2.881  & 10.730  & 18.188  & 10.821  & 3.183  & 4.172  \\
    \midrule
    \multirow{3}[2]{*}{\begin{sideways}Average\end{sideways}} & OWA   & \textcolor[rgb]{ 1,  0,  0}{\textbf{0.856 }} & 0.884  & 0.984  & 0.907  & 0.965  & 0.922  & 8.856  & 1.273  & 1.007  & 0.903  & 0.969  & 0.918  & 0.915  & 0.894  & 0.897  & 1.005  & 1.209  & 0.942  & 0.906  & \textcolor[rgb]{ 0,  0,  1}{0.875 } & 1.127  & 8.039  & 1.209  & 0.925  & 0.924  \\
          & sMAPE & \textcolor[rgb]{ 1,  0,  0}{\textbf{11.781 }} & 11.985  & 13.025  & 12.199  & 12.848  & 12.511  & 76.147  & 16.392  & 13.509  & 12.120  & 12.838  & 12.268  & 12.351  & 12.023  & 12.137  & 13.478  & 14.635  & 12.708  & 12.219  & \textcolor[rgb]{ 0,  0,  1}{11.925 } & 14.673  & 72.619  & 15.344  & 12.489  & 12.471  \\
          & MASE  & \textcolor[rgb]{ 1,  0,  0}{\textbf{1.551 }} & 1.615  & 1.839  & 1.662  & 1.738  & 1.693  & 18.440  & 2.317  & 1.823  & 1.651  & 1.762  & 1.677  & 1.680  & 1.657  & 1.645  & 1.844  & 2.284  & 1.695  & 1.657  & \textcolor[rgb]{ 0,  0,  1}{1.605 } & 2.042  & 16.805  & 2.136  & 1.674  & 1.673  \\
   \toprule
    \multicolumn{2}{c}{$1^{\text{st}}$ Count}                 & \textcolor{red}{\textbf{14}}      & 0         & 0      & 2                                 & 0        & 0                                 & 0                                 & 0      & 1                                & \underline{\textcolor{blue}{3}}       & 0       & 0       & 0 & 0                                 & 0         & 0      & 0                                 & 0        & 0                                 & 1                                 & 0      & 0                                & 0                                 & 0       & 0     \\
     \bottomrule
    \end{tabular}}%

\end{sidewaystable}%

\label{tab:TSGym_vs_Sota_short_full}

\begin{table}[htbp]
  \centering
  \caption{Effects of different meta feature settings on the long-term forecasting task. All metric values are averaged across different prediction lengths. For more details about meta features, refer to section~\ref{appx:meta-features}.}
    \resizebox{\textwidth}{!}{
        \begin{tabular}{c|rr|rr|rr|rr|rr|rr}
        \toprule
        \multicolumn{1}{c}{\textbf{Models}} & \multicolumn{2}{c|}{\textbf{Whole+Sub+Delta}} & \multicolumn{2}{c|}{\textbf{Sub}} & \multicolumn{2}{c|}{\textbf{Sub+Delta}} & \multicolumn{2}{c|}{\textbf{Delta}} & \multicolumn{2}{c|}{\textbf{Whole(default)}} & \multicolumn{2}{c}{\textbf{DUET}} \\
        \midrule
        \multicolumn{1}{c}{\textbf{Metric}} & \multicolumn{1}{r}{\textbf{mse}} & \multicolumn{1}{r|}{\textbf{mae}} & \multicolumn{1}{r}{\textbf{mse}} & \multicolumn{1}{r|}{\textbf{mae}} & \multicolumn{1}{r}{\textbf{mse}} & \multicolumn{1}{r|}{\textbf{mae}} & \multicolumn{1}{r}{\textbf{mse}} & \multicolumn{1}{r|}{\textbf{mae}} & \multicolumn{1}{r}{\textbf{mse}} & \multicolumn{1}{r|}{\textbf{mae}} & \multicolumn{1}{r}{\textbf{mse}} & \multicolumn{1}{r}{\textbf{mae}} \\
        \midrule
        \textbf{ETTm1} & \secondres{0.357} & \secondres{0.383} & 0.363 & 0.388 & \textcolor{red}{\textbf{0.354}} & \textcolor{red}{\textbf{0.380}} & 0.377 & 0.405 & \secondres{0.357} & \secondres{0.383} & 0.407 & 0.409 \\
        \textbf{ETTm2} & 0.273 & 0.329 & 0.269 & 0.329 & \secondres{0.266} & \secondres{0.324} & 0.380  & 0.389 & \textcolor{red}{\textbf{0.261}} & \textcolor{red}{\textbf{0.319}} & 0.296 & 0.338 \\
        \textbf{ETTh1} & \textcolor{red}{\textbf{0.417}} & \textcolor{red}{\textbf{0.429}} & 0.433 & 0.435 & \secondres{0.418} & \secondres{0.433} & 0.558 & 0.496 & 0.426 & 0.440  & 0.433 & 0.437 \\
        \textbf{ETTh2} & 0.375 & 0.407 & 0.362 & 0.402 & \secondres{0.360} & \textcolor{red}{\textbf{0.398}} & 1.256 & 0.744 & \textcolor{red}{\textbf{0.358}} & \secondres{0.400} & 0.380  & 0.403 \\
        \textbf{ECL} & 0.172 & 0.266 & \secondres{0.171} & 0.269 & 0.176 & 0.270  & 0.182 & 0.278 & \textcolor{red}{\textbf{0.170}} & \secondres{0.265} & 0.179 & \textcolor{red}{\textbf{0.262}} \\
        \textbf{Traffic} & \secondres{0.433} & \secondres{0.309} & 0.437 & 0.313 & \textcolor{red}{\textbf{0.432}} & \textcolor{red}{\textbf{0.308}} & 0.587 & 0.368 & 0.435 & 0.313 & 0.797 & 0.427 \\
        \textbf{Weather} & 0.239 & 0.274 & \textcolor{red}{\textbf{0.228}} & \textcolor{red}{\textbf{0.266}} & 0.233 & 0.270  & 0.263 & 0.310  & \secondres{0.229} & \secondres{0.268} & 0.252 & 0.277 \\
        \textbf{Exchange} & 0.406 & 0.429 & 0.408 & 0.429 & \secondres{0.404} & \secondres{0.428} & 0.761 & 0.622 & 0.410  & 0.431 & \textcolor{red}{\textbf{0.322}} & \textcolor{red}{\textbf{0.384}} \\
        \textbf{ILI} & \secondres{2.401} & 1.030  & 3.099 & 1.195 & 2.855 & 1.141 & 2.814 & 1.125 & \textcolor{red}{\textbf{2.233}} & \textcolor{red}{\textbf{1.015}} & 2.640  & \secondres{1.018} \\
        \bottomrule
        \end{tabular}%
    }
  \label{tab:appx_diverse_mf}%
\end{table}%

\subsection{Additional Results of Large Evaluations on Design Choices}
\label{appx:complete_evaluations}

To systematically evaluate our architectural decisions, we conduct detailed ablation studies focusing on 17 component-level analyses, presented separately in Tables~\ref{tab:TSGym_Components1}--\ref{tab:TSGym_Components23} for clarity and due to space constraints. These comparative experiments assess the performance impact of different design choices for each component across nine datasets in the long-term forecasting task. \textbf{Bolded} values indicate the best-performing configuration for each dataset, while the summary row highlights the most frequently superior design choices, with \textcolor{red}{\textbf{red-bolded}} entries denoting the dominant configurations. This fine-grained analysis offers empirical insights to guide component selection in time-series forecasting systems.

\begin{table}
\centering
\caption{Long-term Forecasting Performance of Different Design Choices – Part I (6 Components). Performance of various configurations for 6 Components across multiple datasets, evaluated using best MSE, median, and IQR. \textbf{Bolded} entries indicate the best-performing hyperparameter for each dataset. The last row shows the number of times each configuration achieved the best result, with \textcolor{red}{\textbf{red-bolded}} values highlighting the most frequently superior design.}
\label{tab:TSGym_Components1}
  \resizebox{1\textwidth}{!}{
\setlength{\tabcolsep}{3pt} 
\begin{tabular}{c|c|cc|cc|cccc|cccc|cc|ccc} 
\toprule
\multicolumn{1}{c}{}                          & \multicolumn{1}{c}{}              & \multicolumn{2}{c|}{\textbf{x\_mark}}                                     & 
\multicolumn{2}{c|}{\parbox{1.8cm}{\centering\textbf{multi-granularity}}}                           & 
\multicolumn{4}{c|}{\parbox{1.8cm}{\textbf{Normalization}}}& 
\multicolumn{4}{c|}{\parbox{2cm}{\textbf{Decomposition}}}& 
\multicolumn{2}{c|}{\parbox{2cm}{\centering\textbf{Channel-independent}}}                          & 
\multicolumn{3}{c}{\textbf{Tokenization}}                                                                                                                                     \\ 
\cmidrule{3-19}
\multicolumn{1}{c}{\textbf{dataset}}         & \multicolumn{1}{c}{\textbf{stat}} & \textbf{False}                                    & \textbf{True}         & \textbf{False}                                    & \textbf{True}         & \textbf{DishTS}        & \textbf{None} & \textbf{RevIN}               & \textbf{Stat}          & \textbf{DFT}           & \textbf{MA}                  & \textbf{MoEMA} & \textbf{None}          & \textbf{False}          & \textbf{True}                                    & \parbox{1.5cm}{\centering\textbf{inverted-encoding}} & 
\parbox{1.5cm}{\centering\textbf{series-encoding}} & 
\parbox{1.5cm}{\centering\textbf{series-patching}} \\
\midrule
\multirow{4}{*}{\textbf{ETTm1}}    & \textbf{Best}                     & 0.352                                             & \textbf{0.35}         & \textbf{0.349}                                    & 0.352                 & 0.36                   & 0.362         & \textbf{0.351}               & 0.353                  & \textbf{0.351}         & 0.352                        & 0.354          & 0.354                  & 0.354                   & \textbf{0.35}                                    & 0.354                  & 0.352          & \textbf{0.351}                                    \\
\bottomrule
\end{tabular}}

\end{table}

\begin{table}[htbp]
  \centering
  \caption{Long-term Forecasting Performance of Different Design Choices  – Part II (4 Components) and Part II (7 Components). Same structure and evaluation metrics as Table \ref{tab:TSGym_Components1}.}
  \label{tab:TSGym_Components23}
  
\begin{subtable}[t]{\textwidth}
\centering
\caption{Part II – 4 Components (Backbone, Attention, etc.)}
\label{tab:TSGym_Components2}
  \resizebox{1\textwidth}{!}{
\setlength{\tabcolsep}{3pt} 
\begin{tabular}{c|c|ccc|cccccc|cccc|cccc} 
\toprule
\multicolumn{1}{c}{}                         & \multicolumn{1}{c}{}              & \multicolumn{3}{c|}{\textbf{Backbone}}                               & \multicolumn{6}{c|}{\parbox{2cm}{\textbf{Attention}}}                                                                                                                                    & \multicolumn{4}{c|}{\textbf{Feature-Attention}}                                                                            & \multicolumn{4}{c}{\textbf{Sequence Length}}                                     \\ 
\cmidrule{3-19}
\multicolumn{1}{c}{\textbf{dataset}}         & \multicolumn{1}{c}{\textbf{stat}} & \textbf{GRU}   & \textbf{MLP}                 & \parbox{1.5cm}{\textbf{Trans\-former}} & 
\parbox{1.8cm}{\centering\textbf{auto-cor\-relation}} & 
\parbox{1.5cm}{\centering\textbf{de\-stationary-attention}} & 
\parbox{1.8cm}{\centering\textbf{frequency-enhanced-attention}} & \textbf{null}  & 
\parbox{1.5cm}{\centering\textbf{self-attention}} & 
\parbox{1.5cm}{\centering\textbf{sparse-attention}} & 
\parbox{1.8cm}{\centering\textbf{frequency-enhanced-attention}} & \textbf{null}                & 
\parbox{1.5cm}{\centering\textbf{self-attention}} & 
\parbox{1.5cm}{\centering\textbf{sparse-attention}} & \textbf{192}   & \textbf{48}    & \textbf{512}                 & \textbf{96}     \\ 
\midrule
\multirow{4}{*}{\textbf{ETTm1}}    & \textbf{Best}                     & 0.352          & 0.352                        & \textbf{0.351}       & 0.359                     & 0.382                           & 0.354                                 & \textbf{0.35}  & 0.359                   & 0.354                     & 0.356                                 & \textbf{0.35}                & 0.355                   & 0.36                      & 0.352          & 0.476          & \textbf{0.349}               & 0.38            \\
\bottomrule
\end{tabular}}
\end{subtable}

\begin{subtable}[t]{\textwidth}
\centering
\caption{Part III – 7 Components (d\_model, d\_ff, etc.)}
\label{tab:TSGym_Components3}
  \resizebox{1\textwidth}{!}{
\setlength{\tabcolsep}{3pt} 
\begin{tabular}{c|c|cc|cc|cc|ccc|ccc|cc|cc} 
\toprule
\multicolumn{1}{c}{}                         & \multicolumn{1}{c}{}              & \multicolumn{2}{c|}{\textbf{d\_model}}                             & \multicolumn{2}{c|}{\textbf{d\_ff}}                         & 
\multicolumn{2}{c|}{\parbox{1.8cm}{\centering\textbf{Encoder layers}}}  & \multicolumn{3}{c|}{\textbf{Training Epochs}}                  & \multicolumn{3}{c|}{\textbf{Loss Function}}                    & \multicolumn{2}{c|}{\textbf{Learning Rate}}   & 
\multicolumn{2}{c}{\parbox{2cm}{\centering\textbf{Learning Rate Strategy}}}  \\ 
\cmidrule{3-18}
\multicolumn{1}{c}{\textbf{dataset}}         & \multicolumn{1}{c}{\textbf{stat}} & \textbf{256}                        & \textbf{64}                  & \textbf{1024}                & \textbf{256}                 & \textbf{2}     & \textbf{3}                   & \textbf{10}    & \textbf{20}    & \textbf{50}                  & \textbf{HUBER} & \textbf{MAE}                 & \textbf{MSE}   & \textbf{0.0001}              & \textbf{0.001} & \textbf{null}                & \textbf{type}          \\ 
\midrule
\multirow{4}{*}{\textbf{ETTm1}}    & \textbf{Best}                     & 0.352                               & \textbf{0.35}                & 0.352                        & \textbf{0.35}                & 0.352          & \textbf{0.35}                & 0.352          & \textbf{0.351} & 0.353                        & 0.359          & 0.356                        & \textbf{0.35}  & \textbf{0.35}                & 0.351          & 0.355                        & \textbf{0.35}          \\
\bottomrule
\end{tabular}}
\end{subtable}

\end{table}

\clearpage
\subsubsection{Design Choices Evaluation Results for Long-term Forecasting Using MSE as the Metric}

\begin{figure}[t!]
     \centering

    \begin{subfigure}[t]{0.28\textwidth}
         \centering
         \includegraphics[width=\textwidth]{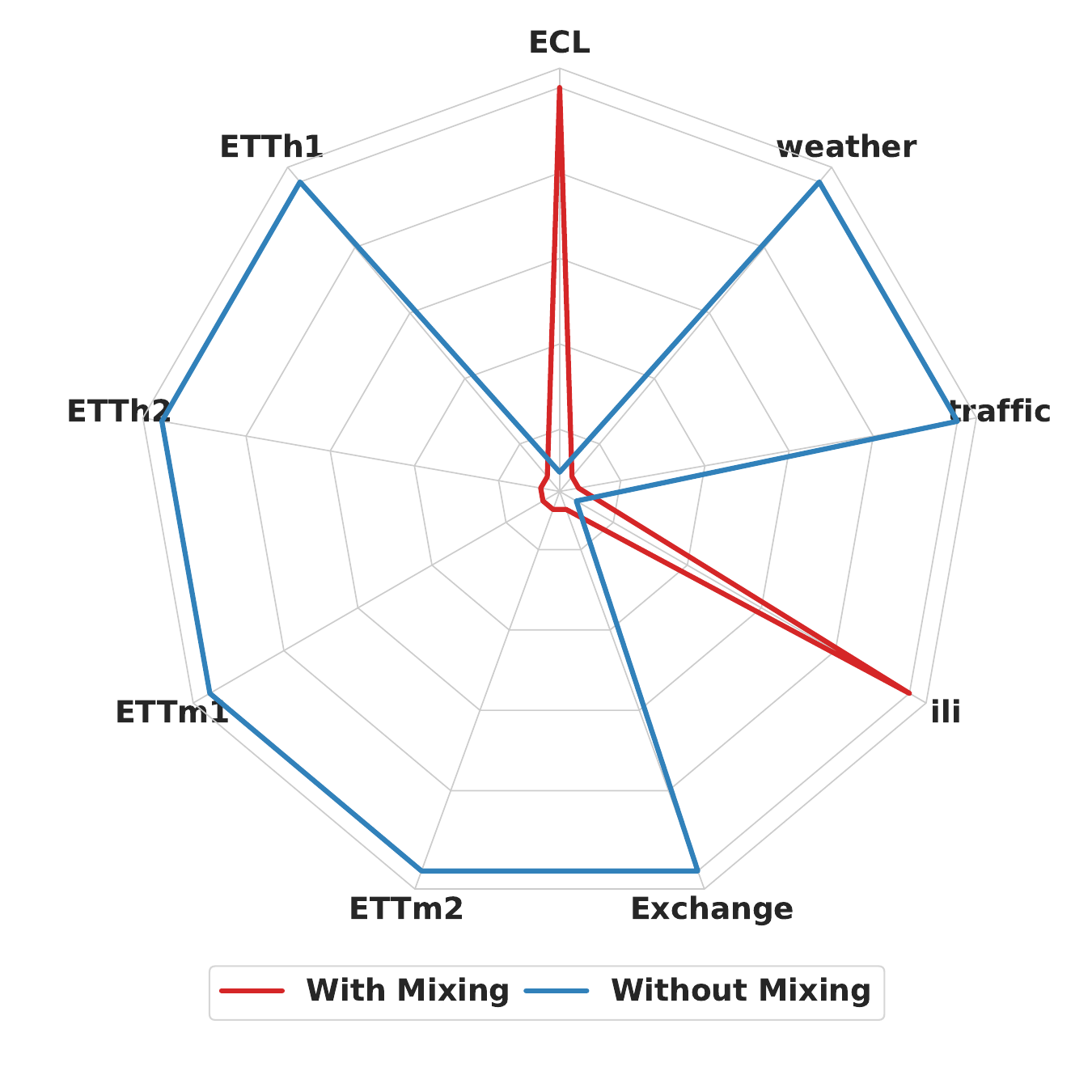}
         \caption{Series Sampling/Mixing}
         \label{fig:exp-appx-mixing-ltf}
     \end{subfigure}
     \hspace{10pt}
    \begin{subfigure}[t]{0.28\textwidth}
         \centering
         \includegraphics[width=\textwidth]{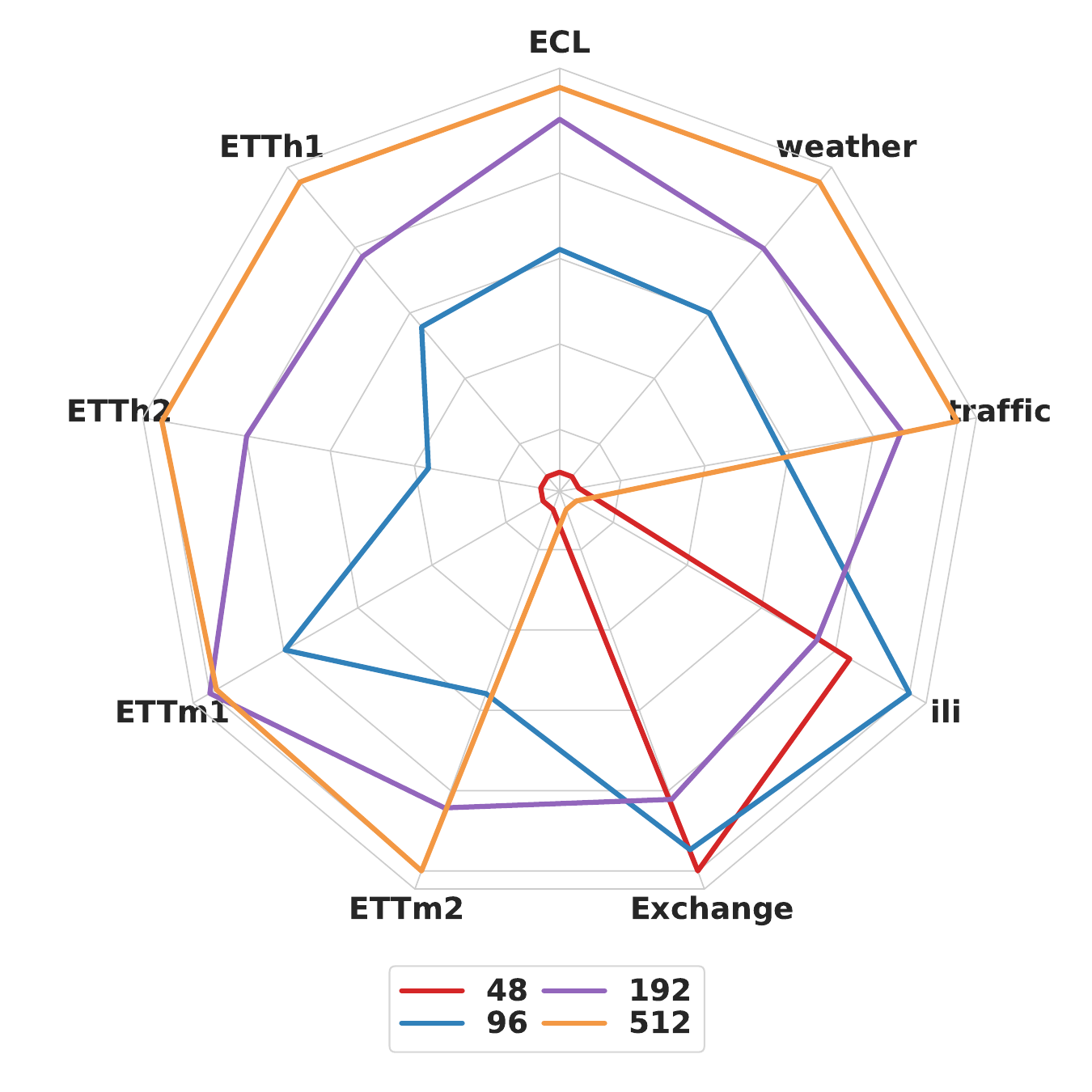}
         \caption{Sequence Length}
         \label{fig:exp-appx-sl-ltf}
     \end{subfigure}
     \hspace{10pt}
    \begin{subfigure}[t]{0.28\textwidth}
         \centering
         \includegraphics[width=\textwidth]{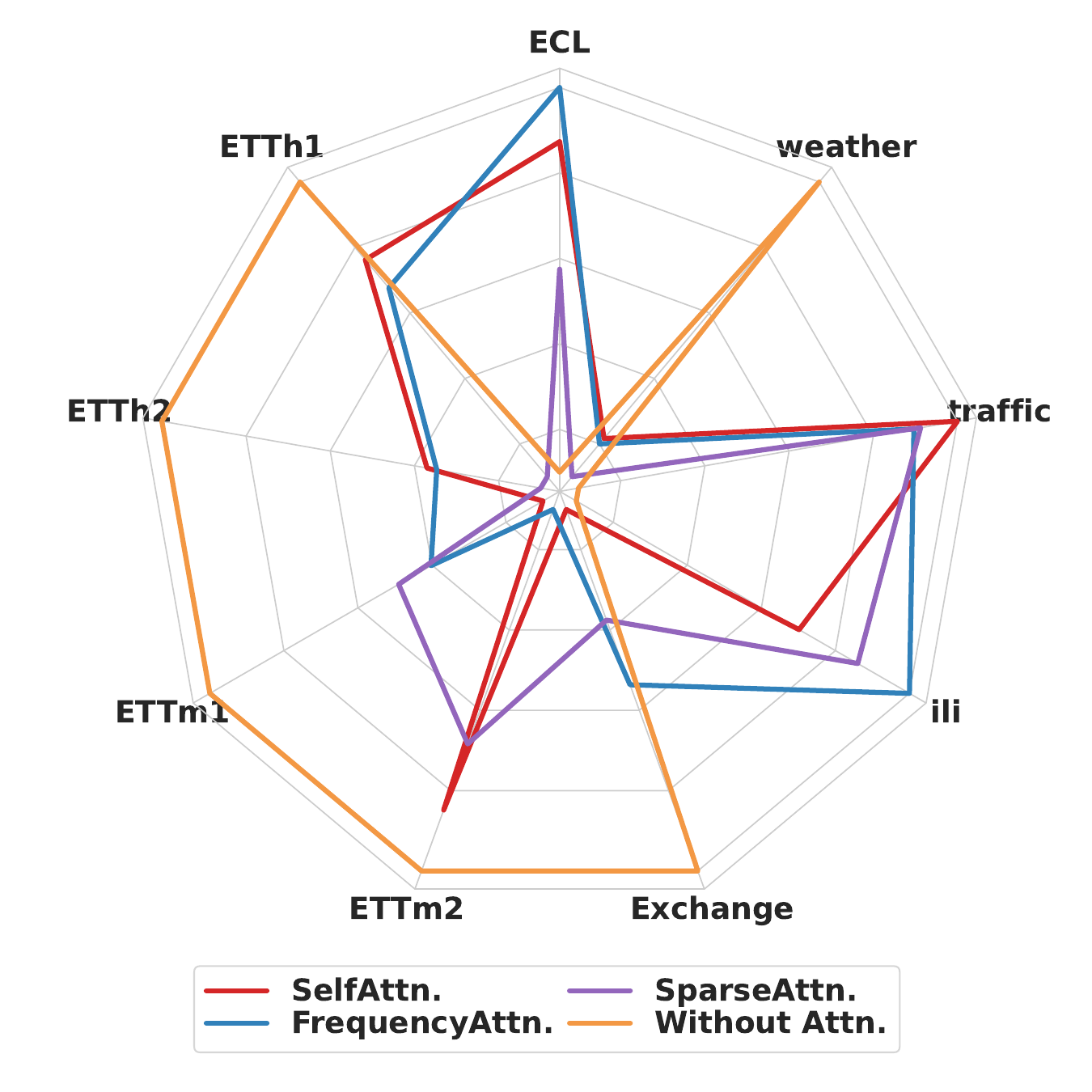}
         \caption{Feature Attention}
         \label{fig:exp-appx-featureattention-ltf}
     \end{subfigure}
     \hspace{10pt}
    \begin{subfigure}[t]{0.28\textwidth}
         \centering
         \includegraphics[width=\textwidth]{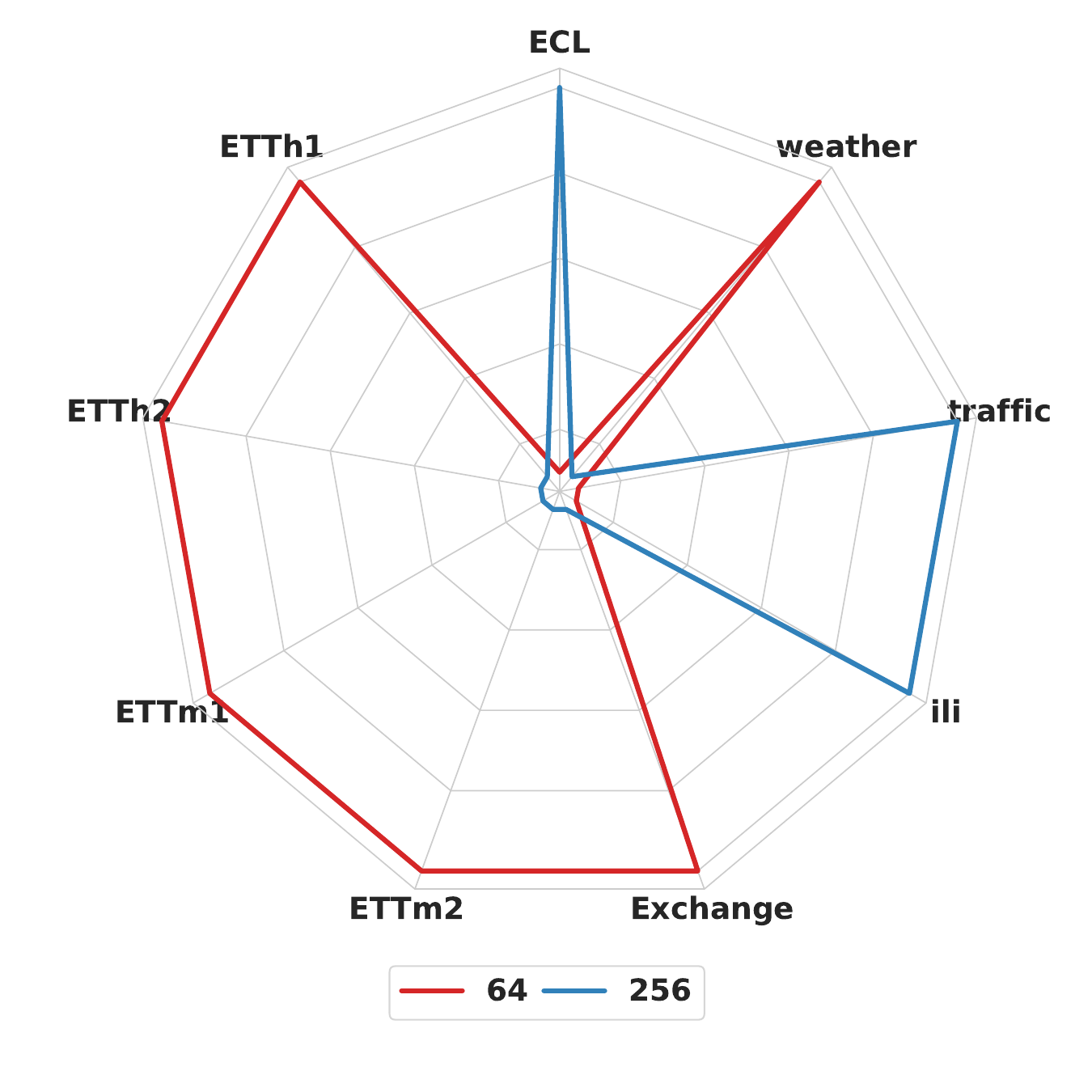}
         \caption{\textit{Hidden Layer Dimensions}}
         \label{fig:exp-appx-dmodel-ltf}
     \end{subfigure}
     \hspace{10pt}
    \begin{subfigure}[t]{0.28\textwidth}
         \centering
         \includegraphics[width=\textwidth]{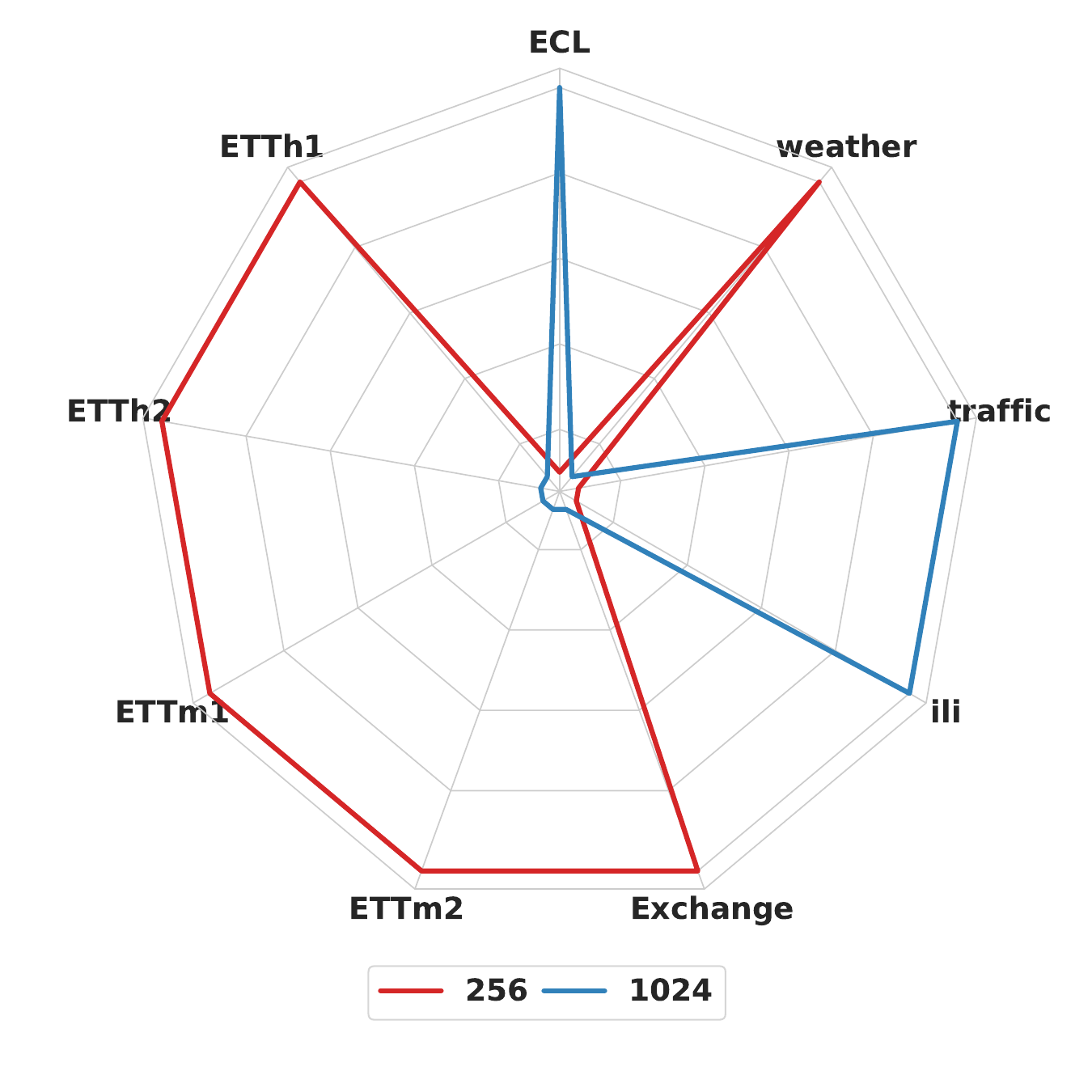}
         \caption{\textit{FCN Layer Dimensions}}
         \label{fig:exp-appx-dff-ltf}
     \end{subfigure}
     \hspace{10pt}
    \begin{subfigure}[t]{0.28\textwidth}
         \centering
         \includegraphics[width=\textwidth]{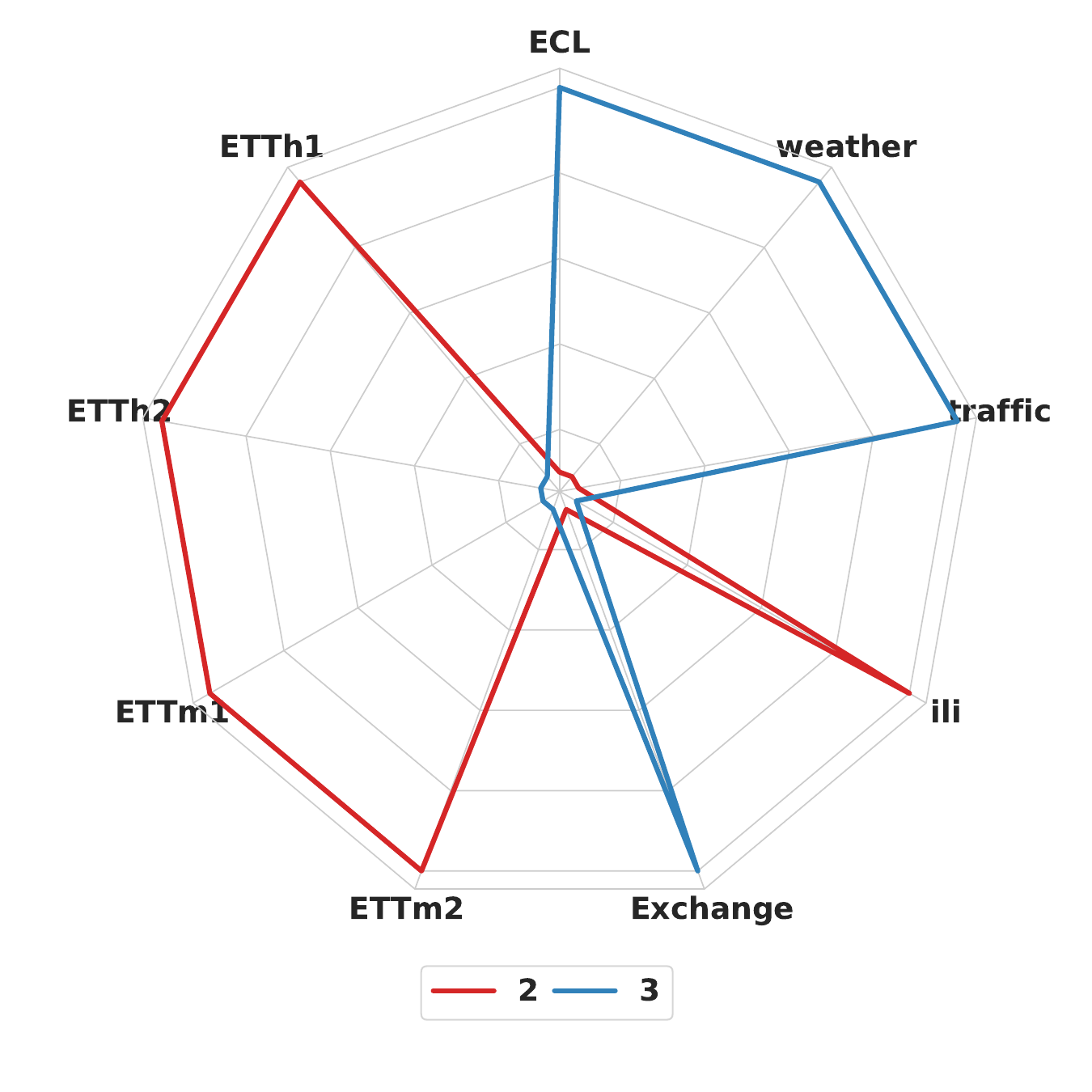}
         \caption{Encoder layers}
         \label{fig:exp-appx-el-ltf}
     \end{subfigure}
     \hspace{10pt}
    \begin{subfigure}[t]{0.28\textwidth}
         \centering
         \includegraphics[width=\textwidth]{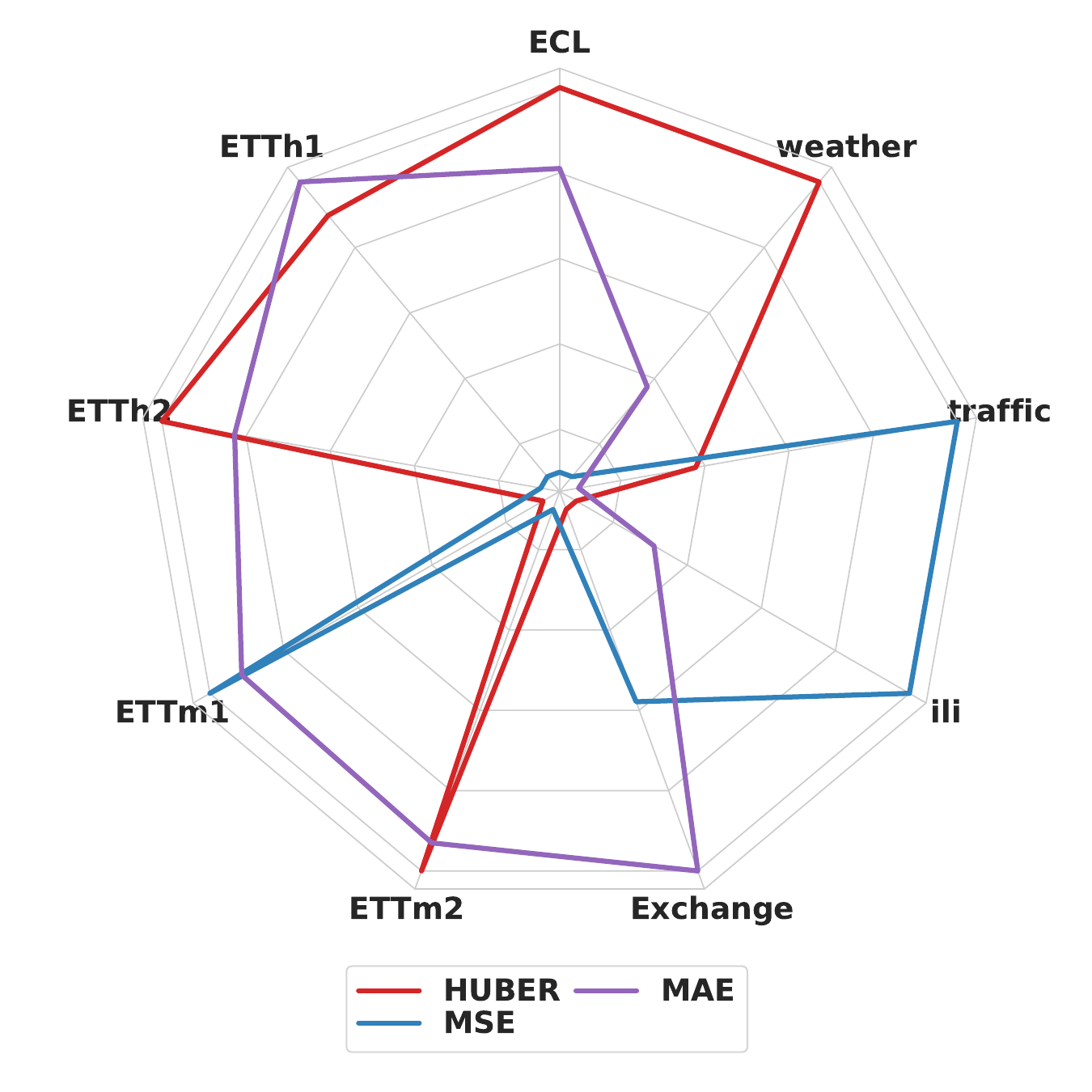}
         \caption{Loss Function}
         \label{fig:exp-appx-lr-loss}
     \end{subfigure}
     \hspace{10pt}
    \begin{subfigure}[t]{0.28\textwidth}
         \centering
         \includegraphics[width=\textwidth]{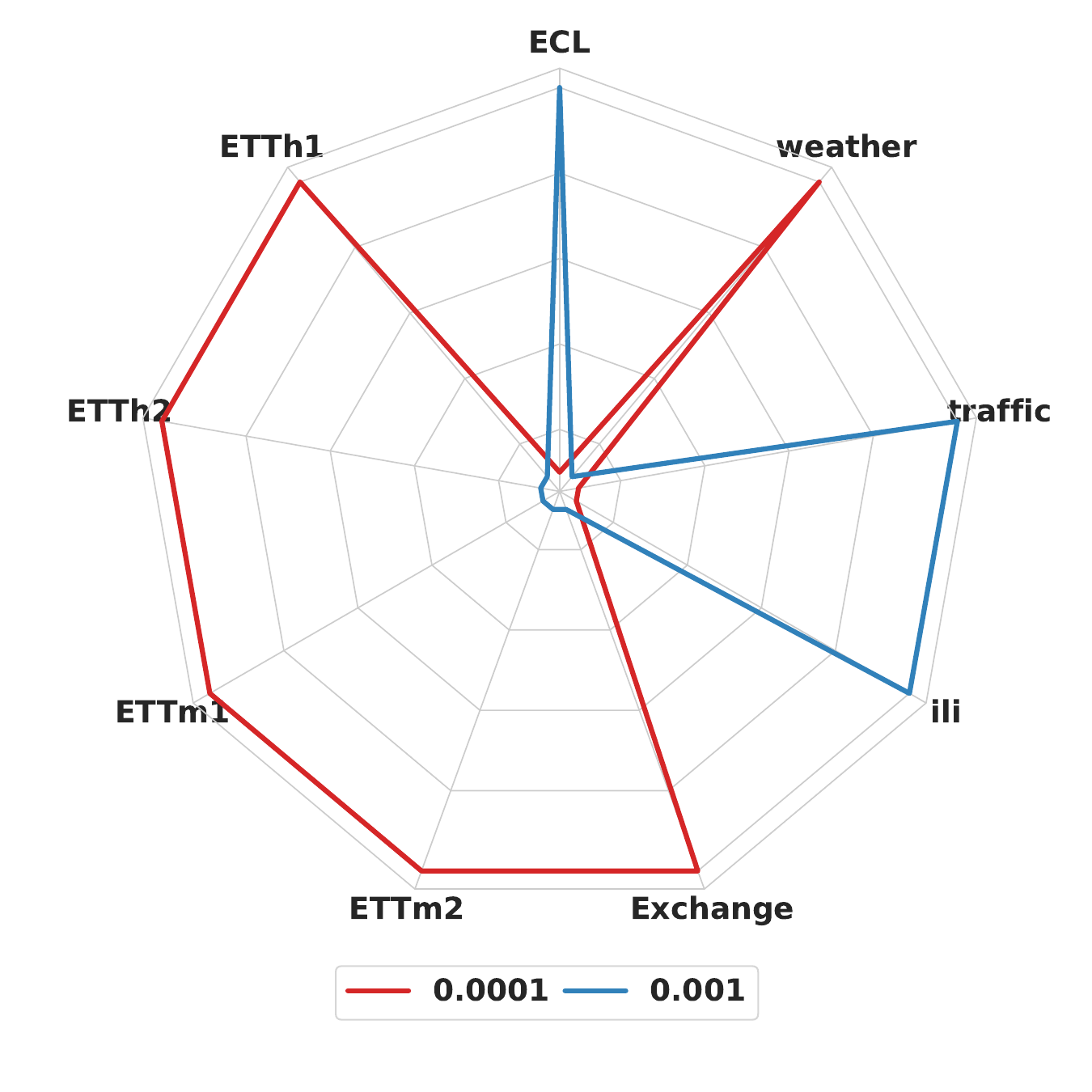}
         \caption{Learning Rate}
         \label{fig:exp-appx-lr-ltf}
     \end{subfigure}
     \hspace{10pt}
    \begin{subfigure}[t]{0.28\textwidth}
         \centering
         \includegraphics[width=\textwidth]{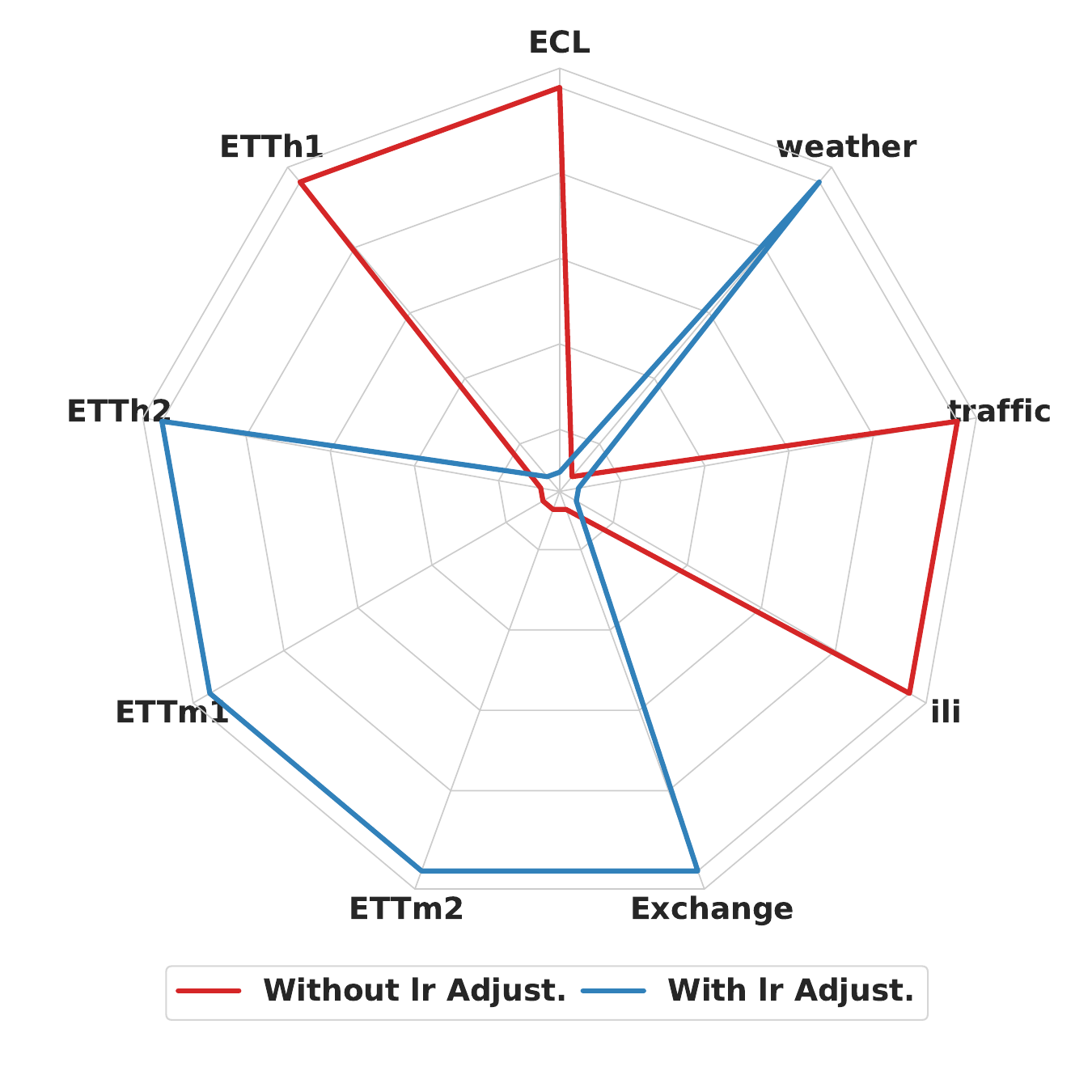}
         \caption{Learning Rate Strategy}
         \label{fig:exp-appx-lrs-ltf}
     \end{subfigure}
     \hspace{10pt}
     \caption{Overall performance across additional design dimensions in long-term forecasting. The results (MSE) are based on the top 25th percentile across all forecasting horizons.}
     \vspace{-0.1in}
     \label{fig:exp-appx-rada-ltf}
\end{figure}

\textbf{Spider Chart Analysis}.
Fig.~\ref{fig:exp-appx-rada-ltf} extends the baseline comparisons presented in Fig.~\ref{fig:exp-rada} by employing multi-dimensional spider charts, where each vertex corresponds to a benchmark dataset. Closer proximity to the outer edge of a vertex indicates better performance of the associated design choice on that particular dataset. These visual representations offer an intuitive understanding of how different architectural decisions influence model effectiveness across diverse forecasting domains. Notably, configurations for components including Series Sampling/Mixing (Fig. \ref{fig:exp-appx-mixing-ltf}), Hidden Layer Dimensions (Fig. \ref{fig:exp-appx-dmodel-ltf}), FCN Layer Dimensions (Fig. \ref{fig:exp-appx-dff-ltf}), Learning Rate (Fig. \ref{fig:exp-appx-lr-ltf}), and Learning Rate Strategy (Fig. \ref{fig:exp-appx-lrs-ltf}) demonstrate similar spatial patterns in the radar charts. Specifically, ECL, ILI, and Traffic datasets exhibit consistent parameter preferences across these components, suggesting intrinsic alignment between their temporal patterns and specific architectural configurations.

\begin{figure}[t!]
     \centering
     \begin{subfigure}[t]{0.28\textwidth}
         \centering
         \includegraphics[width=\textwidth]{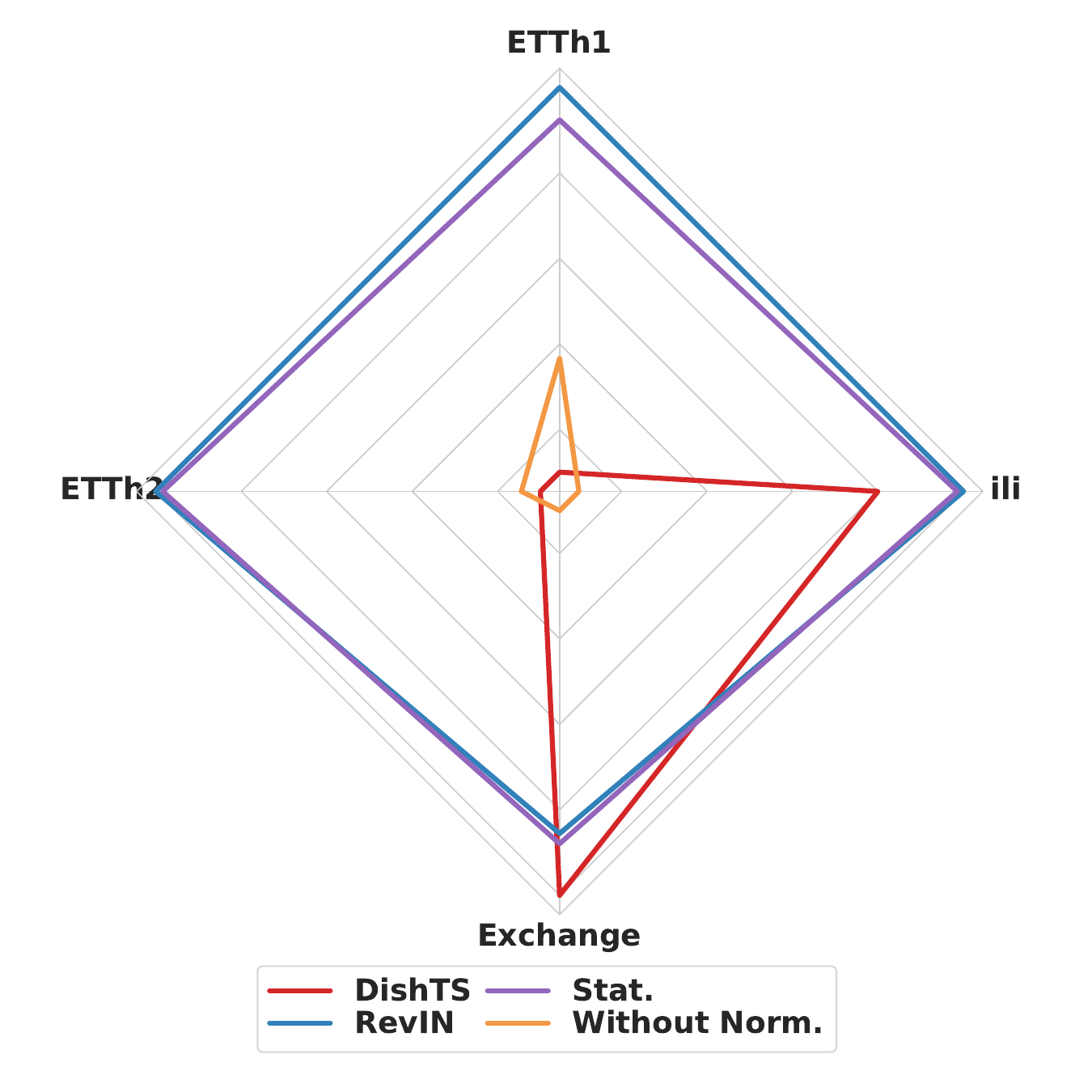}
         \caption{Series Normalization}
         \label{fig:exp-appx-Normalization-llm}
     \end{subfigure}
     \hspace{10pt}
     \begin{subfigure}[t]{0.28\textwidth}
         \centering
         \includegraphics[width=\textwidth]{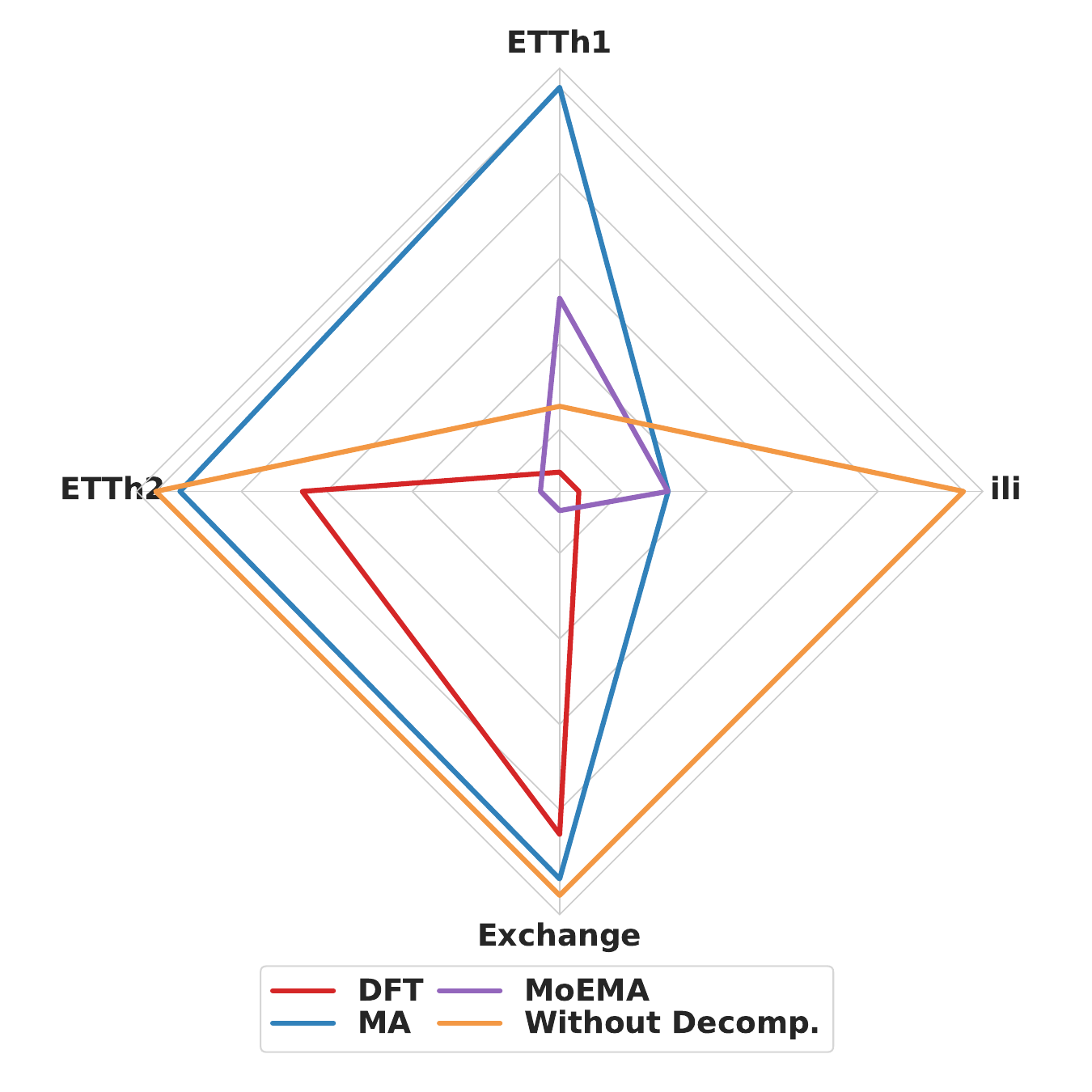}
         \caption{Series Decomposition}
         \label{fig:exp-appx-Decomposition-llm}
     \end{subfigure}
     \hspace{10pt}
    \begin{subfigure}[t]{0.28\textwidth}
         \centering
         \includegraphics[width=\textwidth]{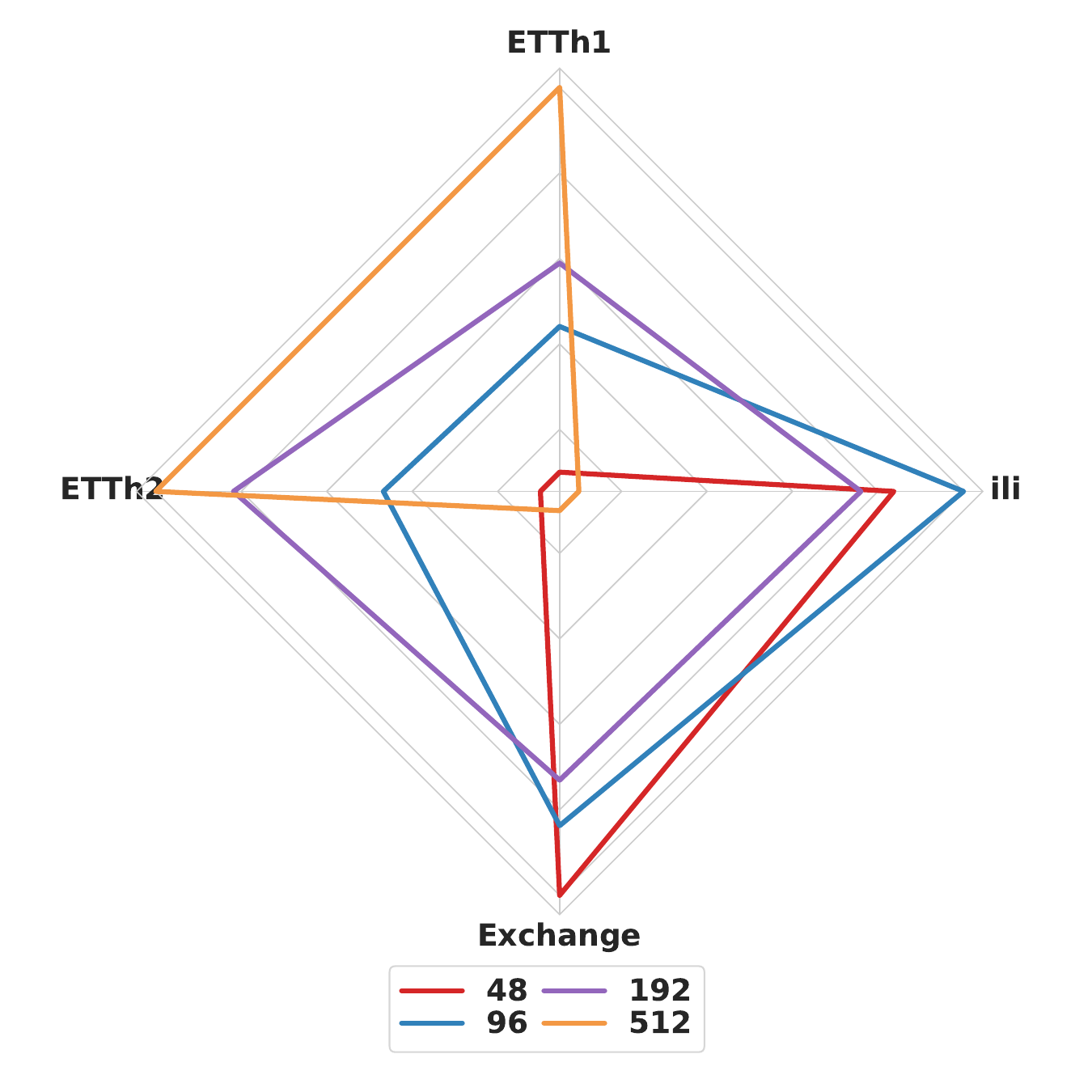}
         \caption{Sequence Length}
         \label{fig:exp-appx-sl-llm}
     \end{subfigure}
     \hspace{10pt}
    \begin{subfigure}[t]{0.28\textwidth}
         \centering
         \includegraphics[width=\textwidth]{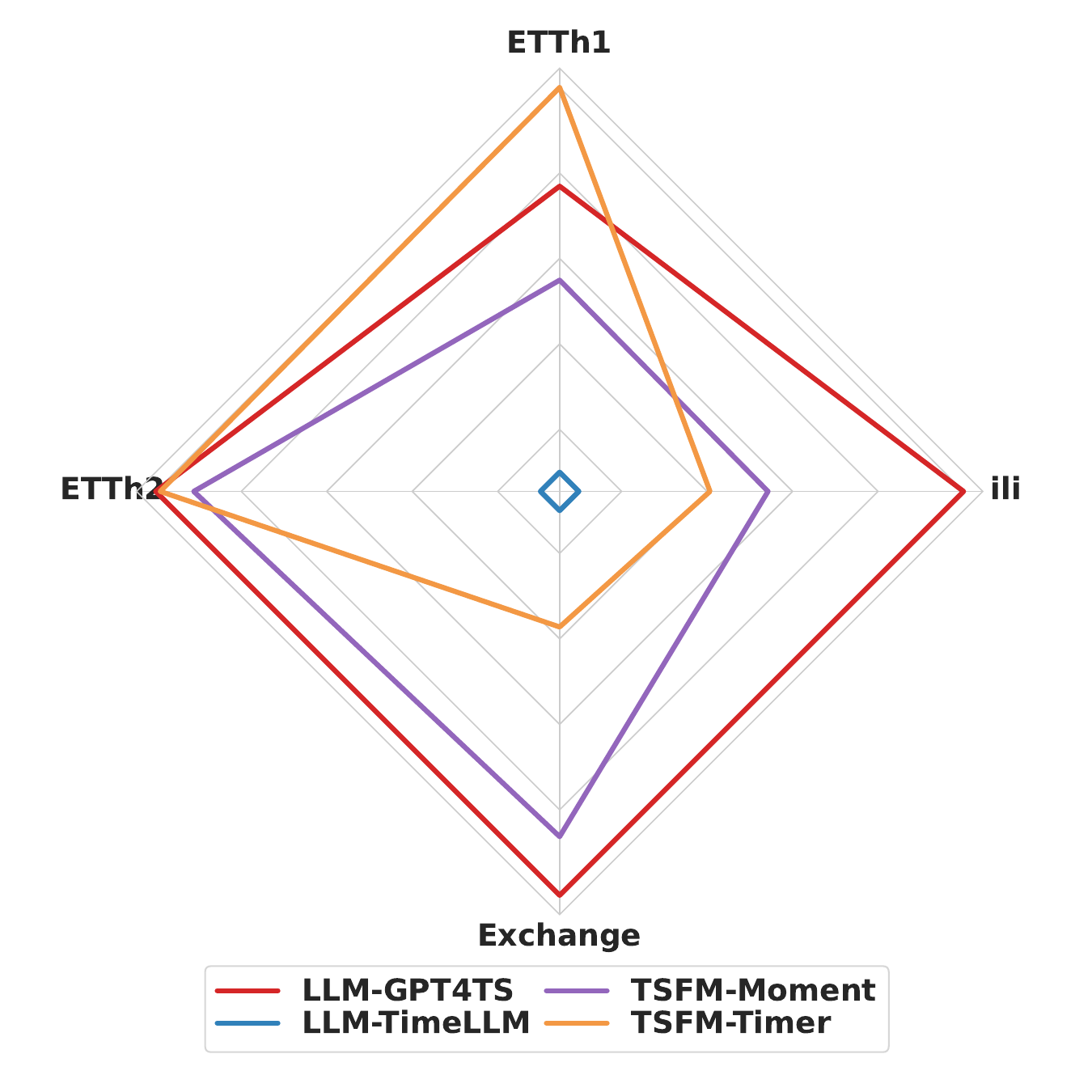}
         \caption{Network Backbone}
         \label{fig:exp-appx-backbone-llm}
     \end{subfigure}
     \hspace{10pt}
    \begin{subfigure}[t]{0.28\textwidth}
         \centering
         \includegraphics[width=\textwidth]{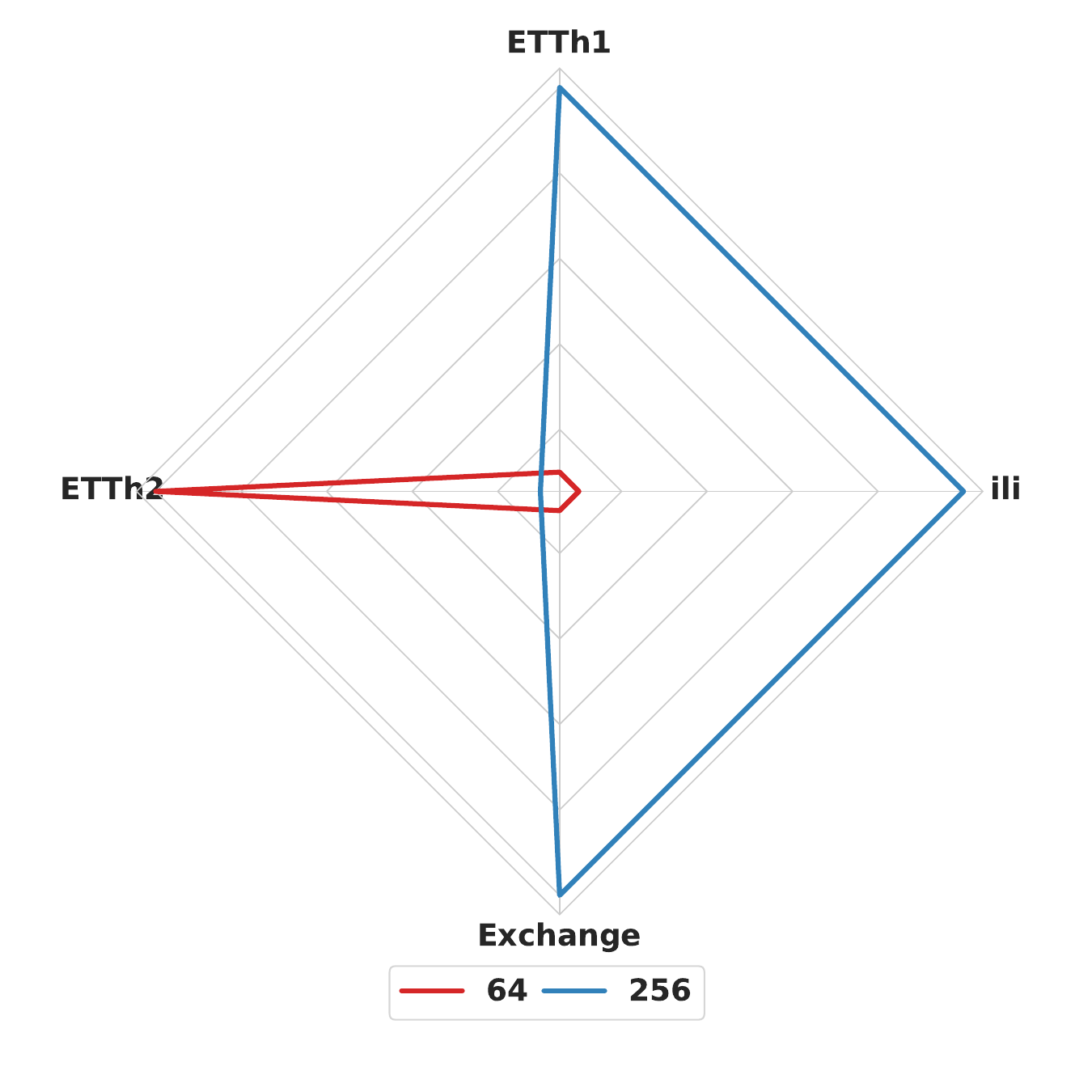}
         \caption{\textit{Hidden Layer Dimensions}}
         \label{fig:exp-appx-dmodel-llm}
     \end{subfigure}
     \hspace{10pt}
    \begin{subfigure}[t]{0.28\textwidth}
         \centering
         \includegraphics[width=\textwidth]{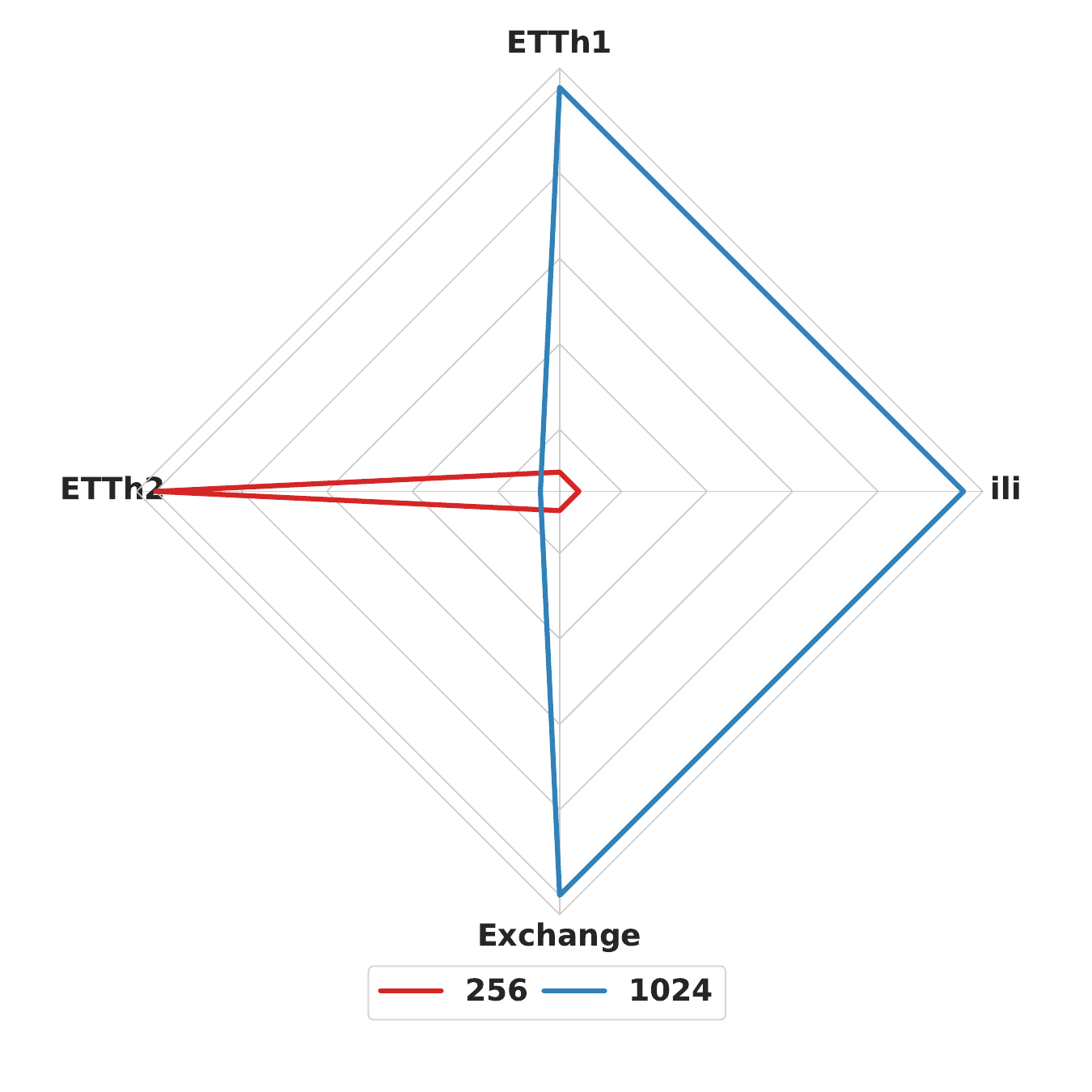}
         \caption{\textit{FCN Layer Dimensions}}
         \label{fig:exp-appx-dff-llm}
     \end{subfigure}
     \hspace{10pt}
    \begin{subfigure}[t]{0.28\textwidth}
         \centering
         \includegraphics[width=\textwidth]{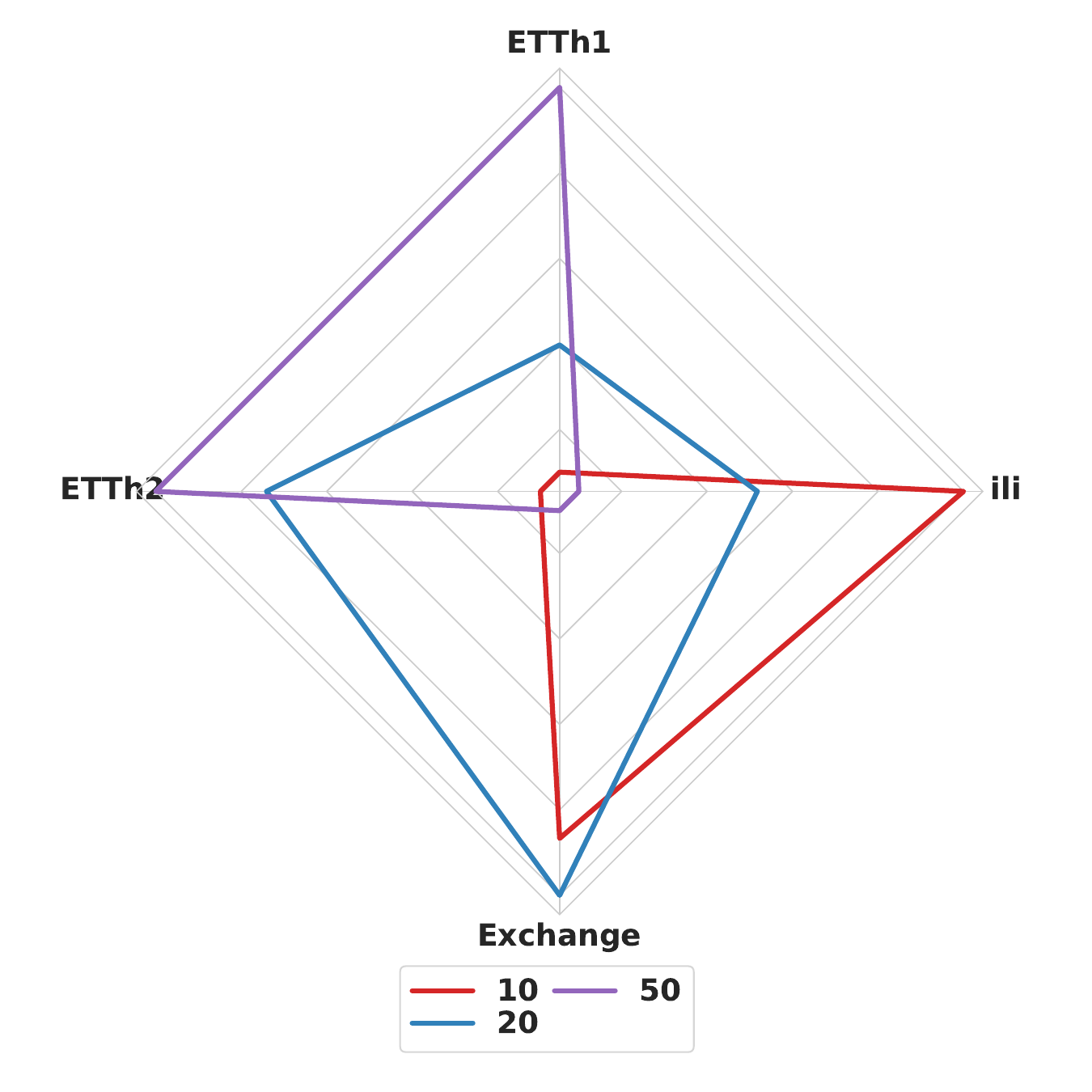}
         \caption{Epochs}
         \label{fig:exp-appx-epochs-llm}
     \end{subfigure}
     \hspace{10pt}
    \begin{subfigure}[t]{0.28\textwidth}
         \centering
         \includegraphics[width=\textwidth]{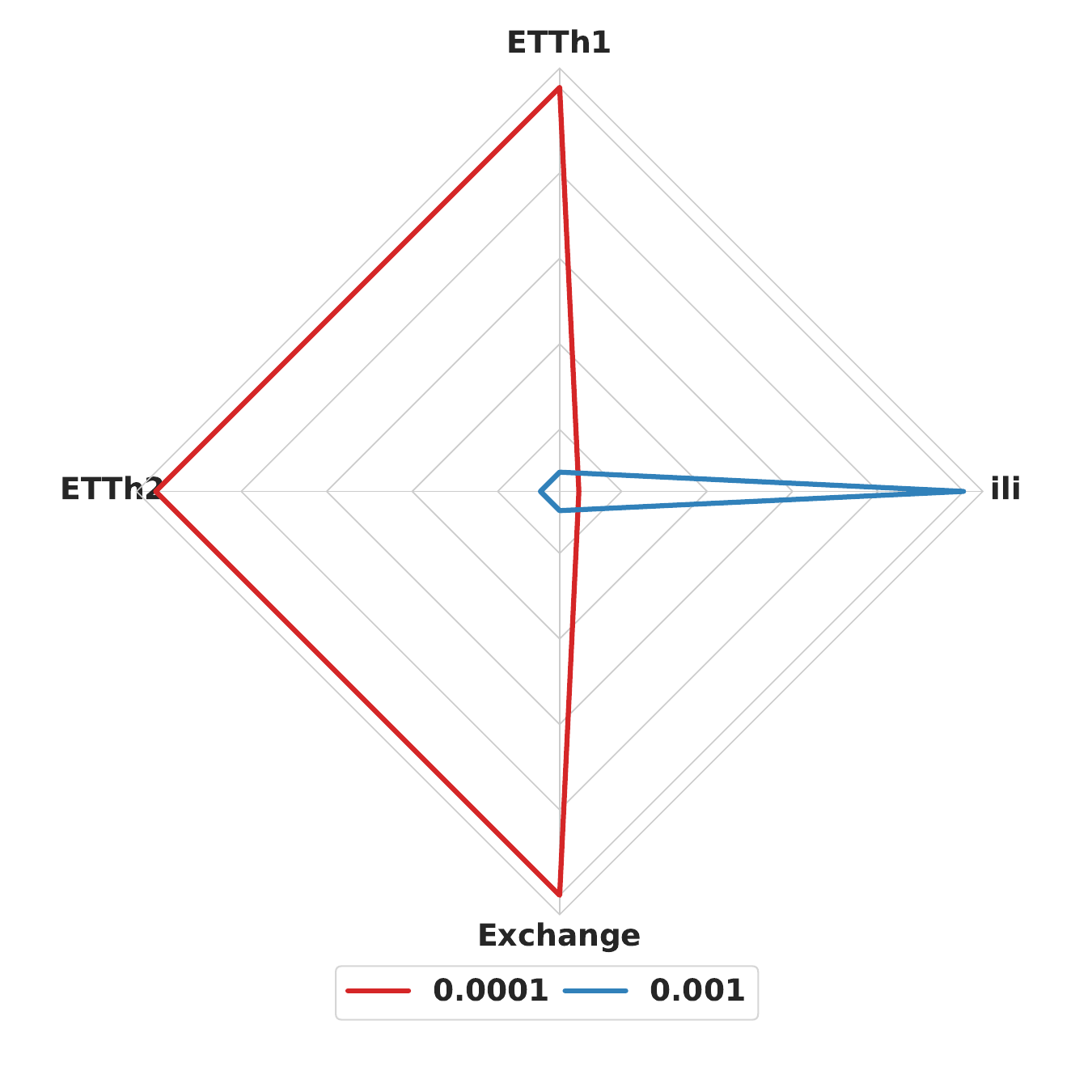}
         \caption{Learning Rate}
         \label{fig:exp-appx-lr-llm}
     \end{subfigure}
     \hspace{10pt}
    \begin{subfigure}[t]{0.28\textwidth}
         \centering
         \includegraphics[width=\textwidth]{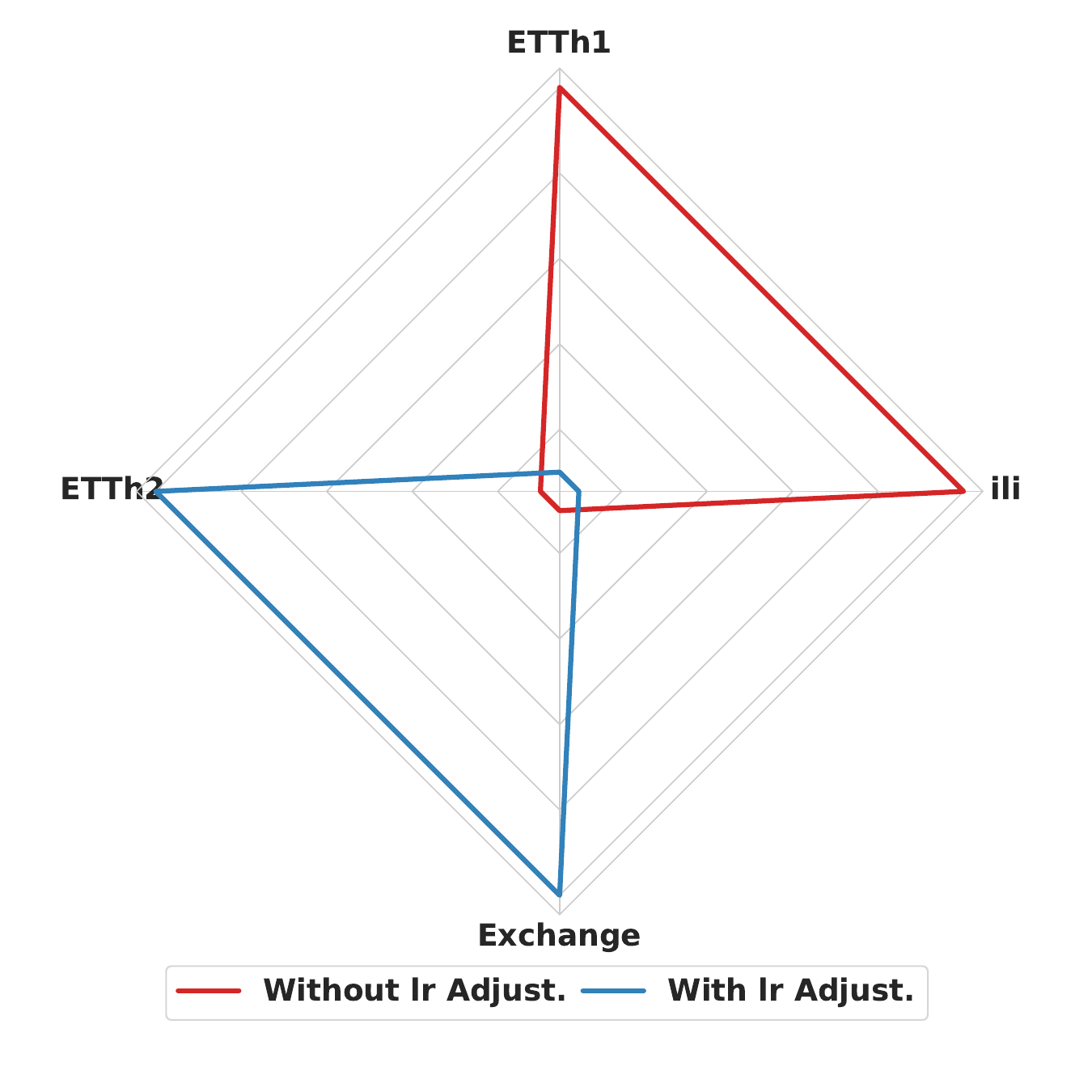}
         \caption{Learning Rate Strategy}
         \label{fig:exp-appx-lrs-llm}
     \end{subfigure}
     \hspace{10pt}
     \caption{Overall performance across all design dimensions when using LLMs or TSFMs in long-term forecasting. The results (MSE) are based on the top 25th percentile across all forecasting horizons.}
     \vspace{-0.1in}
     \label{fig:exp-appx-rada-llm_ltf}
\end{figure}

In addition, Fig.~\ref{fig:exp-appx-rada-llm_ltf} provides a broader evaluation of large-scale time series models, revealing that conventional architectures still maintain a competitive advantage over LLM-based models, especially in domain-specific forecasting tasks where structural inductive biases play a crucial role.

\begin{figure}[t!]
     \centering
     \begin{subfigure}[t]{0.28\textwidth}
         \centering
         \includegraphics[width=\textwidth]{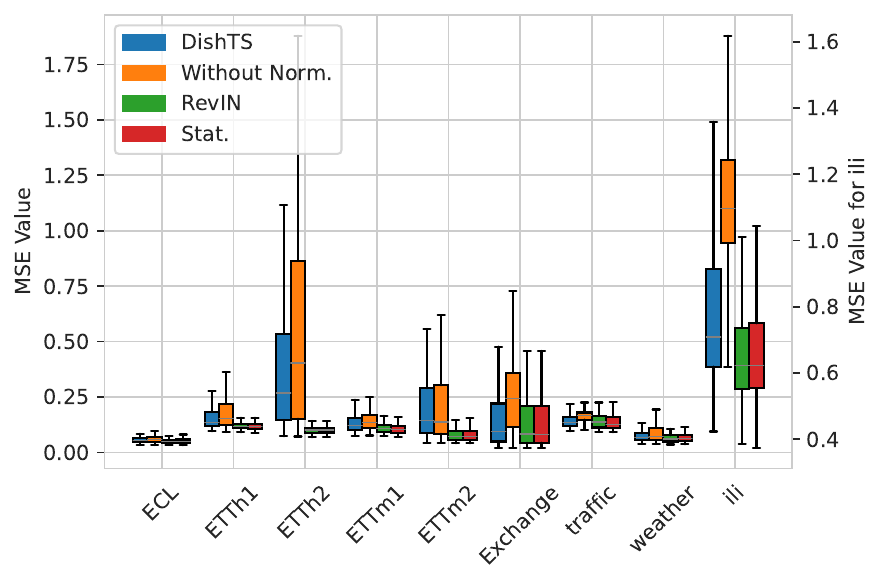}
         \caption{Series Normalization}
         \label{fig:exp-appx-Normalization-ltf_bp}
     \end{subfigure}
     \hspace{10pt}
     \begin{subfigure}[t]{0.28\textwidth}
         \centering
         \includegraphics[width=\textwidth]{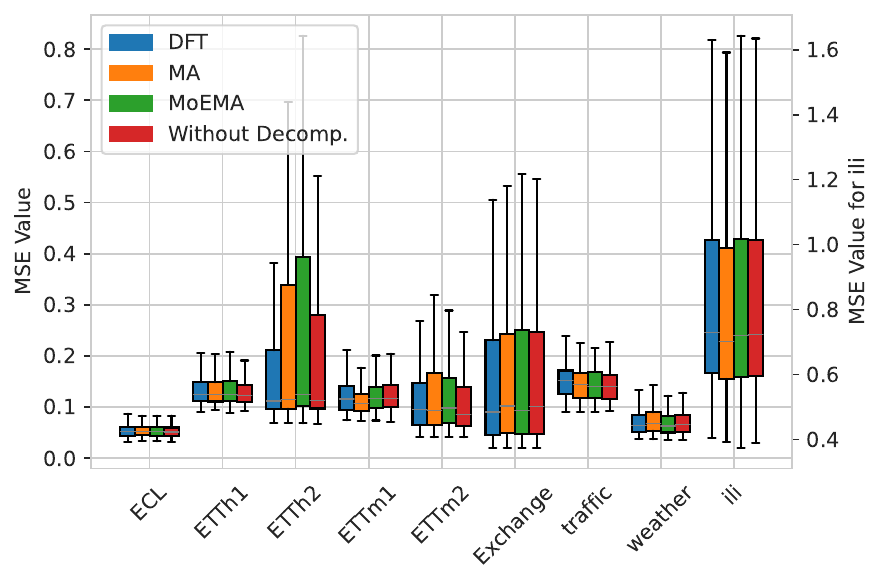}
         \caption{Series Decomposition}
         \label{fig:exp-appx-Decomposition-ltf_bp}
     \end{subfigure}
     \hspace{10pt}
    \begin{subfigure}[t]{0.28\textwidth}
         \centering
         \includegraphics[width=\textwidth]{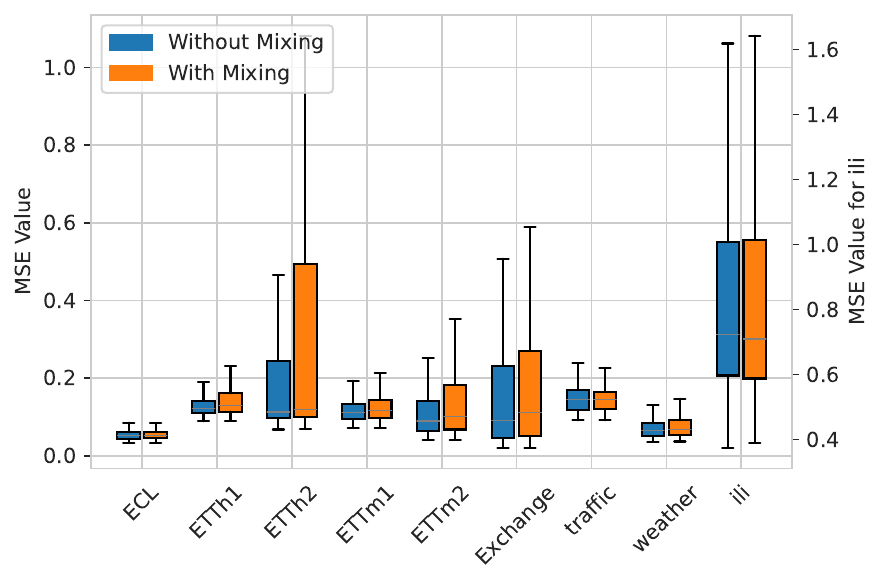}
         \caption{Series Sampling/Mixing}
         \label{fig:exp-appx-mixing-ltf_bp}
     \end{subfigure}
     \hspace{10pt}
    \begin{subfigure}[t]{0.28\textwidth}
         \centering
         \includegraphics[width=\textwidth]{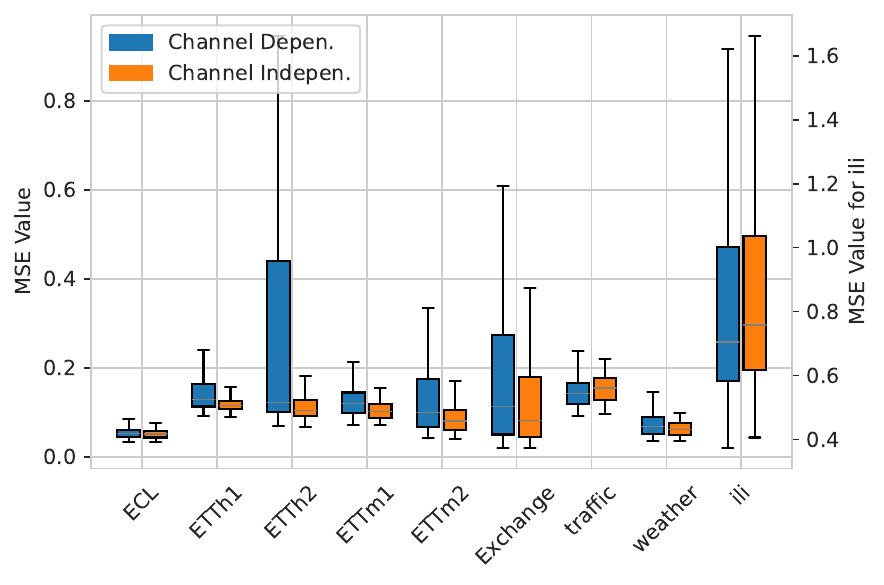}
         \caption{Channel Independent}
         \label{fig:exp-appx-CI-ltf_bp}
     \end{subfigure}
     \hspace{10pt}
    \begin{subfigure}[t]{0.28\textwidth}
         \centering
         \includegraphics[width=\textwidth]{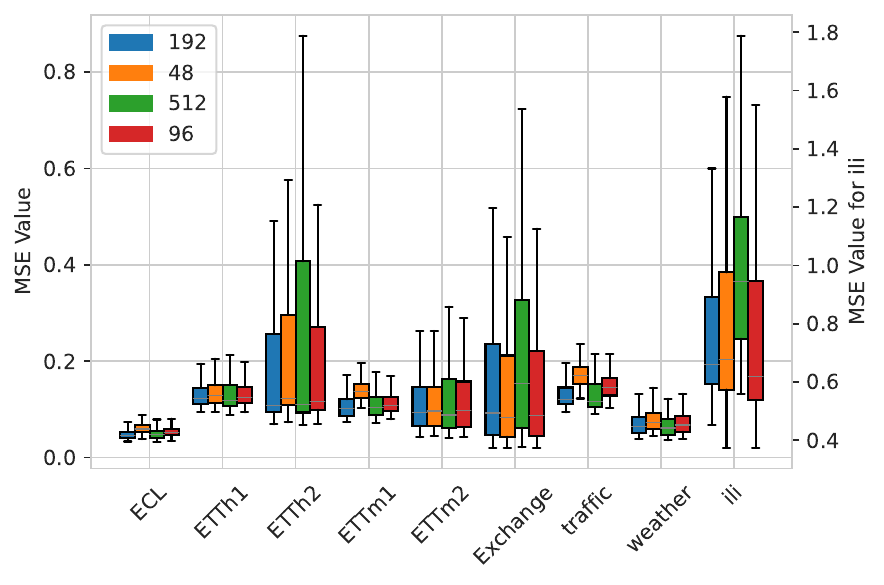}
         \caption{Sequence Length}
         \label{fig:exp-appx-sl-ltf_bp}
     \end{subfigure}
     \hspace{10pt}
    \begin{subfigure}[t]{0.28\textwidth}
         \centering
         \includegraphics[width=\textwidth]{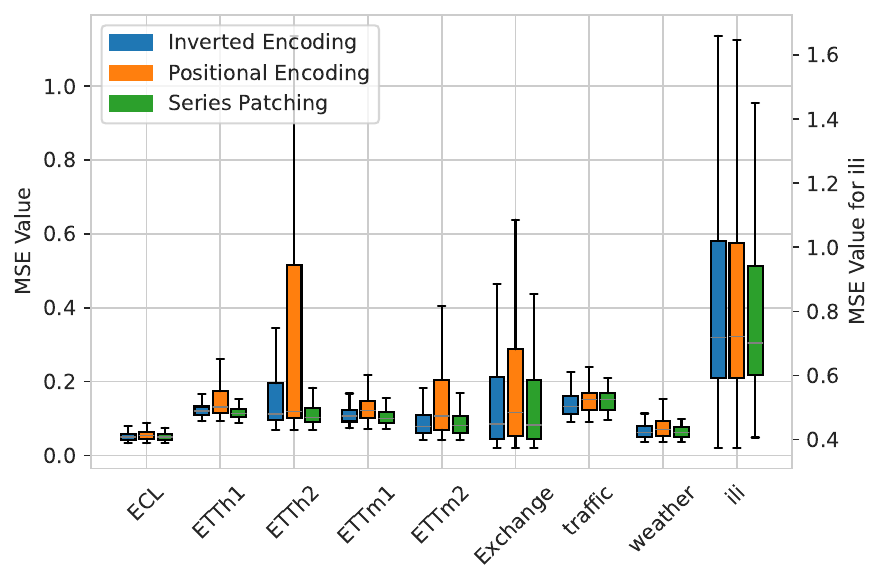}
         \caption{Series Embedding}
         \label{fig:exp-appx-tokenization-ltf_bp}
     \end{subfigure}
     \hspace{10pt}
     \begin{subfigure}[t]{0.28\textwidth}
         \centering
         \includegraphics[width=\textwidth]{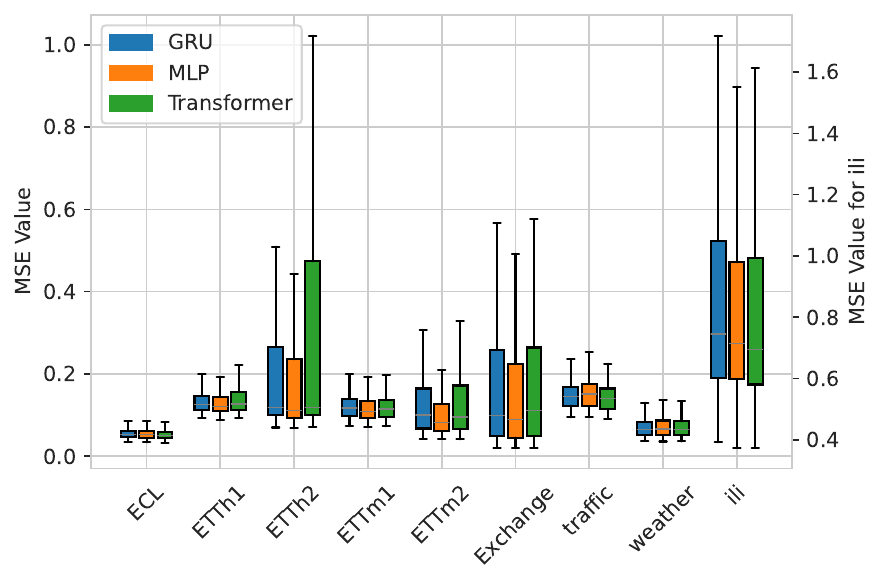}
         \caption{Network Backbone}
         \label{fig:exp-appx-backbone-ltf_bp}
     \end{subfigure}
     \hspace{10pt}
    \begin{subfigure}[t]{0.28\textwidth}
         \centering
         \includegraphics[width=\textwidth]{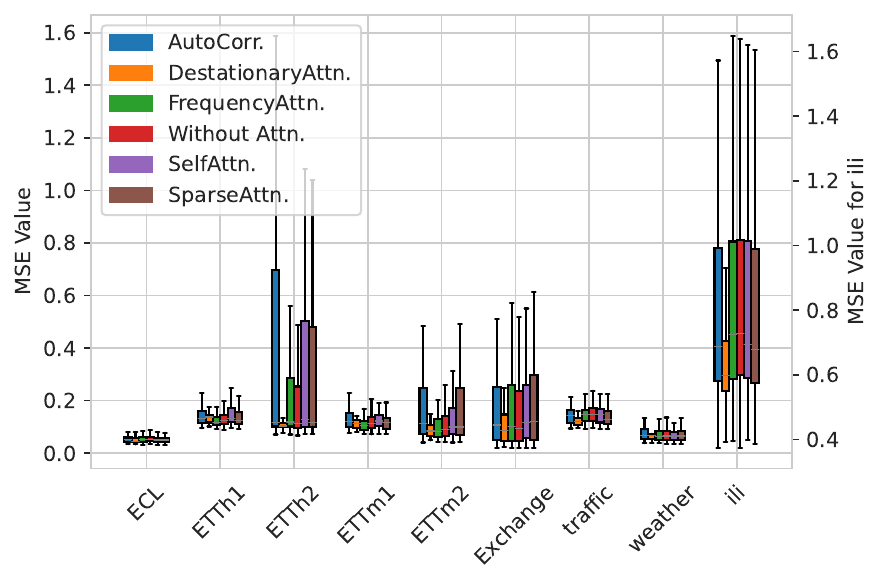}
         \caption{Series Attention}
         \label{fig:exp-appx-attention-ltf_bp}
     \end{subfigure}
     \hspace{10pt}
         \begin{subfigure}[t]{0.28\textwidth}
         \centering
         \includegraphics[width=\textwidth]{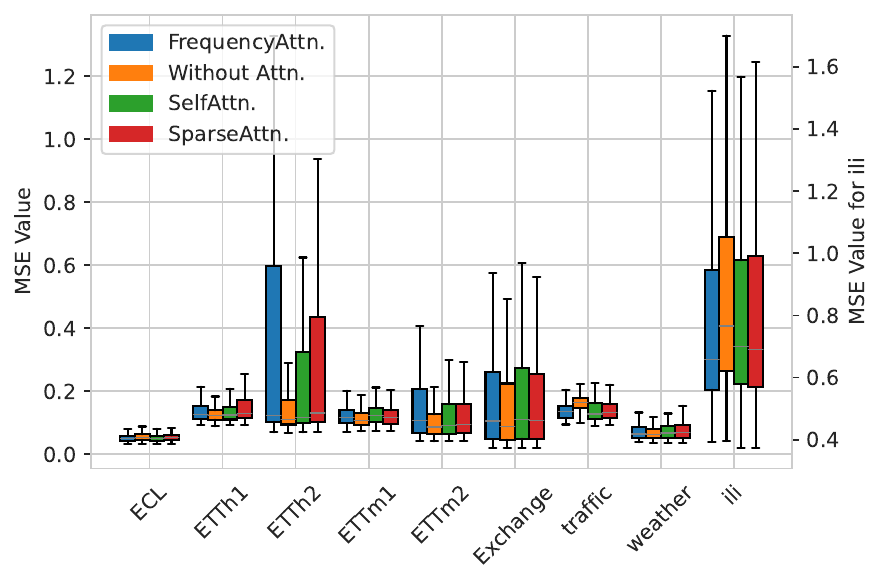}
         \caption{Feature Attention}
         \label{fig:exp-appx-featureattention-ltf_bp}
     \end{subfigure}
     \hspace{10pt}
    \begin{subfigure}[t]{0.28\textwidth}
         \centering
         \includegraphics[width=\textwidth]{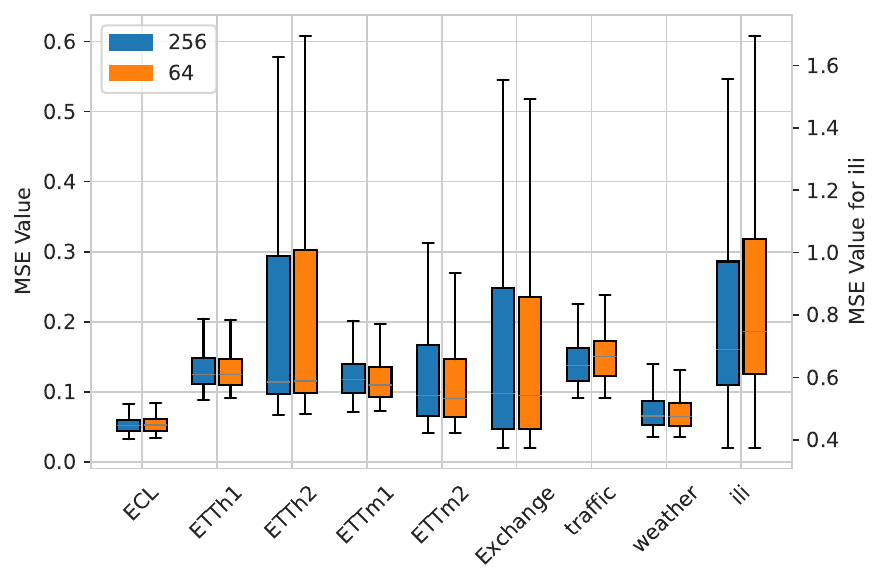}
         \caption{\textit{Hidden Layer Dimensions}}
         \label{fig:exp-appx-dmodel-ltf_bp}
     \end{subfigure}
     \hspace{10pt}
    \begin{subfigure}[t]{0.28\textwidth}
         \centering
         \includegraphics[width=\textwidth]{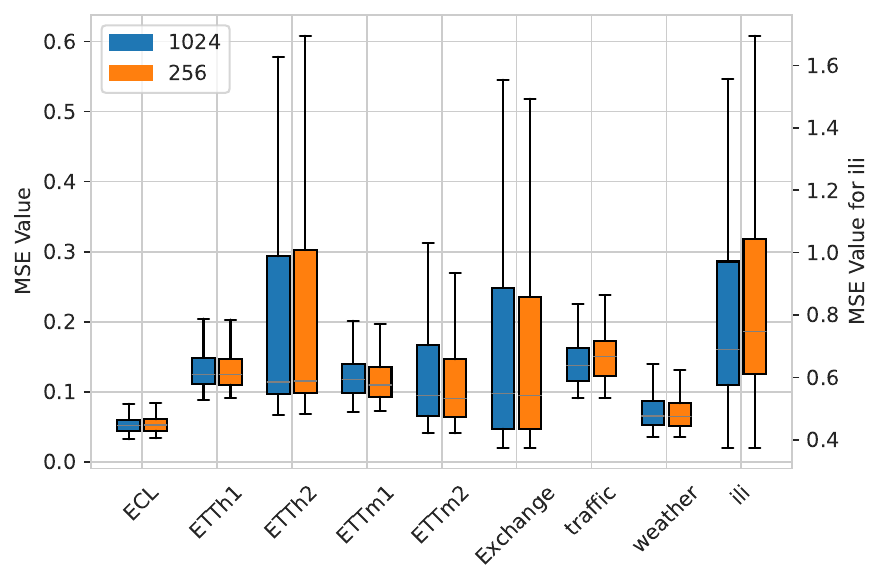}
         \caption{\textit{FCN Layer Dimensions}}
         \label{fig:exp-appx-dff-ltf_bp}
     \end{subfigure}
     \hspace{10pt}
    \begin{subfigure}[t]{0.28\textwidth}
         \centering
         \includegraphics[width=\textwidth]{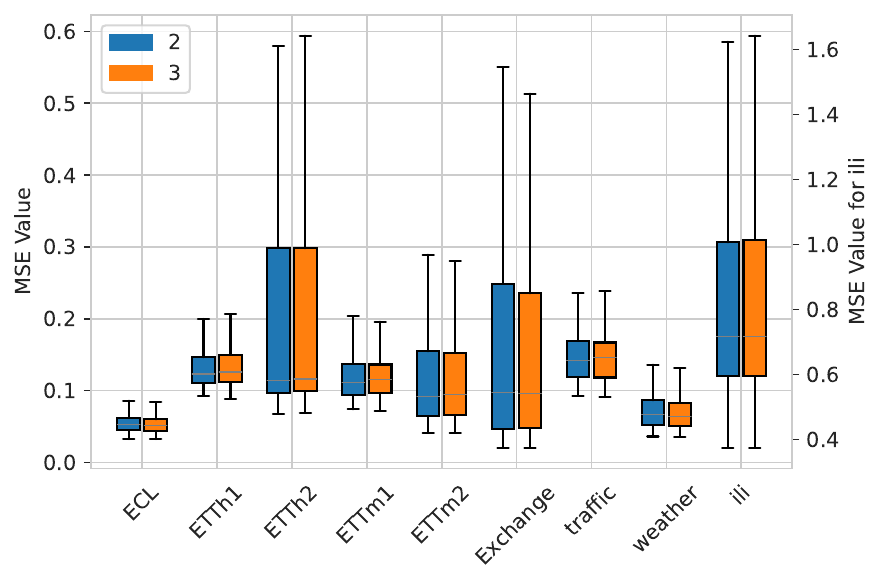}
         \caption{Encoder layers}
         \label{fig:exp-appx-el-ltf_bp}
     \end{subfigure}
     \hspace{10pt}
    \begin{subfigure}[t]{0.28\textwidth}
         \centering
         \includegraphics[width=\textwidth]{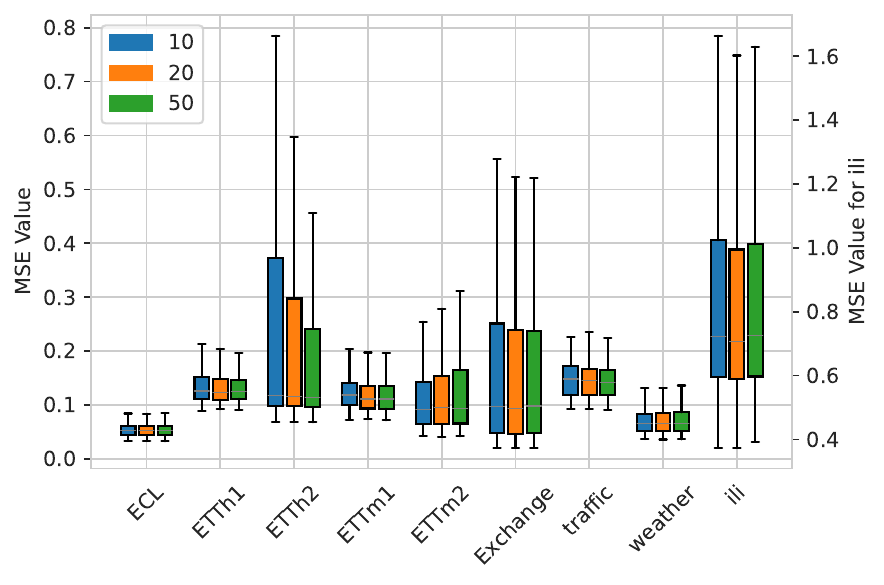}
         \caption{Epochs}
         \label{fig:exp-appx-epochs-ltf_bp}
     \end{subfigure}
     \hspace{10pt}
    \begin{subfigure}[t]{0.28\textwidth}
         \centering
         \includegraphics[width=\textwidth]{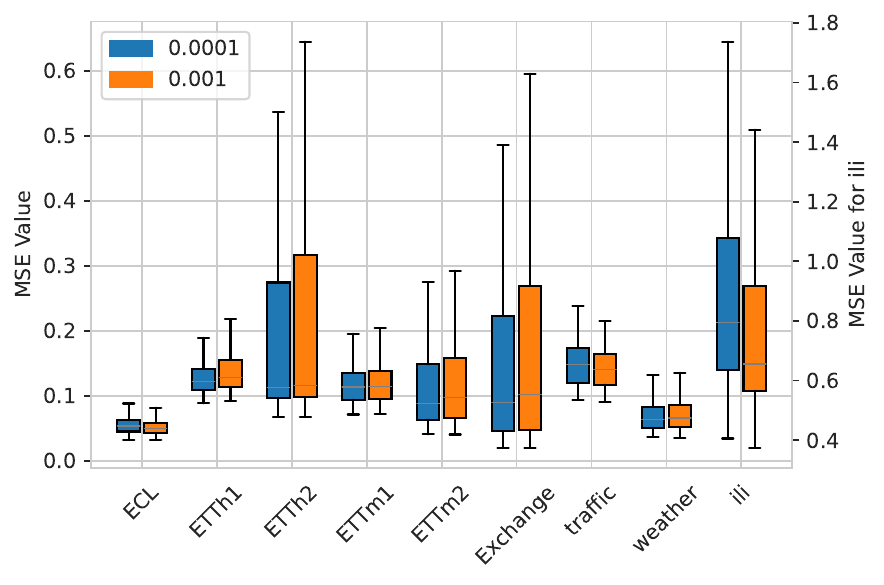}
         \caption{Learning Rate}
         \label{fig:exp-appx-lr-ltf_bp}
     \end{subfigure}
     \hspace{10pt}
    \begin{subfigure}[t]{0.28\textwidth}
         \centering
         \includegraphics[width=\textwidth]{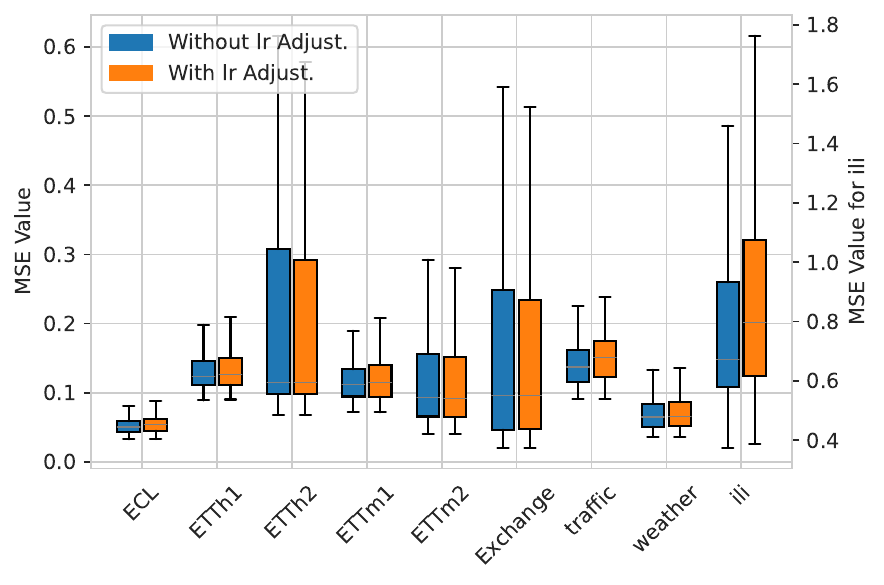}
         \caption{Learning Rate Strategy}
         \label{fig:exp-appx-lrs-ltf_bp}
     \end{subfigure}
     \hspace{10pt}
         \begin{subfigure}[t]{0.28\textwidth}
         \centering
         \includegraphics[width=\textwidth]{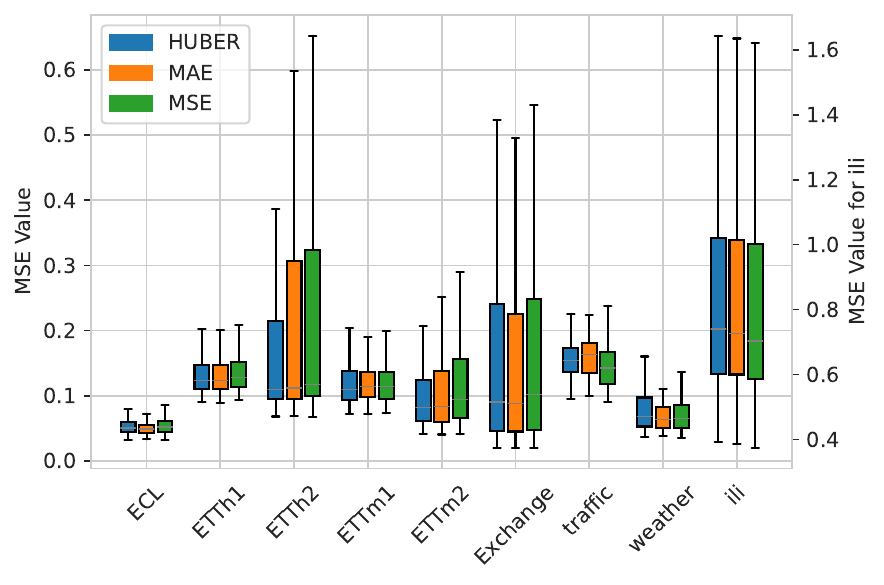}
         \caption{Loss Function}
         \label{fig:exp-appx-lf-ltf_bp}
     \end{subfigure}
     \hspace{10pt}
         \begin{subfigure}[t]{0.28\textwidth}
         \centering
         \includegraphics[width=\textwidth]{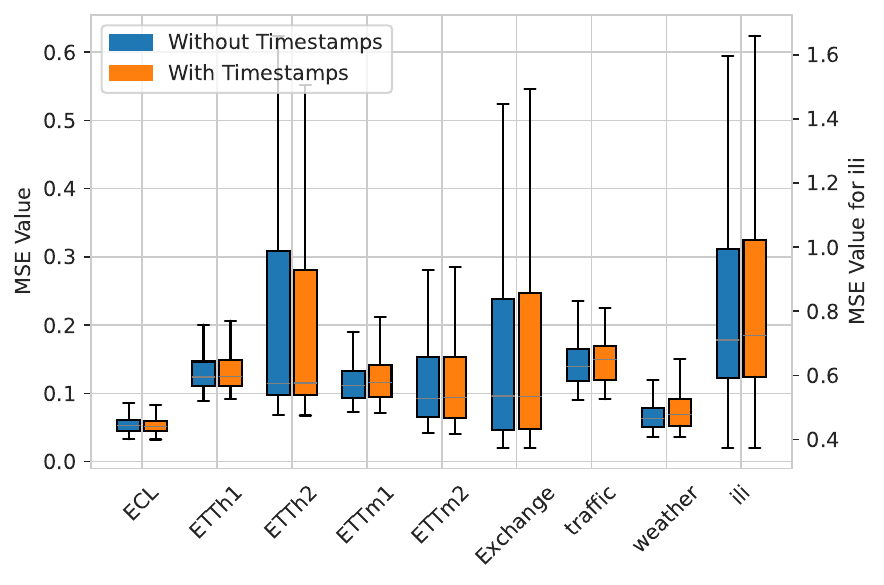}
         \caption{Timestamp}
         \label{fig:exp-appx-xm-ltf_bp}
     \end{subfigure}
     \hspace{10pt}
     \caption{Overall performance across all design dimensions in long-term forecasting. The results (\textbf{MSE}) are averaged across all forecasting horizons. Due to the significantly different value range and variability of the ILI dataset compared to other datasets, its box plot is plotted using the right-hand \textit{y}-axis, while all other datasets share the left-hand \textit{y}-axis.}
     \vspace{-0.1in}
     \label{fig:exp-appx-bp-ltf}
\end{figure}

\begin{figure}[t!]
     \centering
     \begin{subfigure}[t]{0.28\textwidth}
         \centering
         \includegraphics[width=\textwidth]{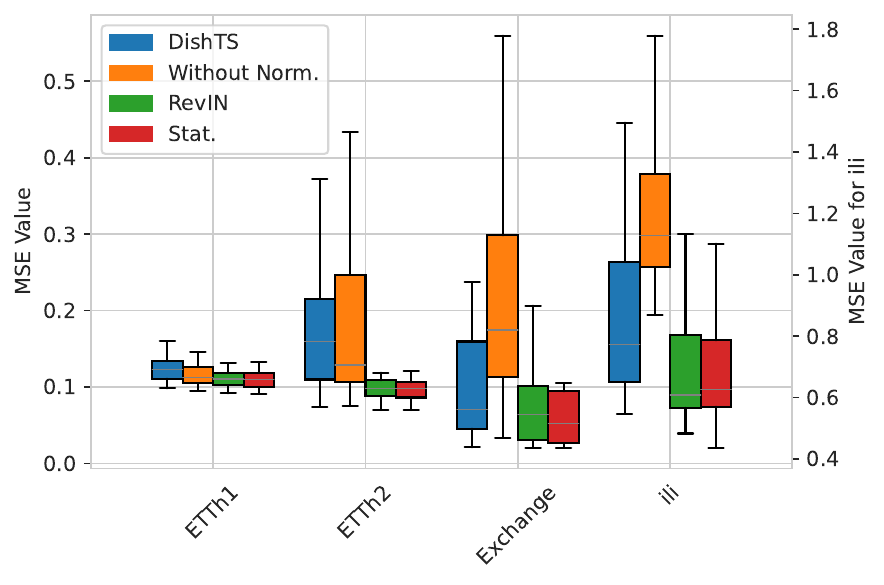}
         \caption{Series Normalization}
         \label{fig:exp-appx-Normalization-llm_bp}
     \end{subfigure}
     \hspace{10pt}
     \begin{subfigure}[t]{0.28\textwidth}
         \centering
         \includegraphics[width=\textwidth]{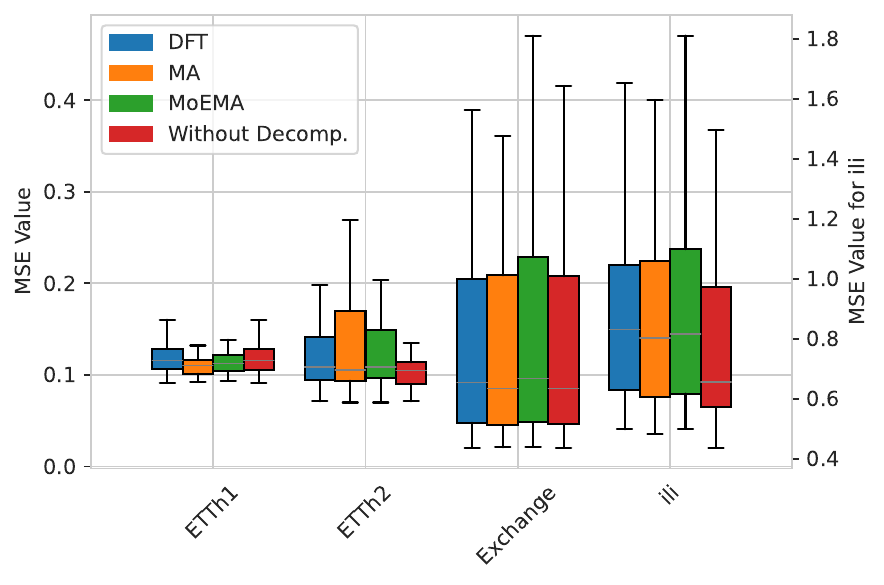}
         \caption{Series Decomposition}
         \label{fig:exp-appx-Decomposition-llm_bp}
     \end{subfigure}
     \hspace{10pt}
    \begin{subfigure}[t]{0.28\textwidth}
         \centering
         \includegraphics[width=\textwidth]{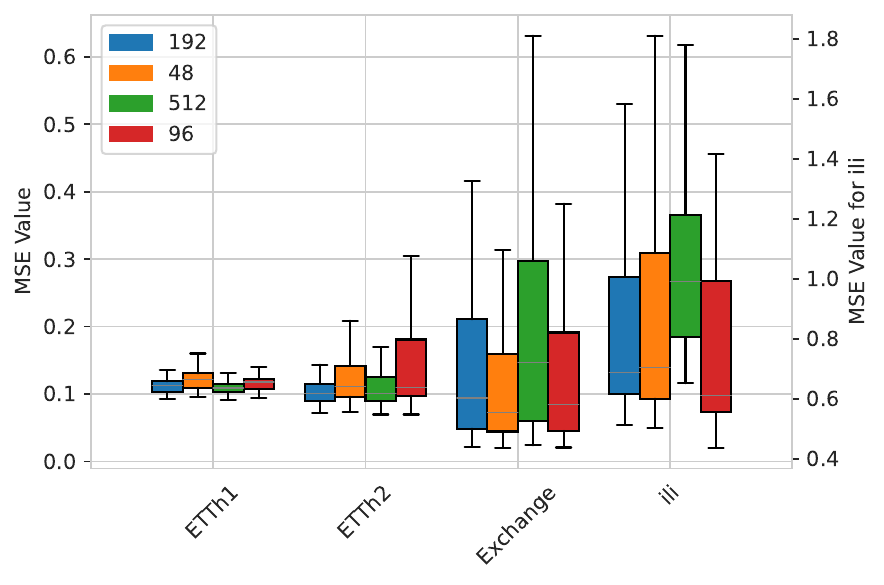}
         \caption{Sequence Length}
         \label{fig:exp-appx-sl-llm}
     \end{subfigure}
     \hspace{10pt}
    \begin{subfigure}[t]{0.28\textwidth}
         \centering
         \includegraphics[width=\textwidth]{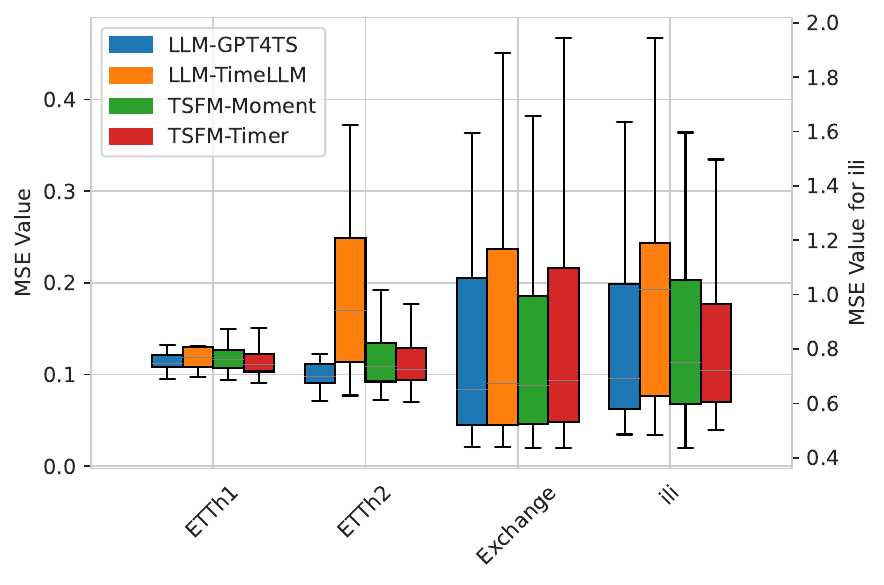}
         \caption{Network Backbone}
         \label{fig:exp-appx-backbone-llm_bp}
     \end{subfigure}
     \hspace{10pt}
    \begin{subfigure}[t]{0.28\textwidth}
         \centering
         \includegraphics[width=\textwidth]{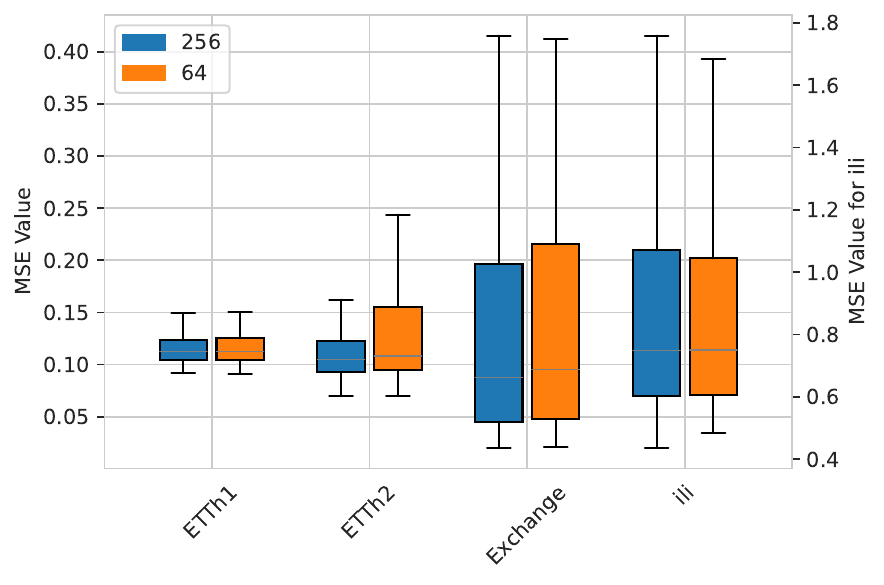}
         \caption{\textit{Hidden Layer Dimensions}}
         \label{fig:exp-appx-dmodel-llm_bp}
     \end{subfigure}
     \hspace{10pt}
    \begin{subfigure}[t]{0.28\textwidth}
         \centering
         \includegraphics[width=\textwidth]{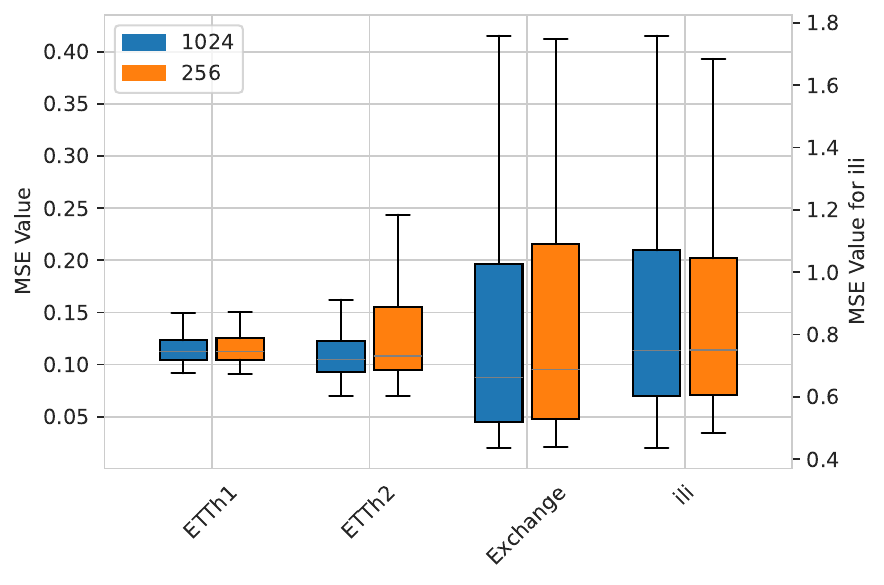}
         \caption{\textit{FCN Layer Dimensions}}
         \label{fig:exp-appx-dff-llm_bp}
     \end{subfigure}
     \hspace{10pt}
    \begin{subfigure}[t]{0.28\textwidth}
         \centering
         \includegraphics[width=\textwidth]{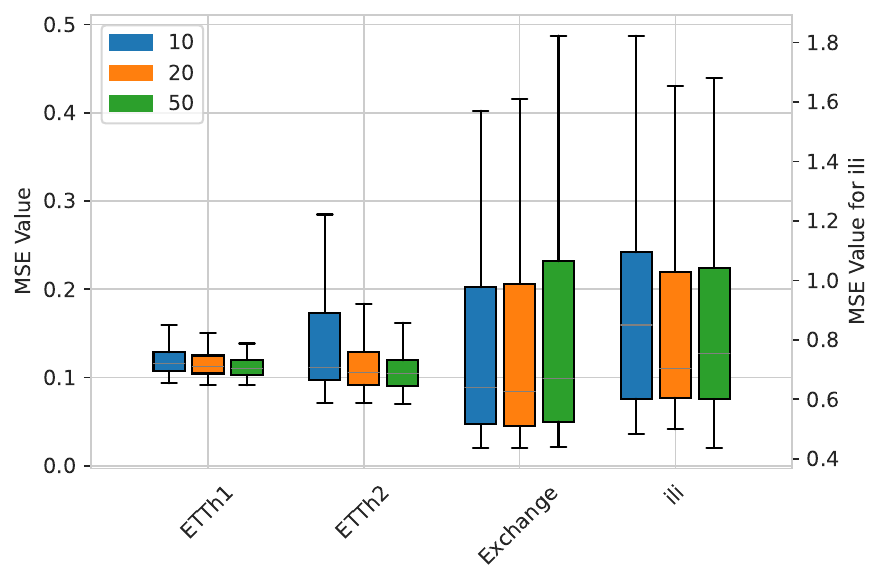}
         \caption{Epochs}
         \label{fig:exp-appx-epochs-llm_bp}
     \end{subfigure}
     \hspace{10pt}
    \begin{subfigure}[t]{0.28\textwidth}
         \centering
         \includegraphics[width=\textwidth]{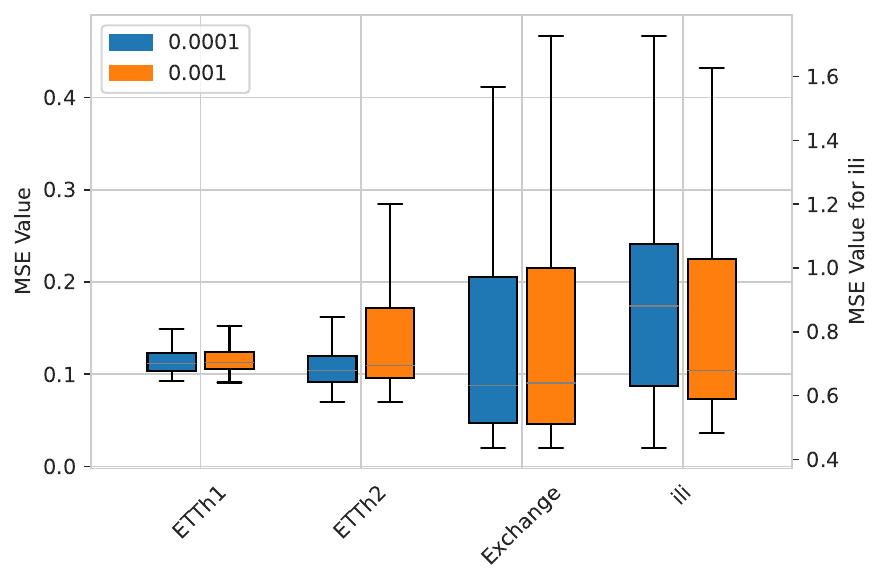}
         \caption{Learning Rate}
         \label{fig:exp-appx-lr-llm_bp}
     \end{subfigure}
     \hspace{10pt}
    \begin{subfigure}[t]{0.28\textwidth}
         \centering
         \includegraphics[width=\textwidth]{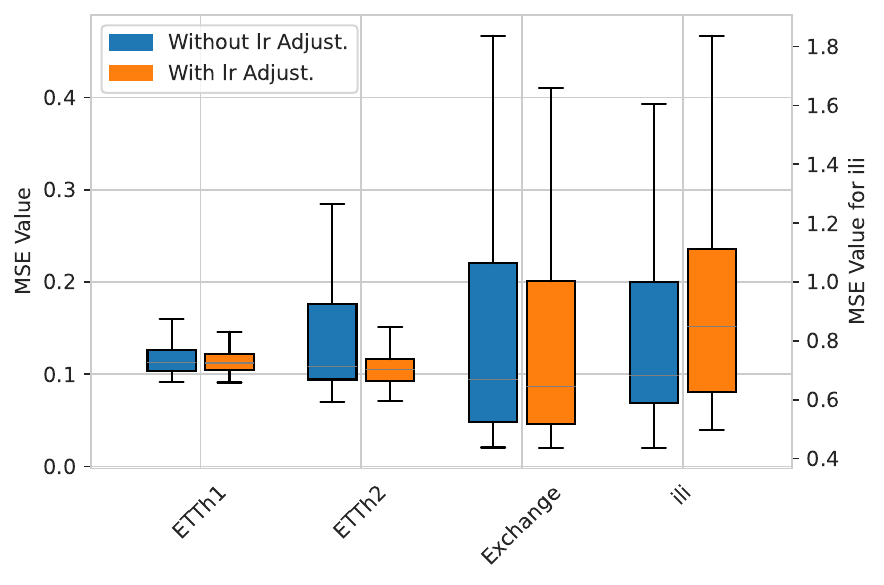}
         \caption{Learning Rate Strategy}
         \label{fig:exp-appx-lrs-llm_bp}
     \end{subfigure}
     \hspace{10pt}
              \begin{subfigure}[t]{0.28\textwidth}
         \centering
         \includegraphics[width=\textwidth]{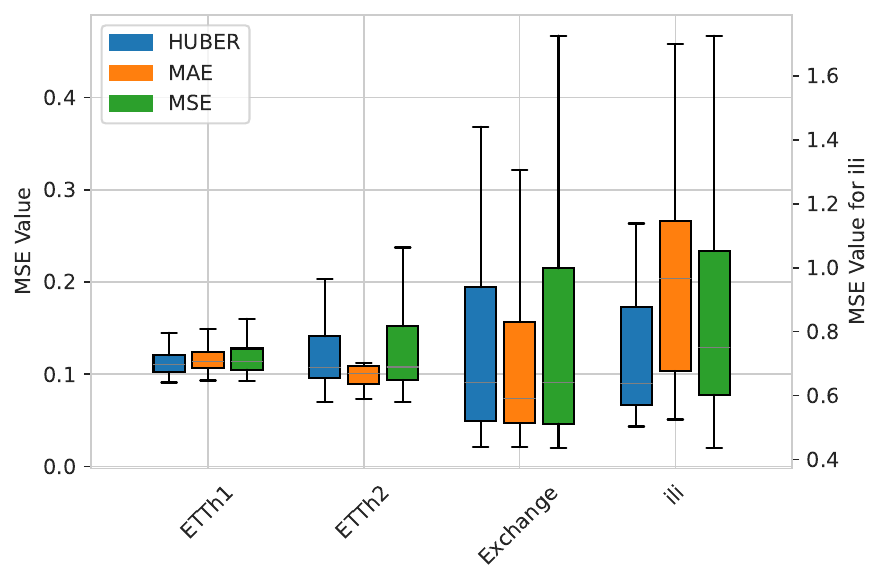}
         \caption{Loss Function}
         \label{fig:exp-appx-lf-llm_bp}
     \end{subfigure}
     \hspace{10pt}
    \begin{subfigure}[t]{0.28\textwidth}
         \centering
         \includegraphics[width=\textwidth]{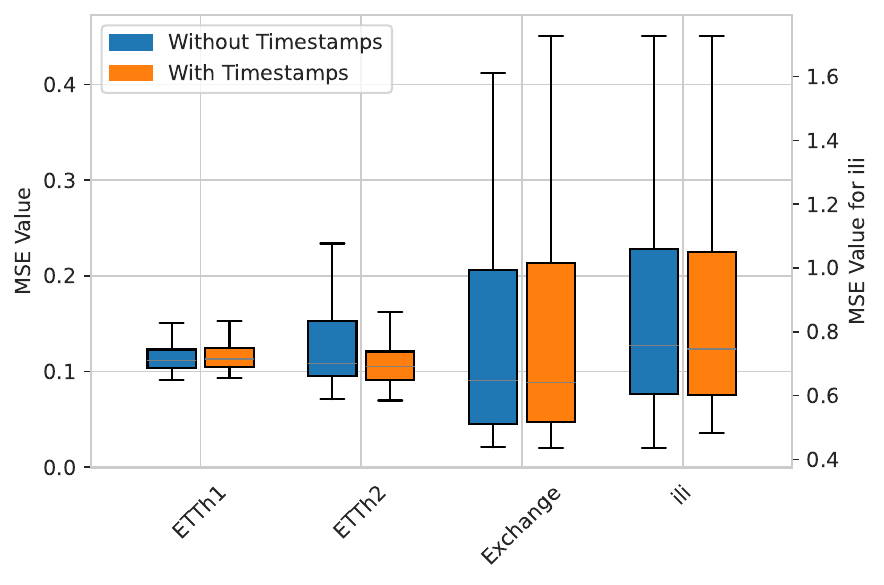}
         \caption{Timestamp}
         \label{fig:exp-appx-xm-llm_bp}
     \end{subfigure}
     \hspace{10pt}
    \begin{subfigure}[t]{0.28\textwidth}
         \centering
         \includegraphics[width=\textwidth]{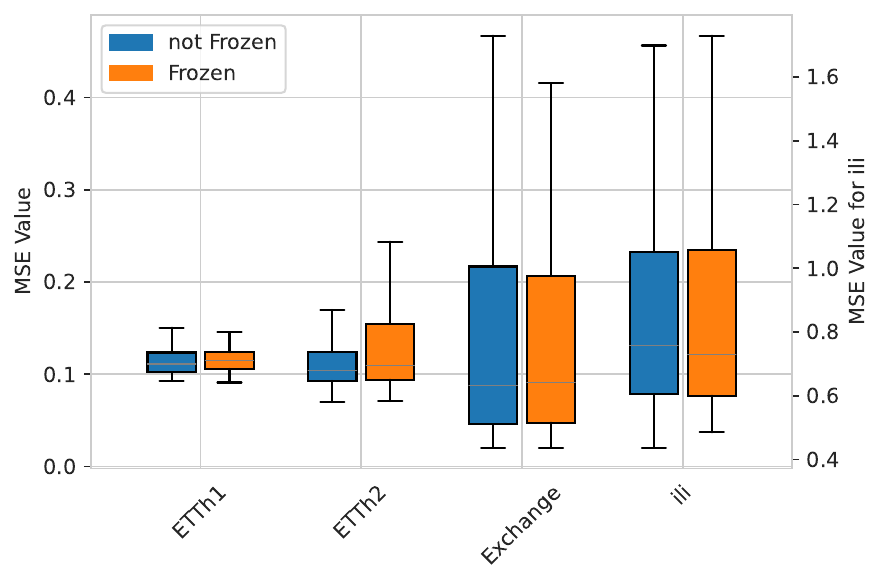}
         \caption{Frozen}
         \label{fig:exp-appx-frozen-llm_bp}
     \end{subfigure}
     \hspace{10pt}
     \caption{Overall performance across all design dimensions when using LLMs or TSFMs in long-term forecasting. The results (\textbf{MSE}) are averaged across all forecasting horizons. Due to the significantly different value range and variability of the ILI dataset compared to other datasets, its box plot is plotted using the right-hand \textit{y}-axis, while all other datasets share the left-hand \textit{y}-axis.}
     \vspace{-0.1in}
     \label{fig:exp-appx-bp-llm}
\end{figure}

\textbf{Box Plots Analysis}.
The impact of various design choices for each architectural component is further illustrated through box plots in Fig.~\ref{fig:exp-appx-bp-ltf} and Fig.~\ref{fig:exp-appx-bp-llm}. These visualizations complement the spider charts by providing a statistical perspective on performance variability and robustness across multiple benchmark datasets. Together, the two forms of analysis offer a comprehensive view of how different configurations affect forecasting accuracy.

\clearpage

\subsubsection{Design Choices Evaluation Results for Long-term Forecasting Using MAE as the Metric}

\begin{figure}[t!]
     \centering
     \begin{subfigure}[t]{0.28\textwidth}
         \centering
         \includegraphics[width=\textwidth]{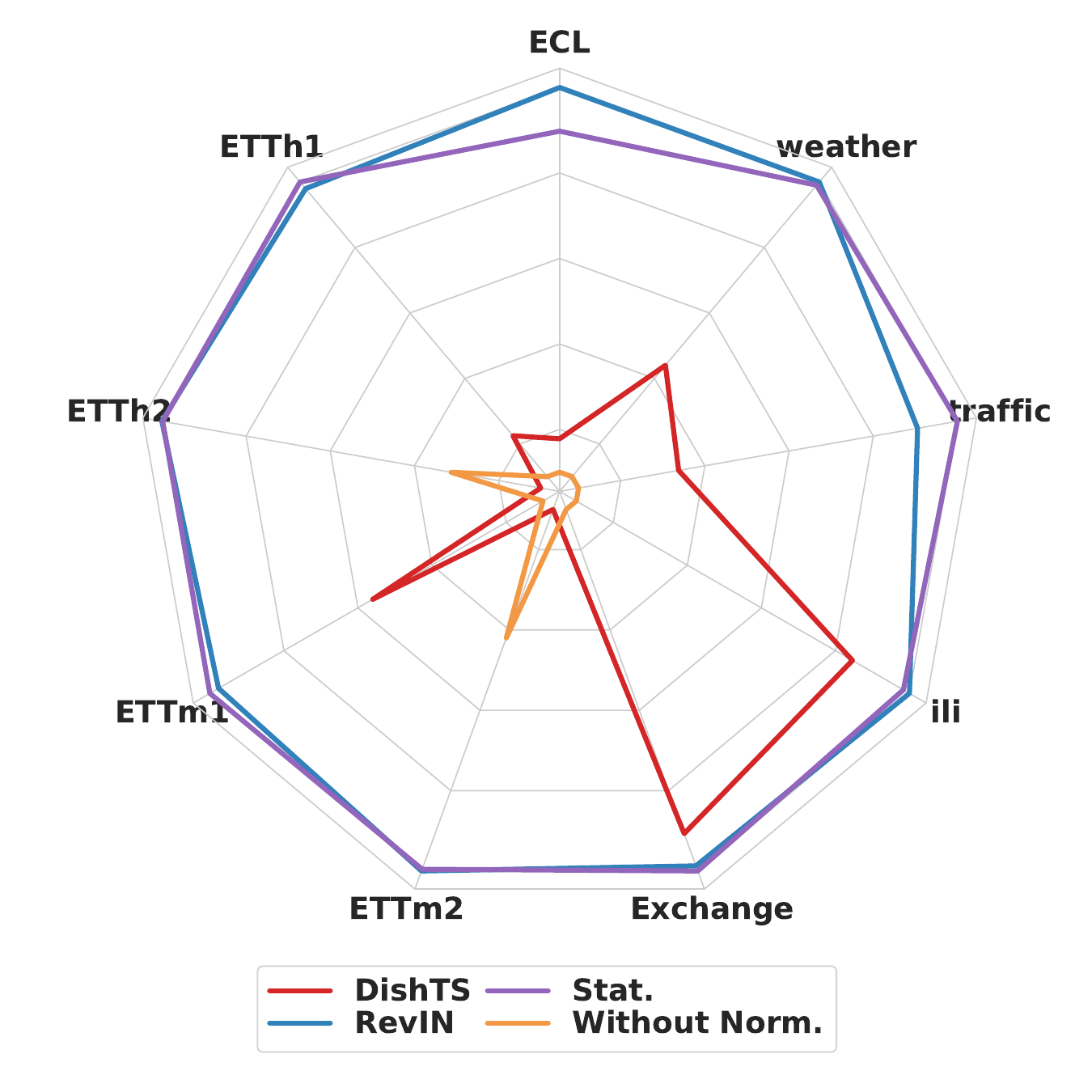}
         \caption{Series Normalization}
         \label{fig:exp-appx-Normalization-mae}
     \end{subfigure}
     \hspace{10pt}
     \begin{subfigure}[t]{0.28\textwidth}
         \centering
         \includegraphics[width=\textwidth]{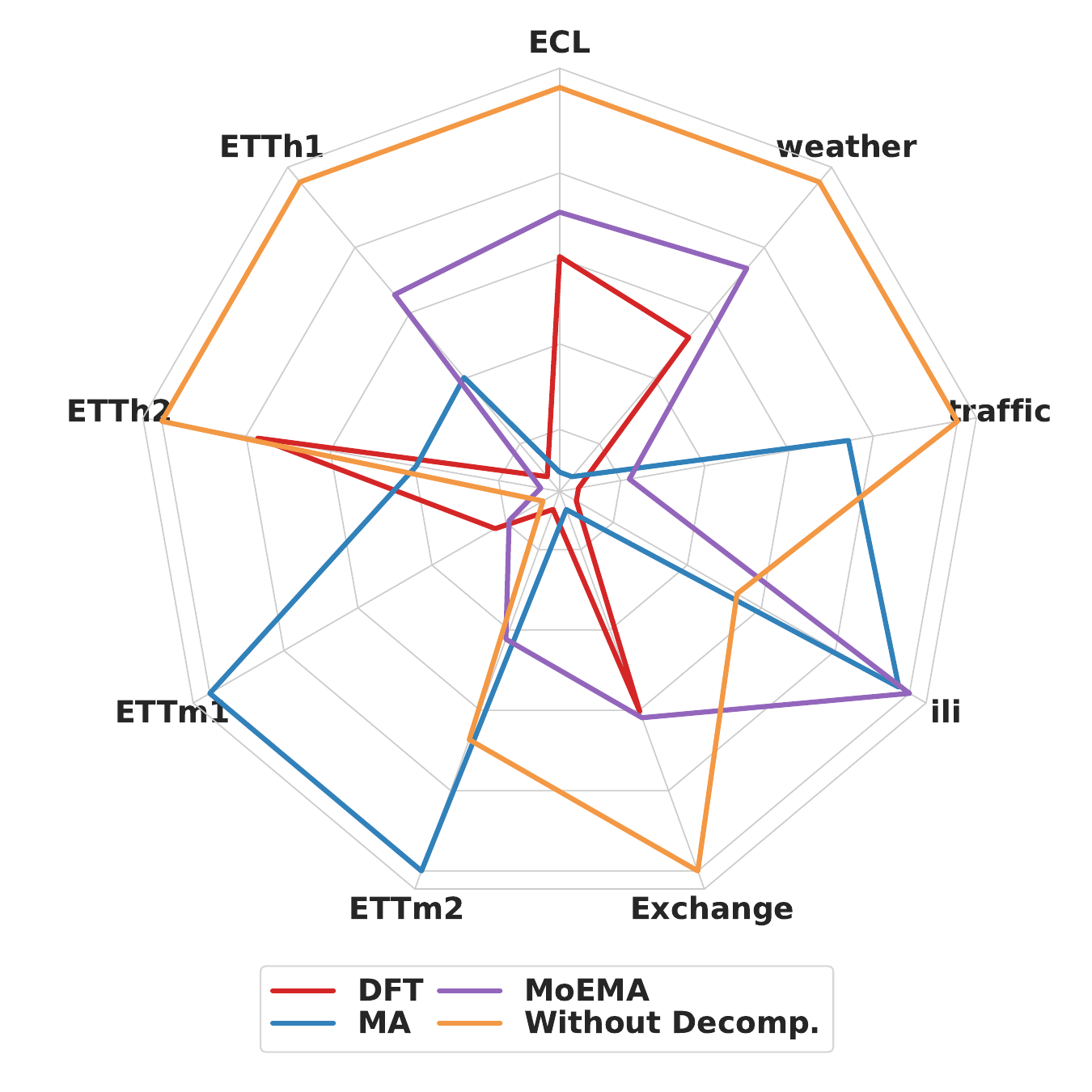}
         \caption{Series Decomposition}
         \label{fig:exp-appx-Decomposition-mae}
     \end{subfigure}
     \hspace{10pt}
    \begin{subfigure}[t]{0.28\textwidth}
         \centering
         \includegraphics[width=\textwidth]{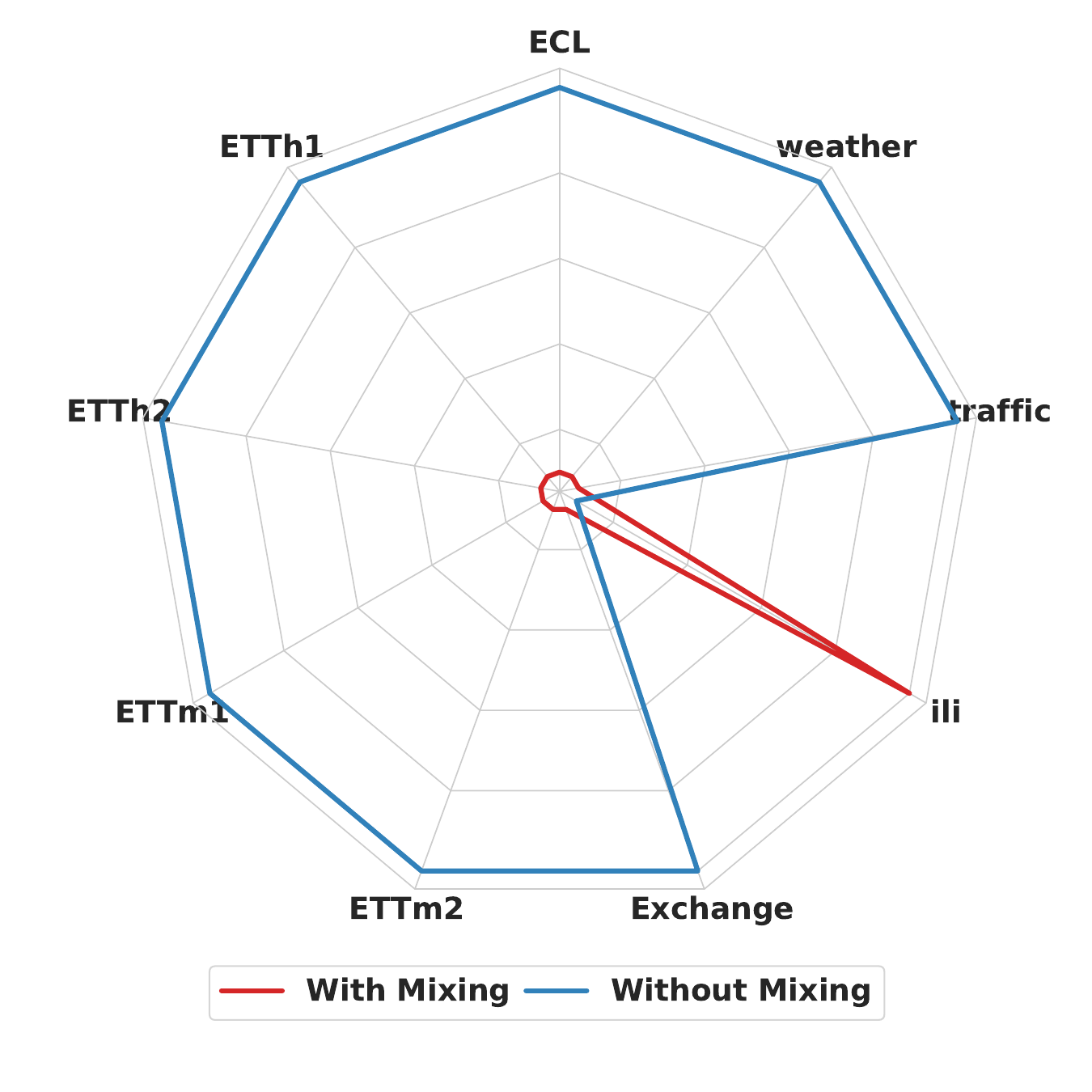}
         \caption{Series Sampling/Mixing}
         \label{fig:exp-appx-mixing-mae}
     \end{subfigure}
     \hspace{10pt}
    \begin{subfigure}[t]{0.28\textwidth}
         \centering
         \includegraphics[width=\textwidth]{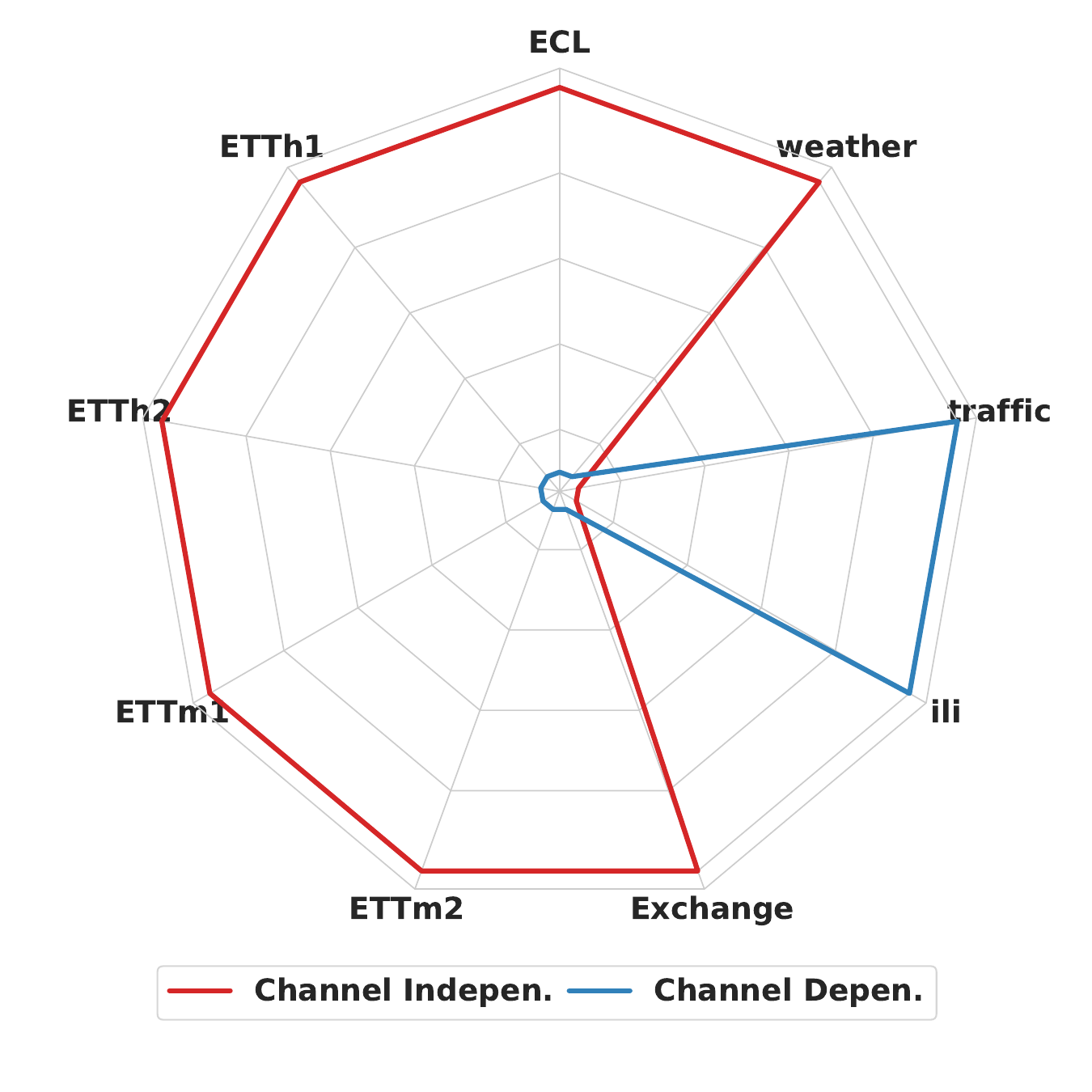}
         \caption{Channel Independent}
         \label{fig:exp-appx-CI-mae}
     \end{subfigure}
     \hspace{10pt}
    \begin{subfigure}[t]{0.28\textwidth}
         \centering
         \includegraphics[width=\textwidth]{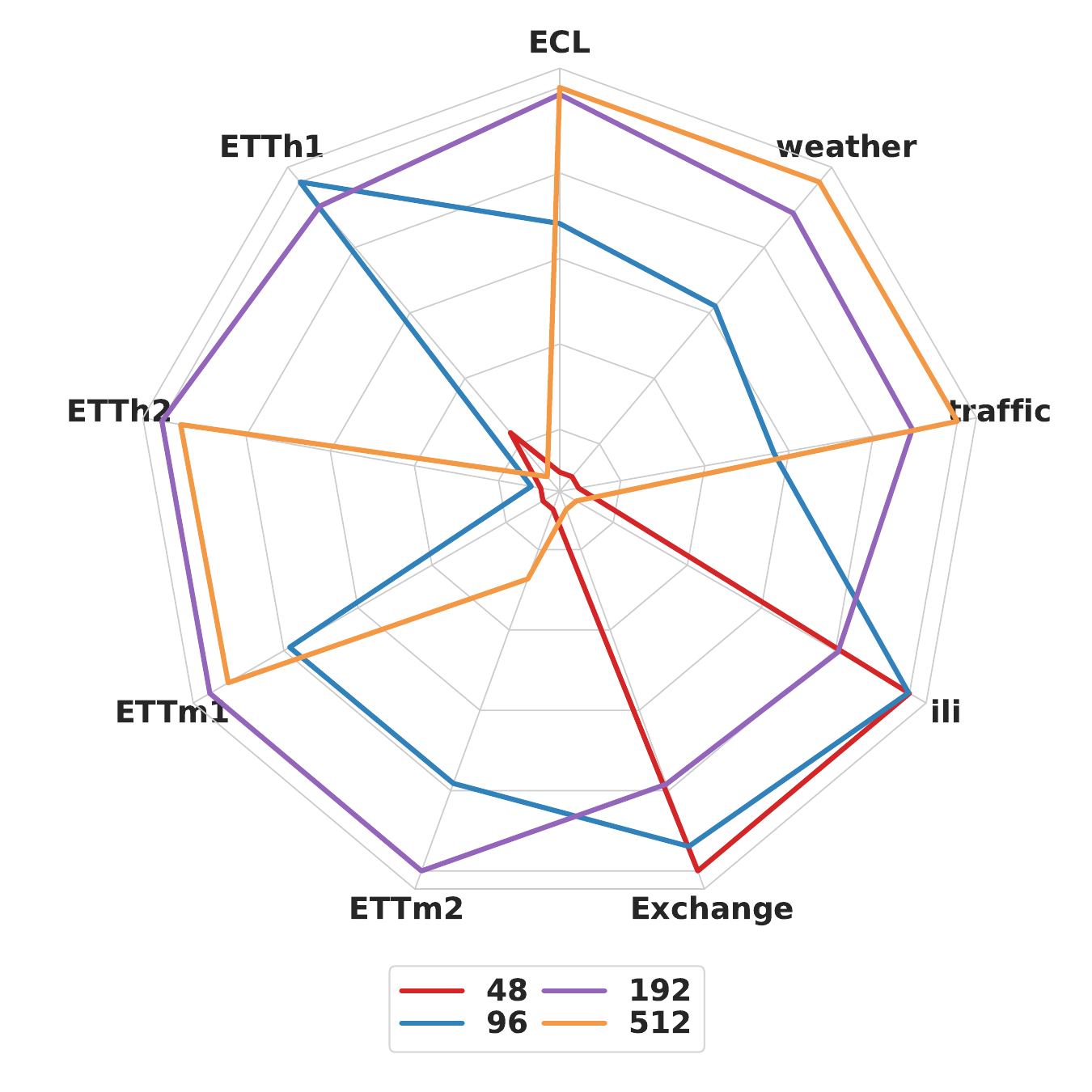}
         \caption{Sequence Length}
         \label{fig:exp-appx-sl-mae}
     \end{subfigure}
     \hspace{10pt}
    \begin{subfigure}[t]{0.28\textwidth}
         \centering
         \includegraphics[width=\textwidth]{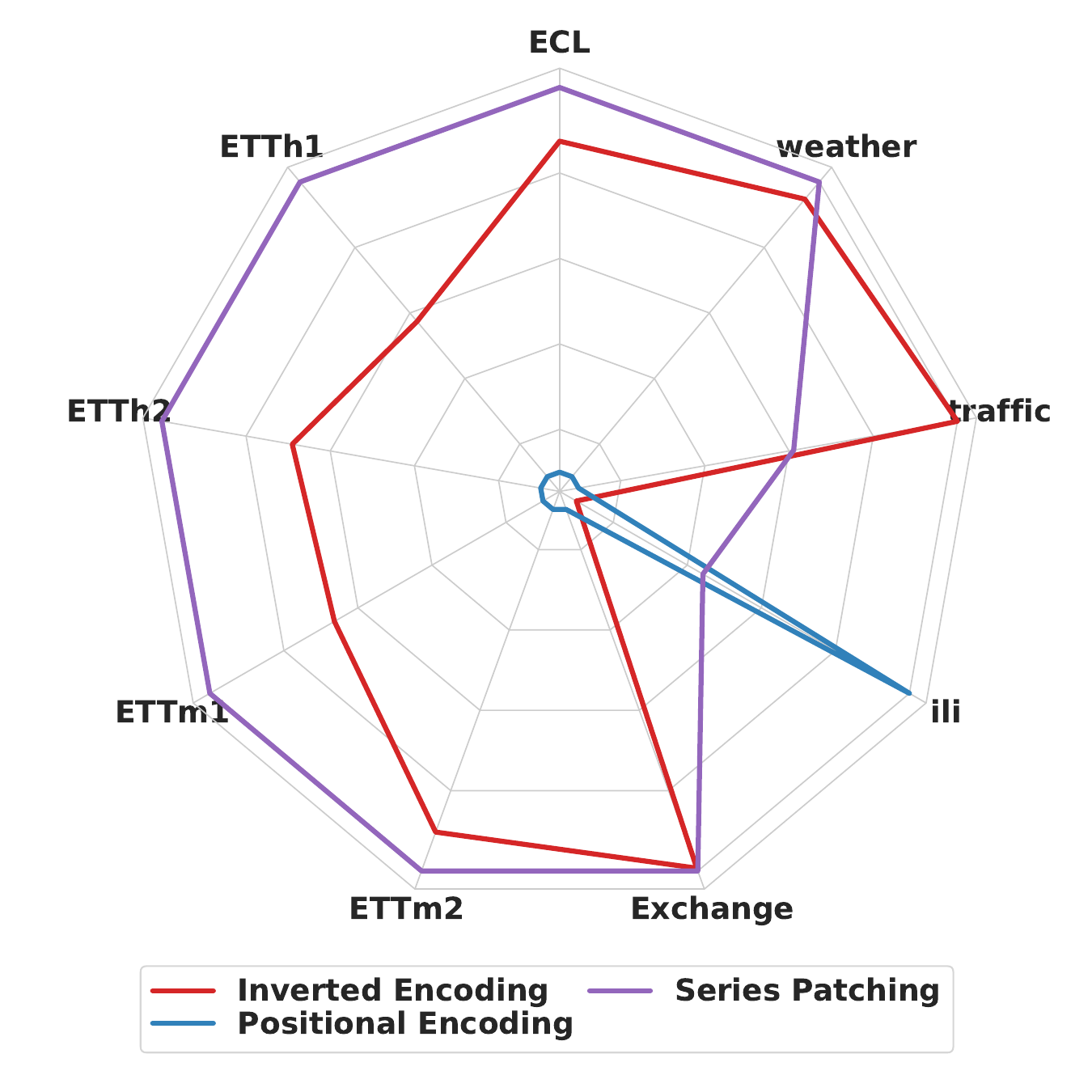}
         \caption{Series Embedding}
         \label{fig:exp-appx-tokenization-mae}
     \end{subfigure}
     \hspace{10pt}
    \begin{subfigure}[t]{0.28\textwidth}
         \centering
         \includegraphics[width=\textwidth]{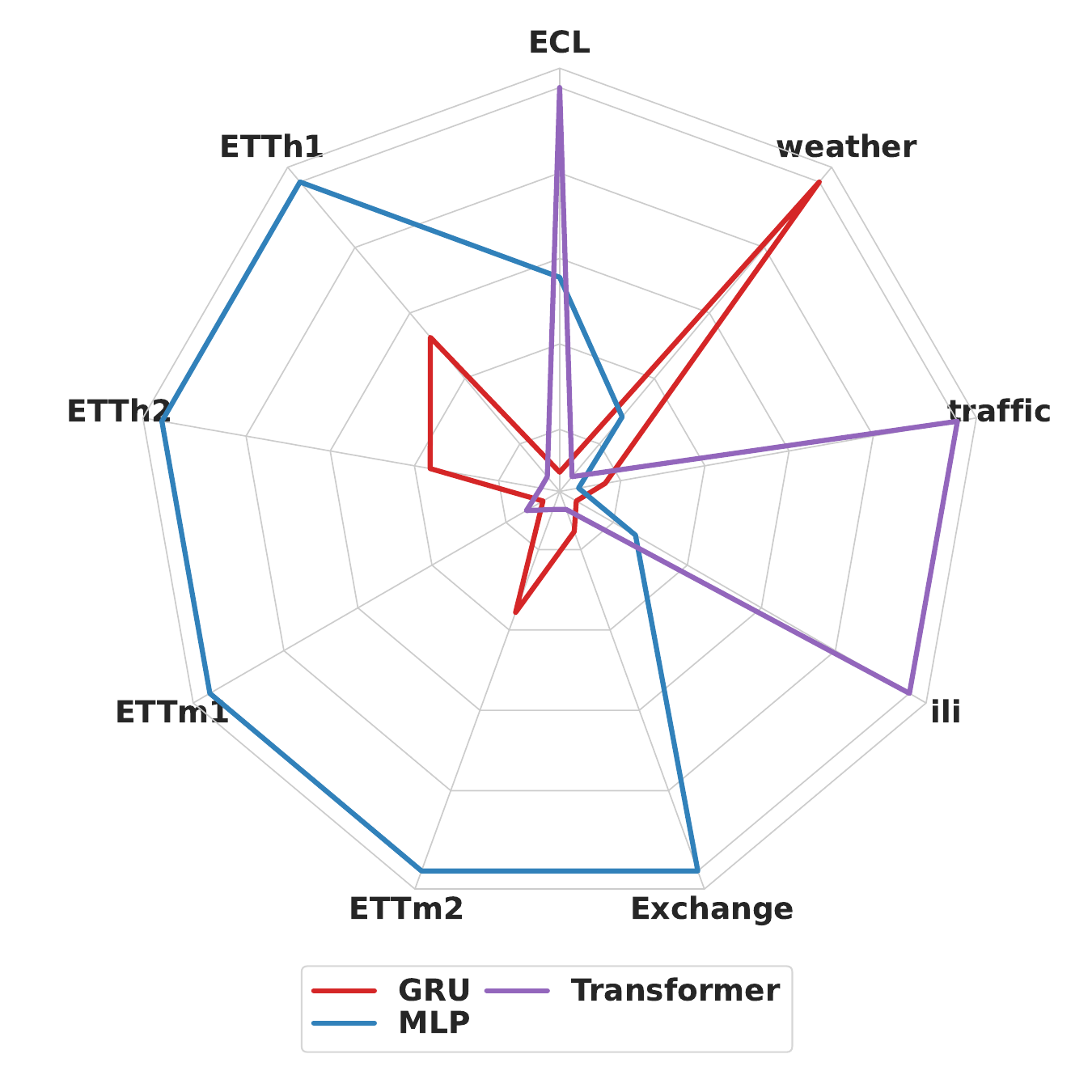}
         \caption{Network Backbone}
         \label{fig:exp-appx-backbone-mae}
     \end{subfigure}
     \hspace{10pt}
    \begin{subfigure}[t]{0.28\textwidth}
         \centering
         \includegraphics[width=\textwidth]{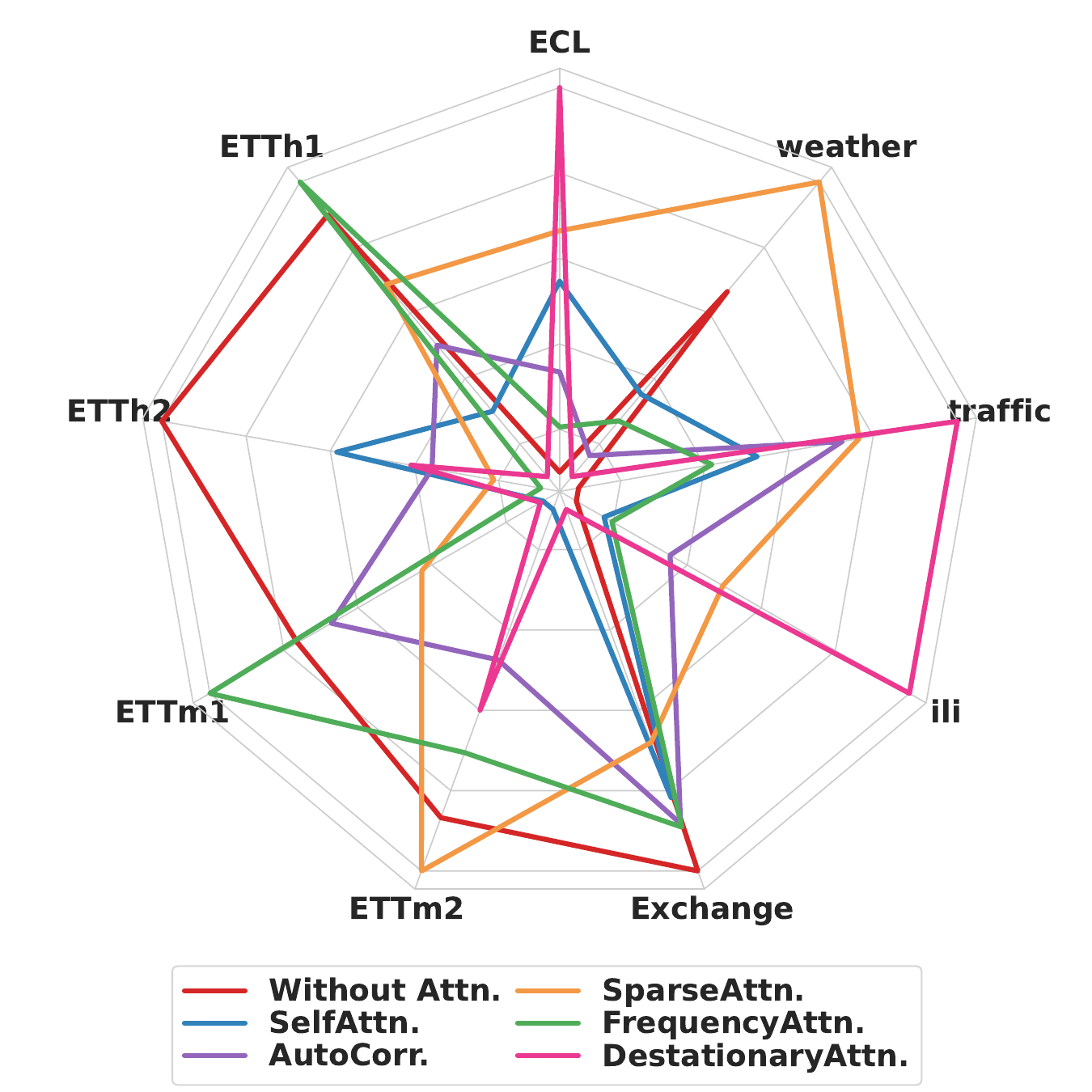}
         \caption{Series Attention}
         \label{fig:exp-appx-attention-mae}
     \end{subfigure}
     \hspace{10pt}
    \begin{subfigure}[t]{0.28\textwidth}
         \centering
         \includegraphics[width=\textwidth]{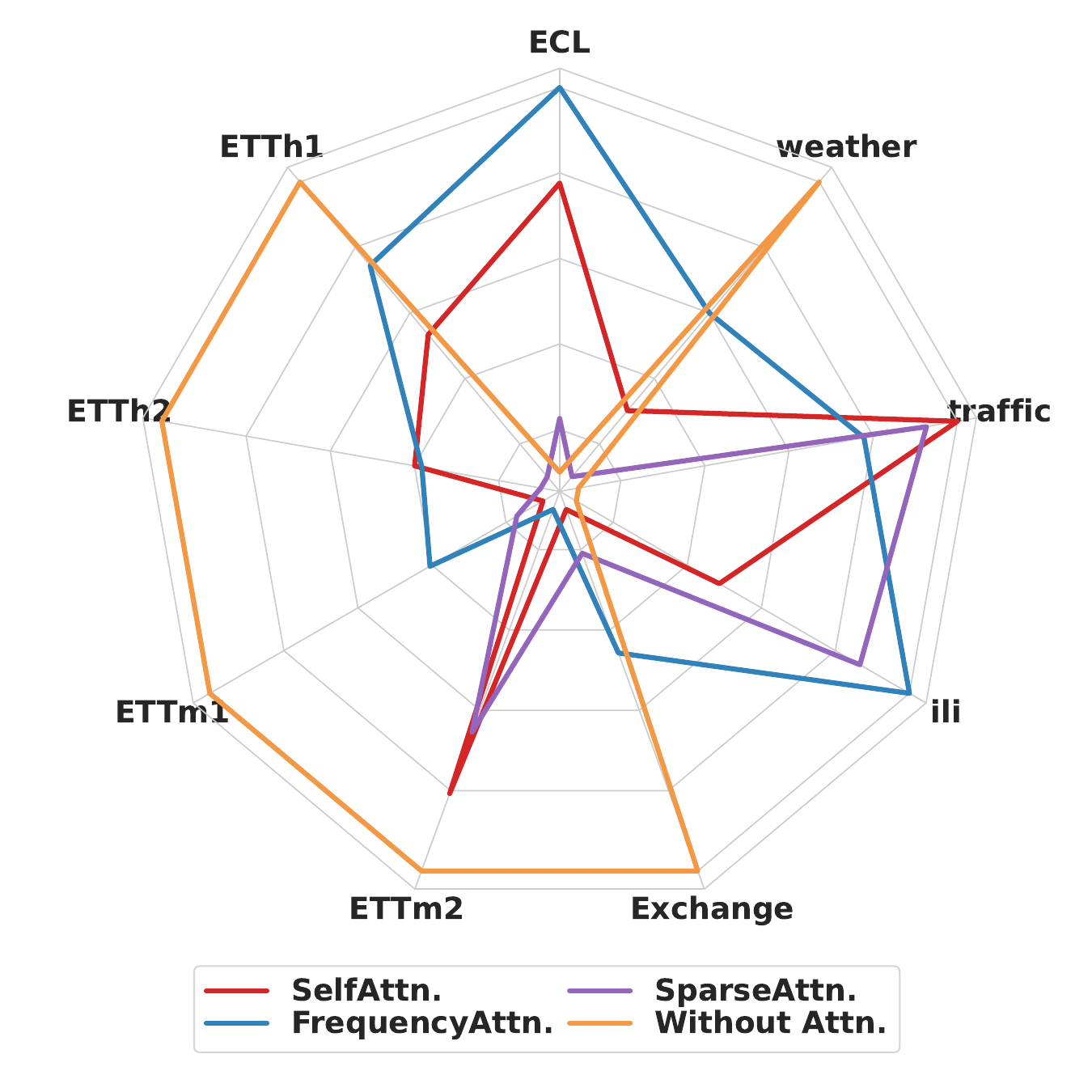}
         \caption{Feature Attention}
         \label{fig:exp-appx-featureattention-mae}
     \end{subfigure}
     \hspace{10pt}
    \begin{subfigure}[t]{0.28\textwidth}
         \centering
         \includegraphics[width=\textwidth]{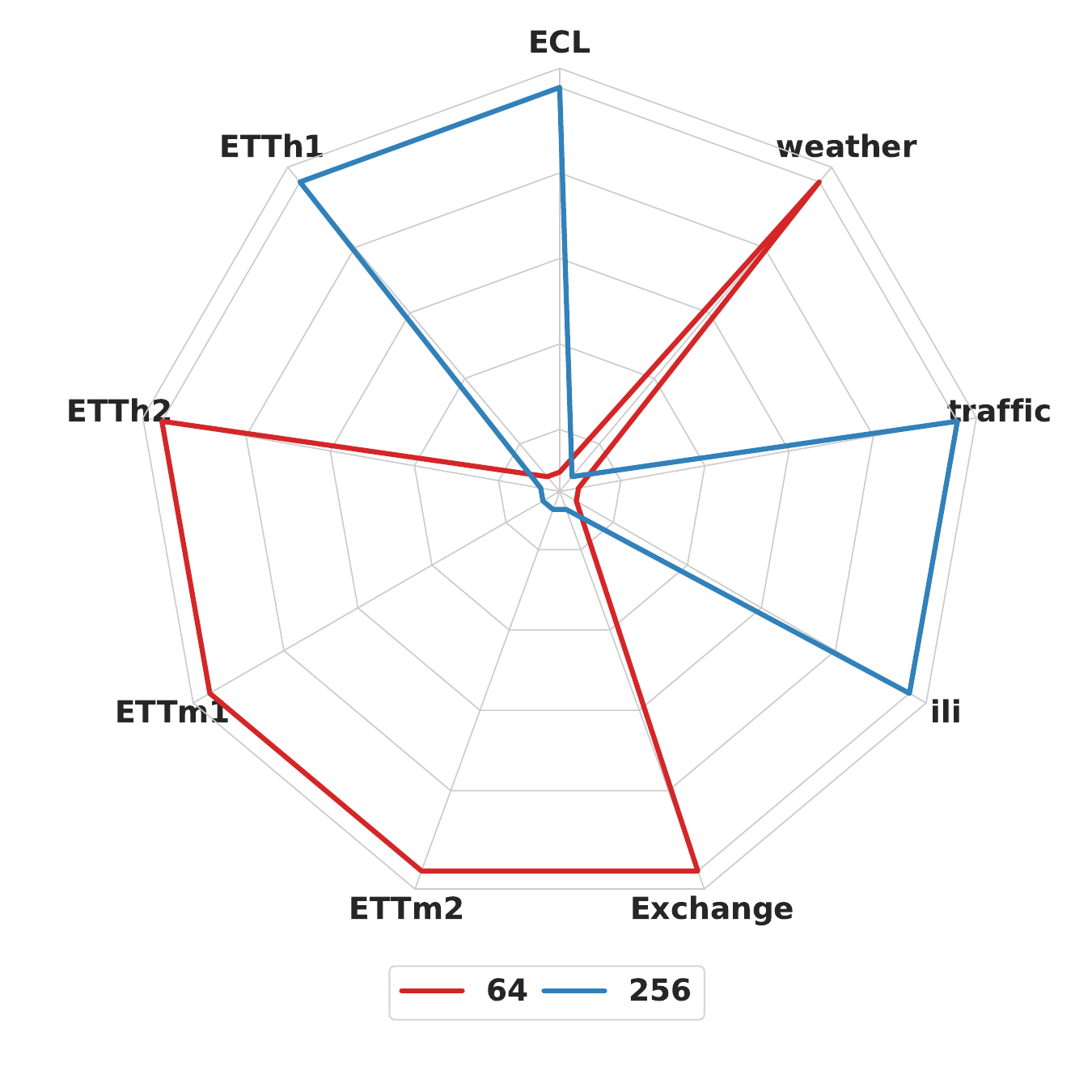}
         \caption{\textit{Hidden Layer Dimensions}}
         \label{fig:exp-appx-dmodel-mae}
     \end{subfigure}
     \hspace{10pt}
    \begin{subfigure}[t]{0.28\textwidth}
         \centering
         \includegraphics[width=\textwidth]{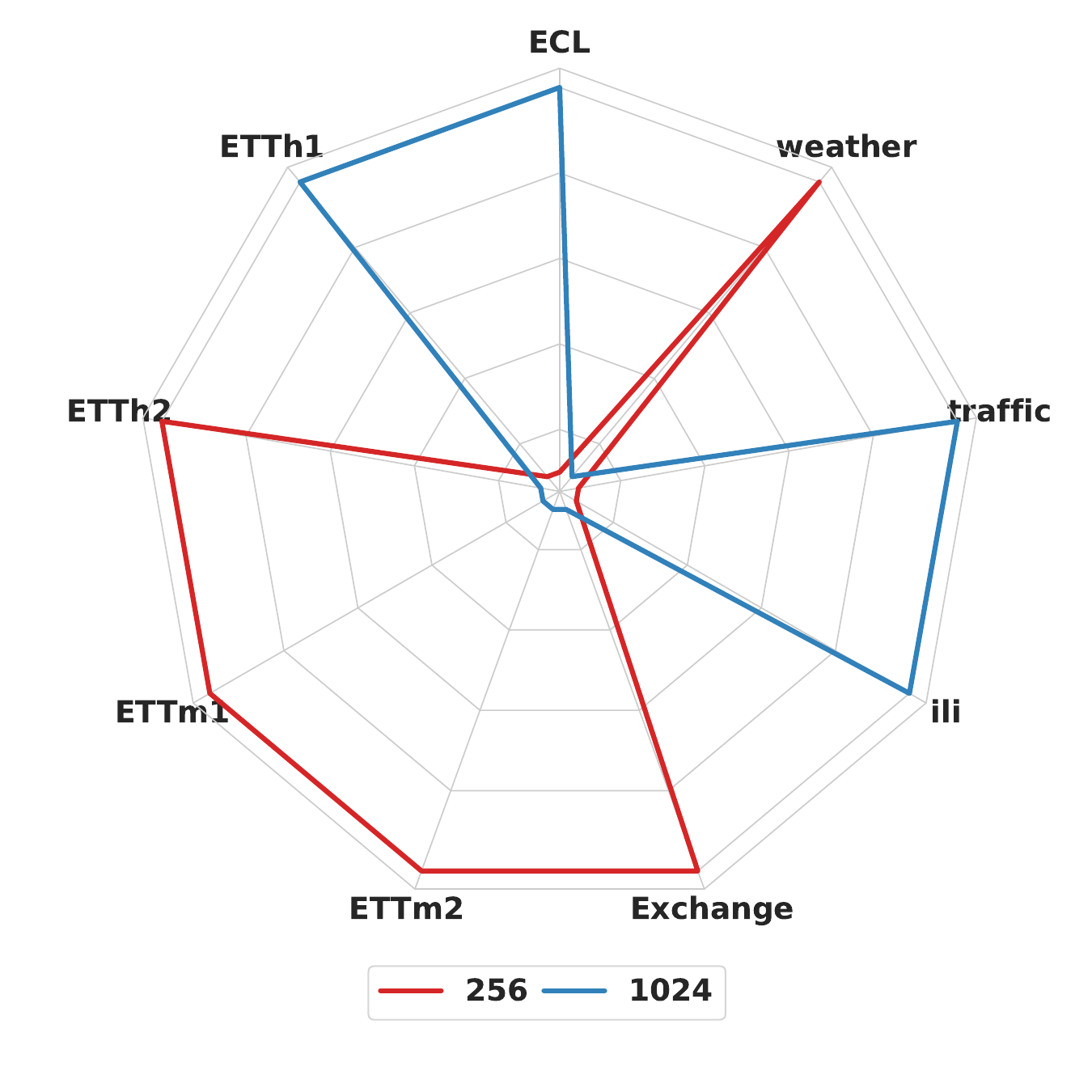}
         \caption{\textit{FCN Layer Dimensions}}
         \label{fig:exp-appx-dff-mae}
     \end{subfigure}
     \hspace{10pt}
    \begin{subfigure}[t]{0.28\textwidth}
         \centering
         \includegraphics[width=\textwidth]{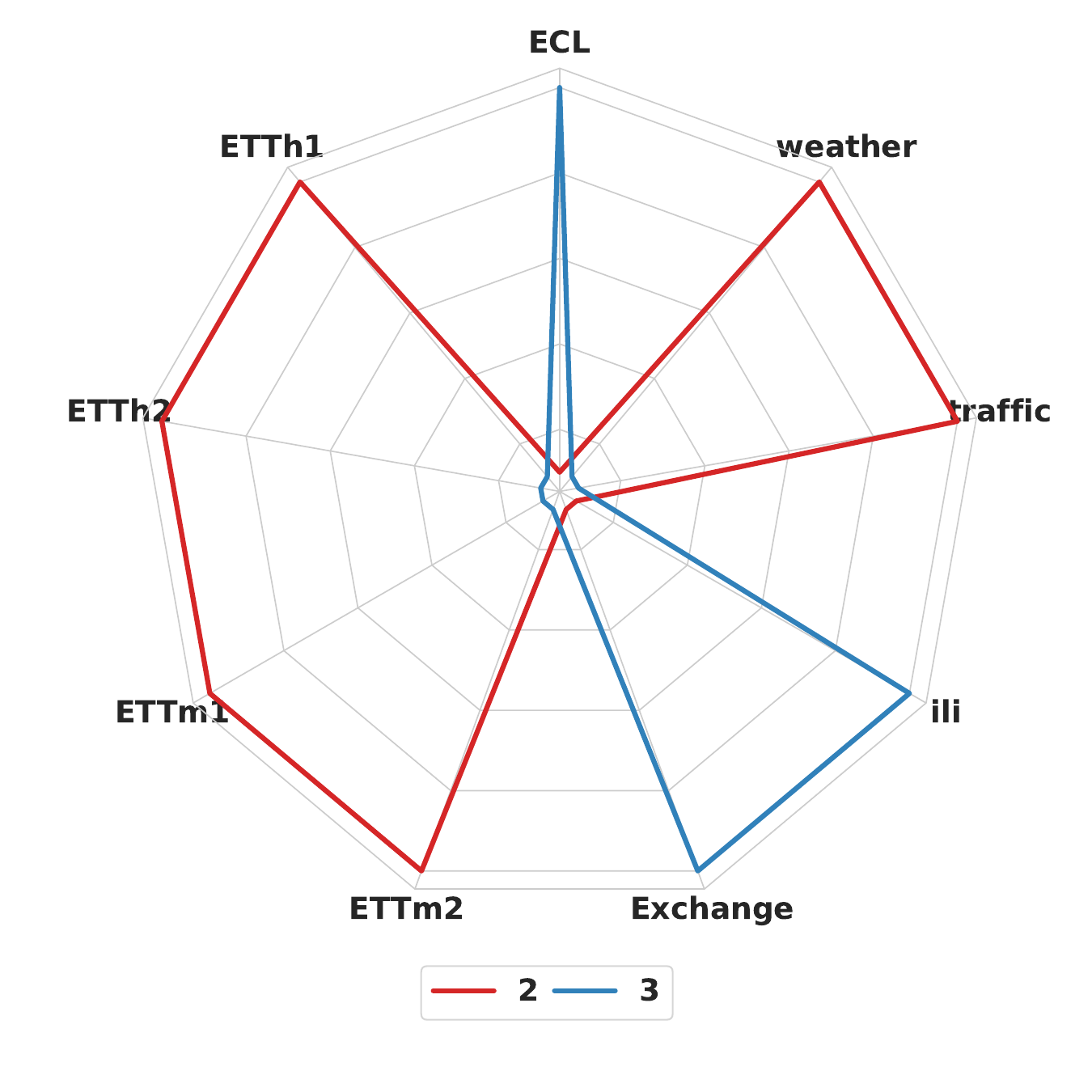}
         \caption{Encoder layers}
         \label{fig:exp-appx-el-mae}
     \end{subfigure}
     \hspace{10pt}
    \begin{subfigure}[t]{0.28\textwidth}
         \centering
         \includegraphics[width=\textwidth]{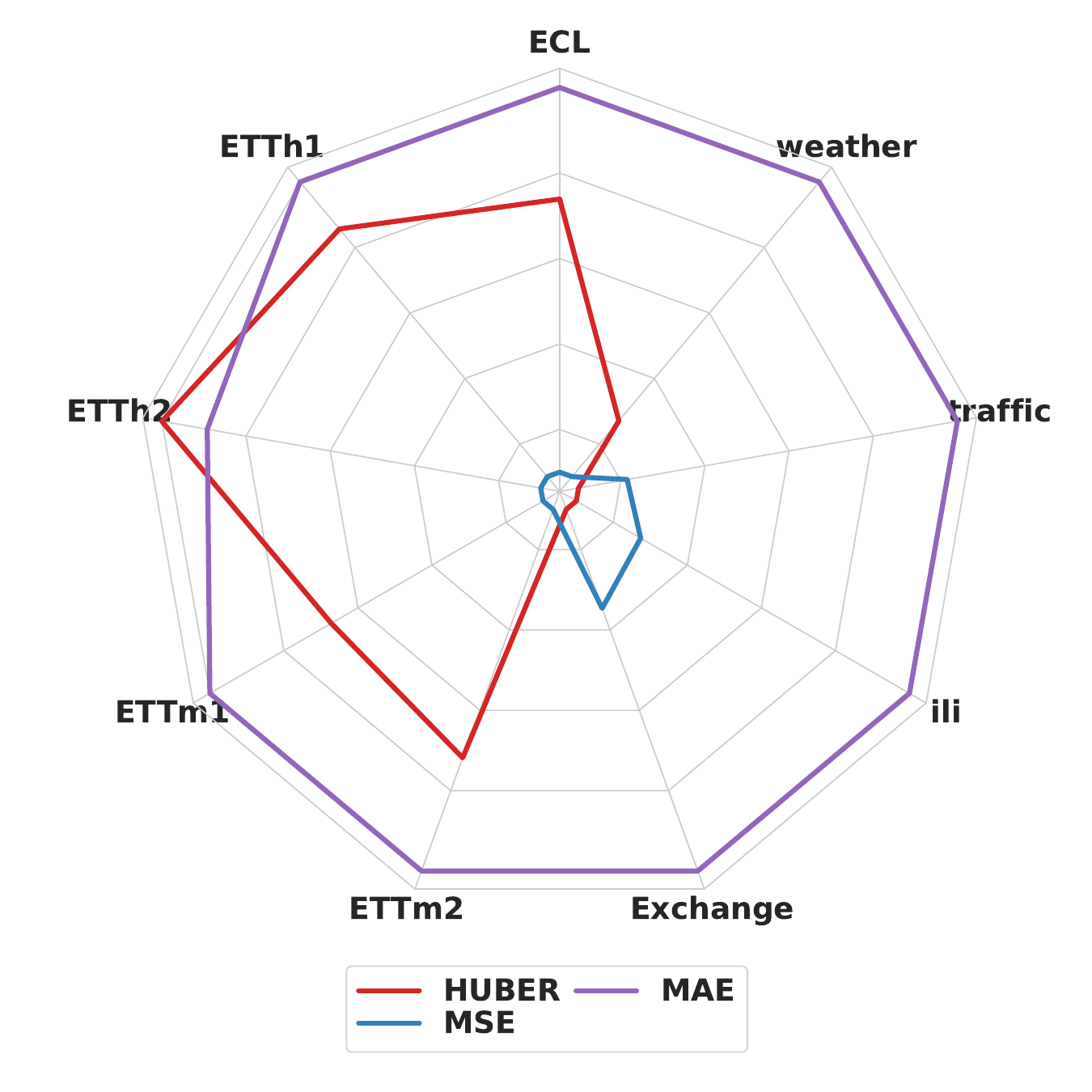}
         \caption{Loss Funtion}
         \label{fig:exp-appx-loss-mae}
     \end{subfigure}
     \hspace{10pt}
    \begin{subfigure}[t]{0.28\textwidth}
         \centering
         \includegraphics[width=\textwidth]{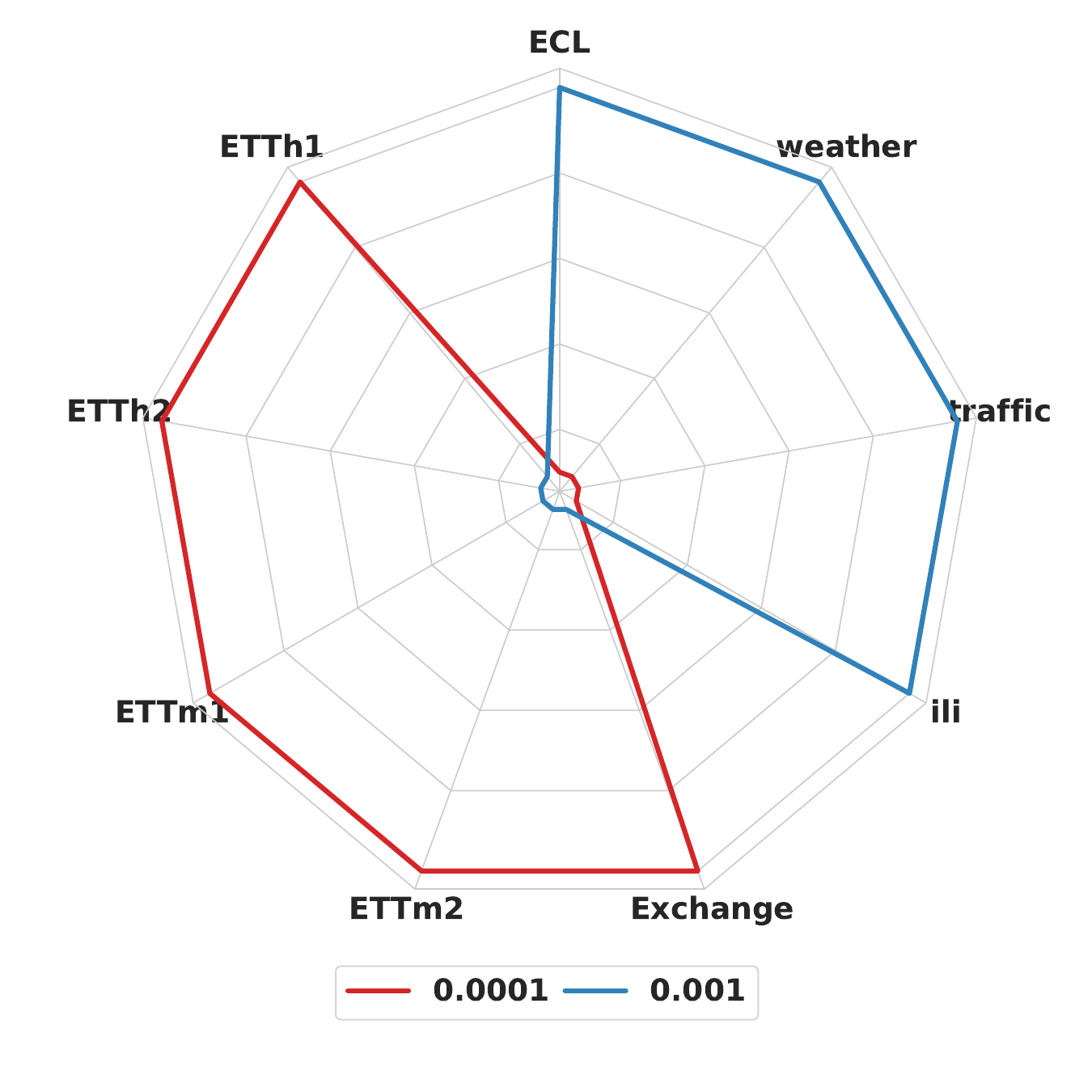}
         \caption{Learning Rate}
         \label{fig:exp-appx-lr-mae}
     \end{subfigure}
     \hspace{10pt}
    \begin{subfigure}[t]{0.28\textwidth}
         \centering
         \includegraphics[width=\textwidth]{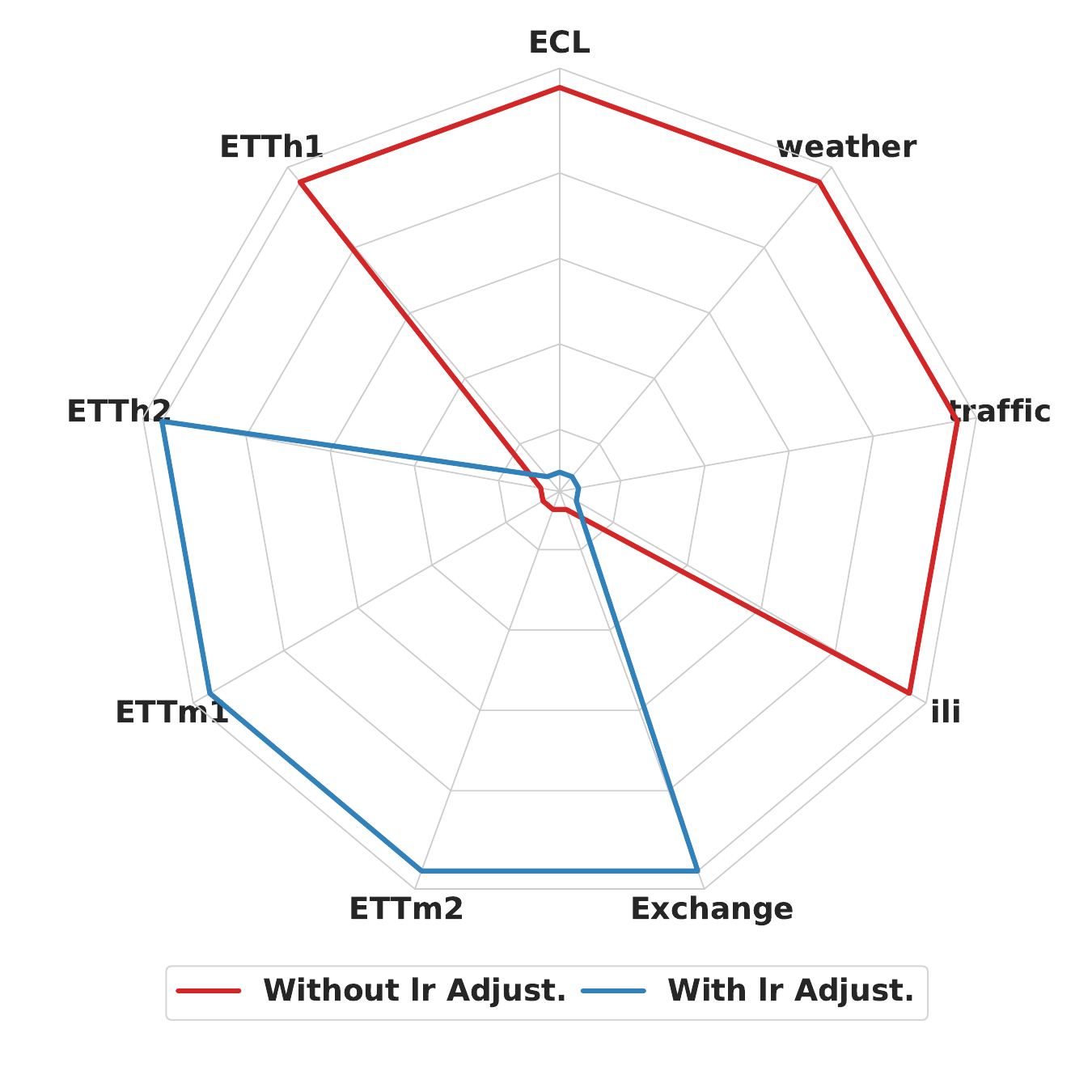}
         \caption{Learning Rate Strategy}
         \label{fig:exp-appx-lrs-mae}
     \end{subfigure}
     \hspace{10pt}
     \caption{Overall performance across key design dimensions in long-term forecasting. The results (\textbf{MAE}) are based on the top 25th percentile across all forecasting horizons.}
     \vspace{-0.1in}
     \label{fig:exp-appx-rada-mae}
\end{figure}

\begin{figure}[t!]
     \centering
     \begin{subfigure}[t]{0.28\textwidth}
         \centering
         \includegraphics[width=\textwidth]{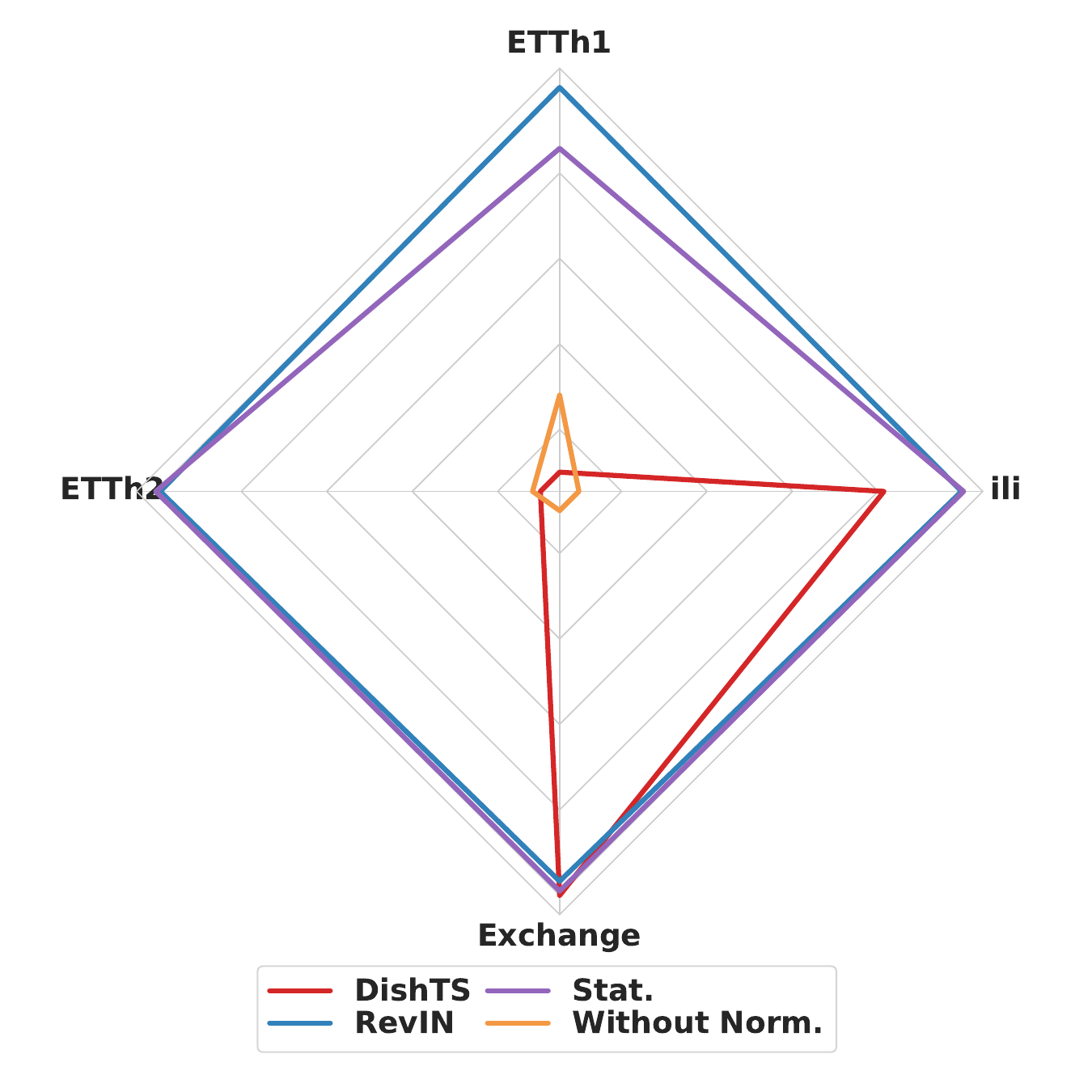}
         \caption{Series Normalization}
         \label{fig:exp-appx-Normalization-llm-mae}
     \end{subfigure}
     \hspace{10pt}
     \begin{subfigure}[t]{0.28\textwidth}
         \centering
         \includegraphics[width=\textwidth]{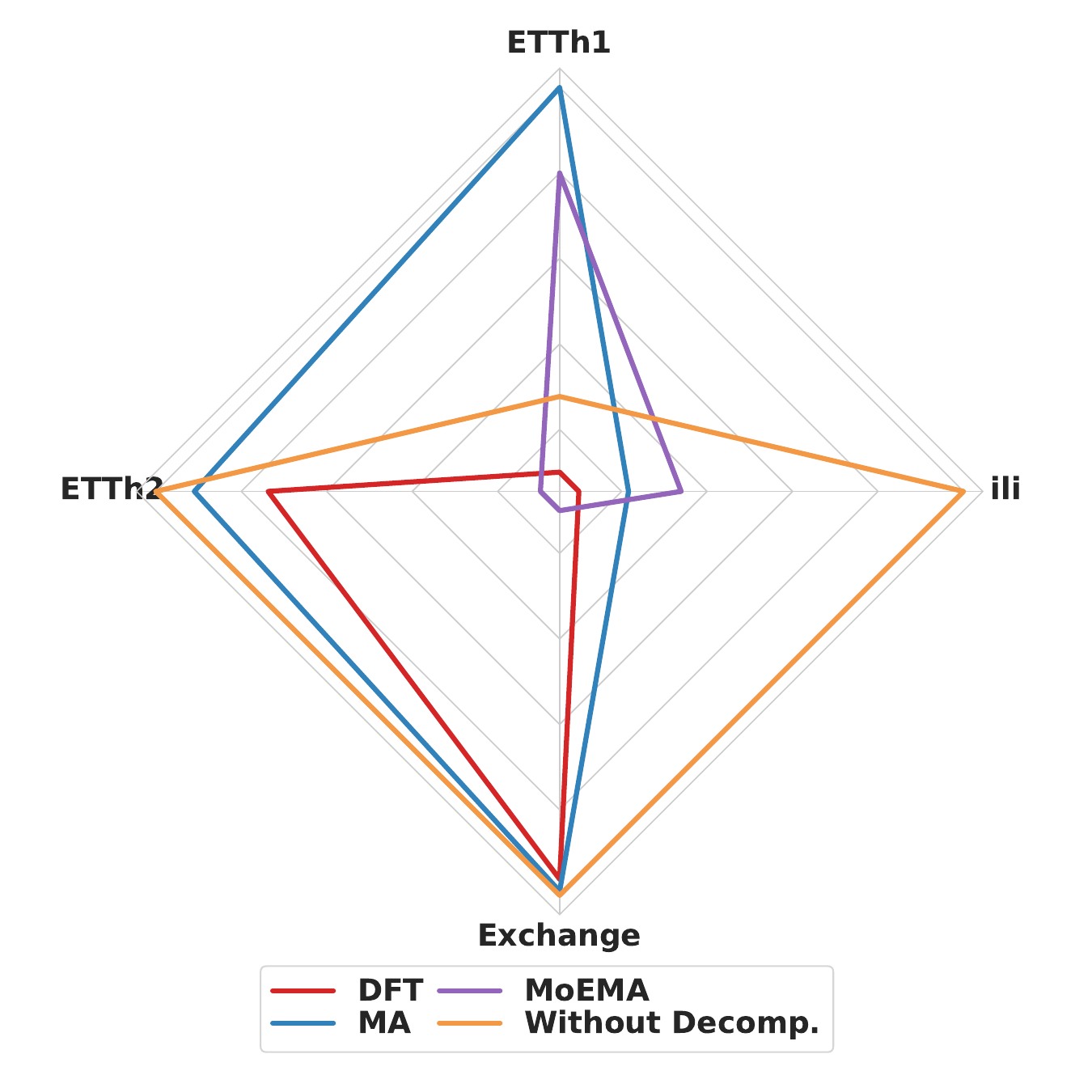}
         \caption{Series Decomposition}
         \label{fig:exp-appx-Decomposition-llm-mae}
     \end{subfigure}
     \hspace{10pt}
    \begin{subfigure}[t]{0.28\textwidth}
         \centering
         \includegraphics[width=\textwidth]{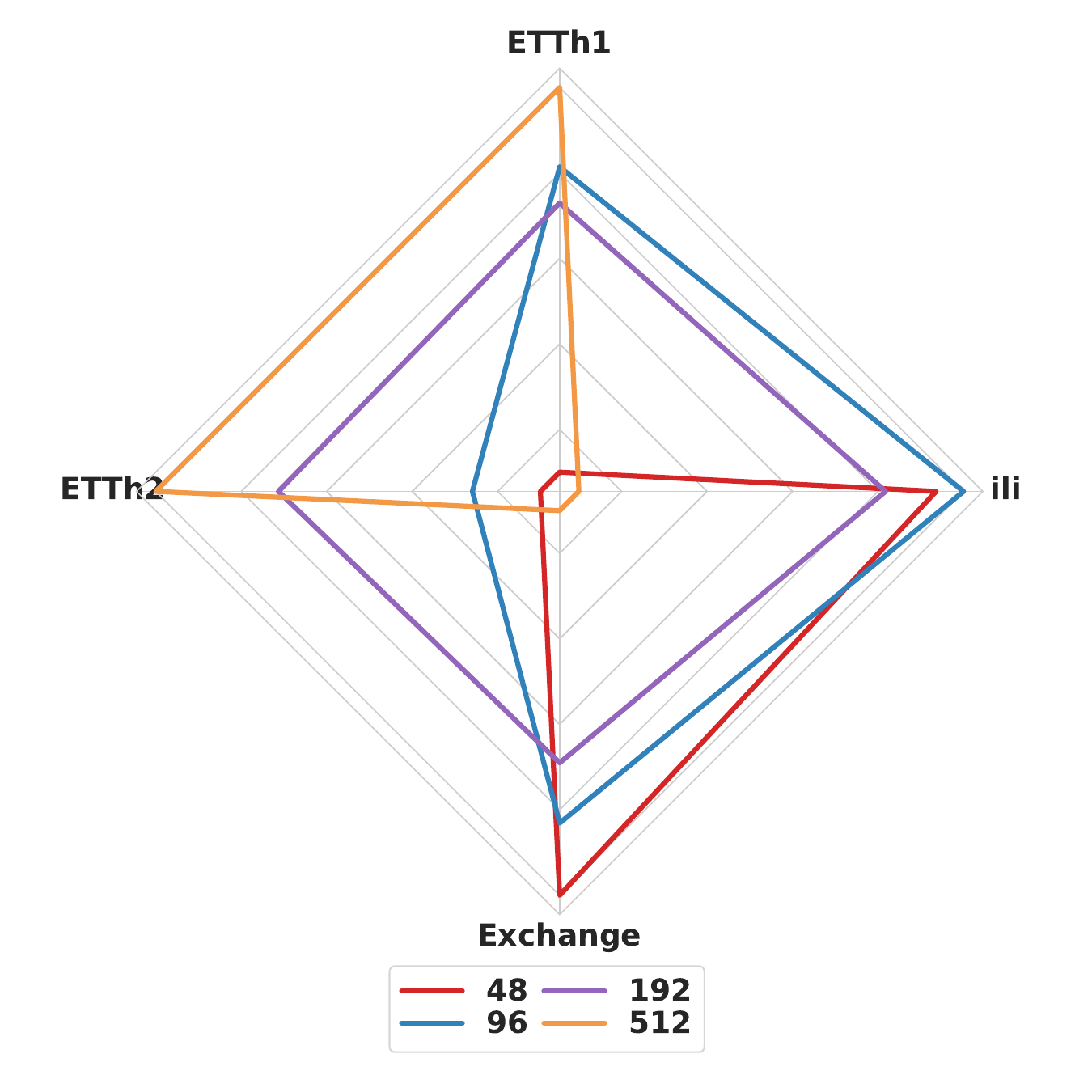}
         \caption{Sequence Length}
         \label{fig:exp-appx-sl-llm-mae}
     \end{subfigure}
     \hspace{10pt}
    \begin{subfigure}[t]{0.28\textwidth}
         \centering
         \includegraphics[width=\textwidth]{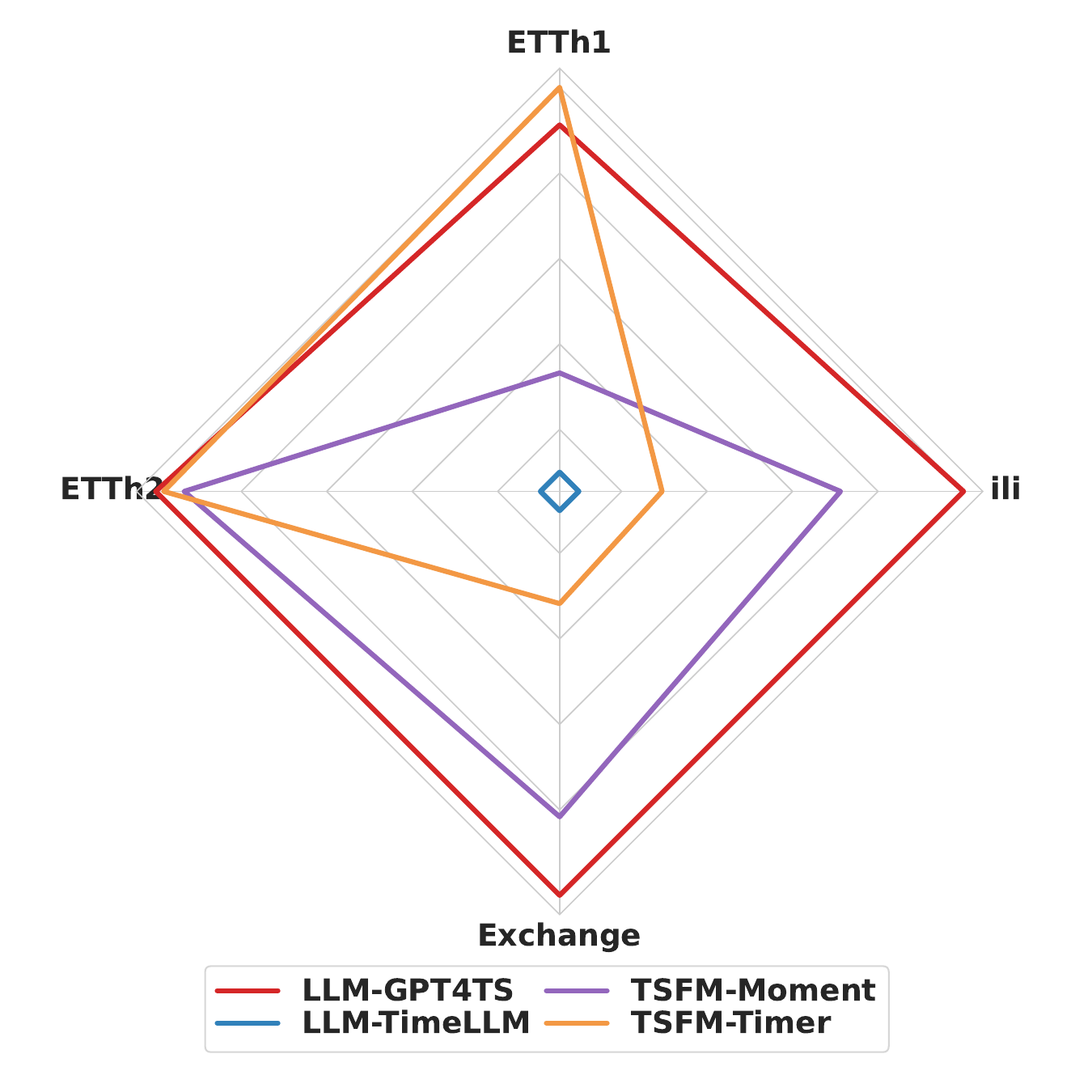}
         \caption{Network Backbone}
         \label{fig:exp-appx-backbone-llm-mae}
     \end{subfigure}
     \hspace{10pt}
    \begin{subfigure}[t]{0.28\textwidth}
         \centering
         \includegraphics[width=\textwidth]{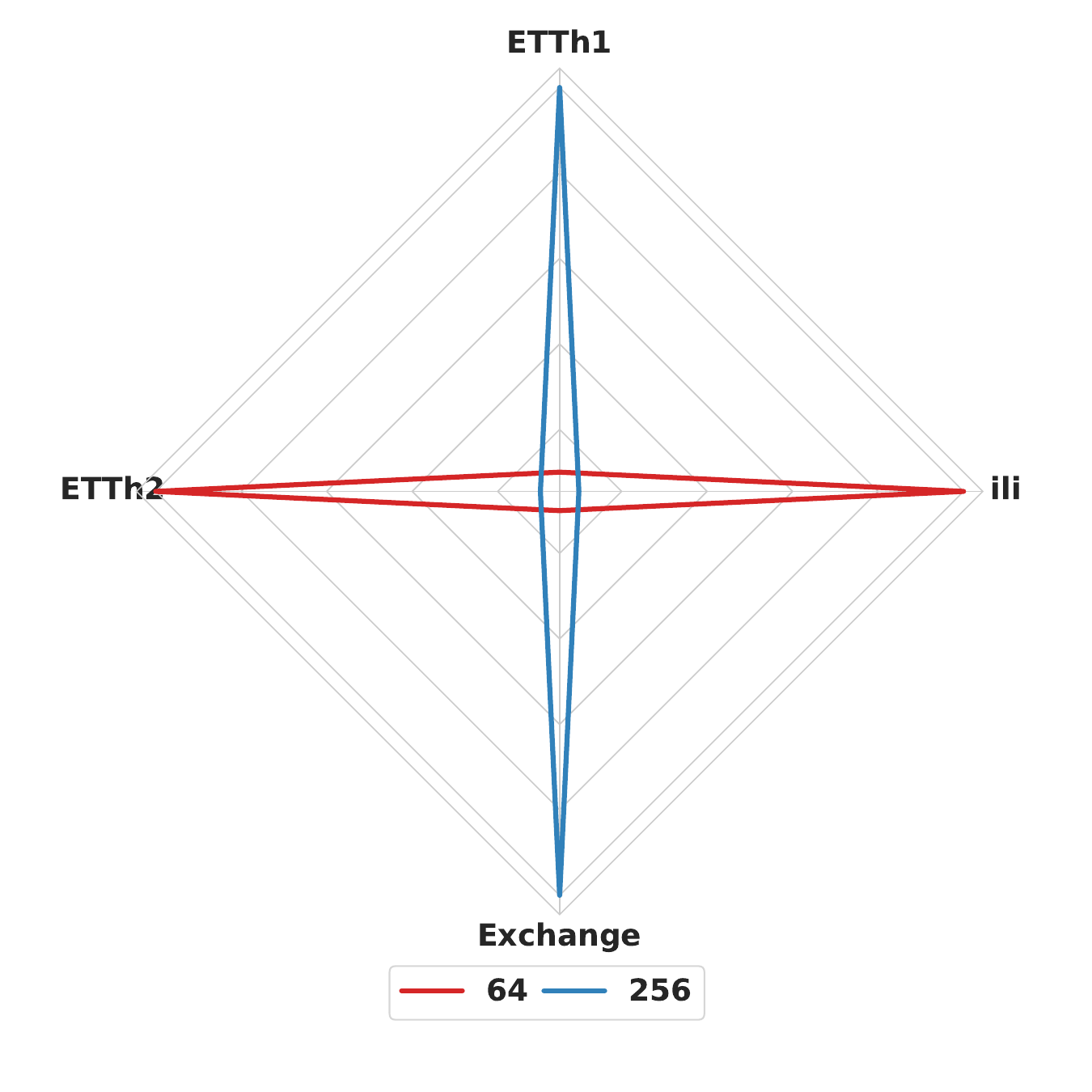}
         \caption{\textit{Hidden Layer Dimensions}}
         \label{fig:exp-appx-dmodel-llm-mae}
     \end{subfigure}
     \hspace{10pt}
    \begin{subfigure}[t]{0.28\textwidth}
         \centering
         \includegraphics[width=\textwidth]{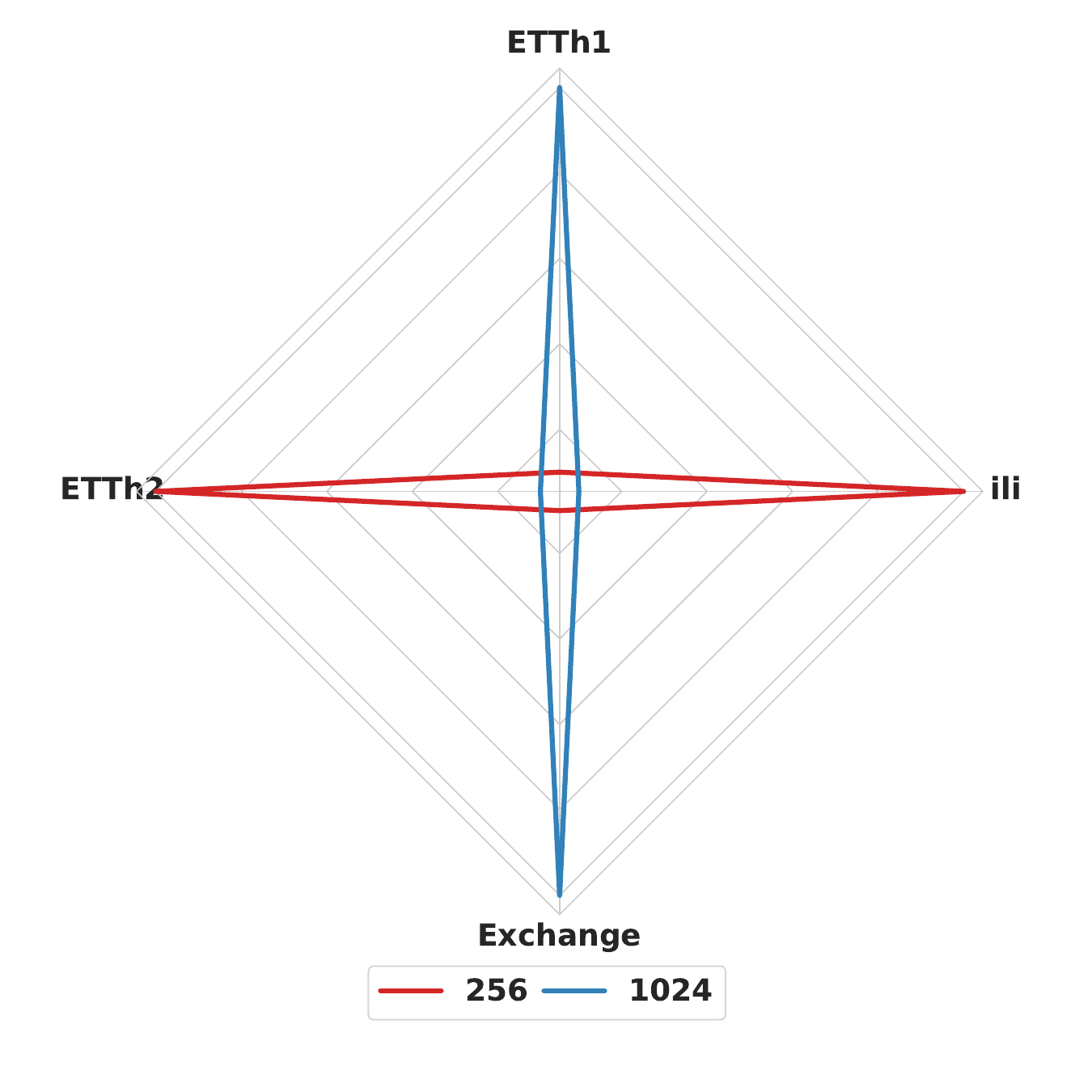}
         \caption{\textit{FCN Layer Dimensions}}
         \label{fig:exp-appx-dff-llm-mae}
     \end{subfigure}
     \hspace{10pt}
    \begin{subfigure}[t]{0.28\textwidth}
         \centering
         \includegraphics[width=\textwidth]{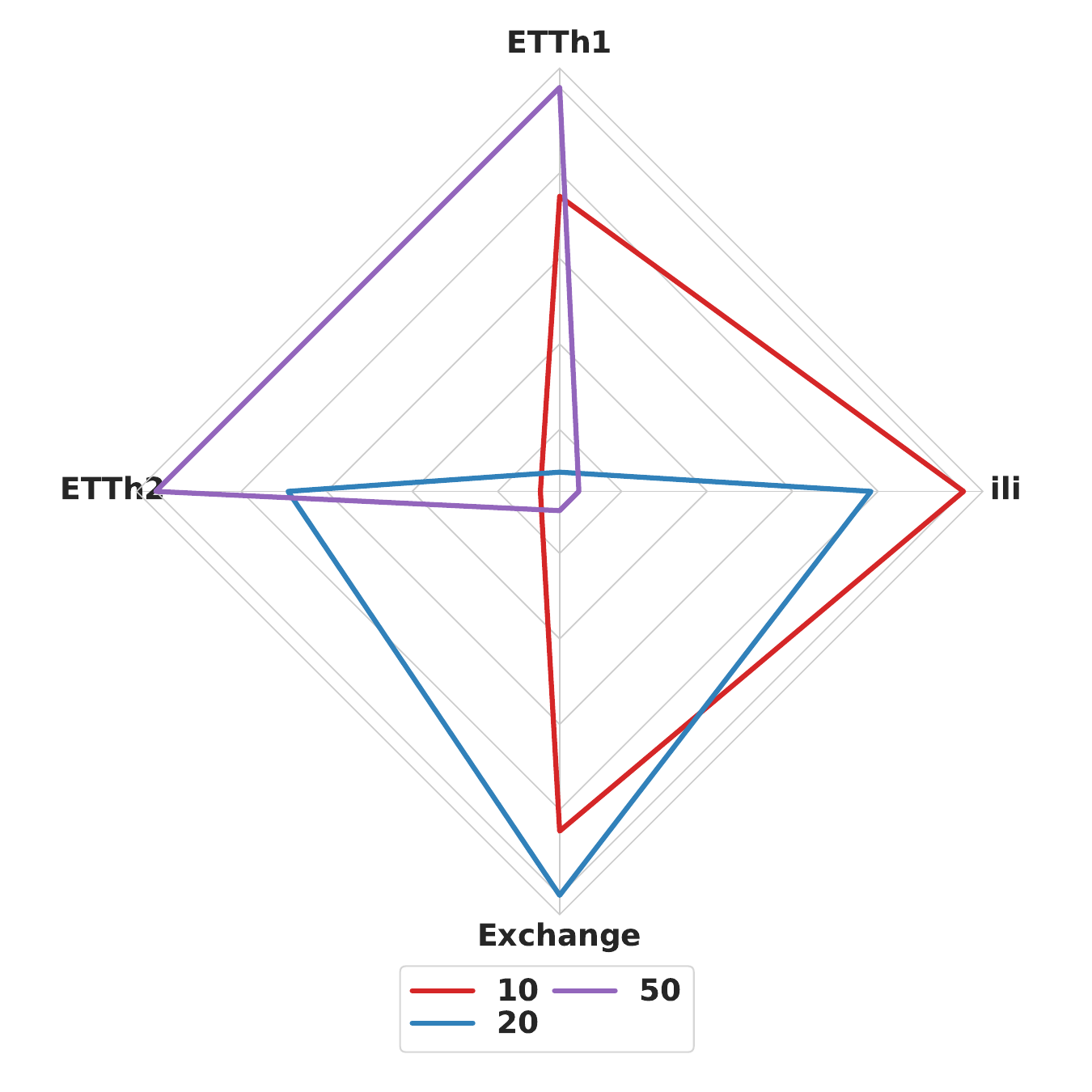}
         \caption{Epochs}
         \label{fig:exp-appx-epochs-llm-mae}
     \end{subfigure}
     \hspace{10pt}
    \begin{subfigure}[t]{0.28\textwidth}
         \centering
         \includegraphics[width=\textwidth]{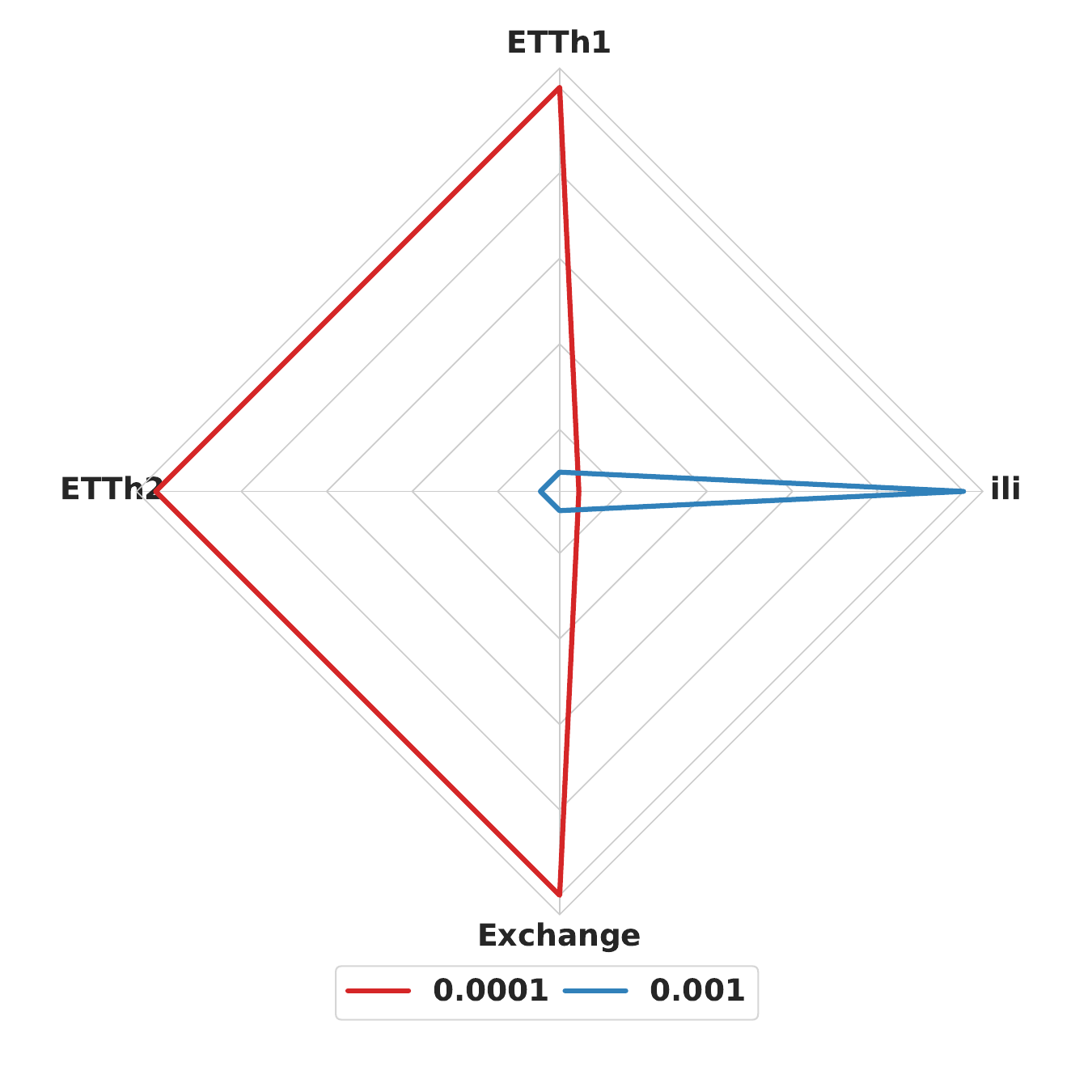}
         \caption{Learning Rate}
         \label{fig:exp-appx-lr-llm-mae}
     \end{subfigure}
     \hspace{10pt}
    \begin{subfigure}[t]{0.28\textwidth}
         \centering
         \includegraphics[width=\textwidth]{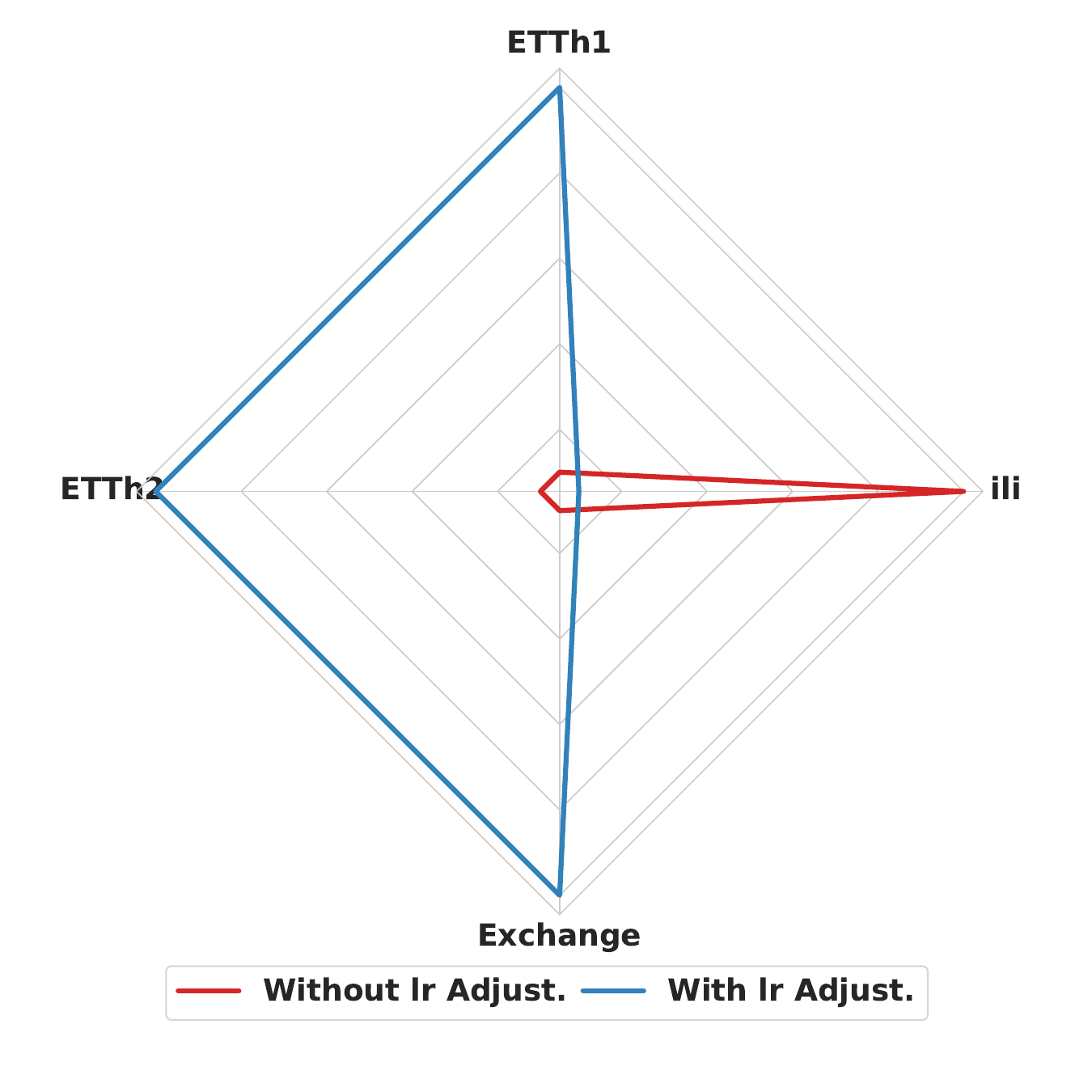}
         \caption{Learning Rate Strategy}
         \label{fig:exp-appx-lrs-llm-mae}
     \end{subfigure}
     \hspace{10pt}
     \caption{Overall performance across all design dimensions when using LLMs or TSFMs in long-term forecasting. The results (\textbf{MAE}) are based on the top 25th percentile across all forecasting horizons.}
     \vspace{-0.1in}
     \label{fig:exp-appx-rada-llm-mae}
\end{figure}

For the MAE-based performance evaluation, we analyze the effects of different design choices using both spider charts and box plots (Fig.~\ref{fig:exp-appx-rada-mae} and Fig.~\ref{fig:exp-appx-rada-llm-mae}). These visualizations complement the MSE-based analysis and confirm the generalizability of our findings across error metrics. In particular, normalization methods such as RevIN and Stationary consistently achieve the lowest MAE values, underscoring their effectiveness in mitigating non-stationarity. Similarly, decomposition strategies exhibit selective benefits: MA-based methods improve predictions on datasets like ETTh1 and ETTm2, while raw-series modeling remains more effective on ECL and Traffic, where decomposition tends to degrade performance.

Beyond preprocessing, MAE evaluations further validate the consistency of our architectural insights. Channel-independent designs retain strong performance across most datasets, except on Traffic and ILI, where localized dependencies dominate. Tokenization methods show stable ranking across both metrics, with patch-wise encoding consistently outperforming point-wise approaches. Notably, complex architectures such as Transformers provide only marginal gains over MLPs in certain cases (e.g., Traffic), suggesting that their benefits may not justify the added complexity. Overall, the alignment between MAE and MSE results reinforces the robustness of our design principles, demonstrating that the observed patterns are not metric-specific but instead reflect core relationships between architecture and forecasting performance.


\begin{figure}[t!]
     \centering
     \begin{subfigure}[t]{0.28\textwidth}
         \centering
         \includegraphics[width=\textwidth]{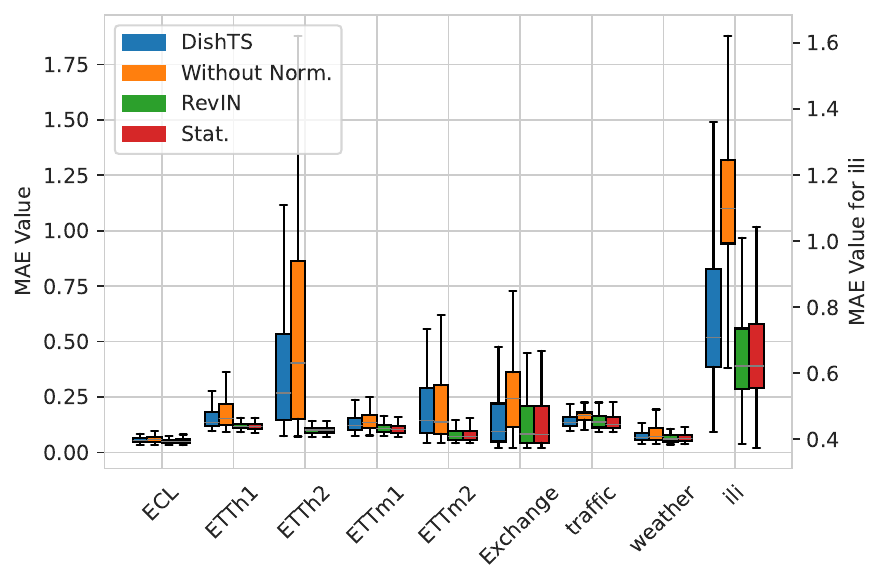}
         \caption{Series Normalization}
         \label{fig:exp-appx-Normalization-ltf_bp_mae}
     \end{subfigure}
     \hspace{10pt}
     \begin{subfigure}[t]{0.28\textwidth}
         \centering
         \includegraphics[width=\textwidth]{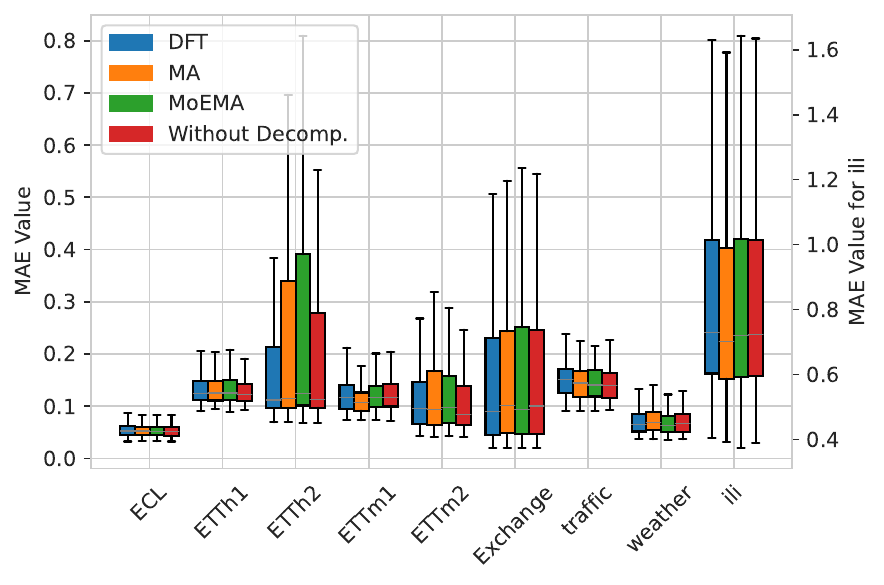}
         \caption{Series Decomposition}
         \label{fig:exp-appx-Decomposition-ltf_bp_mae}
     \end{subfigure}
     \hspace{10pt}
    \begin{subfigure}[t]{0.28\textwidth}
         \centering
         \includegraphics[width=\textwidth]{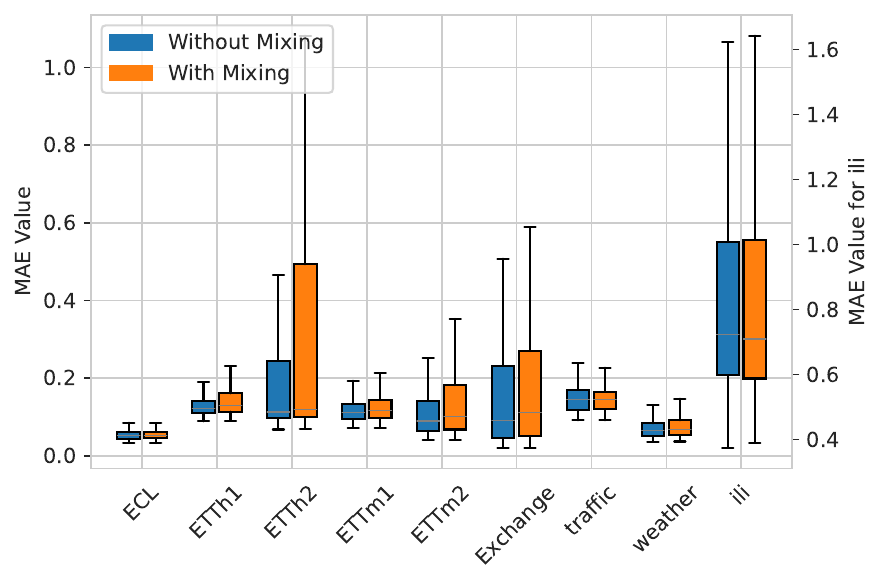}
         \caption{Series Sampling/Mixing}
         \label{fig:exp-appx-mixing-ltf_bp_mae}
     \end{subfigure}
     \hspace{10pt}
    \begin{subfigure}[t]{0.28\textwidth}
         \centering
         \includegraphics[width=\textwidth]{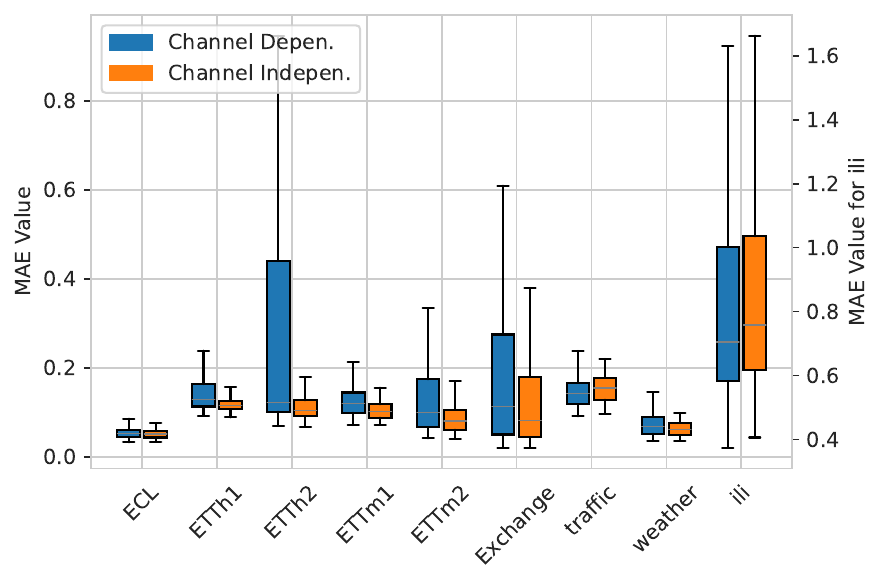}
         \caption{Channel Independent}
         \label{fig:exp-appx-CI-ltf_bp_mae}
     \end{subfigure}
     \hspace{10pt}
    \begin{subfigure}[t]{0.28\textwidth}
         \centering
         \includegraphics[width=\textwidth]{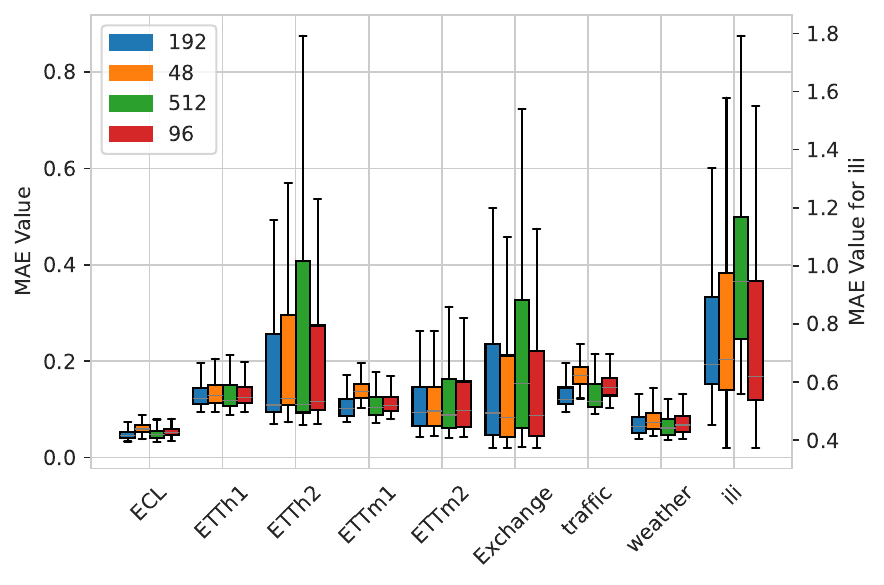}
         \caption{Sequence Length}
         \label{fig:exp-appx-sl-ltf_bp_mae}
     \end{subfigure}
     \hspace{10pt}
    \begin{subfigure}[t]{0.28\textwidth}
         \centering
         \includegraphics[width=\textwidth]{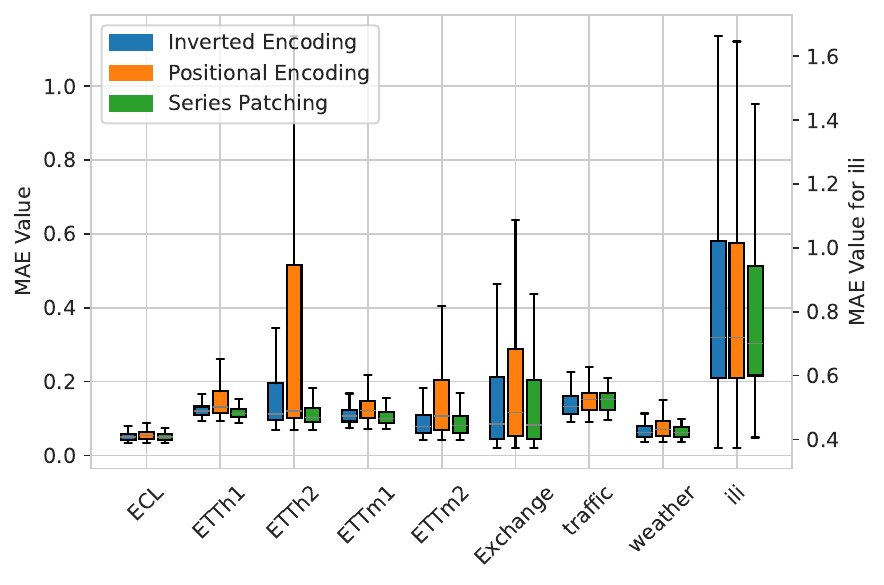}
         \caption{Series Embedding}
         \label{fig:exp-appx-tokenization-ltf_bp_mae}
     \end{subfigure}
     \hspace{10pt}
     \begin{subfigure}[t]{0.28\textwidth}
         \centering
         \includegraphics[width=\textwidth]{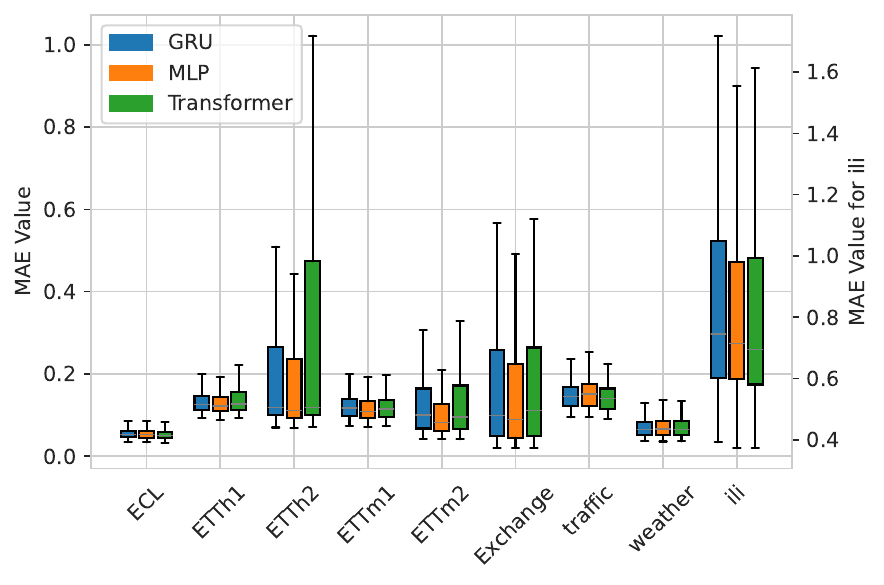}
         \caption{Network Backbone}
         \label{fig:exp-appx-backbone-ltf_bp_mae}
     \end{subfigure}
     \hspace{10pt}
    \begin{subfigure}[t]{0.28\textwidth}
         \centering
         \includegraphics[width=\textwidth]{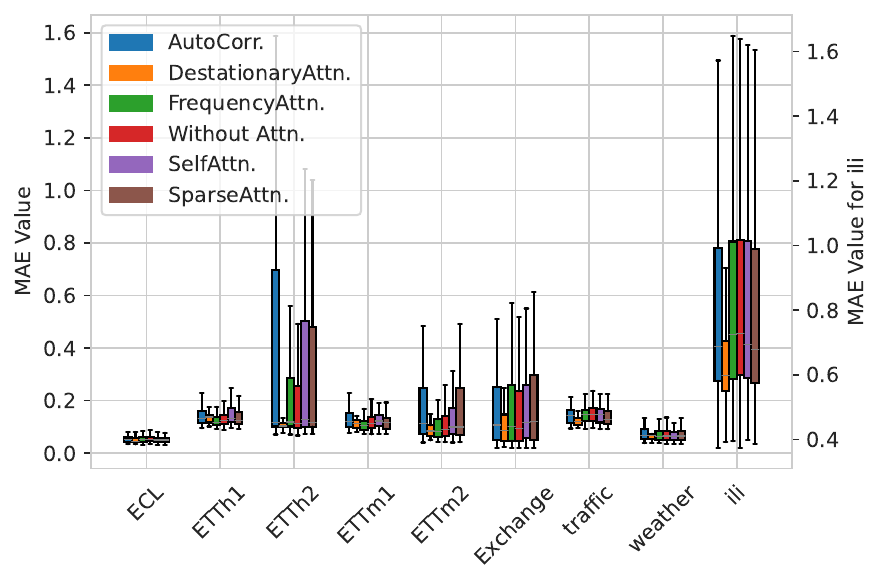}
         \caption{Series Attention}
         \label{fig:exp-appx-attention-ltf_bp_mae}
     \end{subfigure}
     \hspace{10pt}
         \begin{subfigure}[t]{0.28\textwidth}
         \centering
         \includegraphics[width=\textwidth]{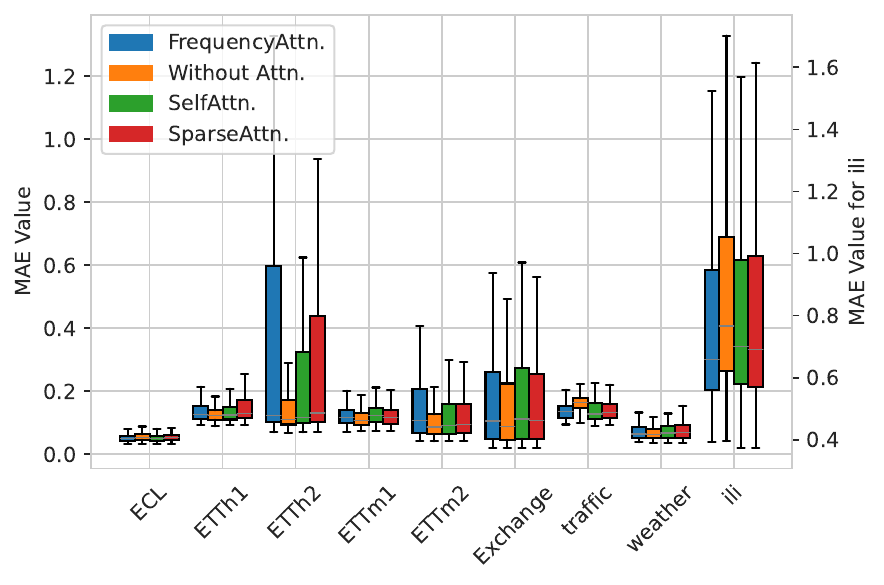}
         \caption{Feature Attention}
         \label{fig:exp-appx-featureattention-ltf_bp_mae}
     \end{subfigure}
     \hspace{10pt}
    \begin{subfigure}[t]{0.28\textwidth}
         \centering
         \includegraphics[width=\textwidth]{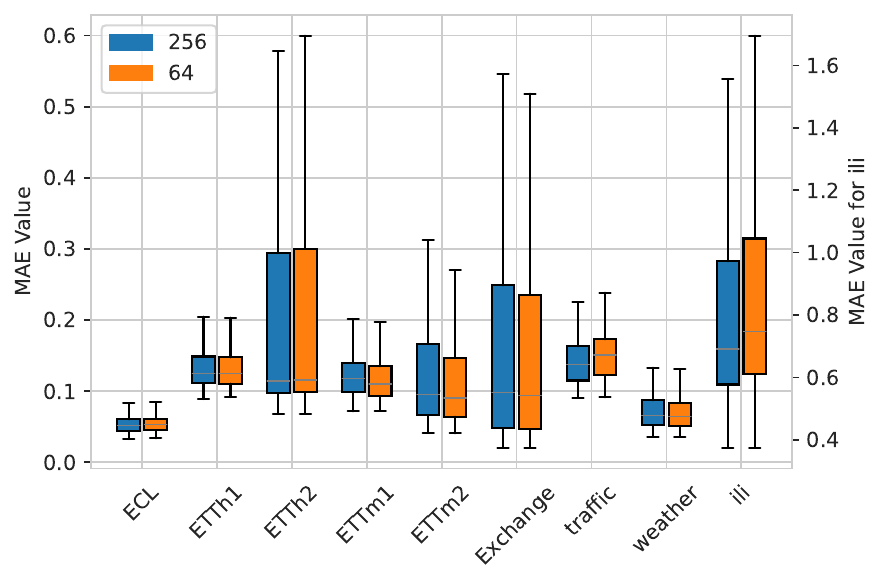}
         \caption{\textit{Hidden Layer Dimensions}}
         \label{fig:exp-appx-dmodel-ltf_bp_mae}
     \end{subfigure}
     \hspace{10pt}
    \begin{subfigure}[t]{0.28\textwidth}
         \centering
         \includegraphics[width=\textwidth]{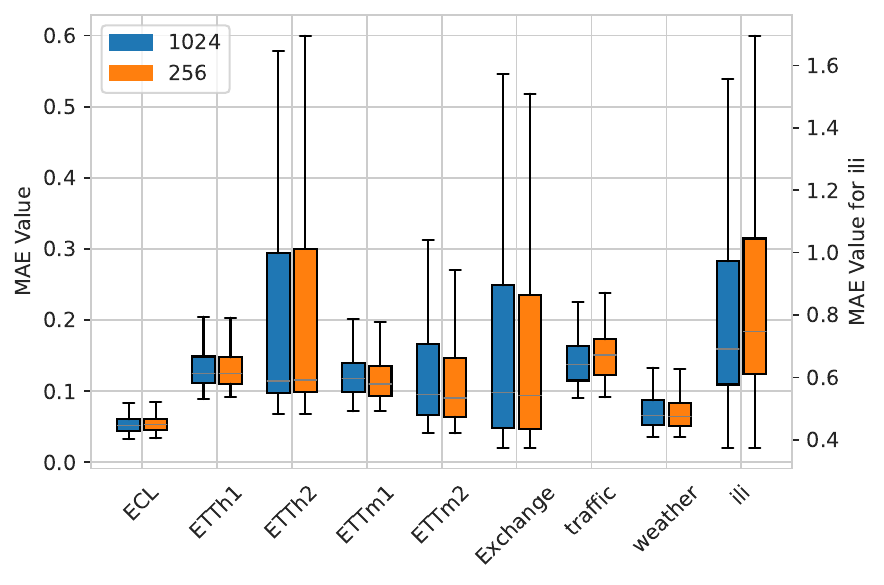}
         \caption{\textit{FCN Layer Dimensions}}
         \label{fig:exp-appx-dff-ltf_bp_mae}
     \end{subfigure}
     \hspace{10pt}
    \begin{subfigure}[t]{0.28\textwidth}
         \centering
         \includegraphics[width=\textwidth]{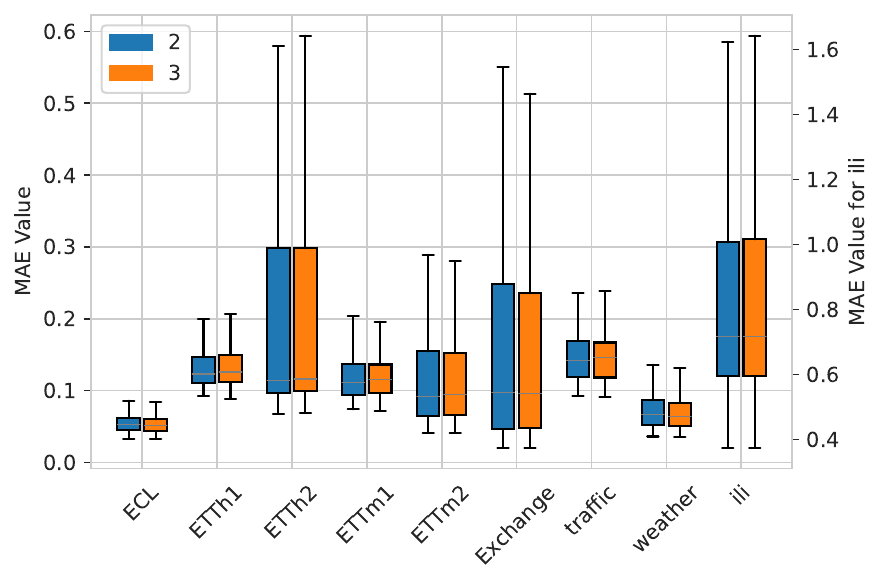}
         \caption{Encoder layers}
         \label{fig:exp-appx-el-ltf_bp_mae}
     \end{subfigure}
     \hspace{10pt}
    \begin{subfigure}[t]{0.28\textwidth}
         \centering
         \includegraphics[width=\textwidth]{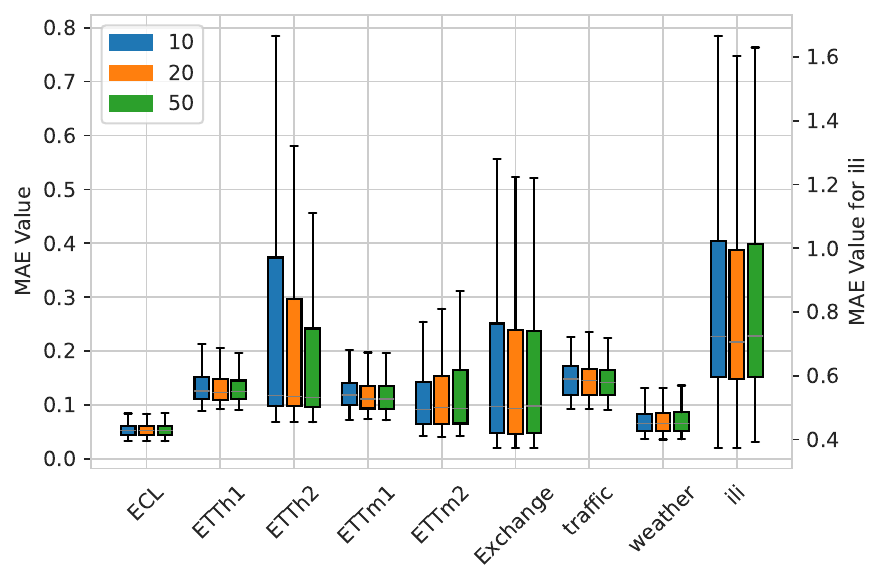}
         \caption{Epochs}
         \label{fig:exp-appx-epochs-ltf_bp_mae}
     \end{subfigure}
     \hspace{10pt}
    \begin{subfigure}[t]{0.28\textwidth}
         \centering
         \includegraphics[width=\textwidth]{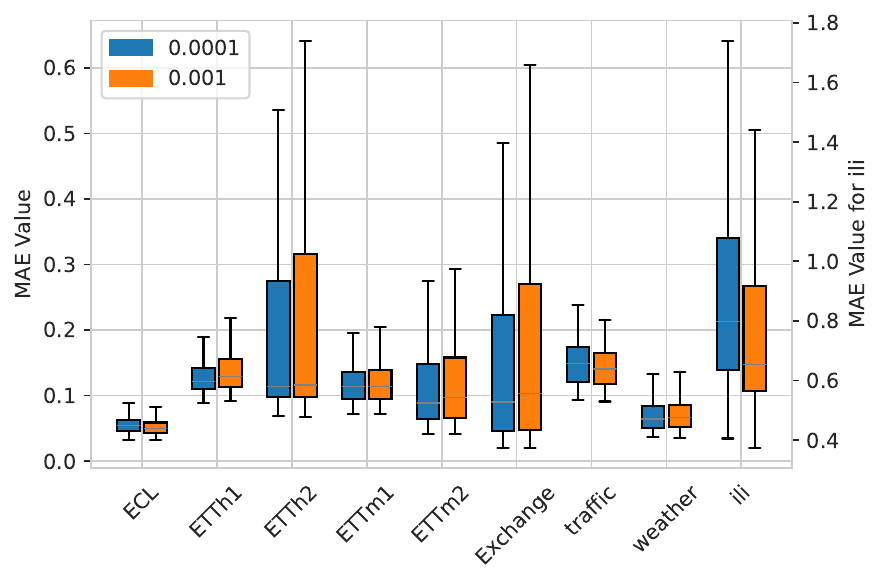}
         \caption{Learning Rate}
         \label{fig:exp-appx-lr-ltf_bp_mae}
     \end{subfigure}
     \hspace{10pt}
    \begin{subfigure}[t]{0.28\textwidth}
         \centering
         \includegraphics[width=\textwidth]{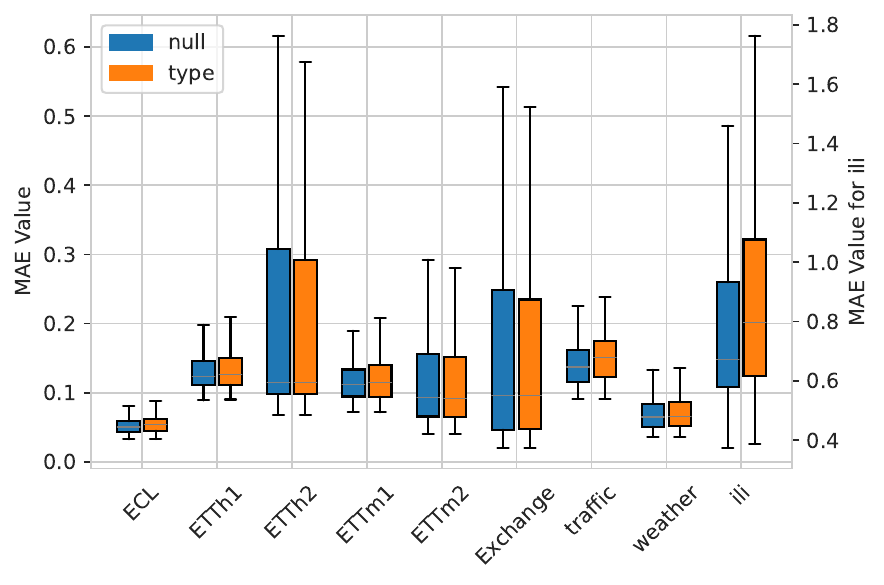}
         \caption{Learning Rate Strategy}
         \label{fig:exp-appx-lrs-ltf_bp_mae}
     \end{subfigure}
     \hspace{10pt}
         \begin{subfigure}[t]{0.28\textwidth}
         \centering
         \includegraphics[width=\textwidth]{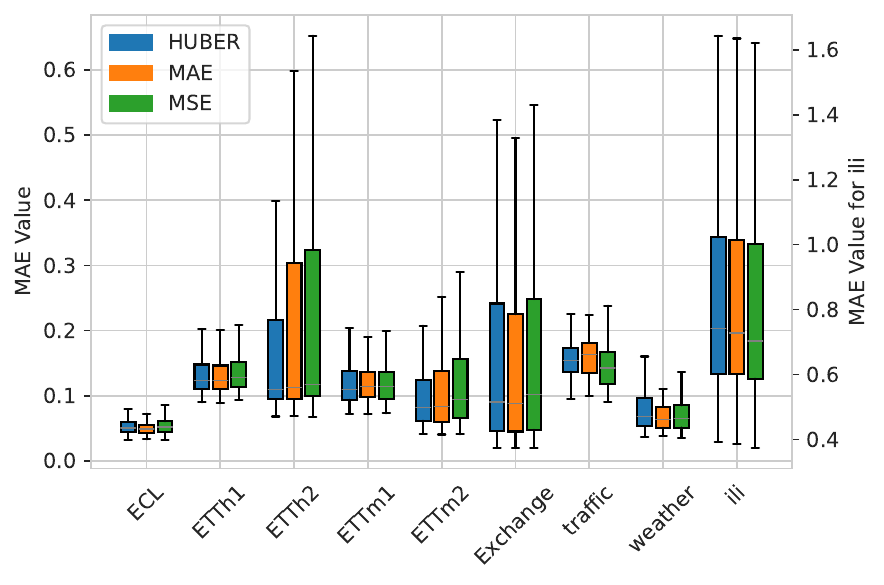}
         \caption{Loss Function}
         \label{fig:exp-appx-lf-ltf_bp_mae}
     \end{subfigure}
     \hspace{10pt}
         \begin{subfigure}[t]{0.28\textwidth}
         \centering
         \includegraphics[width=\textwidth]{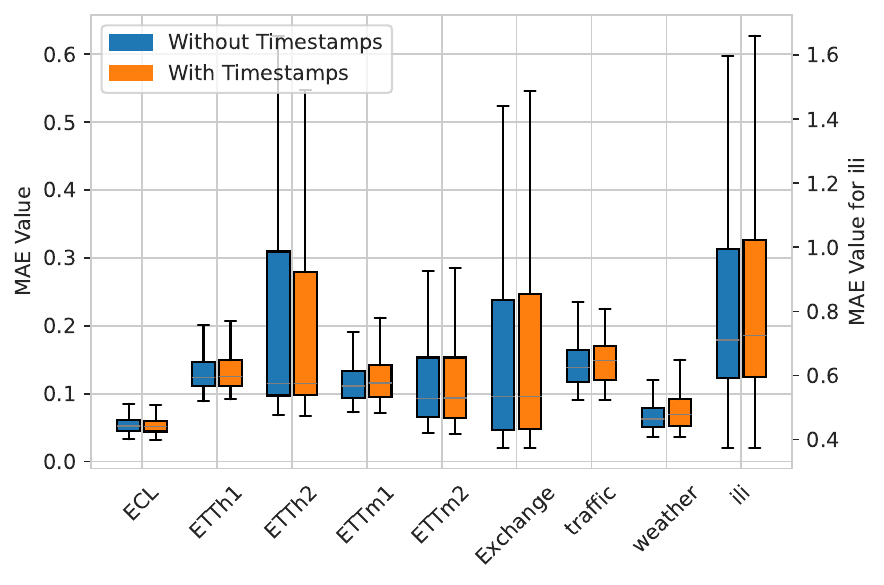}
         \caption{Timestamp}
         \label{fig:exp-appx-xm-ltf_bp_mae}
     \end{subfigure}
     \hspace{10pt}
     \caption{Overall performance across all design dimensions in long-term forecasting. The results (\textbf{MAE}) are averaged across all forecasting horizons.}
     \vspace{-0.1in}
     \label{fig:exp-appx-bp-ltf_mae}
\end{figure}

\begin{figure}[t!]
     \centering
     \begin{subfigure}[t]{0.28\textwidth}
         \centering
         \includegraphics[width=\textwidth]{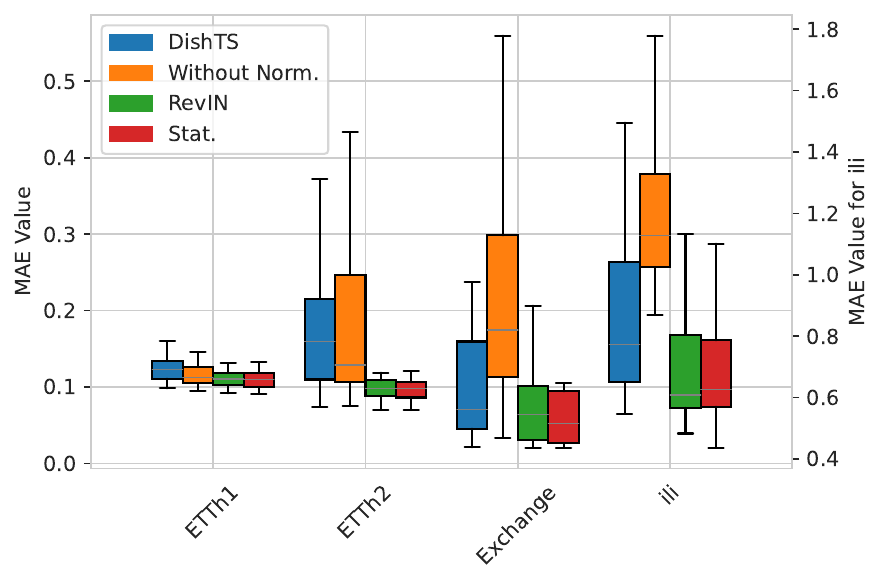}
         \caption{Series Normalization}
         \label{fig:exp-appx-Normalization-llm_bp_mae}
     \end{subfigure}
     \hspace{10pt}
     \begin{subfigure}[t]{0.28\textwidth}
         \centering
         \includegraphics[width=\textwidth]{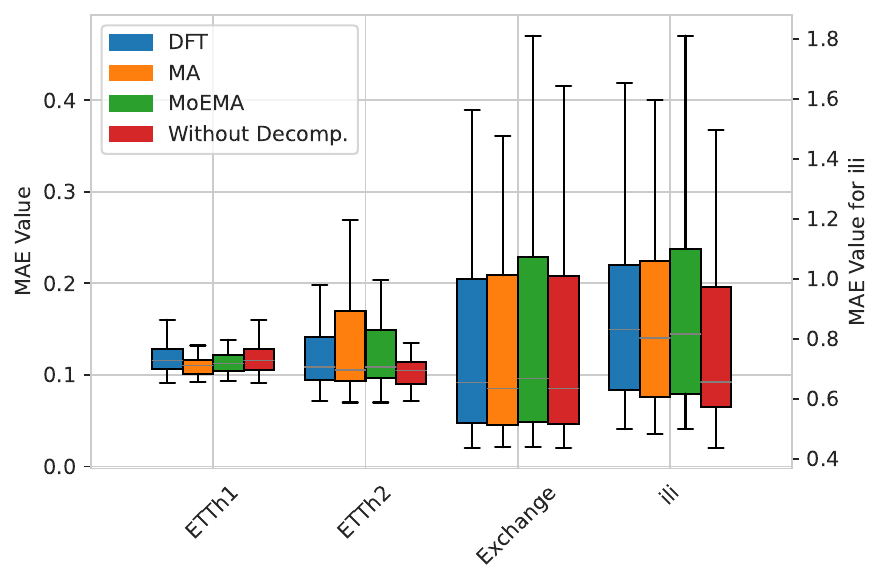}
         \caption{Series Decomposition}
         \label{fig:exp-appx-Decomposition-llm_bp_mae}
     \end{subfigure}
     \hspace{10pt}
    \begin{subfigure}[t]{0.28\textwidth}
         \centering
         \includegraphics[width=\textwidth]{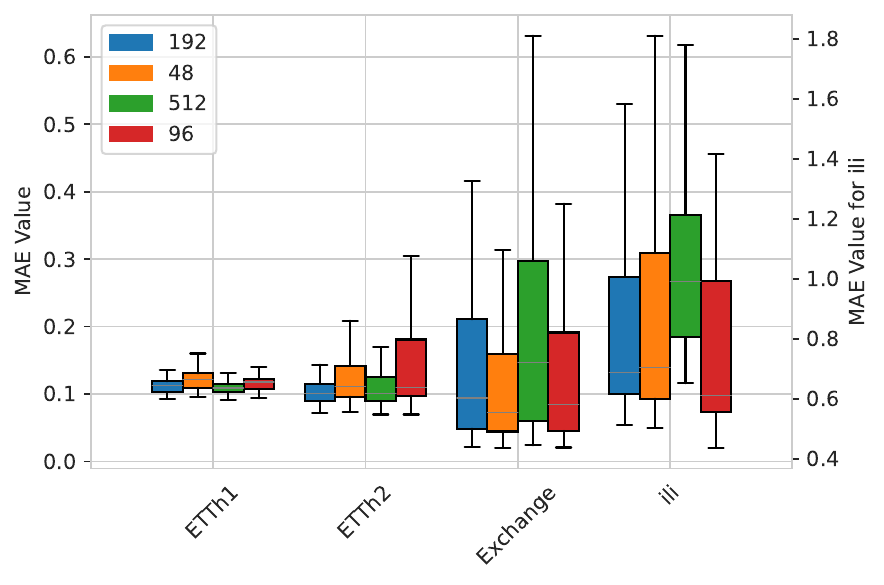}
         \caption{Sequence Length}
         \label{fig:exp-appx-sl-llm_bp_mae}
     \end{subfigure}
     \hspace{10pt}
    \begin{subfigure}[t]{0.28\textwidth}
         \centering
         \includegraphics[width=\textwidth]{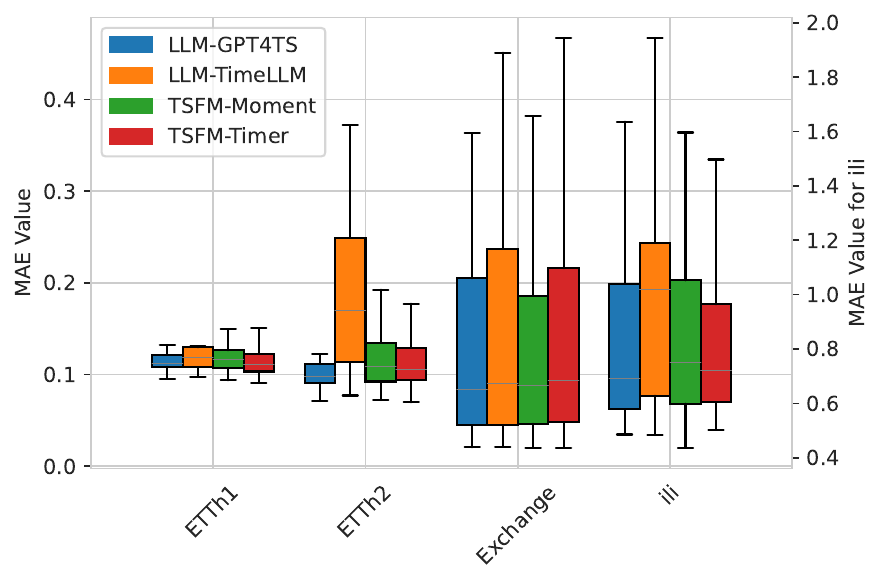}
         \caption{Network Backbone}
         \label{fig:exp-appx-backbone-llm_bp_mae}
     \end{subfigure}
     \hspace{10pt}
    \begin{subfigure}[t]{0.28\textwidth}
         \centering
         \includegraphics[width=\textwidth]{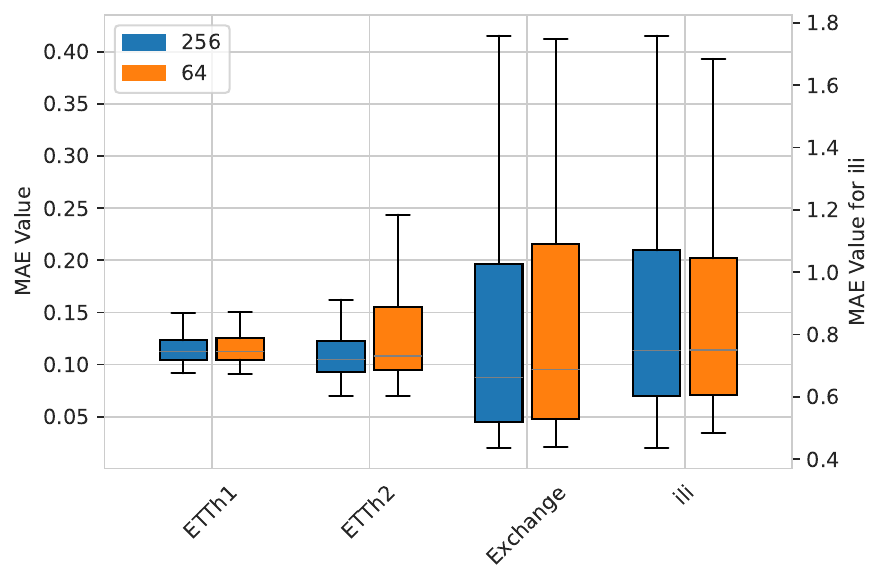}
         \caption{\textit{Hidden Layer Dimensions}}
         \label{fig:exp-appx-dmodel-llm_bp_mae}
     \end{subfigure}
     \hspace{10pt}
    \begin{subfigure}[t]{0.28\textwidth}
         \centering
         \includegraphics[width=\textwidth]{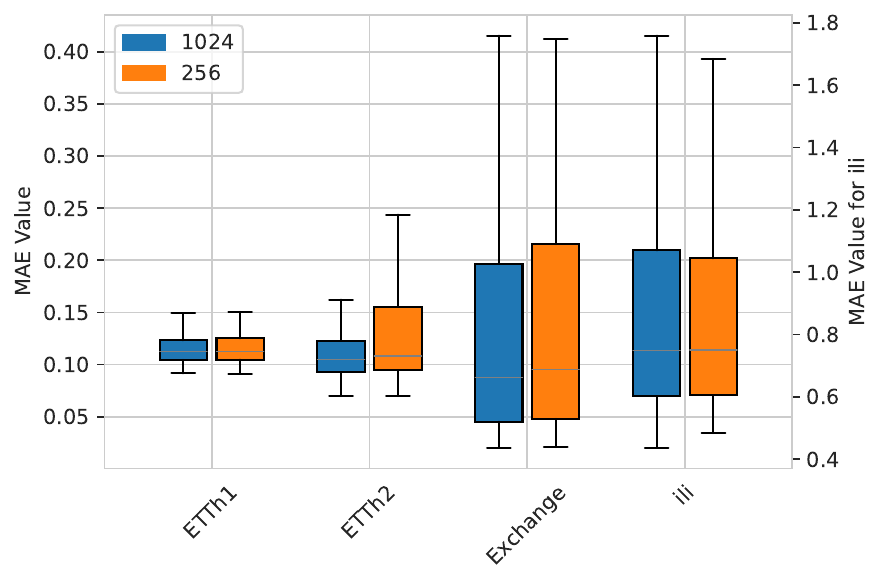}
         \caption{\textit{FCN Layer Dimensions}}
         \label{fig:exp-appx-dff-llm_bp_mae}
     \end{subfigure}
     \hspace{10pt}
    \begin{subfigure}[t]{0.28\textwidth}
         \centering
         \includegraphics[width=\textwidth]{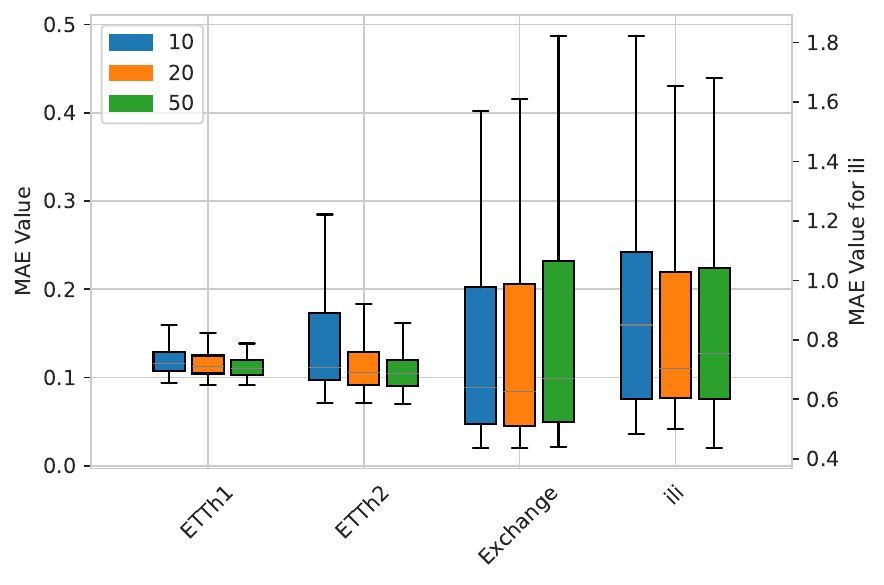}
         \caption{Epochs}
         \label{fig:exp-appx-epochs-llm_bp_mae}
     \end{subfigure}
     \hspace{10pt}
    \begin{subfigure}[t]{0.28\textwidth}
         \centering
         \includegraphics[width=\textwidth]{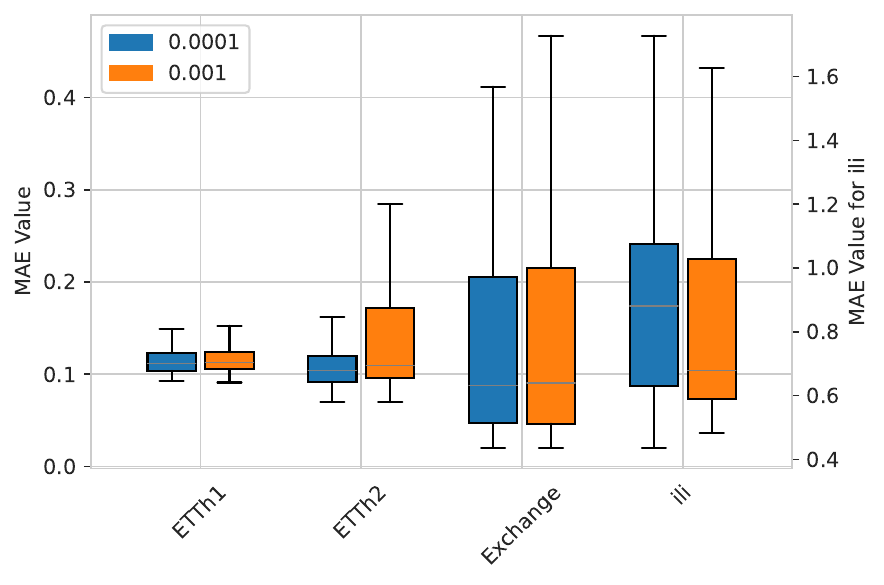}
         \caption{Learning Rate}
         \label{fig:exp-appx-lr-llm_bp_mae}
     \end{subfigure}
     \hspace{10pt}
    \begin{subfigure}[t]{0.28\textwidth}
         \centering
         \includegraphics[width=\textwidth]{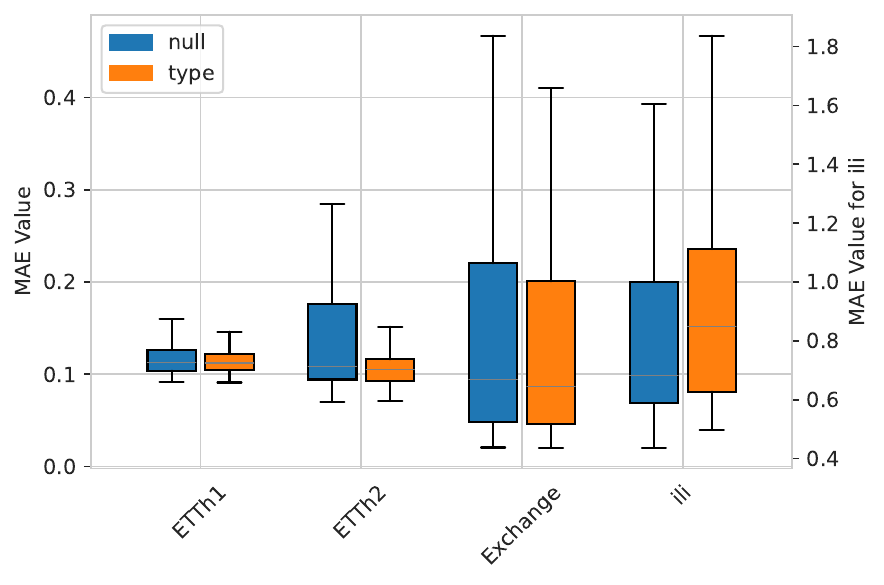}
         \caption{Learning Rate Strategy}
         \label{fig:exp-appx-lrs-llm_bp_mae}
     \end{subfigure}
     \hspace{10pt}
         \begin{subfigure}[t]{0.28\textwidth}
         \centering
         \includegraphics[width=\textwidth]{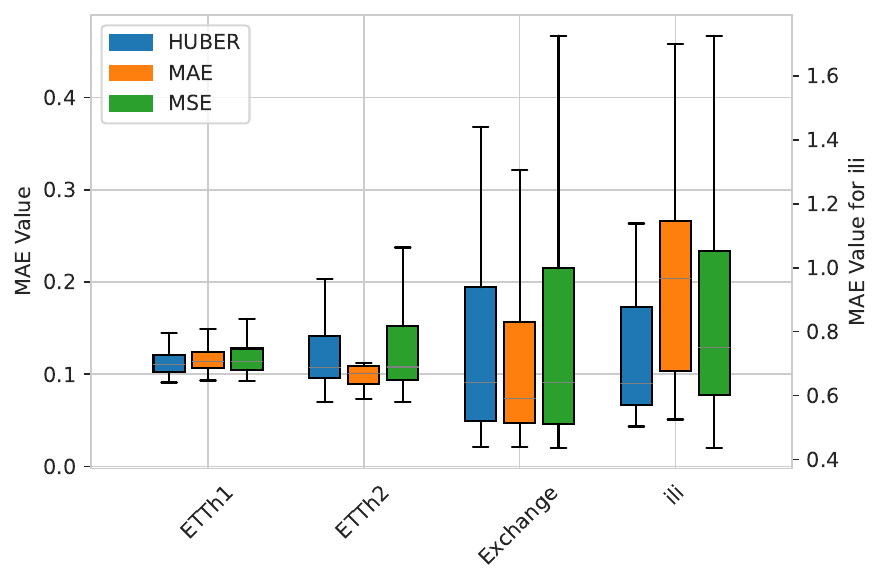}
         \caption{Loss Function}
         \label{fig:exp-appx-lf-llm_bp_mae}
     \end{subfigure}
     \hspace{10pt}
    \begin{subfigure}[t]{0.28\textwidth}
         \centering
         \includegraphics[width=\textwidth]{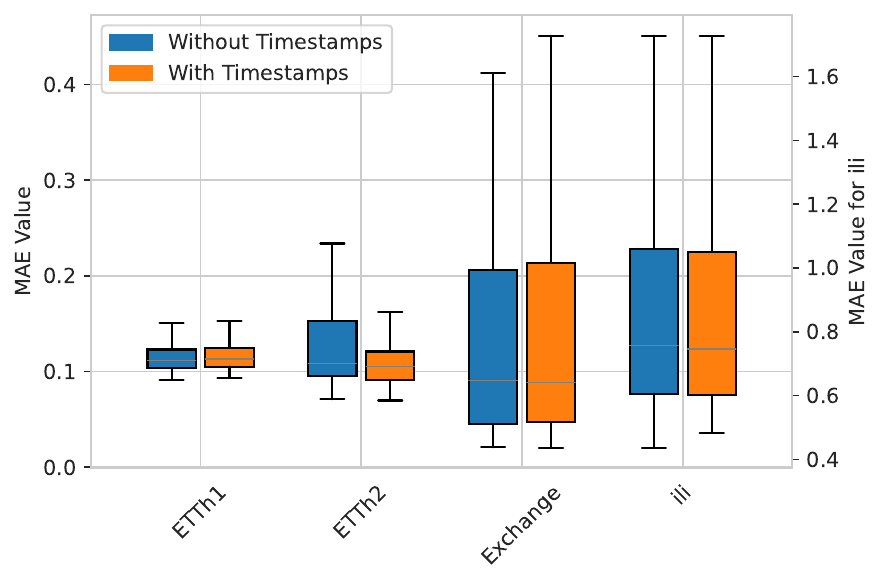}
         \caption{Timestamp}
         \label{fig:exp-appx-xm-llm_bp_mae}
     \end{subfigure}
     \hspace{10pt}
    \begin{subfigure}[t]{0.28\textwidth}
         \centering
         \includegraphics[width=\textwidth]{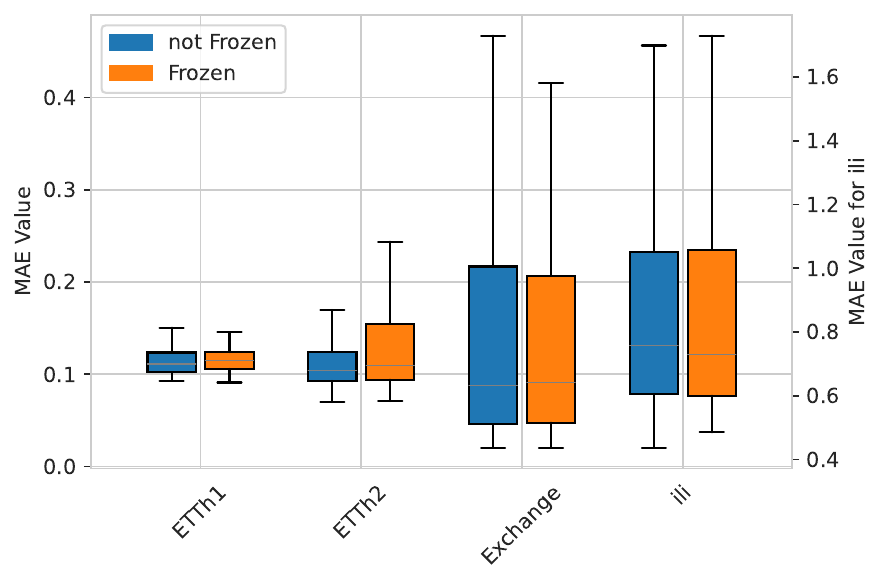}
         \caption{Frozen}
         \label{fig:exp-appx-frozen-llm_bp_mae}
     \end{subfigure}
     \hspace{10pt}
     \caption{Overall performance across all design dimensions when using LLMs or TSFMs in long-term forecasting. The results (\textbf{MAE}) are averaged across all forecasting horizons.}
     \vspace{-0.1in}
     \label{fig:exp-appx-bp-llm_mae}
\end{figure}

\clearpage

\subsection{Complete Evaluation Results of Short-term Forecasting Using MASE, OWA and sMAPE as the Metric}

\begin{figure}[t!]
     \centering
     \begin{subfigure}[t]{0.28\textwidth}
         \centering
         \includegraphics[width=\textwidth]{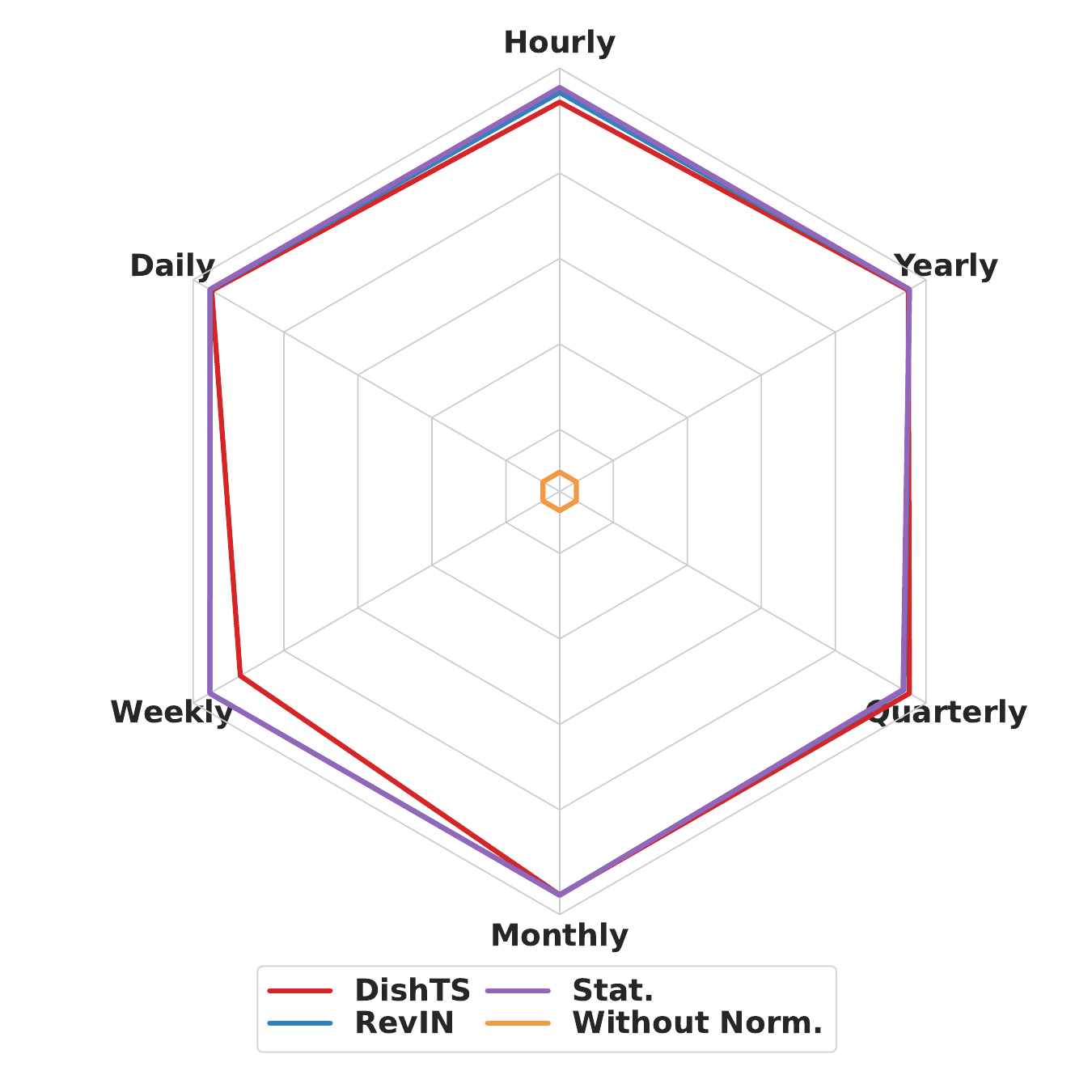}
         \caption{Series Normalization}
         \label{fig:exp-appx-Normalization-stf}
     \end{subfigure}
     \hspace{10pt}
     \begin{subfigure}[t]{0.28\textwidth}
         \centering
         \includegraphics[width=\textwidth]{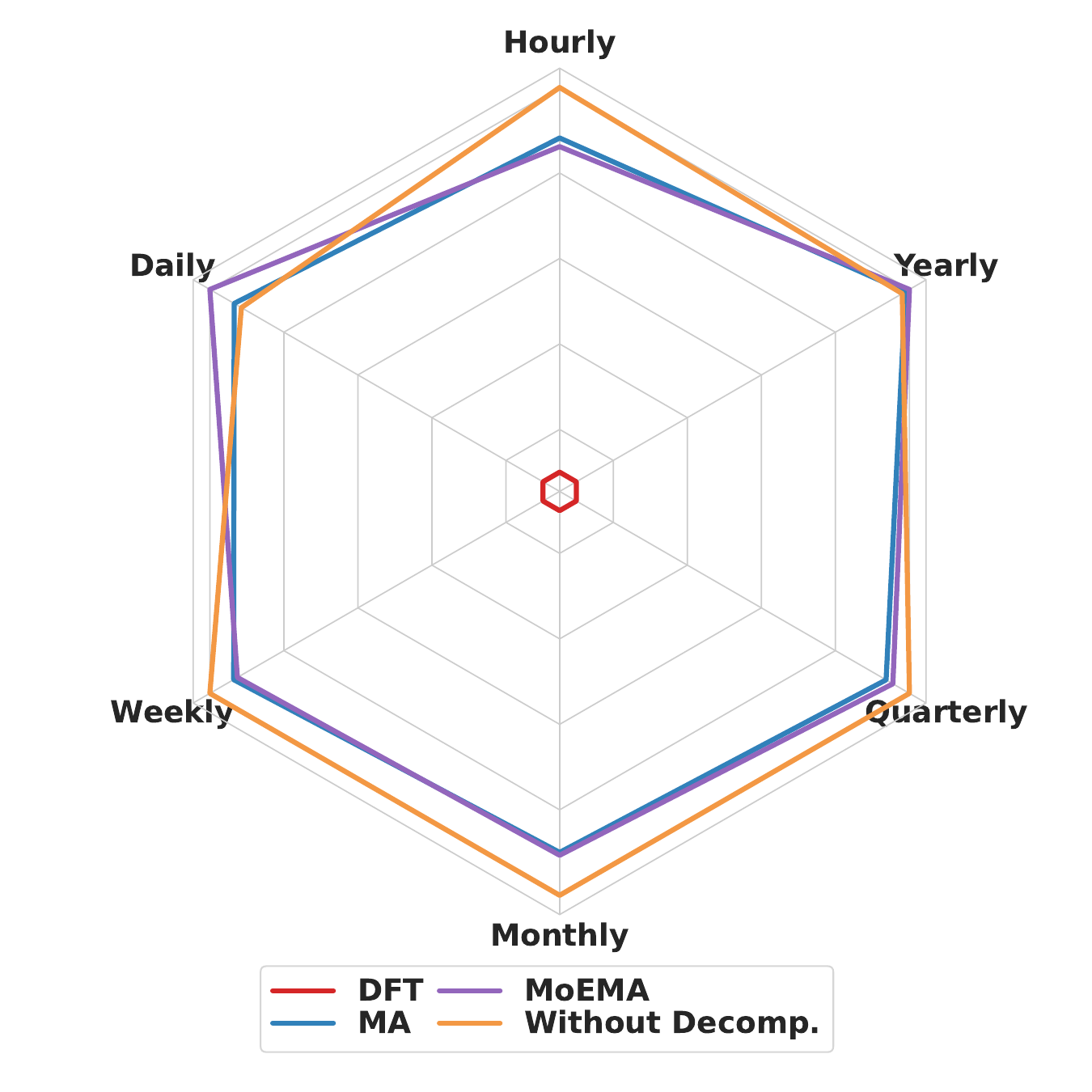}
         \caption{Series Decomposition}
         \label{fig:exp-appx-Decomposition-stf}
     \end{subfigure}
     \hspace{10pt}
    \begin{subfigure}[t]{0.28\textwidth}
         \centering
         \includegraphics[width=\textwidth]{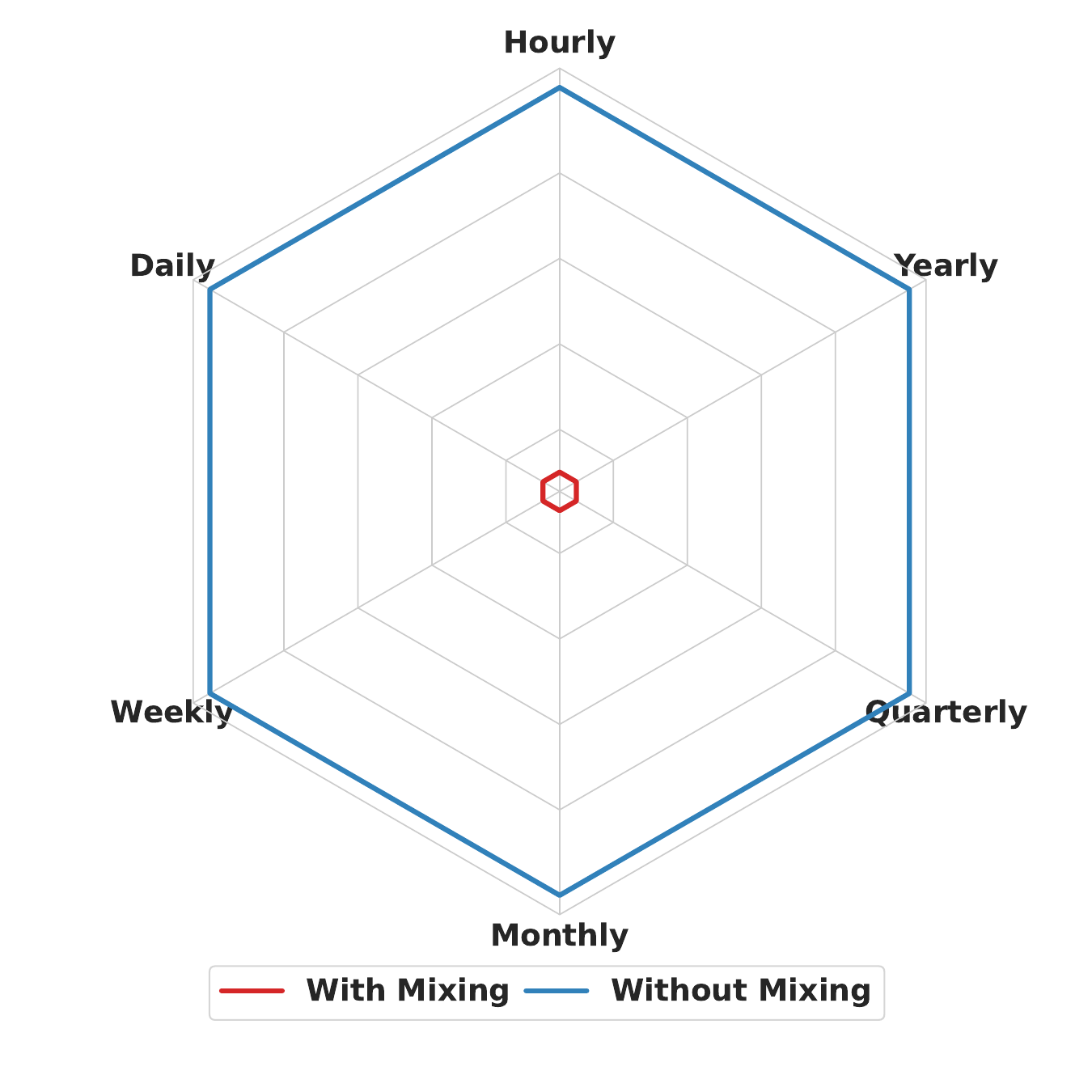}
         \caption{Series Sampling/Mixing}
         \label{fig:exp-appx-mixing-stf}
     \end{subfigure}
     \hspace{10pt}
    \begin{subfigure}[t]{0.28\textwidth}
         \centering
         \includegraphics[width=\textwidth]{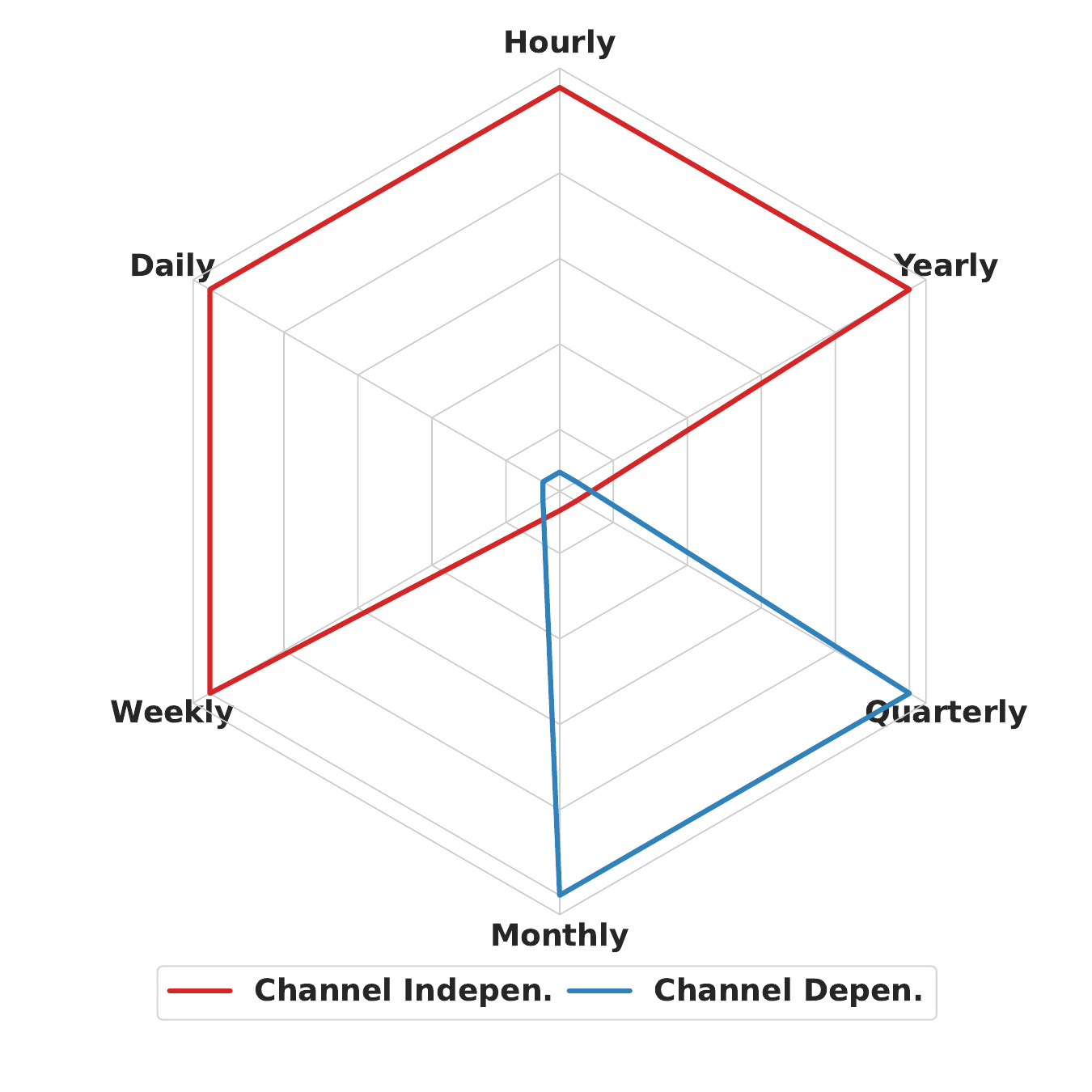}
         \caption{Channel Independent}
         \label{fig:exp-appx-CI-stf}
     \end{subfigure}
     \hspace{10pt}
    \begin{subfigure}[t]{0.28\textwidth}
         \centering
         \includegraphics[width=\textwidth]{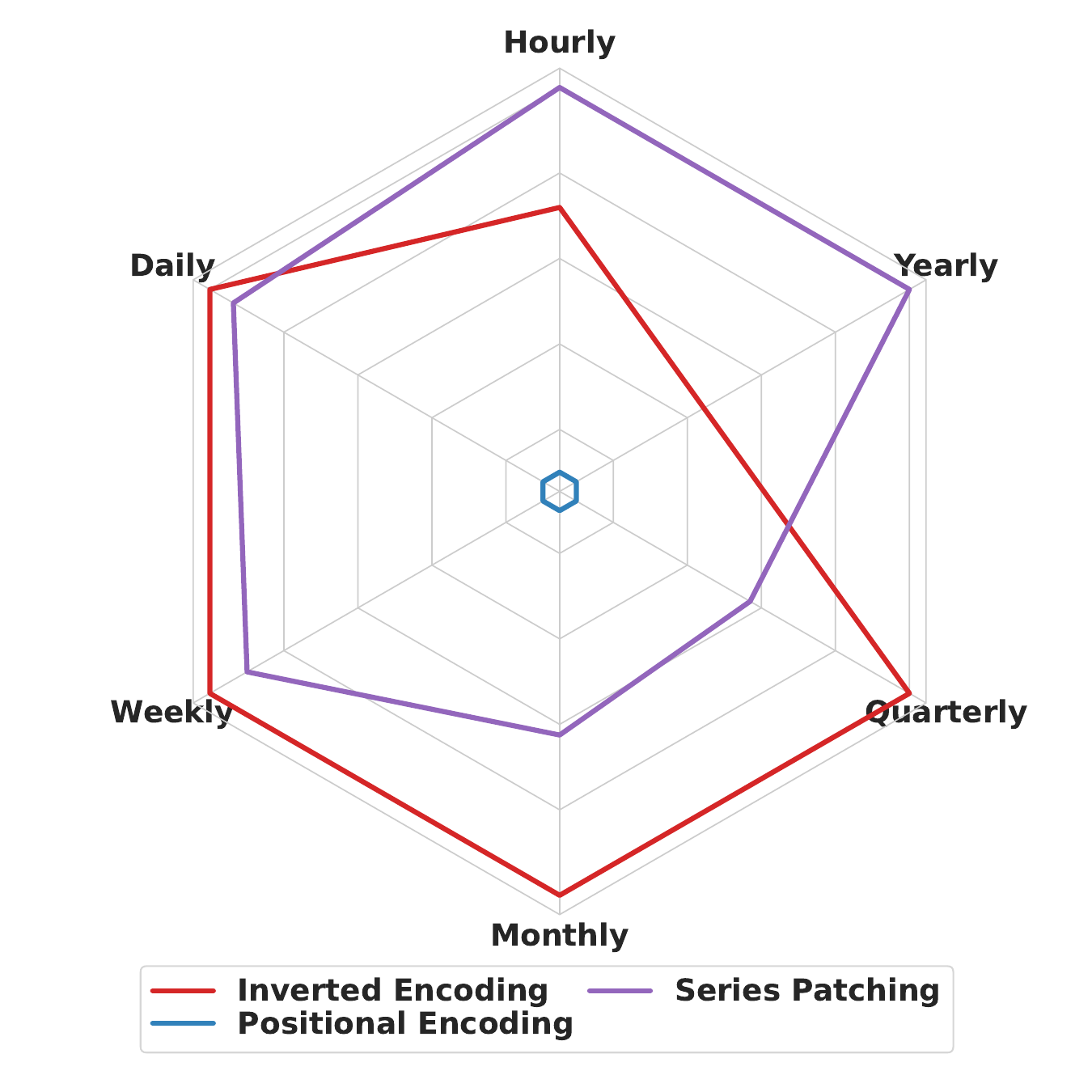}
         \caption{Series Embedding}
         \label{fig:exp-appx-tokenization-stf}
     \end{subfigure}
     \hspace{10pt}
    \begin{subfigure}[t]{0.28\textwidth}
         \centering
         \includegraphics[width=\textwidth]{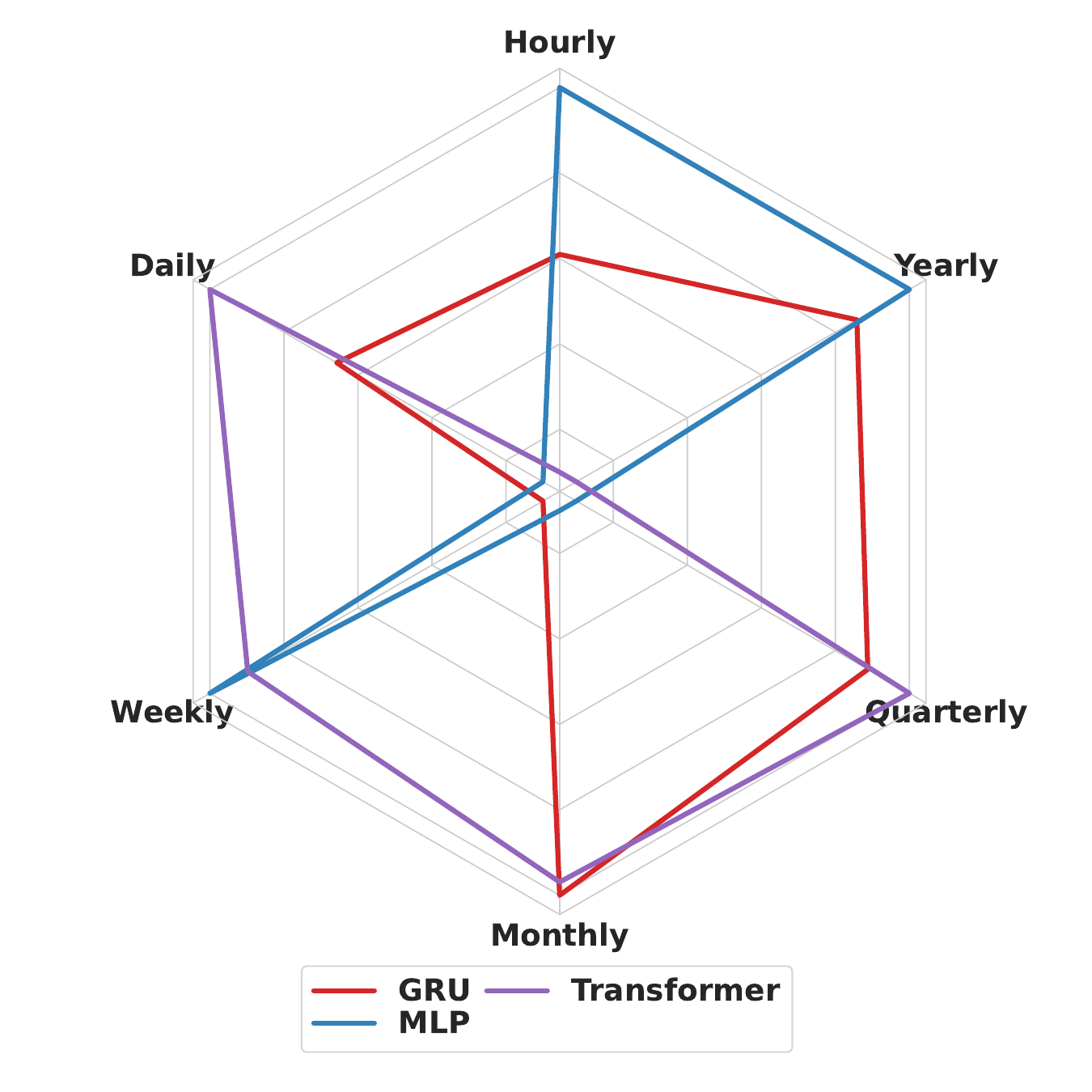}
         \caption{Network Backbone}
         \label{fig:exp-appx-backbone-stf}
     \end{subfigure}
     \hspace{10pt}
    \begin{subfigure}[t]{0.28\textwidth}
         \centering
         \includegraphics[width=\textwidth]{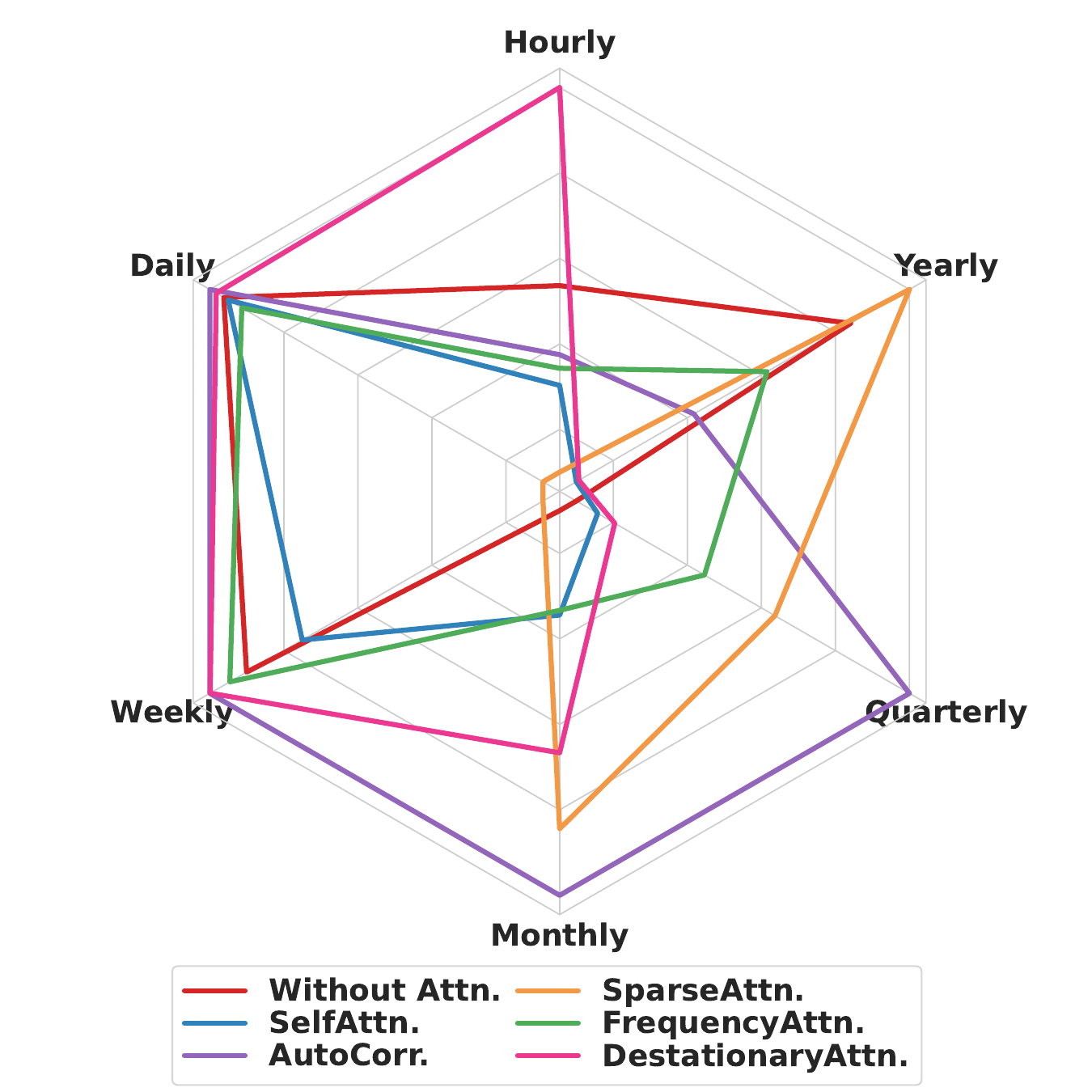}
         \caption{Series Attention}
         \label{fig:exp-appx-attention-stf}
     \end{subfigure}
     \hspace{10pt}
    \begin{subfigure}[t]{0.28\textwidth}
         \centering
         \includegraphics[width=\textwidth]{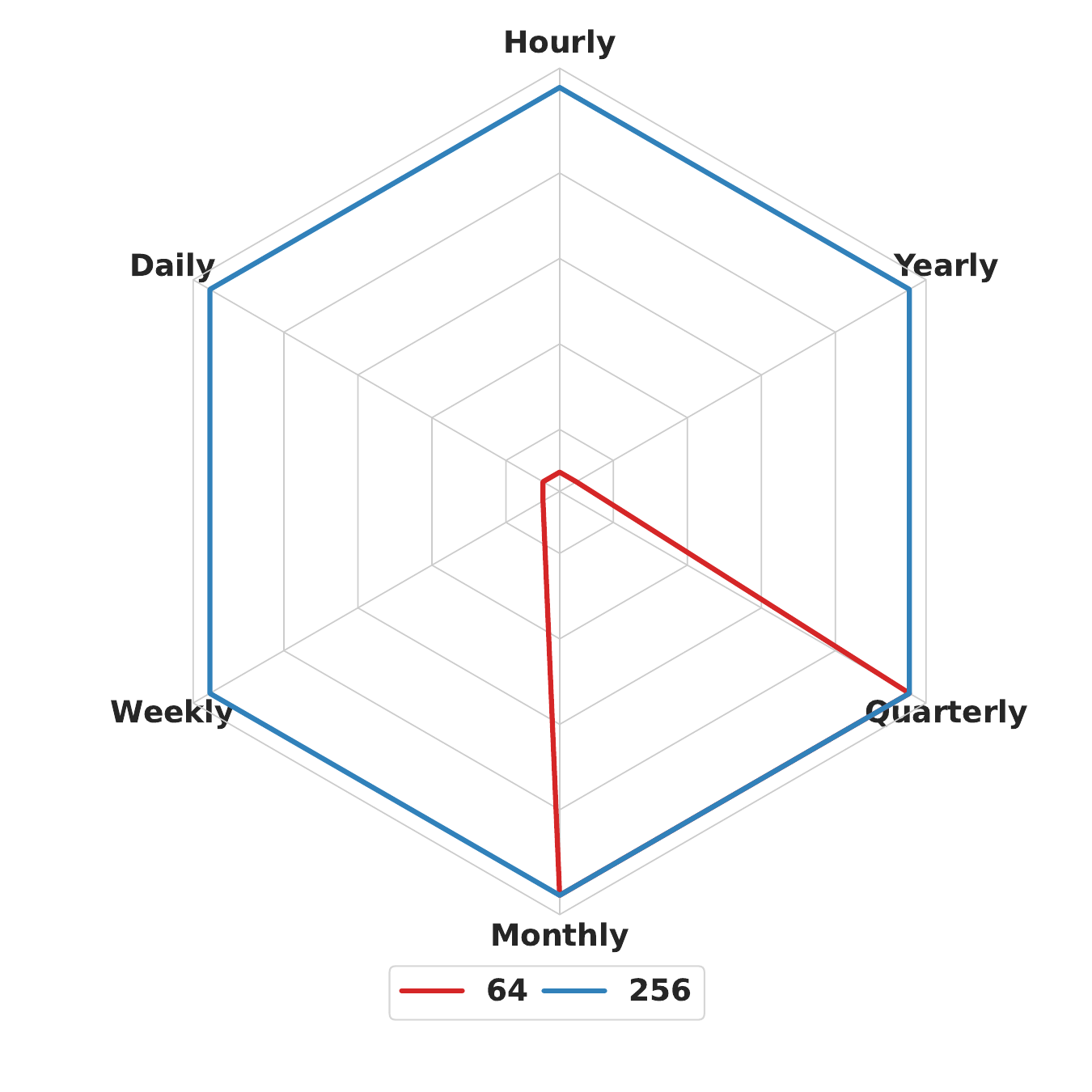}
         \caption{\textit{Hidden Layer Dimensions}}
         \label{fig:exp-appx-dmodel-stf}
     \end{subfigure}
     \hspace{10pt}
    \begin{subfigure}[t]{0.28\textwidth}
         \centering
         \includegraphics[width=\textwidth]{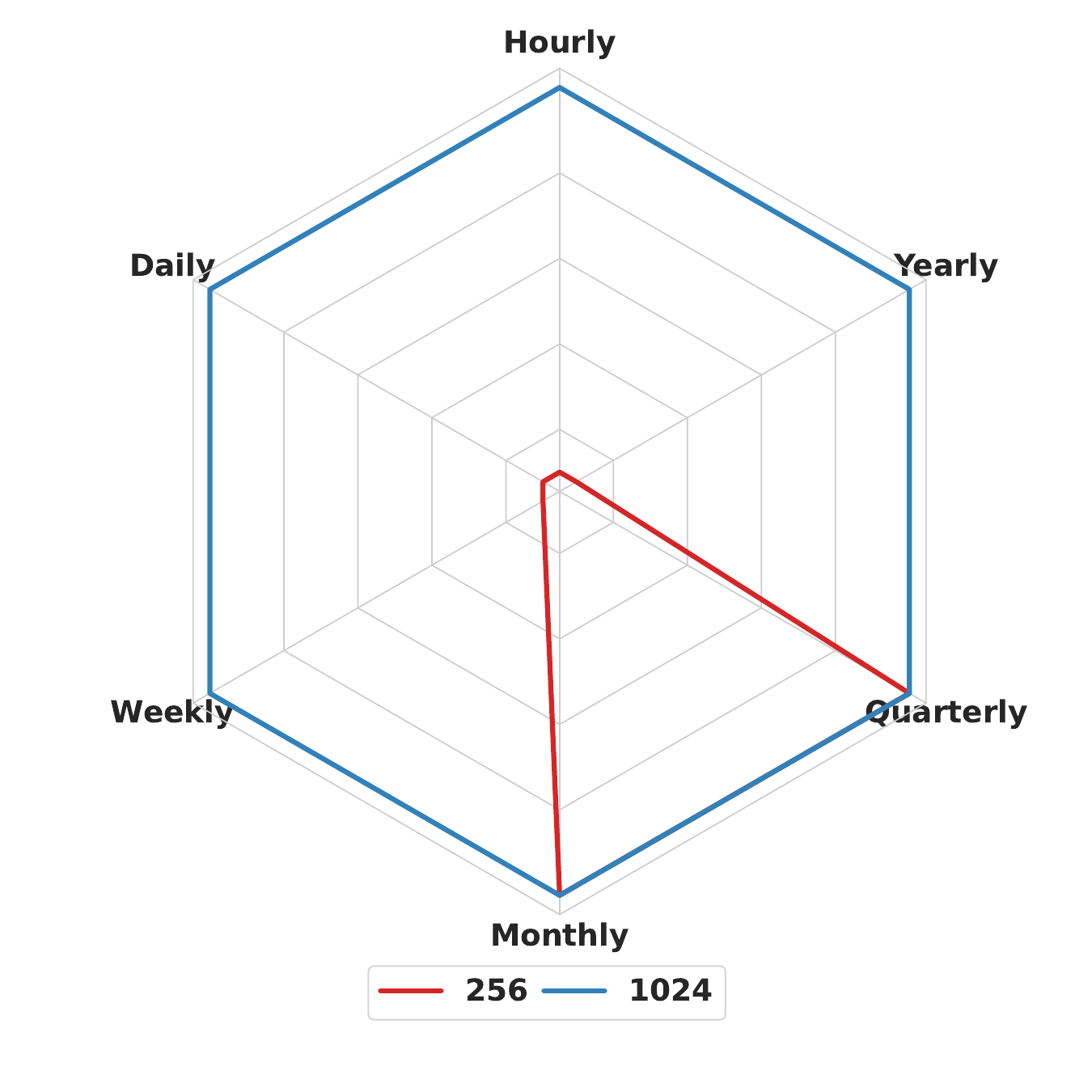}
         \caption{\textit{FCN Layer Dimensions}}
         \label{fig:exp-appx-dff-stf}
     \end{subfigure}
     \hspace{10pt}
    \begin{subfigure}[t]{0.28\textwidth}
         \centering
         \includegraphics[width=\textwidth]{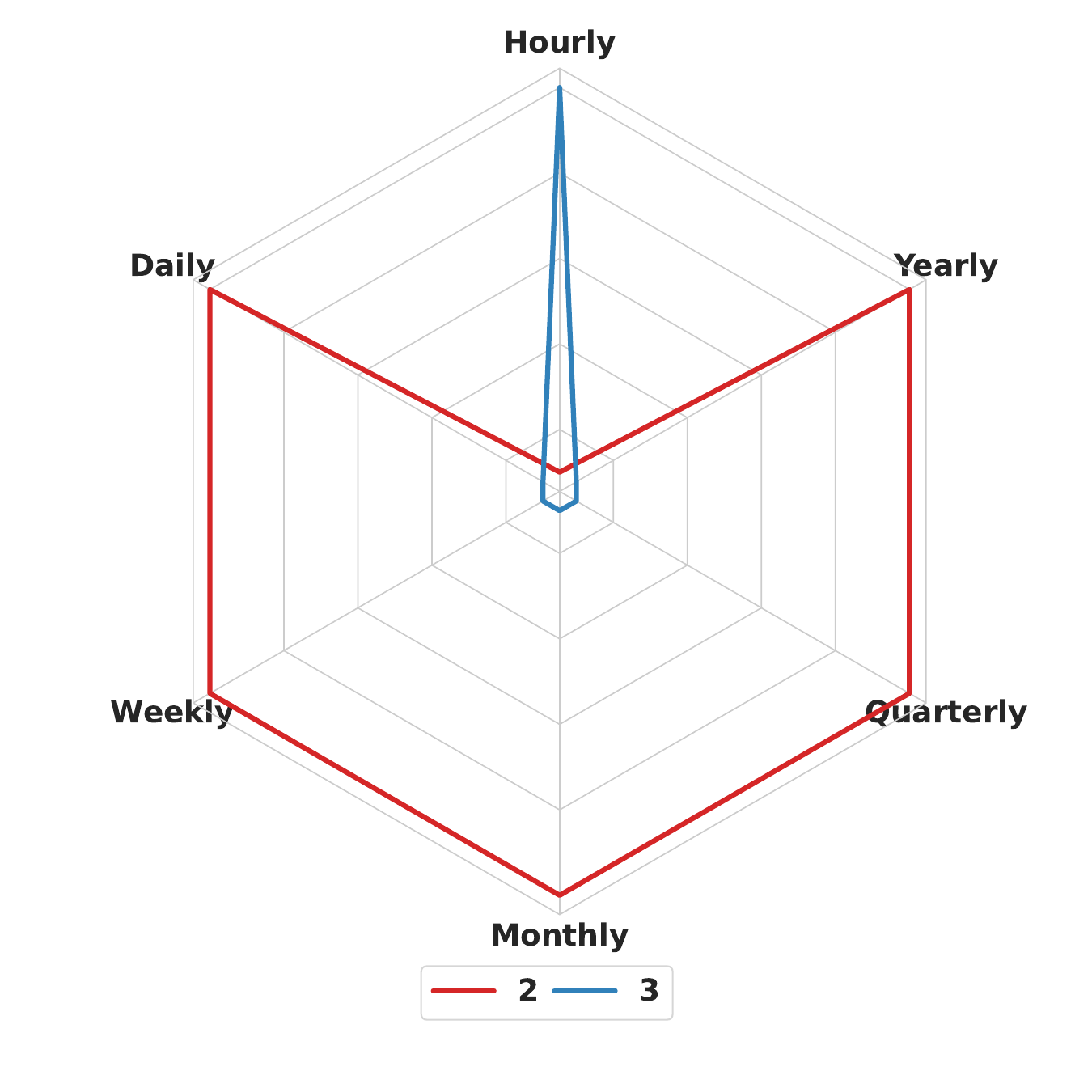}
         \caption{Encoder layers}
         \label{fig:exp-appx-el-stf}
     \end{subfigure}
     \hspace{10pt}
    \begin{subfigure}[t]{0.28\textwidth}
         \centering
         \includegraphics[width=\textwidth]{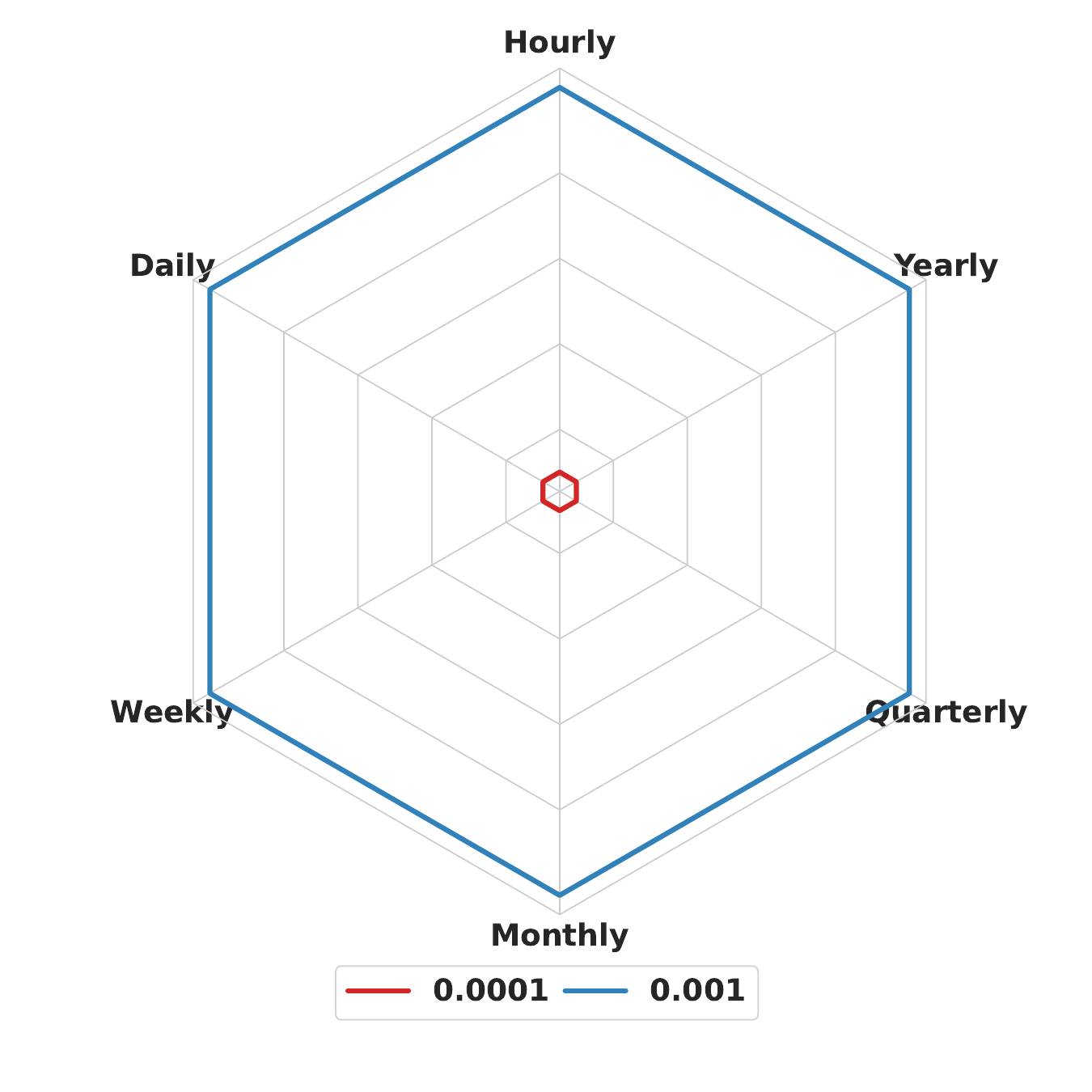}
         \caption{Learning Rate}
         \label{fig:exp-appx-lr-stf}
     \end{subfigure}
     \hspace{10pt}
    \begin{subfigure}[t]{0.28\textwidth}
         \centering
         \includegraphics[width=\textwidth]{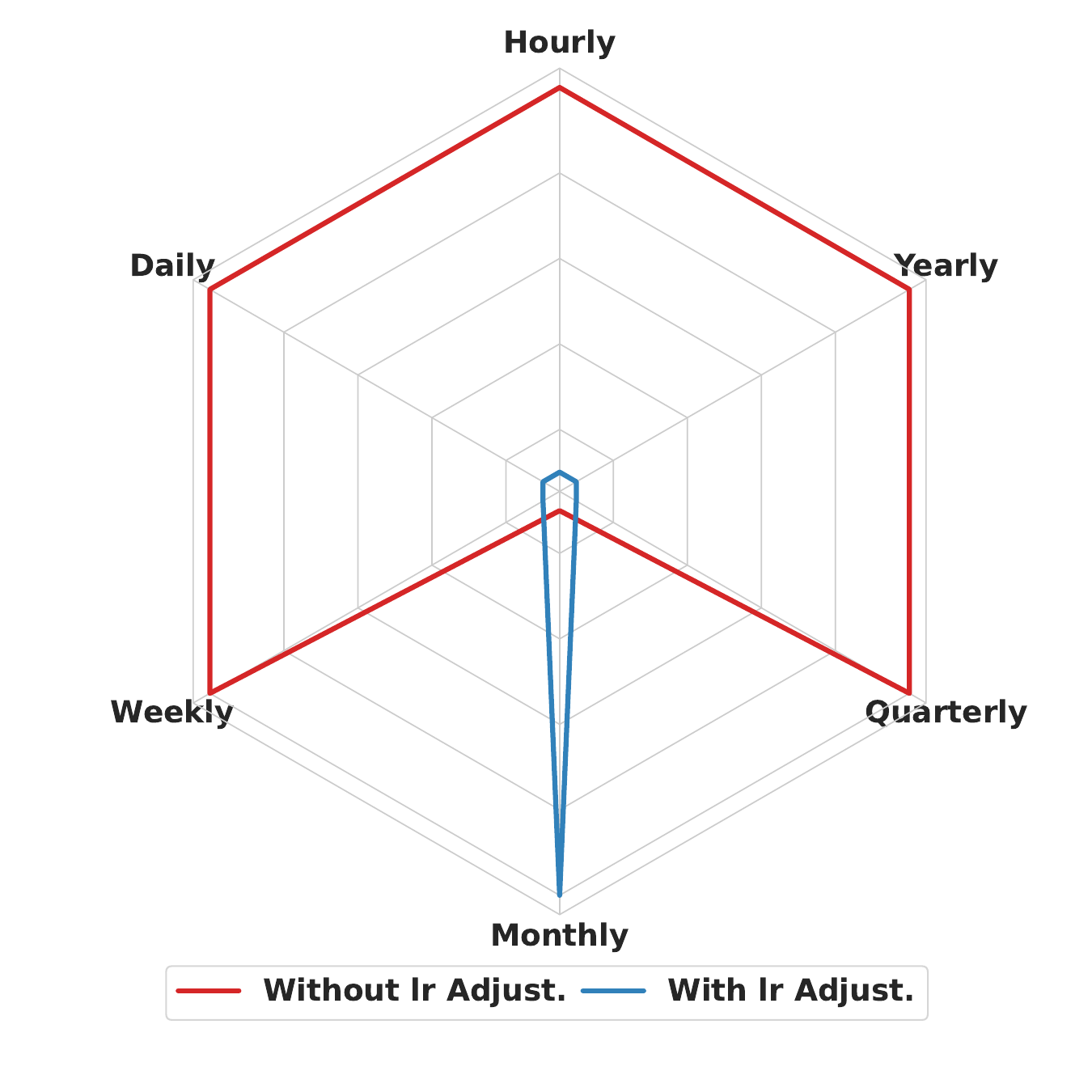}
         \caption{Learning Rate Strategy}
         \label{fig:exp-appx-lrs-stf}
     \end{subfigure}
     \hspace{10pt}
     \caption{Overall performance across all design dimensions in short-term forecasting. The results (\textbf{MASE}) are based on the top 25th percentile across all forecasting horizons.}
     \vspace{-0.1in}
     \label{fig:exp-appx-rada-stf_mase}
\end{figure}

\begin{figure}[t!]
     \centering
     \begin{subfigure}[t]{0.28\textwidth}
         \centering
         \includegraphics[width=\textwidth]{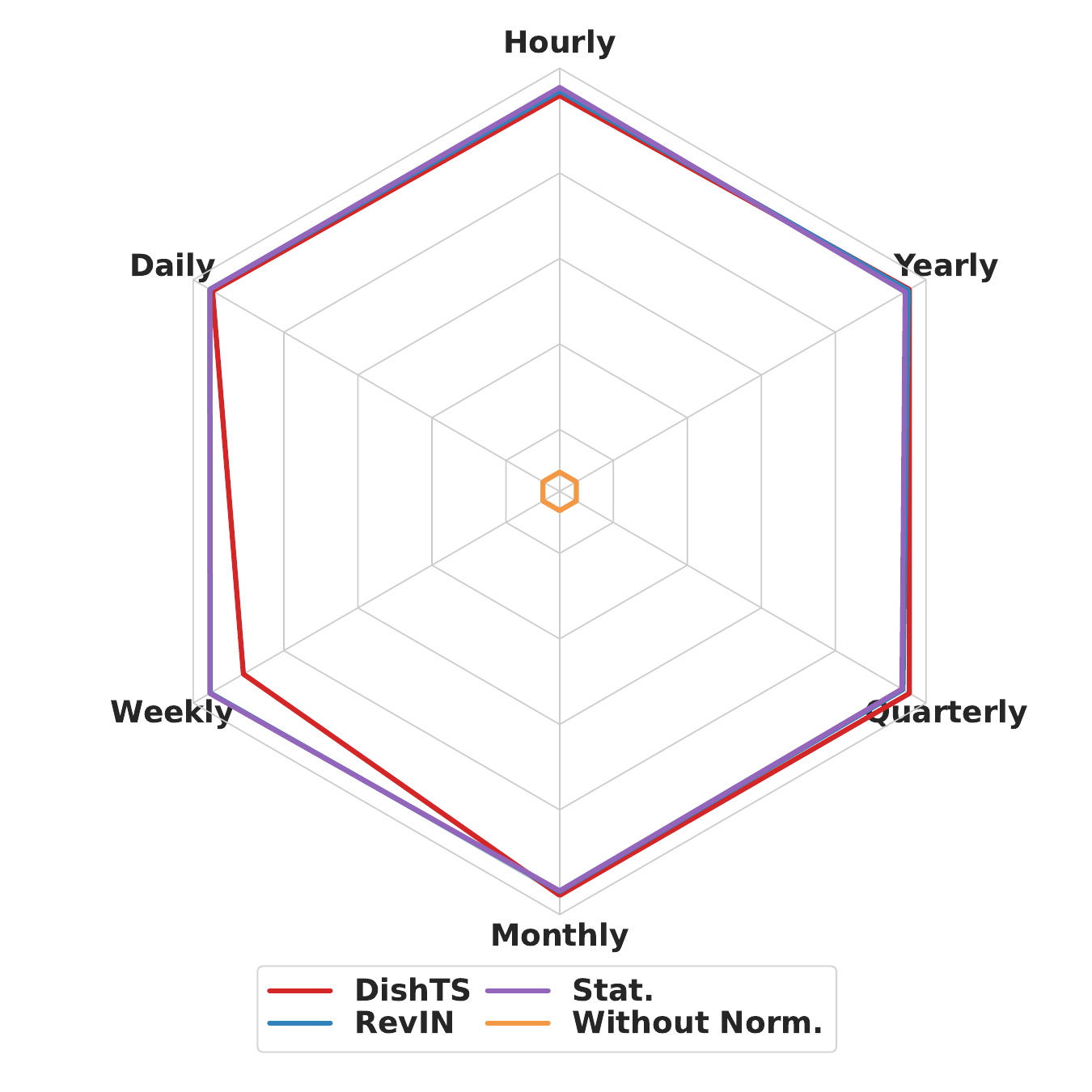}
         \caption{Series Normalization}
         \label{fig:exp-appx-Normalization-stf_owa}
     \end{subfigure}
     \hspace{10pt}
     \begin{subfigure}[t]{0.28\textwidth}
         \centering
         \includegraphics[width=\textwidth]{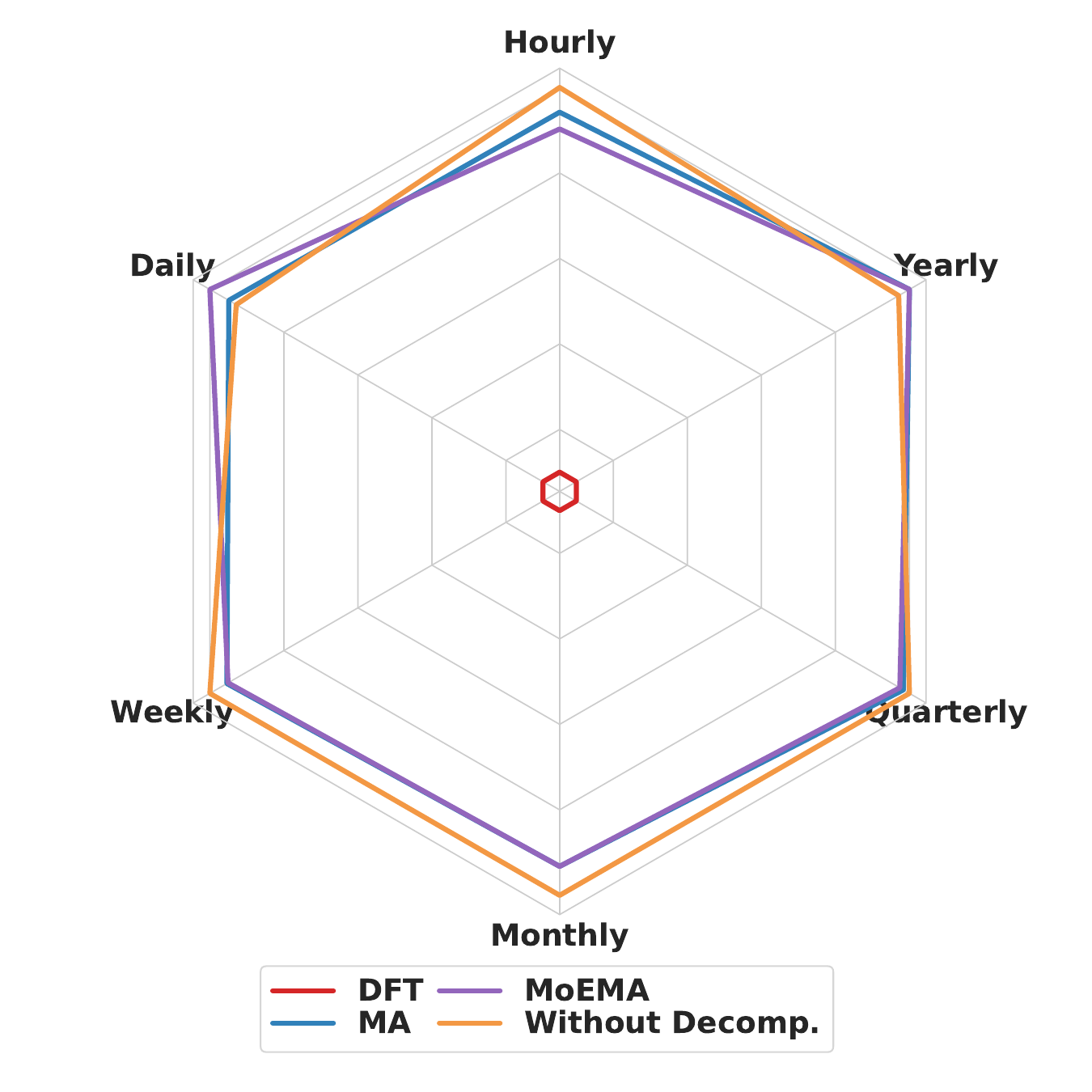}
         \caption{Series Decomposition}
         \label{fig:exp-appx-Decomposition-stf_owa}
     \end{subfigure}
     \hspace{10pt}
    \begin{subfigure}[t]{0.28\textwidth}
         \centering
         \includegraphics[width=\textwidth]{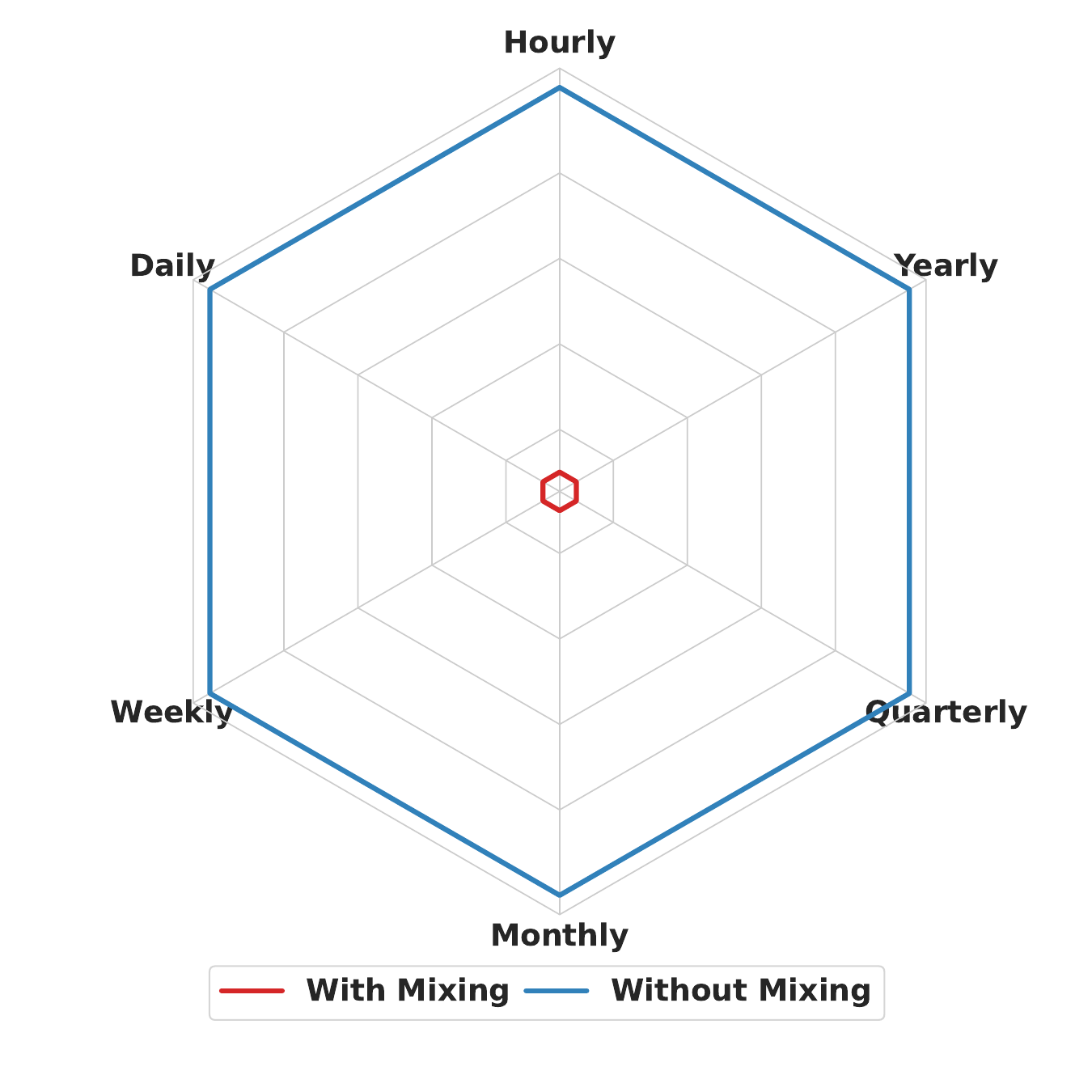}
         \caption{Series Sampling/Mixing}
         \label{fig:exp-appx-mixing-stf_owa}
     \end{subfigure}
     \hspace{10pt}
    \begin{subfigure}[t]{0.28\textwidth}
         \centering
         \includegraphics[width=\textwidth]{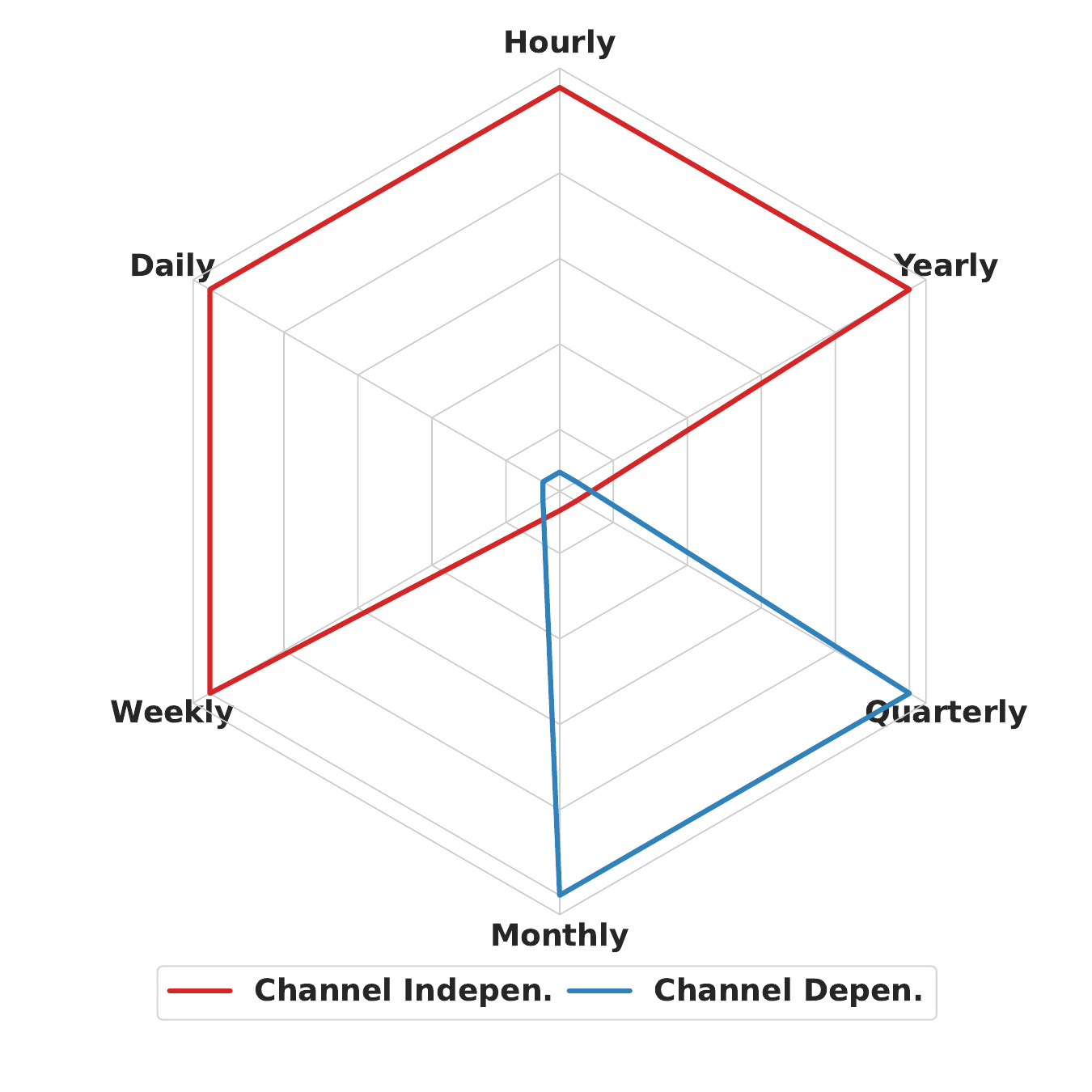}
         \caption{Channel Independent}
         \label{fig:exp-appx-CI-stf_owa}
     \end{subfigure}
     \hspace{10pt}
    \begin{subfigure}[t]{0.28\textwidth}
         \centering
         \includegraphics[width=\textwidth]{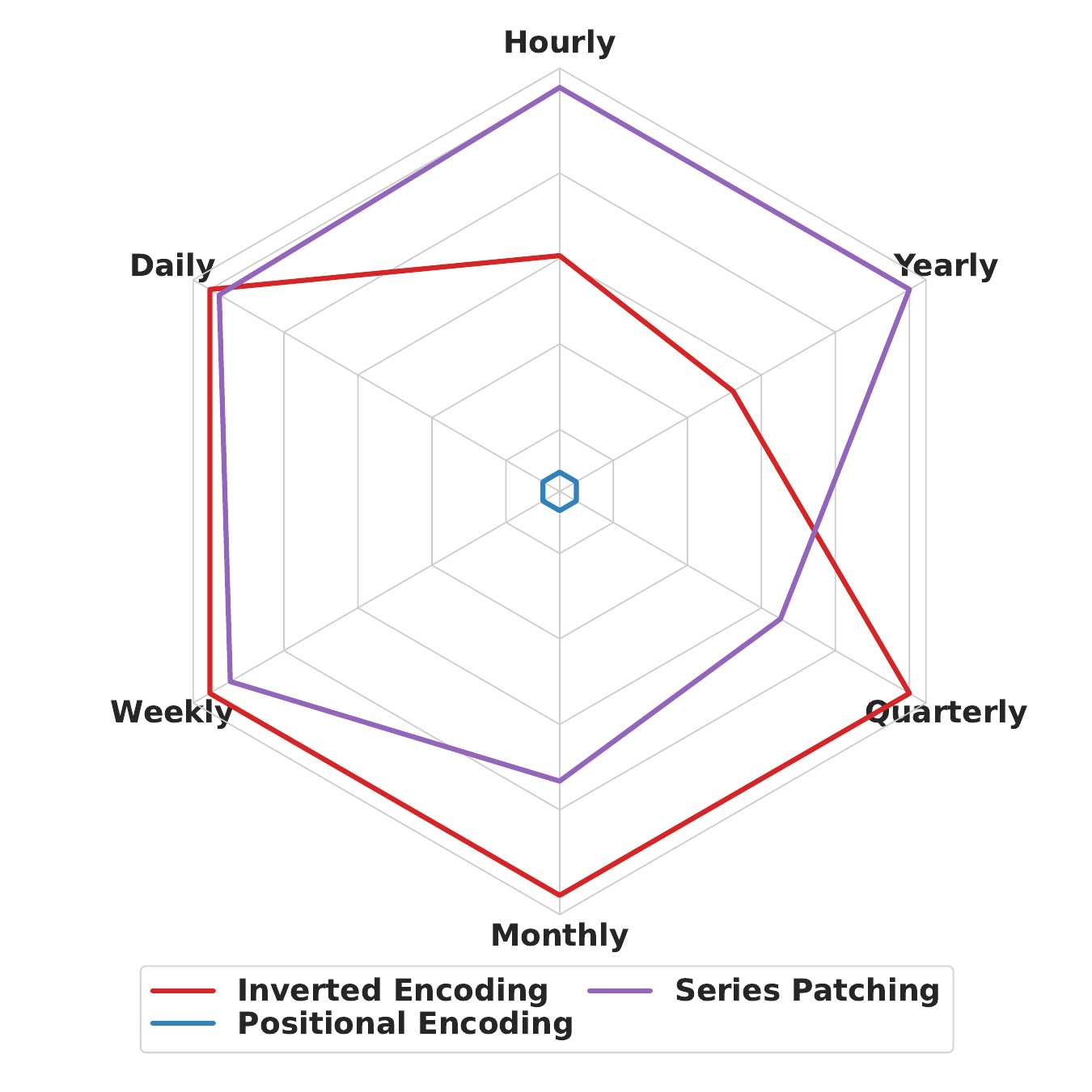}
         \caption{Series Embedding}
         \label{fig:exp-appx-tokenization-stf_owa}
     \end{subfigure}
     \hspace{10pt}
    \begin{subfigure}[t]{0.28\textwidth}
         \centering
         \includegraphics[width=\textwidth]{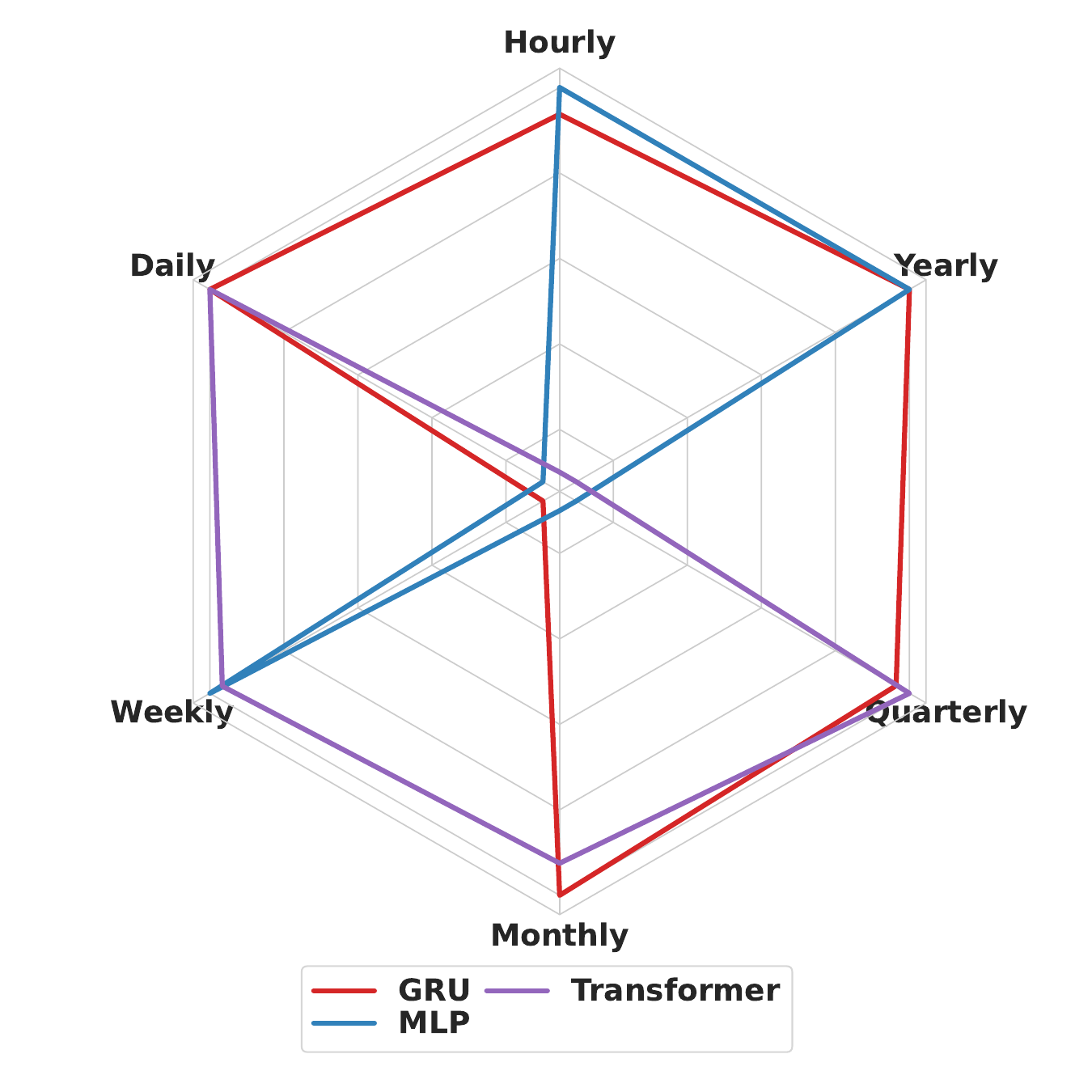}
         \caption{Network Backbone}
         \label{fig:exp-appx-backbone-stf_owa}
     \end{subfigure}
     \hspace{10pt}
    \begin{subfigure}[t]{0.28\textwidth}
         \centering
         \includegraphics[width=\textwidth]{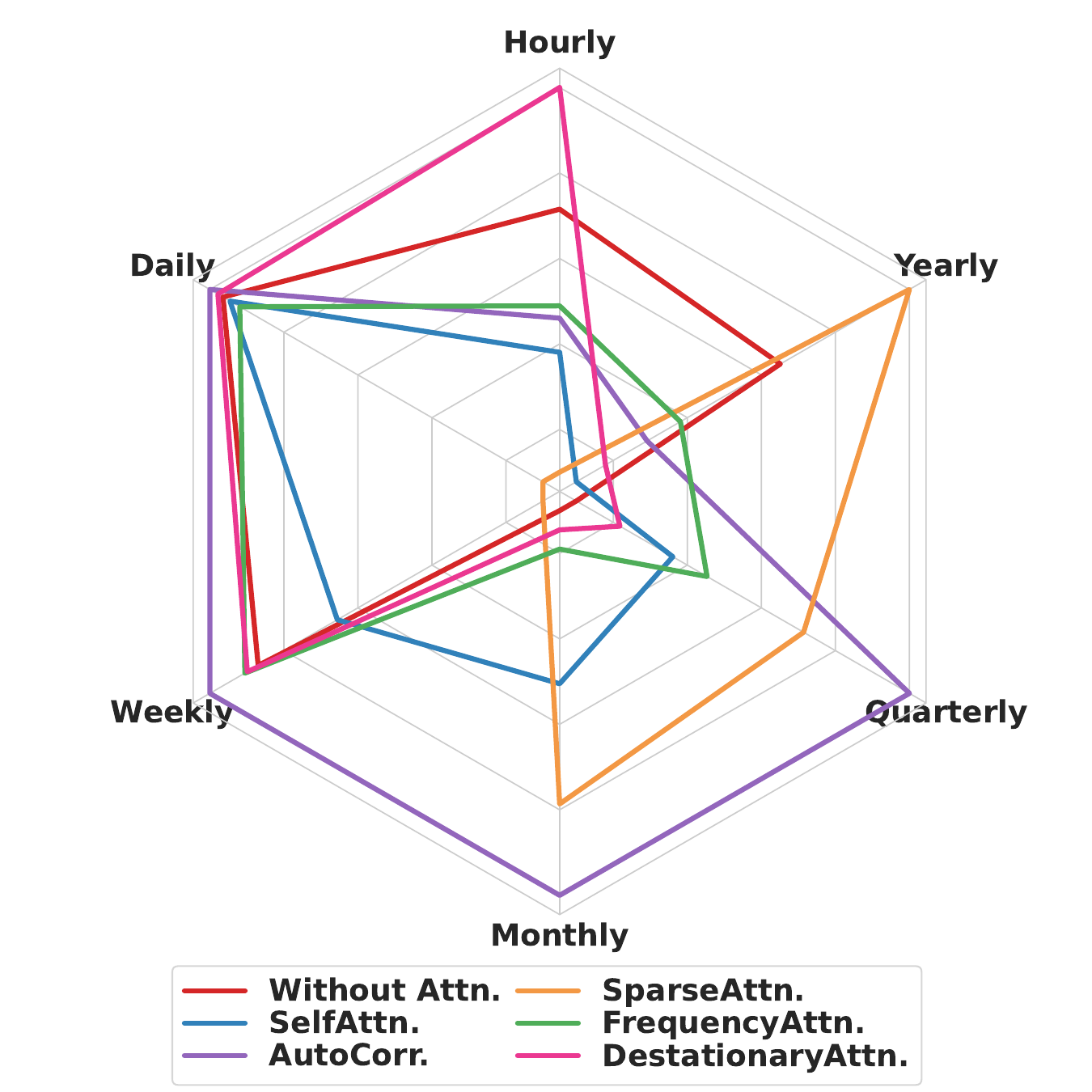}
         \caption{Series Attention}
         \label{fig:exp-appx-attention-stf_owa}
     \end{subfigure}
     \hspace{10pt}
    \begin{subfigure}[t]{0.28\textwidth}
         \centering
         \includegraphics[width=\textwidth]{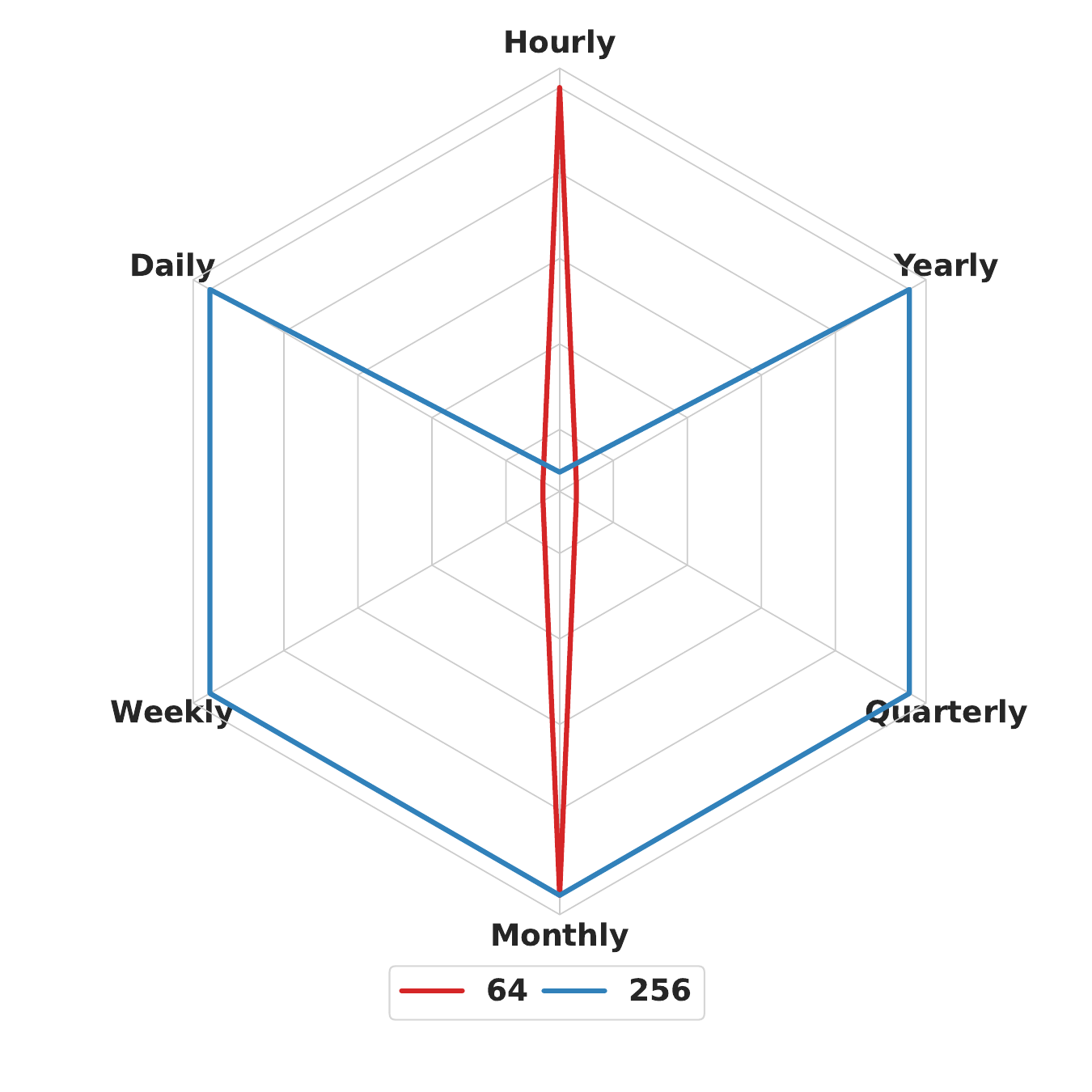}
         \caption{\textit{Hidden Layer Dimensions}}
         \label{fig:exp-appx-dmodel-stf_owa}
     \end{subfigure}
     \hspace{10pt}
    \begin{subfigure}[t]{0.28\textwidth}
         \centering
         \includegraphics[width=\textwidth]{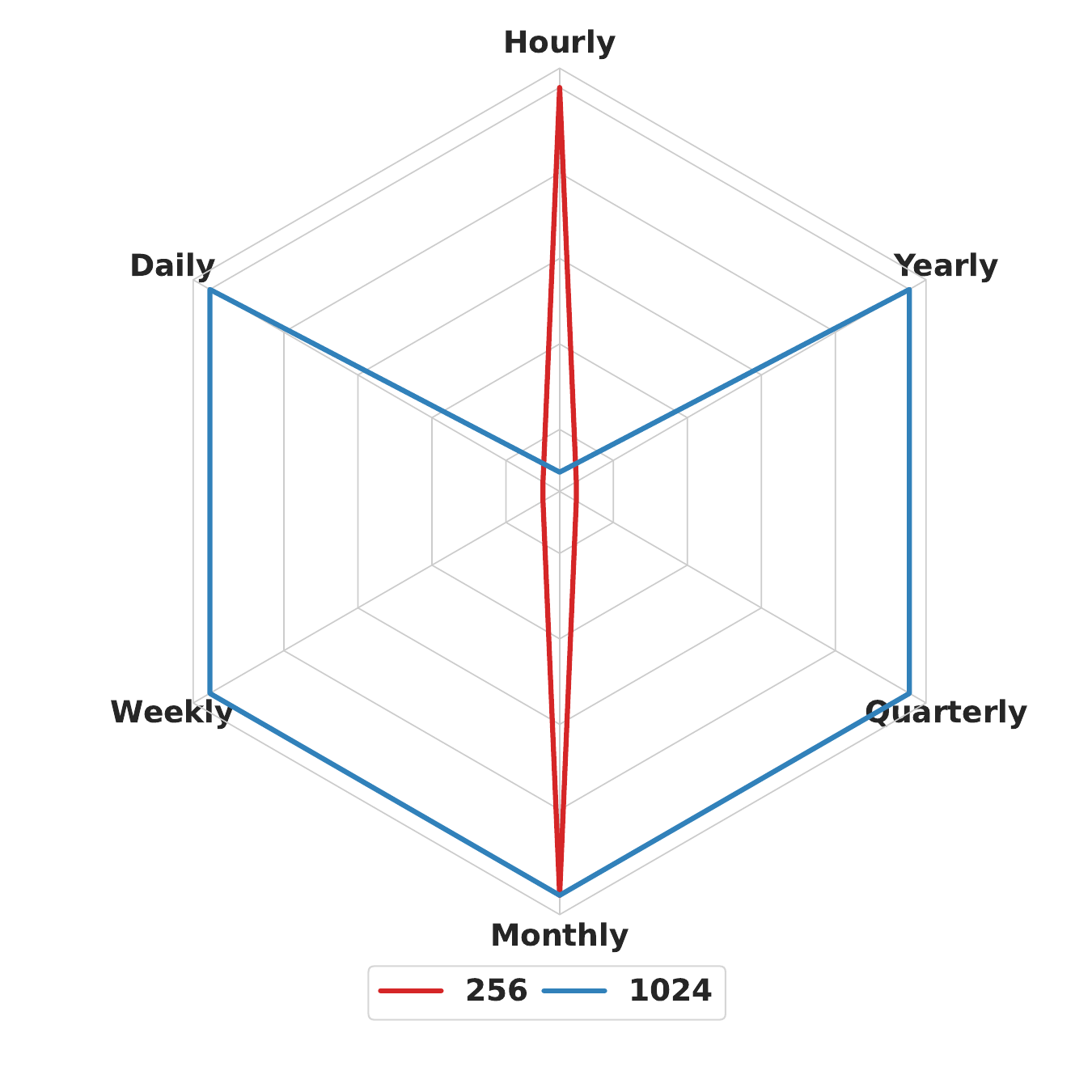}
         \caption{\textit{FCN Layer Dimensions}}
         \label{fig:exp-appx-dff-stf_owa}
     \end{subfigure}
     \hspace{10pt}
    \begin{subfigure}[t]{0.28\textwidth}
         \centering
         \includegraphics[width=\textwidth]{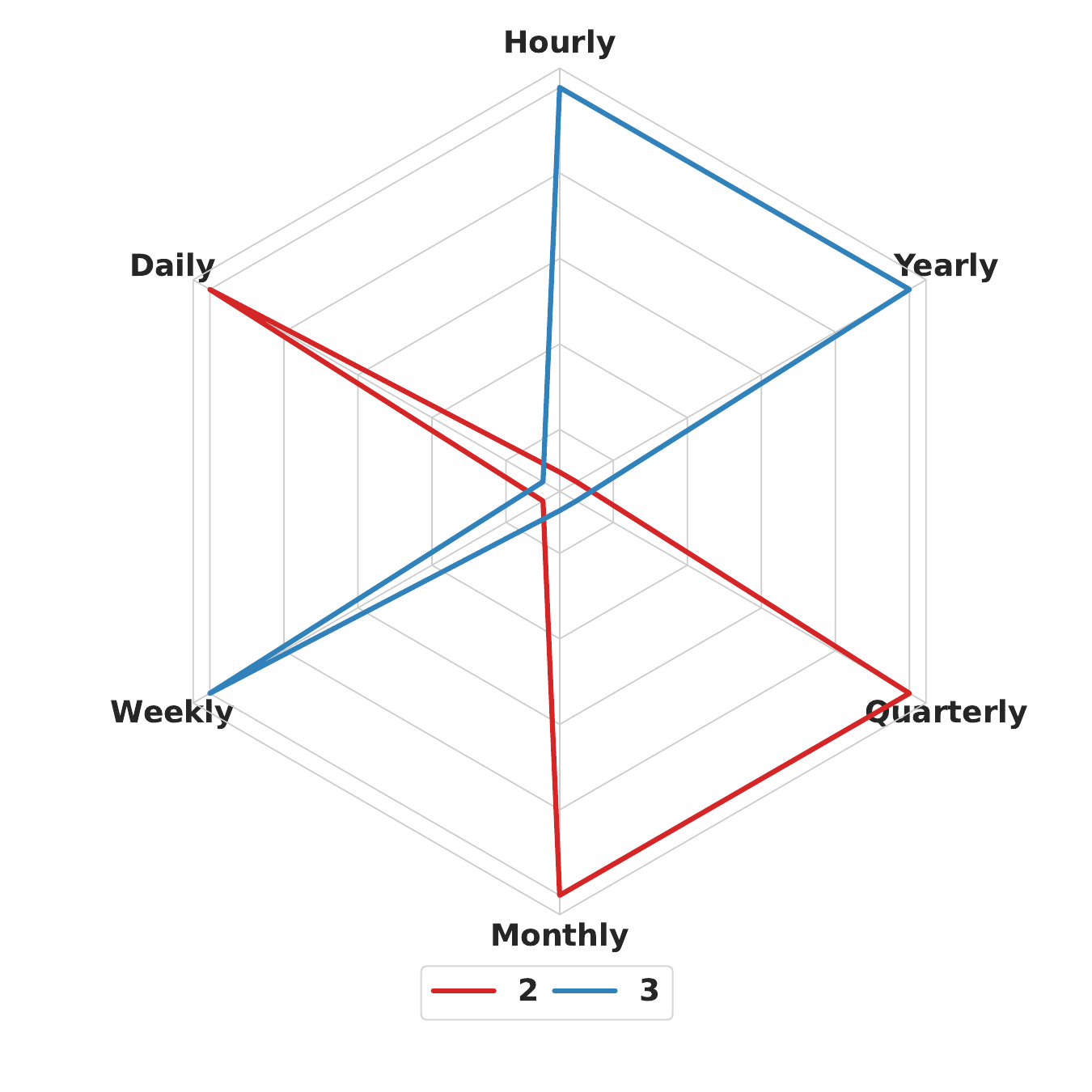}
         \caption{Encoder layers}
         \label{fig:exp-appx-el-stf_owa}
     \end{subfigure}
     \hspace{10pt}
    \begin{subfigure}[t]{0.28\textwidth}
         \centering
         \includegraphics[width=\textwidth]{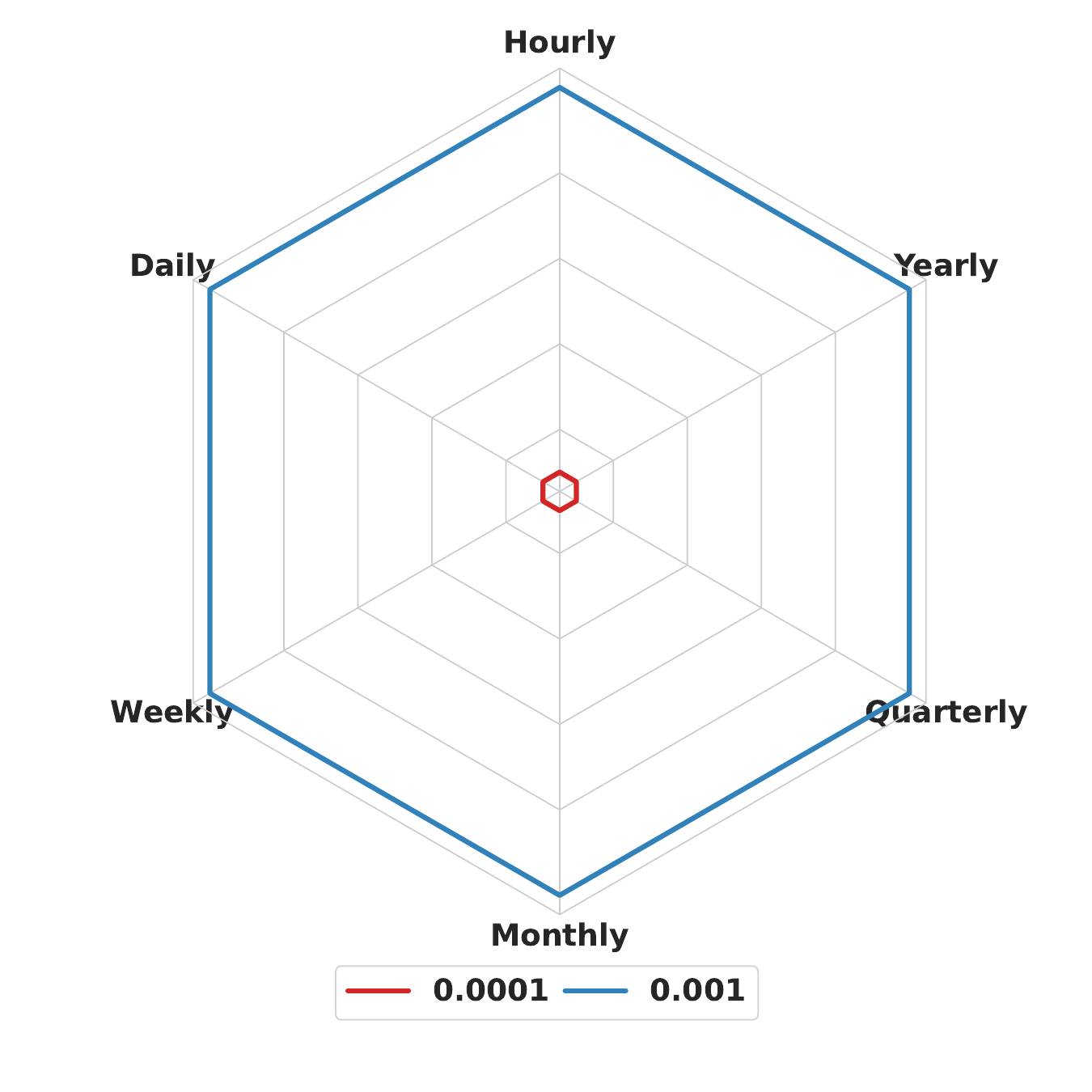}
         \caption{Learning Rate}
         \label{fig:exp-appx-lr-stf_owa}
     \end{subfigure}
     \hspace{10pt}
    \begin{subfigure}[t]{0.28\textwidth}
         \centering
         \includegraphics[width=\textwidth]{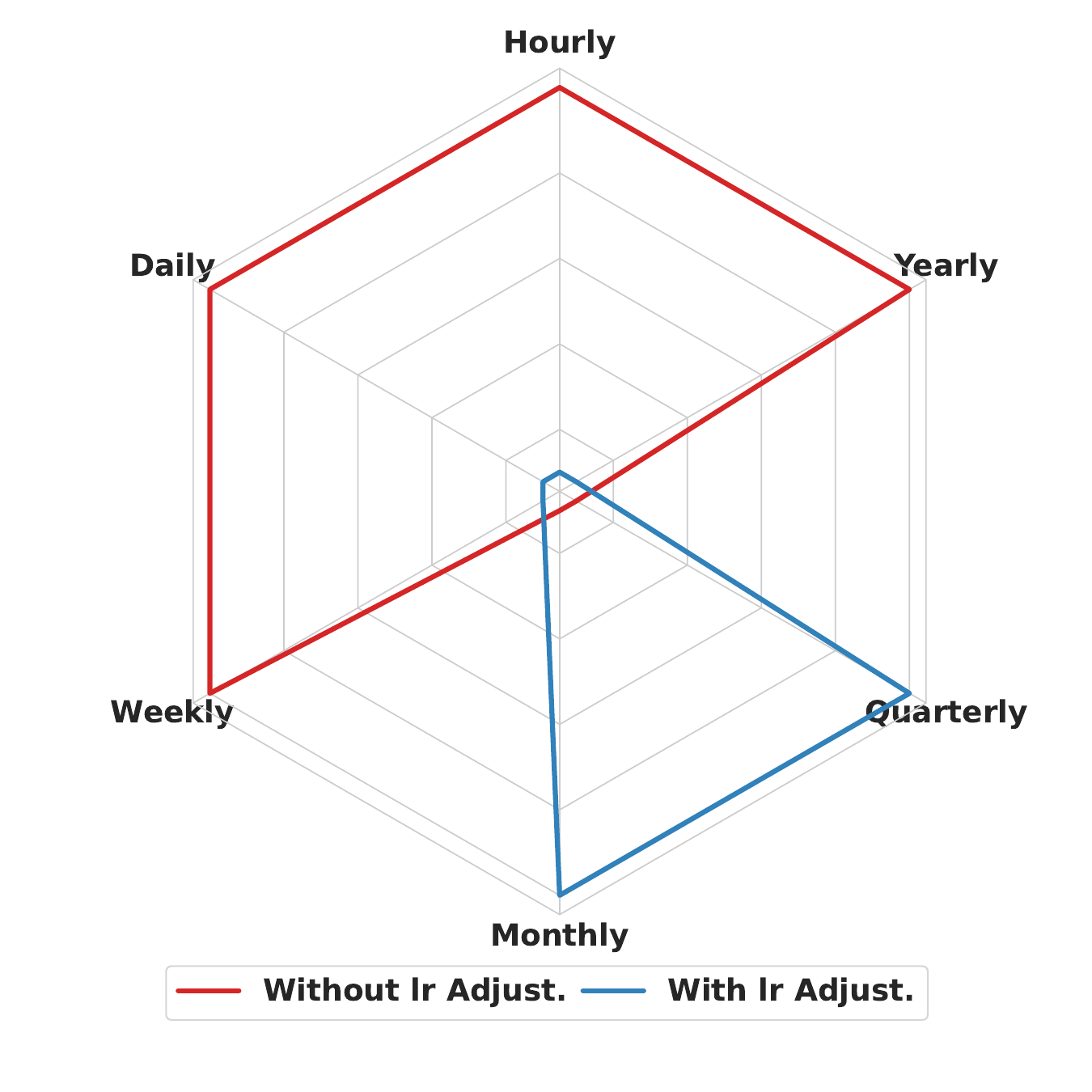}
         \caption{Learning Rate Strategy}
         \label{fig:exp-appx-lrs-stf_owa}
     \end{subfigure}
     \hspace{10pt}
     \caption{Overall performance across all design dimensions in short-term forecasting. The results (\textbf{OWA}) are based on the top 25th percentile across all forecasting horizons.}
     \vspace{-0.1in}
     \label{fig:exp-appx-rada-stf_owa}
\end{figure}

\begin{figure}[t!]
     \centering
     \begin{subfigure}[t]{0.28\textwidth}
         \centering
         \includegraphics[width=\textwidth]{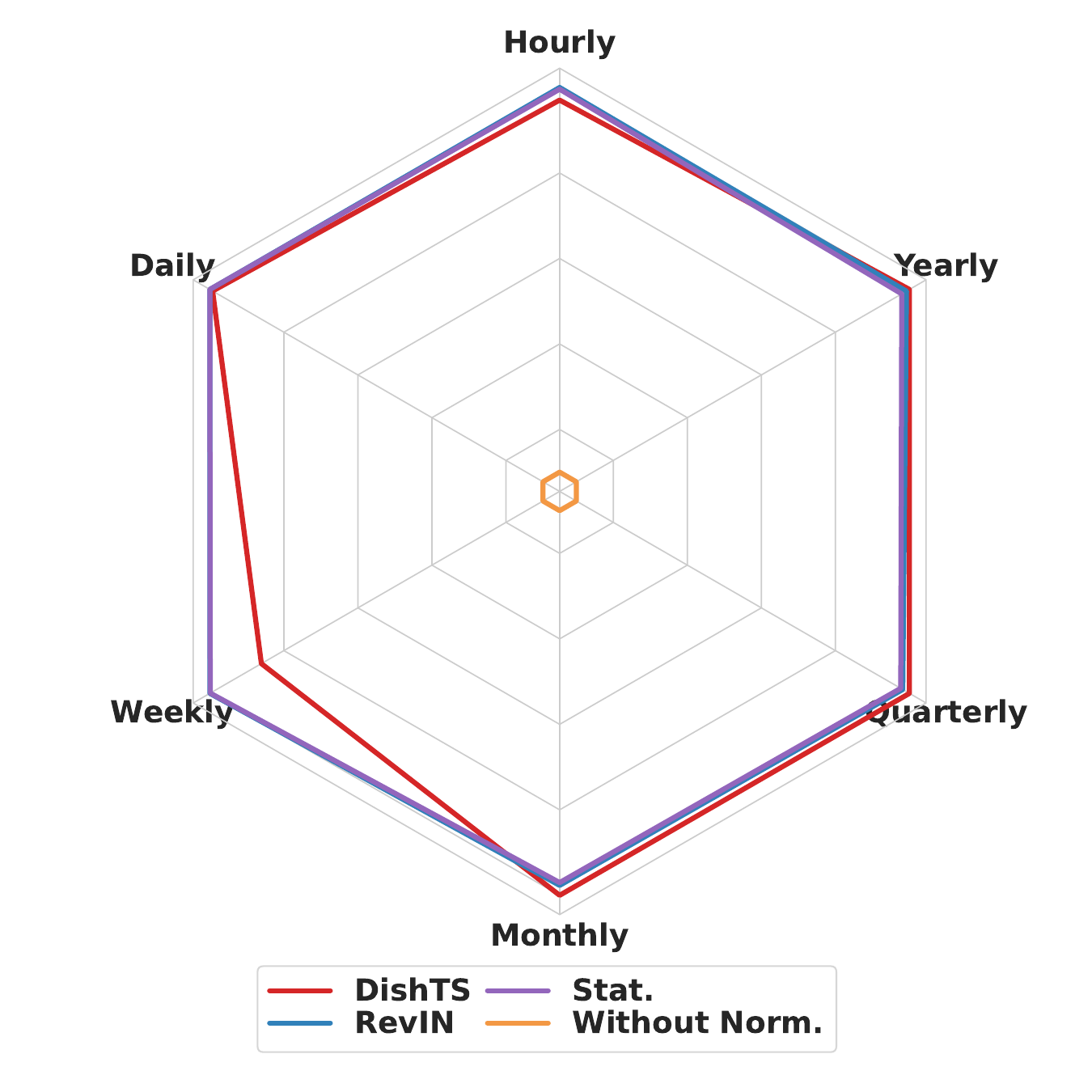}
         \caption{Series Normalization}
         \label{fig:exp-appx-Normalization-stf_smape}
     \end{subfigure}
     \hspace{10pt}
     \begin{subfigure}[t]{0.28\textwidth}
         \centering
         \includegraphics[width=\textwidth]{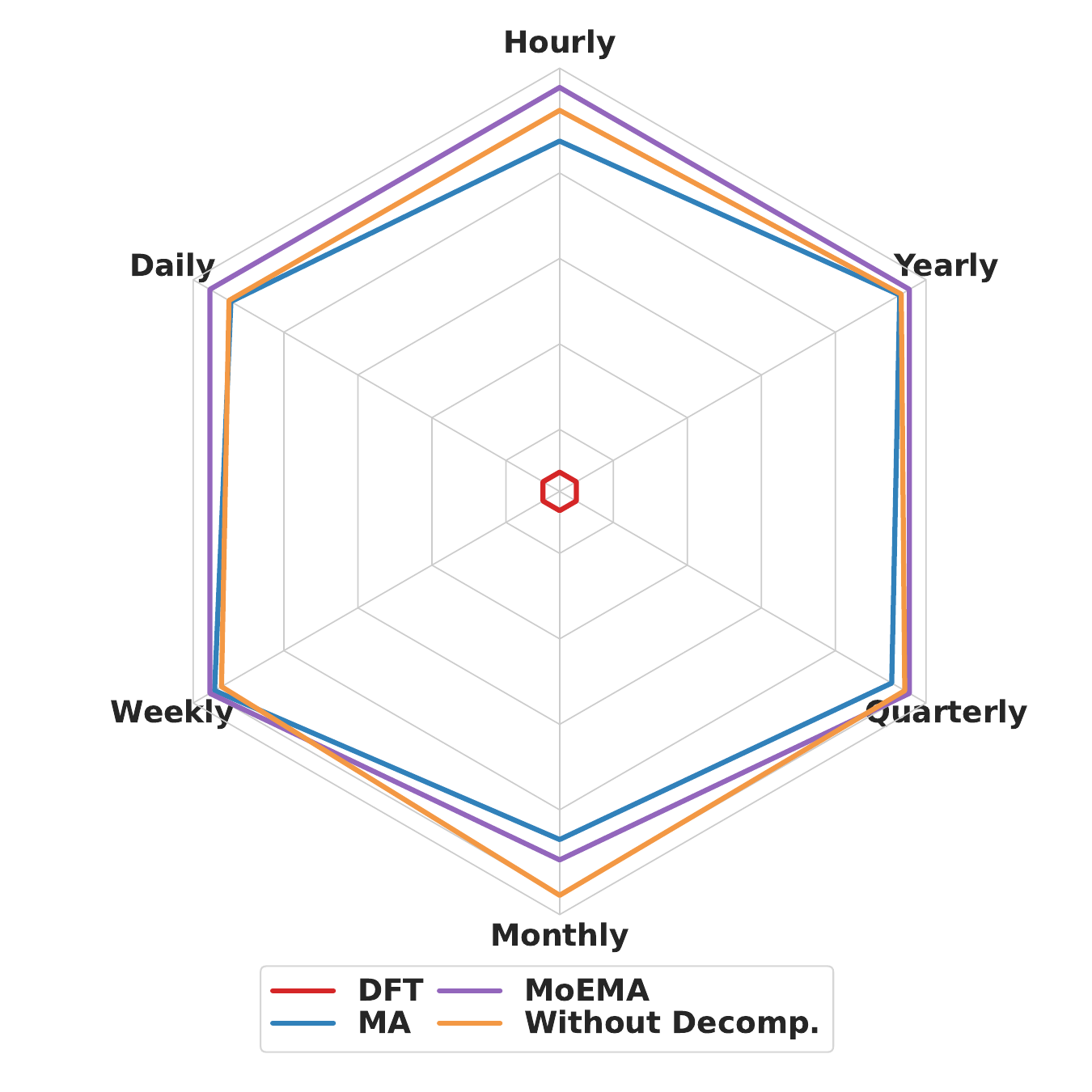}
         \caption{Series Decomposition}
         \label{fig:exp-appx-Decomposition-stf_smape}
     \end{subfigure}
     \hspace{10pt}
    \begin{subfigure}[t]{0.28\textwidth}
         \centering
         \includegraphics[width=\textwidth]{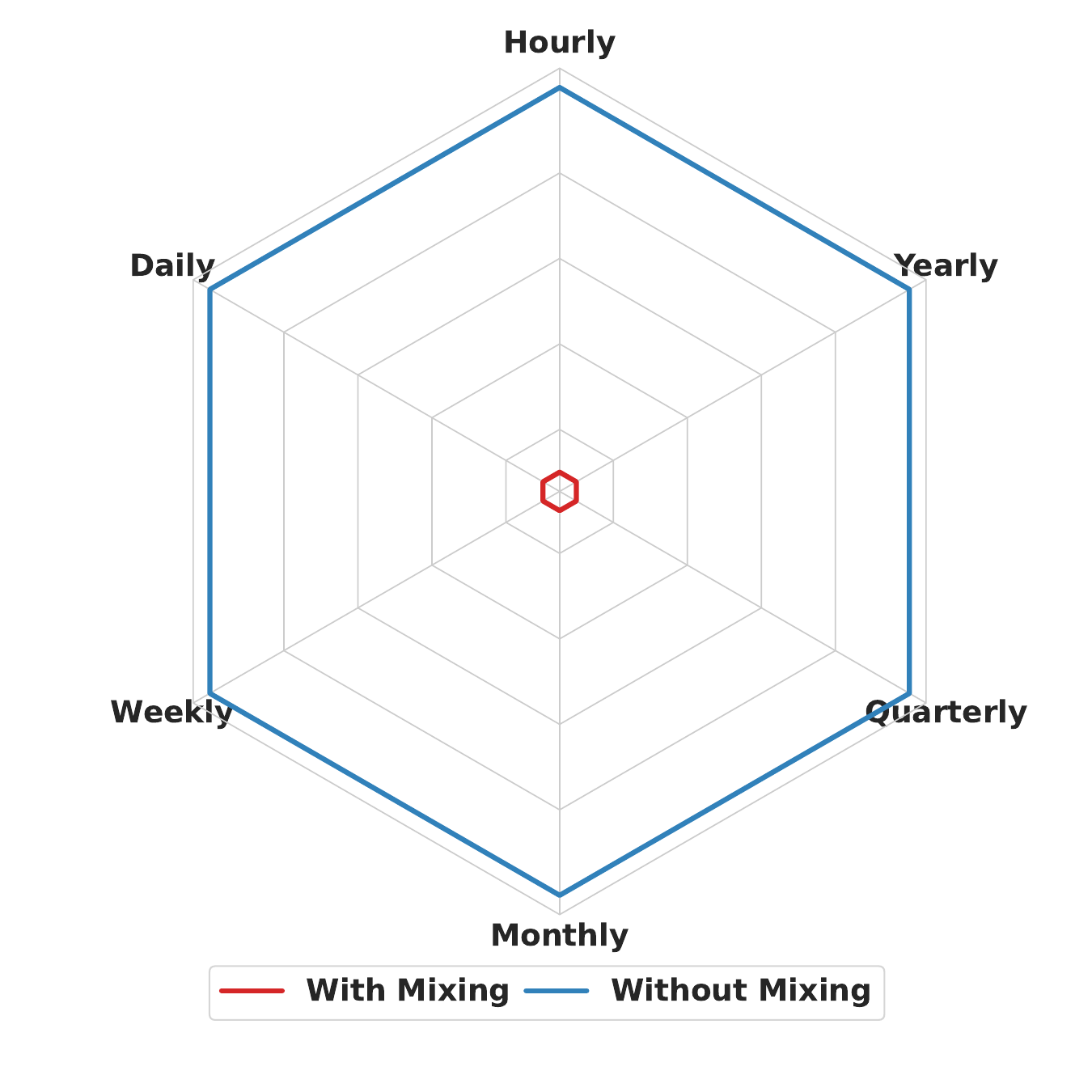}
         \caption{Series Sampling/Mixing}
         \label{fig:exp-appx-mixing-stf_smape}
     \end{subfigure}
     \hspace{10pt}
    \begin{subfigure}[t]{0.28\textwidth}
         \centering
         \includegraphics[width=\textwidth]{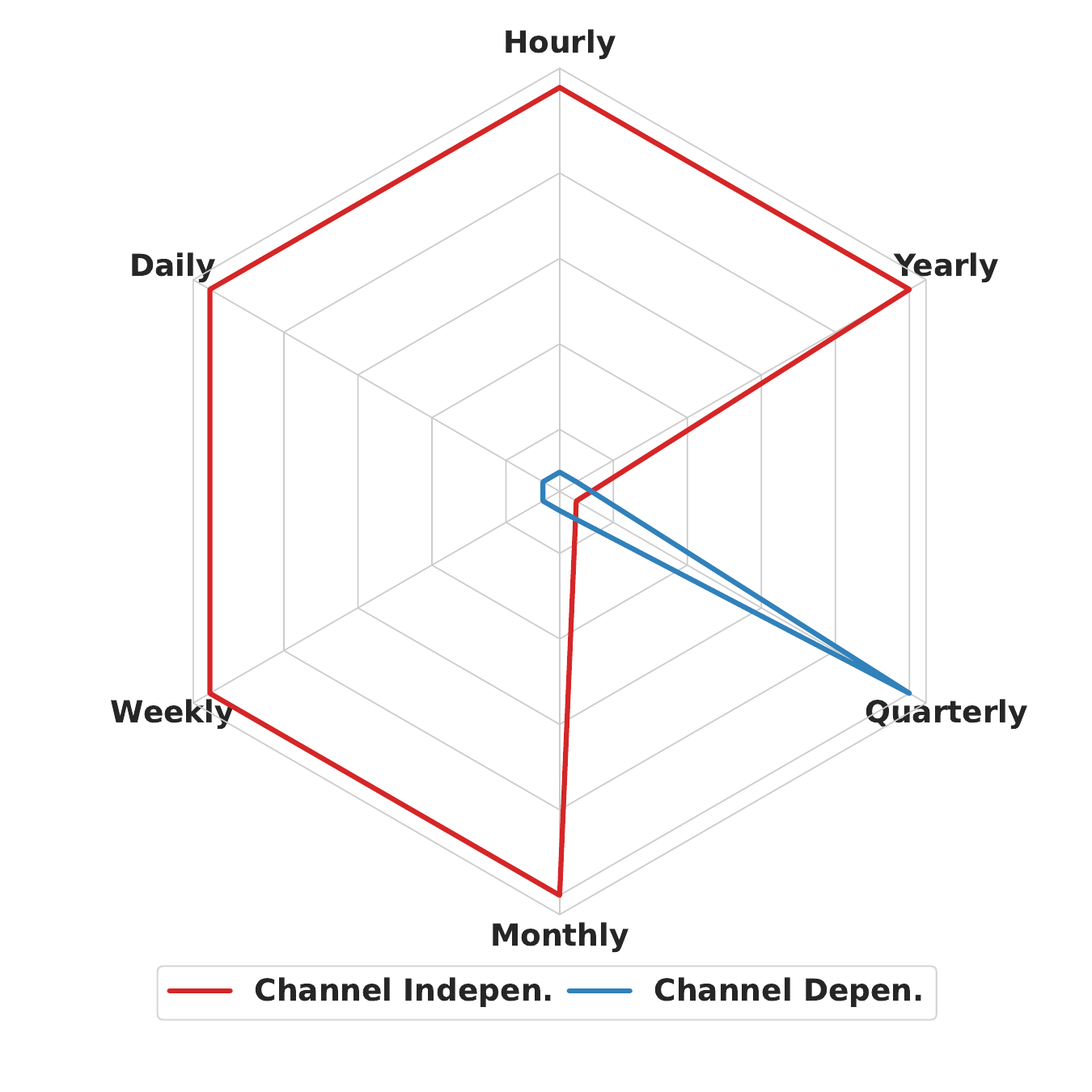}
         \caption{Channel Independent}
         \label{fig:exp-appx-CI-stf_smape}
     \end{subfigure}
     \hspace{10pt}
    \begin{subfigure}[t]{0.28\textwidth}
         \centering
         \includegraphics[width=\textwidth]{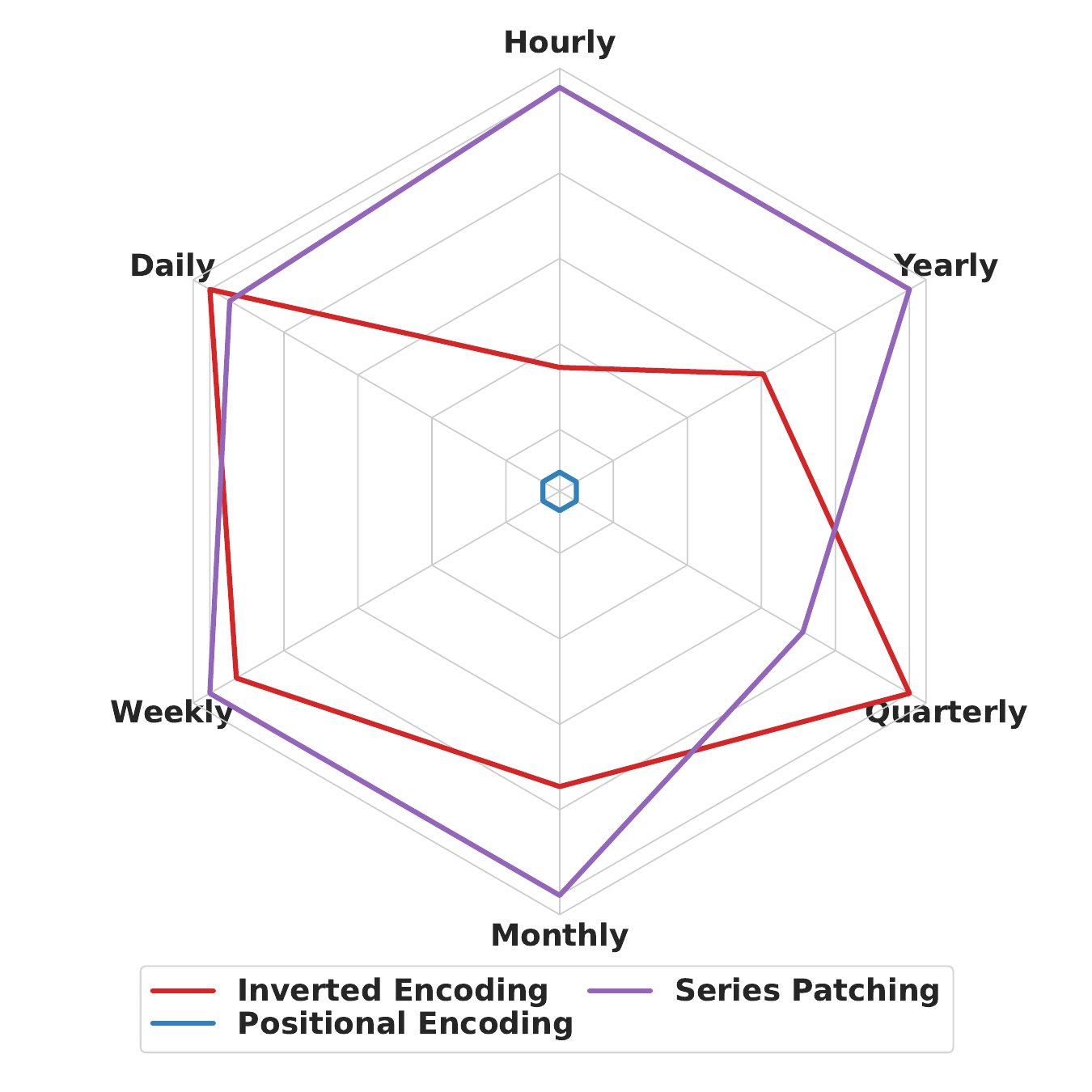}
         \caption{Series Embedding}
         \label{fig:exp-appx-tokenization-stf_smape}
     \end{subfigure}
     \hspace{10pt}
    \begin{subfigure}[t]{0.28\textwidth}
         \centering
         \includegraphics[width=\textwidth]{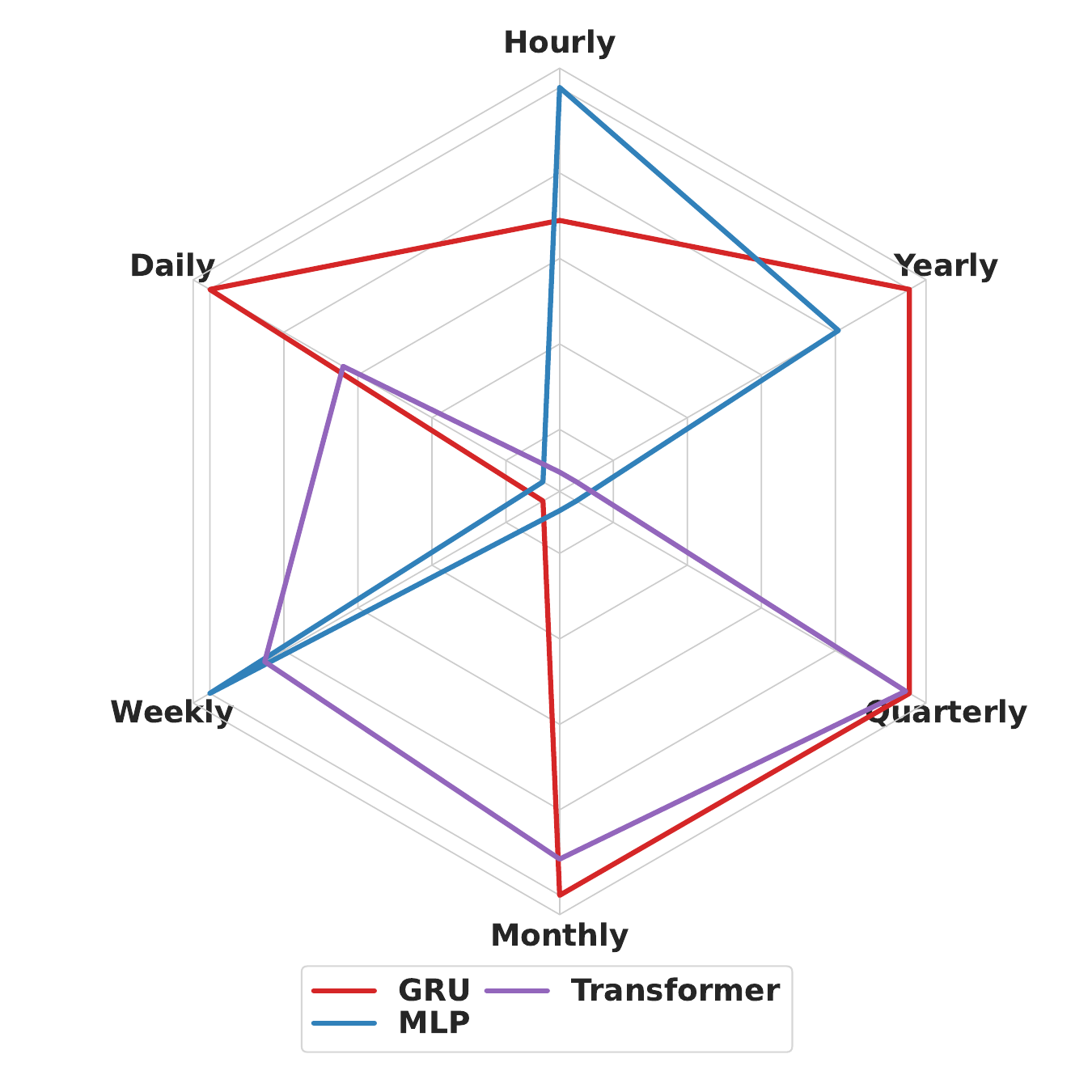}
         \caption{Network Backbone}
         \label{fig:exp-appx-backbone-stf_smape}
     \end{subfigure}
     \hspace{10pt}
    \begin{subfigure}[t]{0.28\textwidth}
         \centering
         \includegraphics[width=\textwidth]{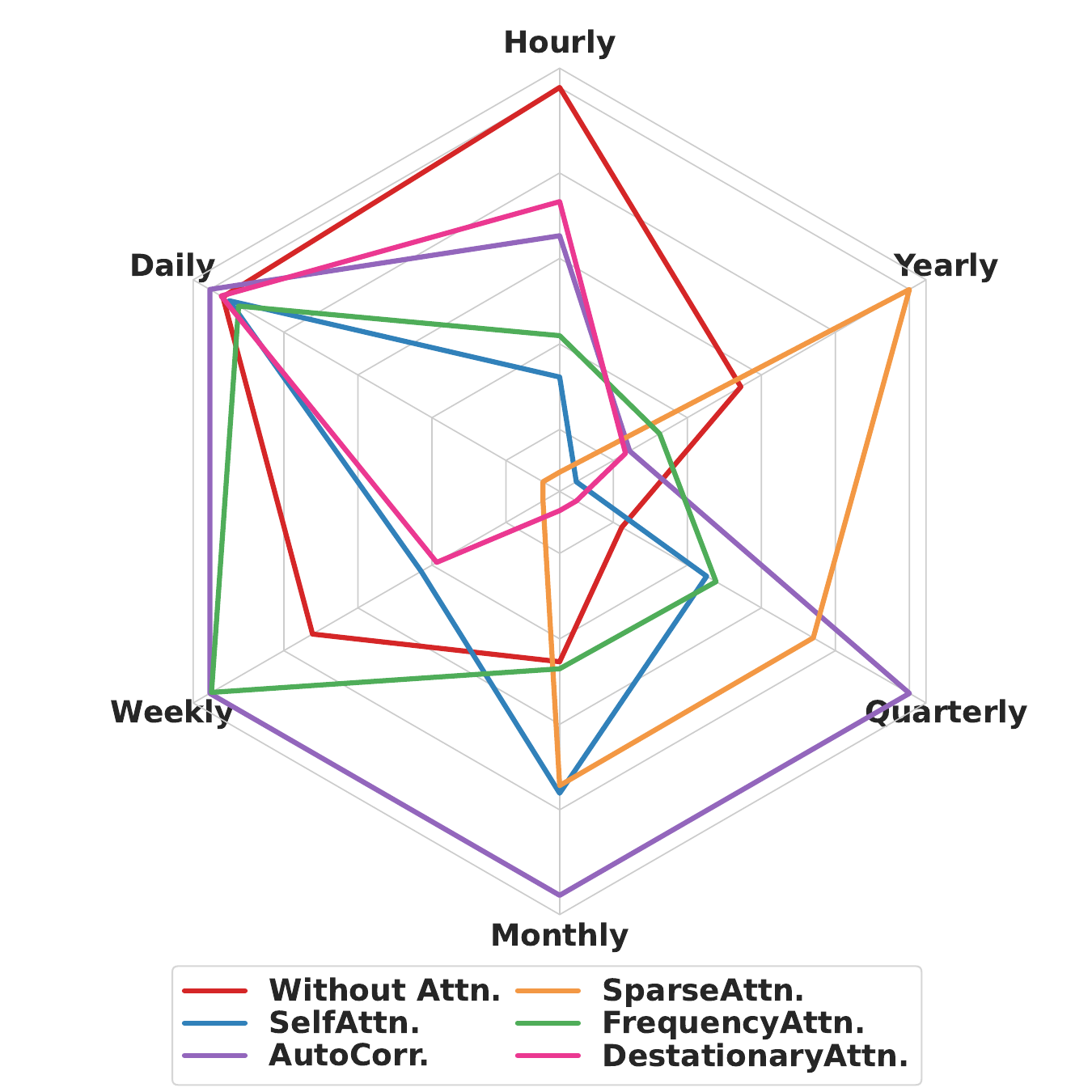}
         \caption{Series Attention}
         \label{fig:exp-appx-attention-stf_smape}
     \end{subfigure}
     \hspace{10pt}
    \begin{subfigure}[t]{0.28\textwidth}
         \centering
         \includegraphics[width=\textwidth]{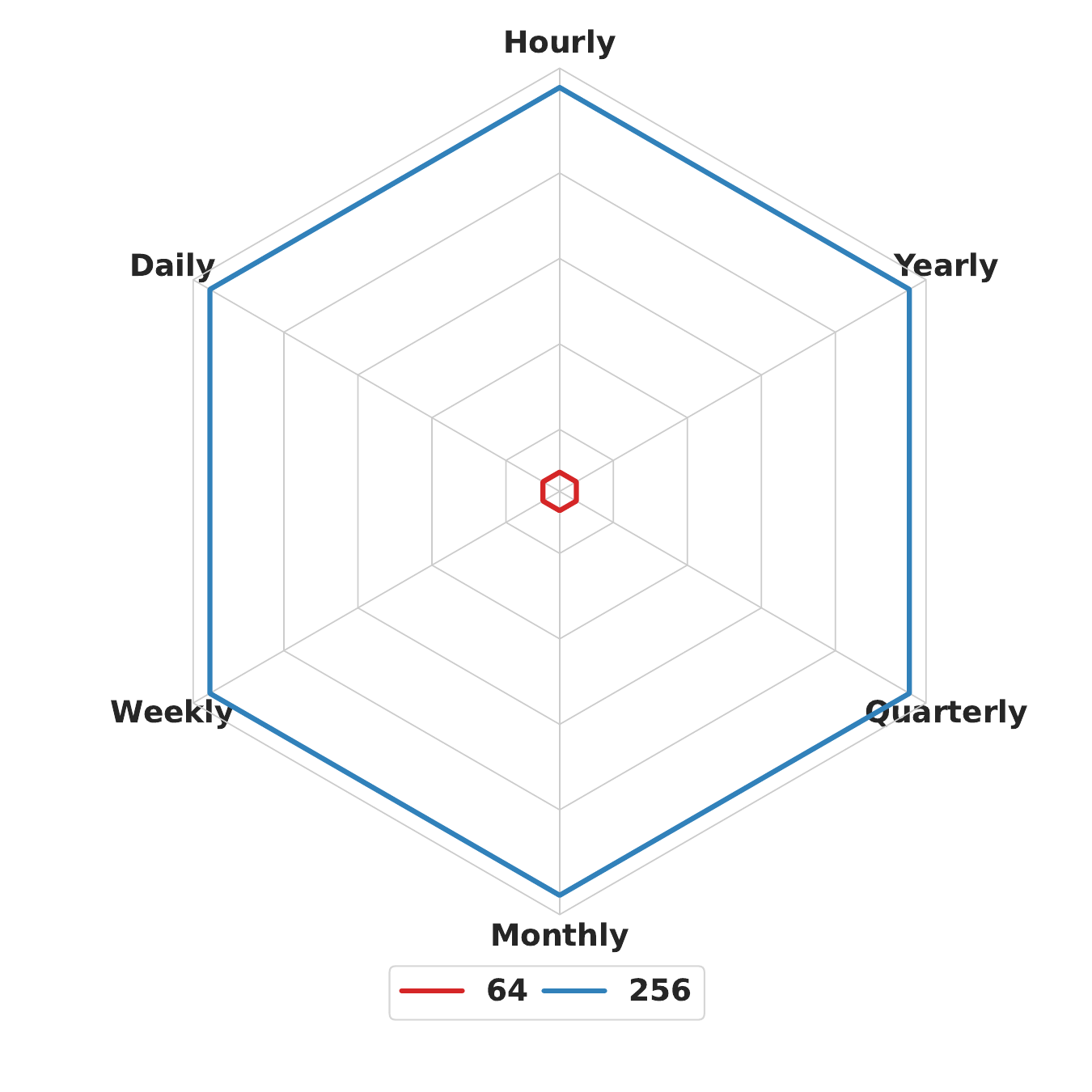}
         \caption{\textit{Hidden Layer Dimensions}}
         \label{fig:exp-appx-dmodel-stf_smape}
     \end{subfigure}
     \hspace{10pt}
    \begin{subfigure}[t]{0.28\textwidth}
         \centering
         \includegraphics[width=\textwidth]{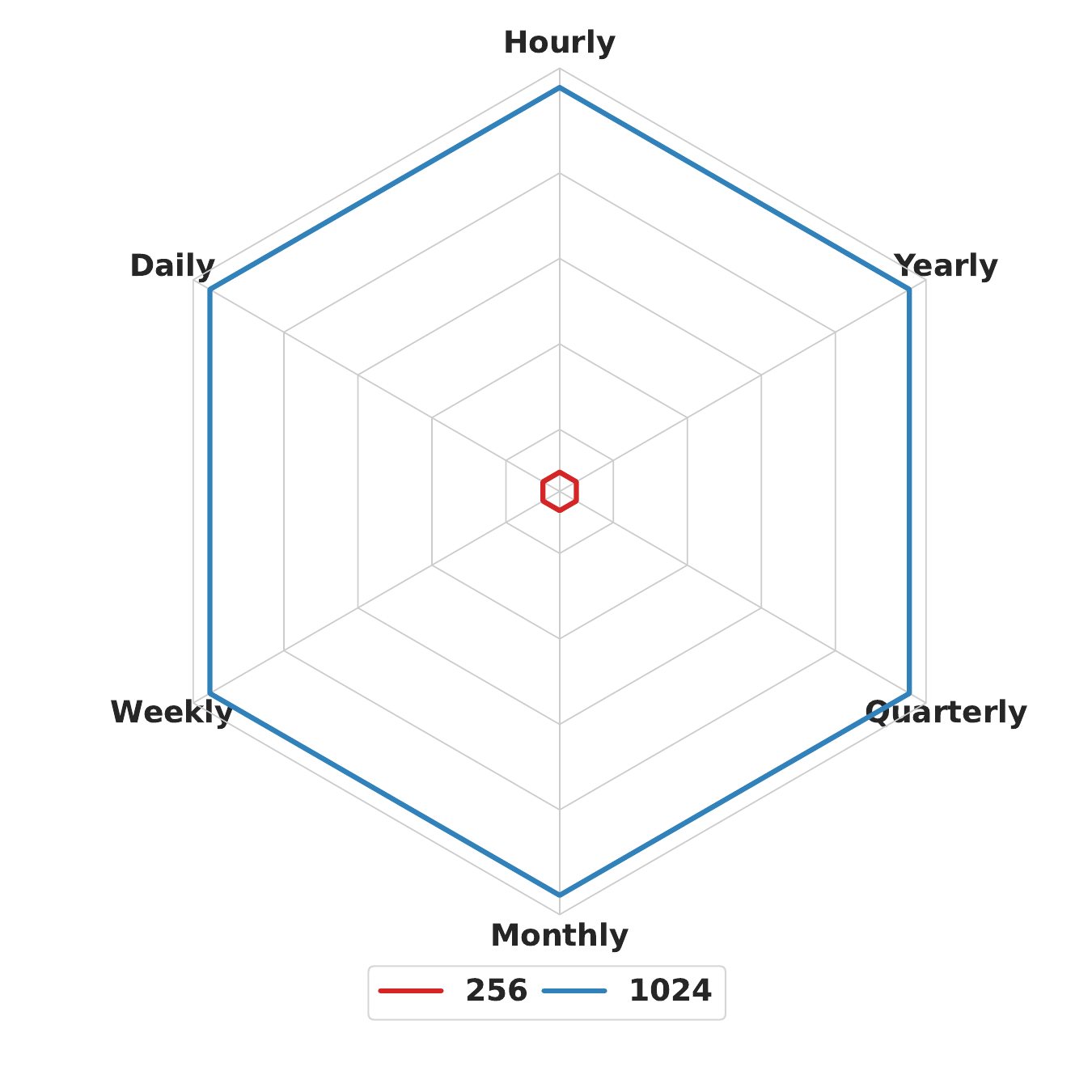}
         \caption{\textit{FCN Layer Dimensions}}
         \label{fig:exp-appx-dff-stf_smape}
     \end{subfigure}
     \hspace{10pt}
    \begin{subfigure}[t]{0.28\textwidth}
         \centering
         \includegraphics[width=\textwidth]{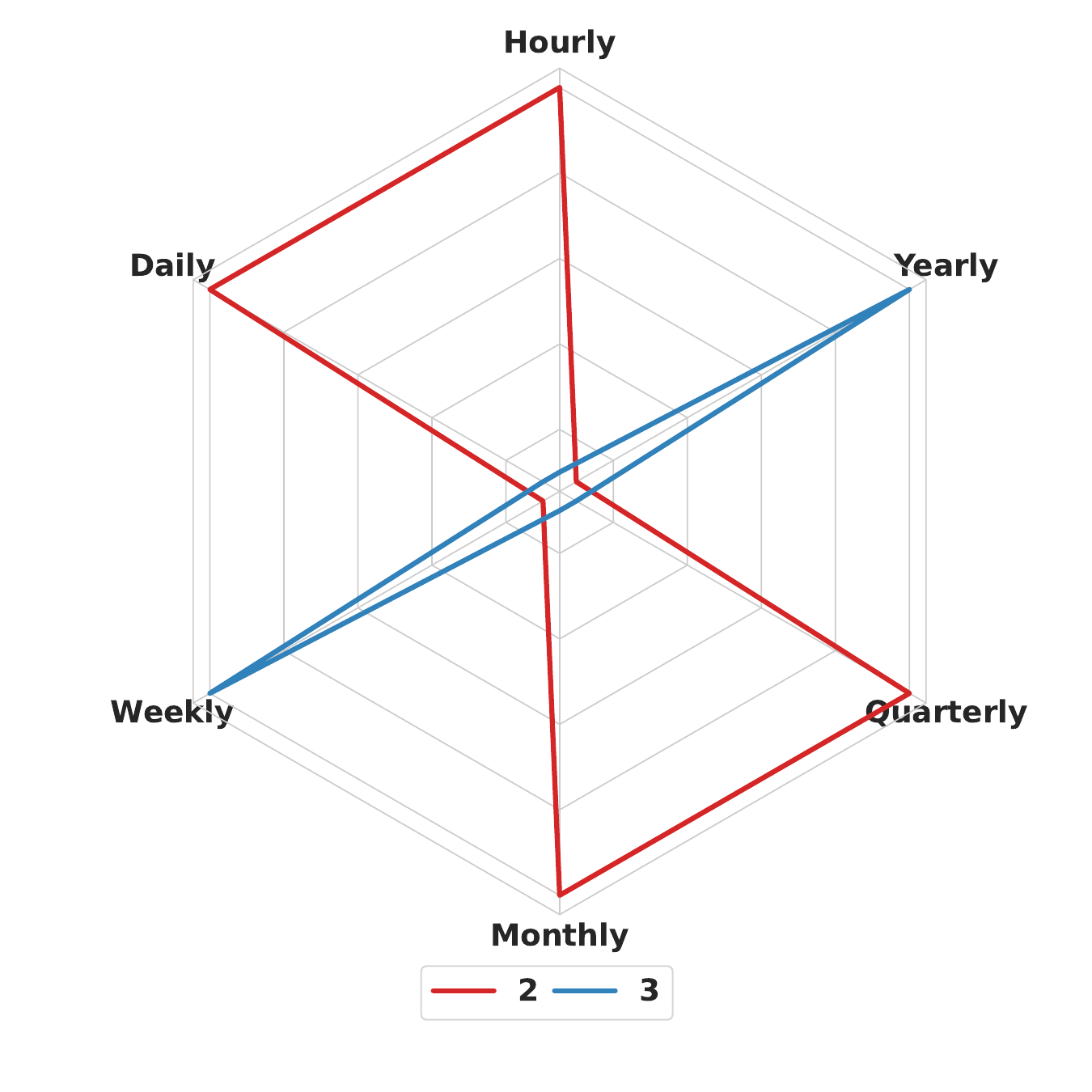}
         \caption{Encoder layers}
         \label{fig:exp-appx-el-stf_smape}
     \end{subfigure}
     \hspace{10pt}
    \begin{subfigure}[t]{0.28\textwidth}
         \centering
         \includegraphics[width=\textwidth]{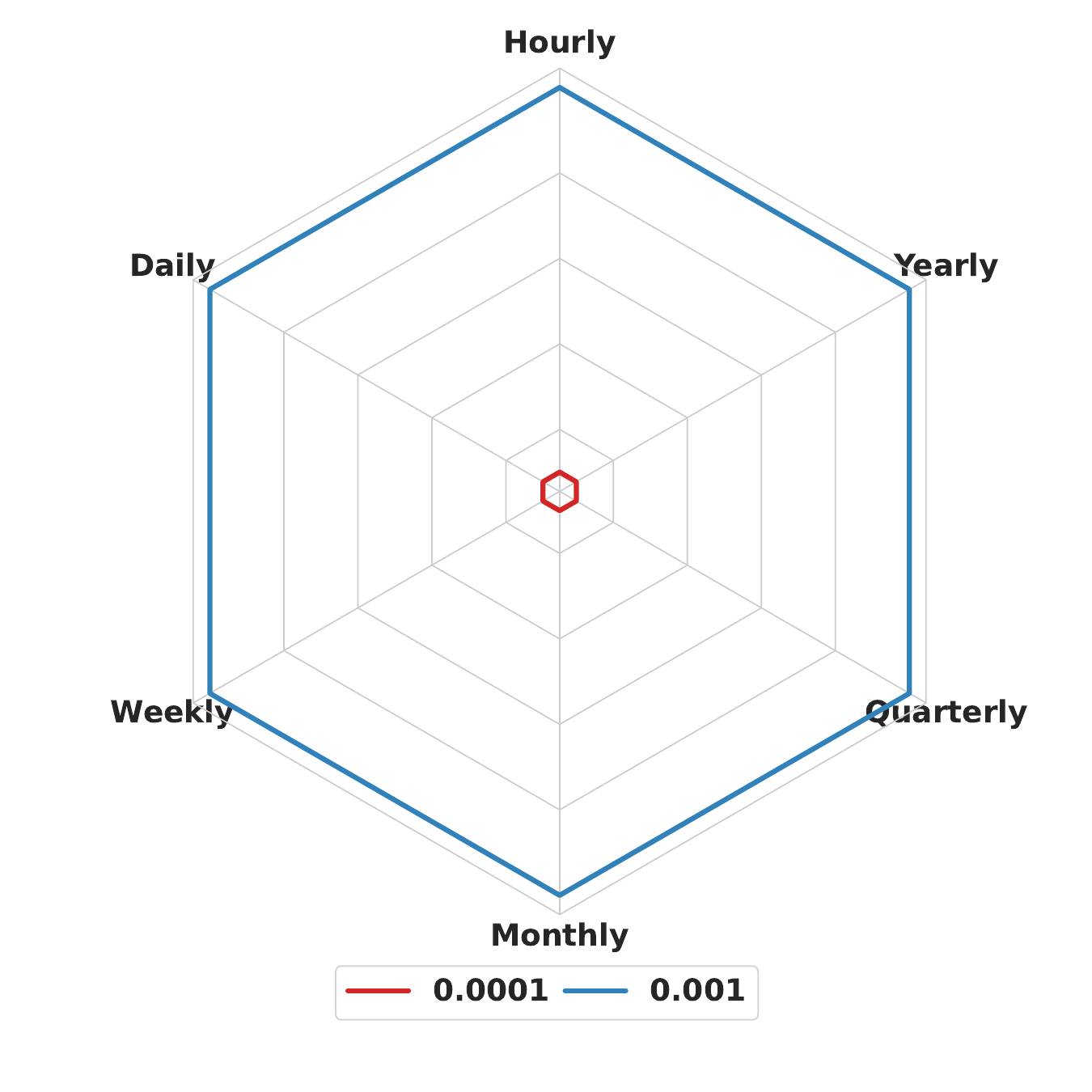}
         \caption{Learning Rate}
         \label{fig:exp-appx-lr-stf_smape}
     \end{subfigure}
     \hspace{10pt}
    \begin{subfigure}[t]{0.28\textwidth}
         \centering
         \includegraphics[width=\textwidth]{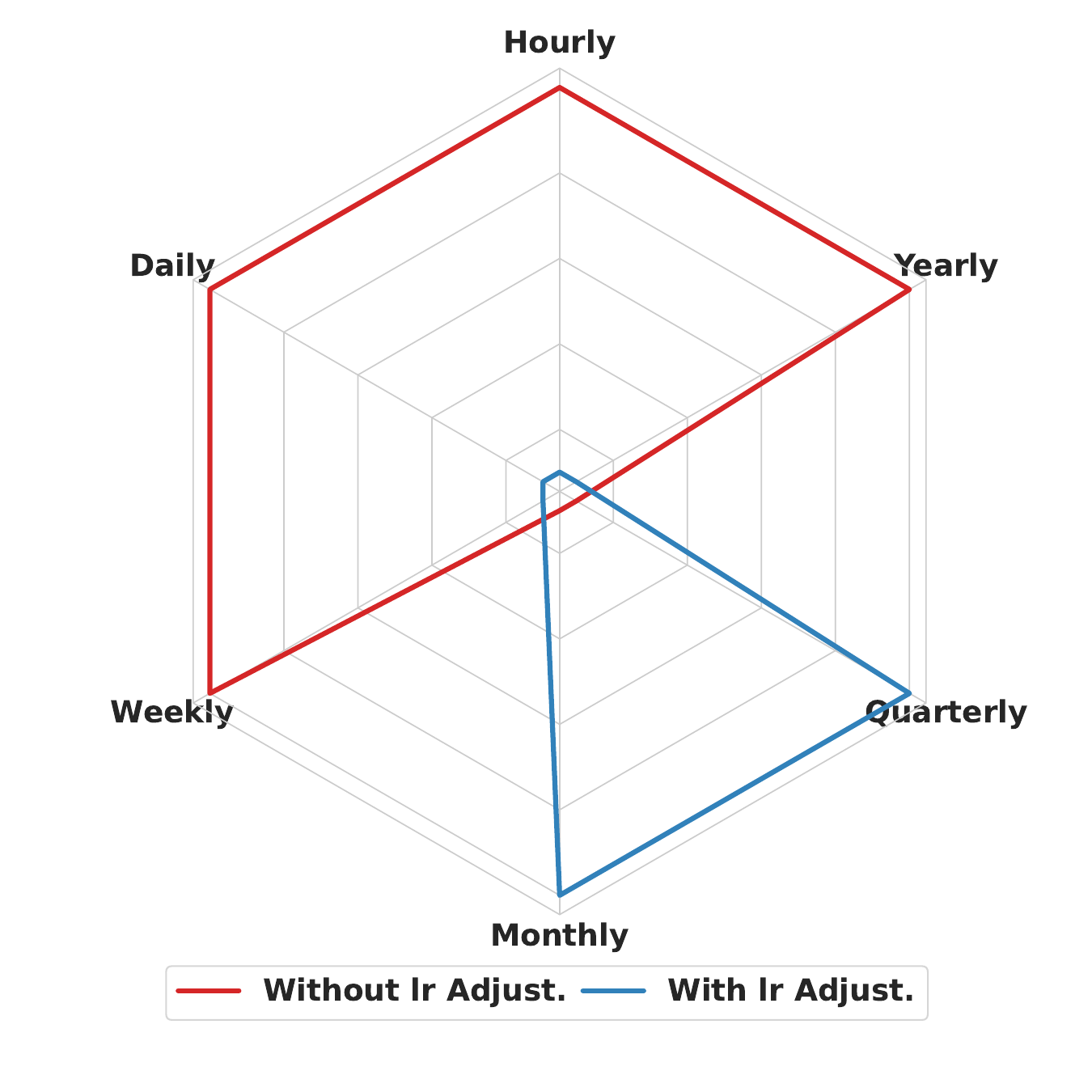}
         \caption{Learning Rate Strategy}
         \label{fig:exp-appx-lrs-stf_smape}
     \end{subfigure}
     \hspace{10pt}
     \caption{Overall performance across all design dimensions in short-term forecasting. The results (\textbf{SMAPE}) are based on the top 25th percentile across all forecasting horizons.}
     \vspace{-0.1in}
     \label{fig:exp-appx-rada-stf_smape}
\end{figure}

For short-term forecasting, we comprehensively evaluate different design dimensions using both spider charts and box plots. The spider charts—shown in Figure~\ref{fig:exp-appx-rada-stf_mase}, Figure~\ref{fig:exp-appx-rada-stf_owa}, and Figure~\ref{fig:exp-appx-rada-stf_smape}—visualize performance across datasets, with each vertex representing a benchmark dataset. Closer proximity to a vertex indicates stronger performance of a particular design choice in that dataset.

\begin{figure}[t!]
     \centering
     \begin{subfigure}[t]{0.28\textwidth}
         \centering
         \includegraphics[width=\textwidth]{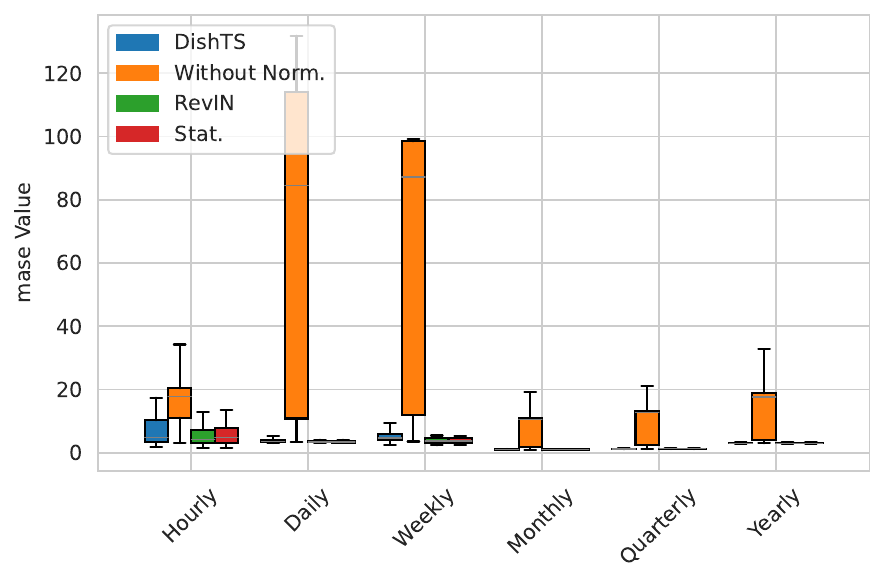}
         \caption{Series Normalization}
         \label{fig:exp-appx-Normalization-stf_bp}
     \end{subfigure}
     \hspace{10pt}
     \begin{subfigure}[t]{0.28\textwidth}
         \centering
         \includegraphics[width=\textwidth]{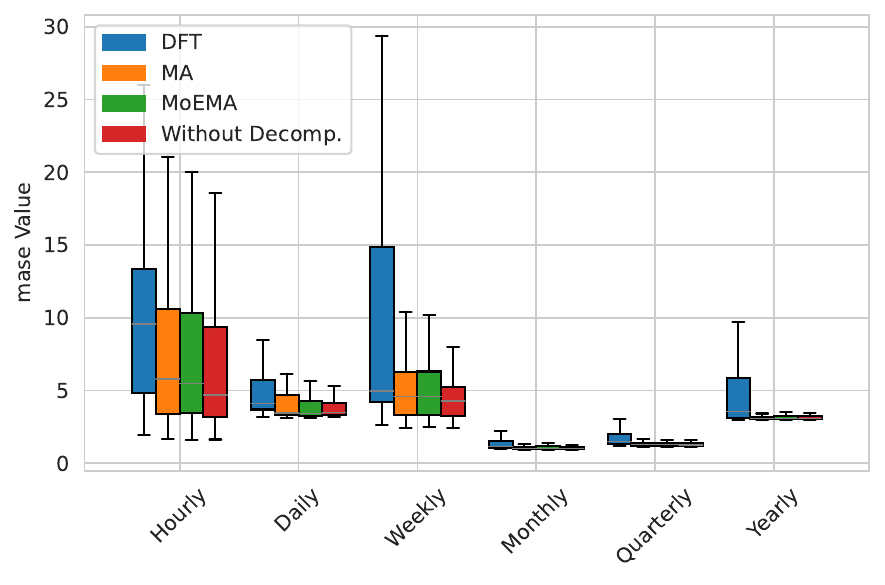}
         \caption{Series Decomposition}
         \label{fig:exp-appx-Decomposition-stf_bp}
     \end{subfigure}
     \hspace{10pt}
    \begin{subfigure}[t]{0.28\textwidth}
         \centering
         \includegraphics[width=\textwidth]{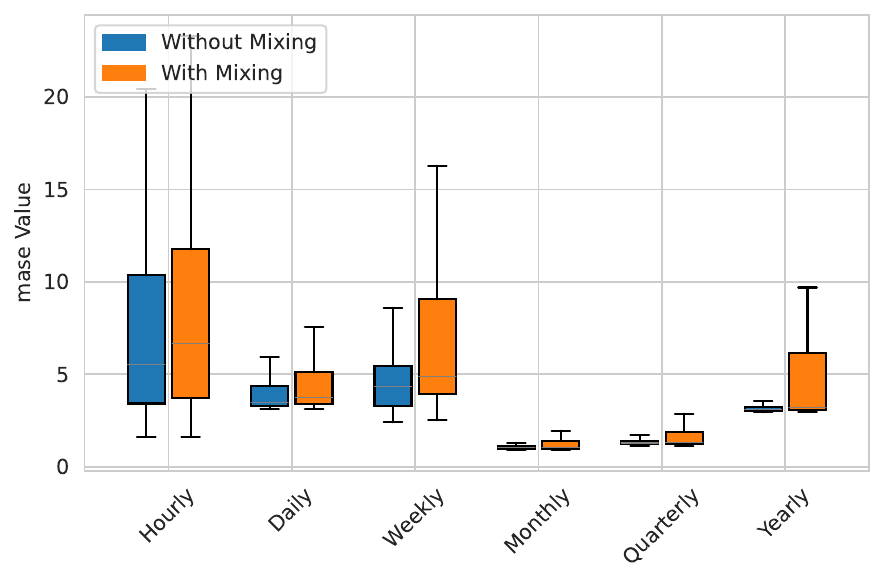}
         \caption{Series Sampling/Mixing}
         \label{fig:exp-appx-mixing-stf_bp}
     \end{subfigure}
     \hspace{10pt}
    \begin{subfigure}[t]{0.28\textwidth}
         \centering
         \includegraphics[width=\textwidth]{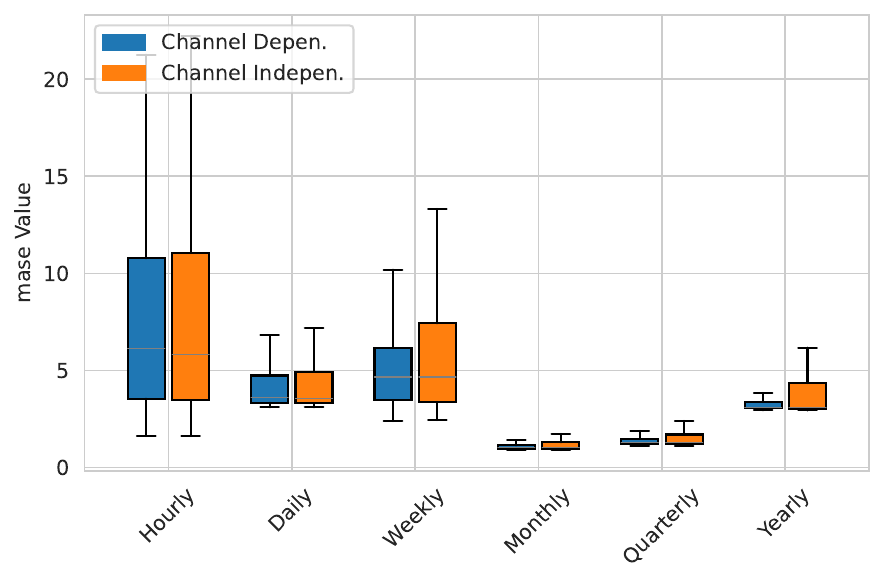}
         \caption{Channel Independent}
         \label{fig:exp-appx-CI-stf_bp}
     \end{subfigure}
     \hspace{10pt}
    \begin{subfigure}[t]{0.28\textwidth}
         \centering
         \includegraphics[width=\textwidth]{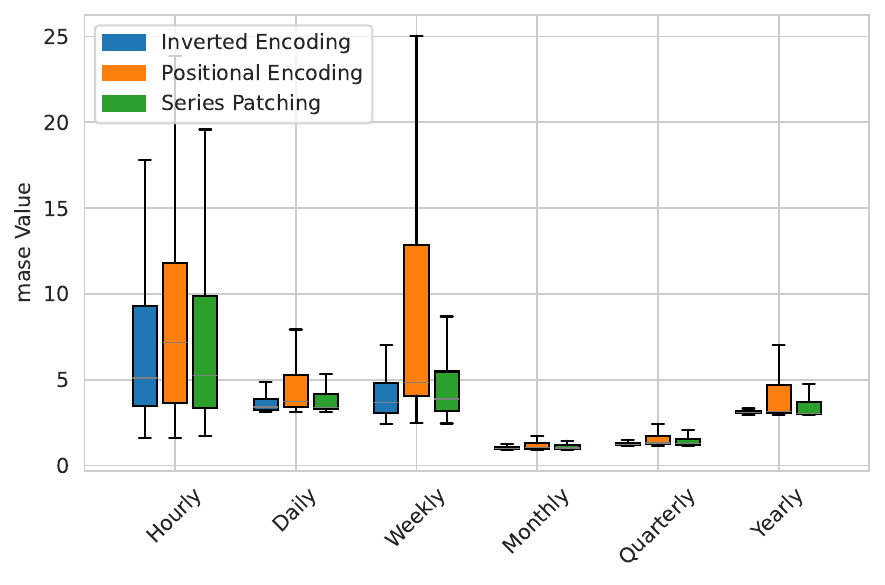}
         \caption{Series Embedding}
         \label{fig:exp-appx-tokenization-stf_bp}
     \end{subfigure}
     \hspace{10pt}
    \begin{subfigure}[t]{0.28\textwidth}
         \centering
         \includegraphics[width=\textwidth]{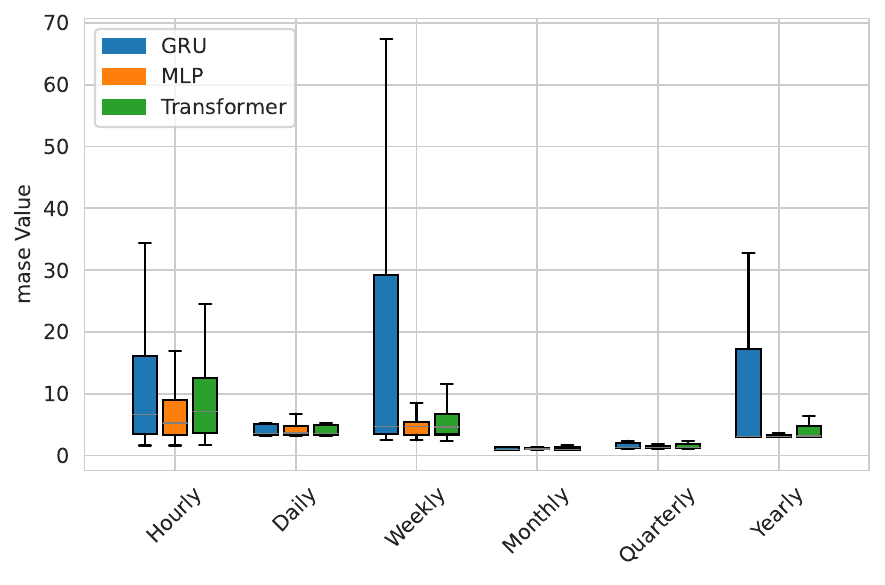}
         \caption{Network Backbone}
         \label{fig:exp-appx-backbone-stf_bp}
     \end{subfigure}
     \hspace{10pt}
    \begin{subfigure}[t]{0.28\textwidth}
         \centering
         \includegraphics[width=\textwidth]{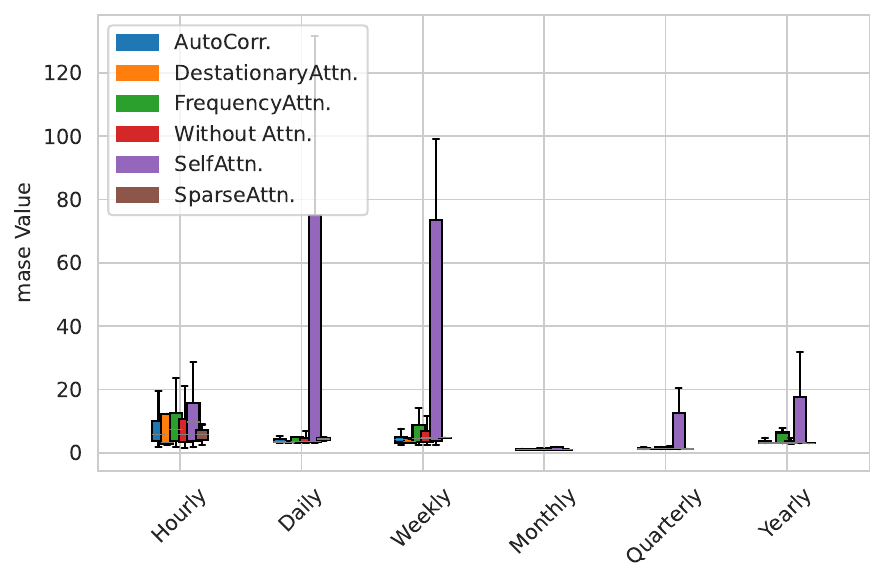}
         \caption{Series Attention}
         \label{fig:exp-appx-attention-stf_bp}
     \end{subfigure}
     \hspace{10pt}
    \begin{subfigure}[t]{0.28\textwidth}
         \centering
         \includegraphics[width=\textwidth]{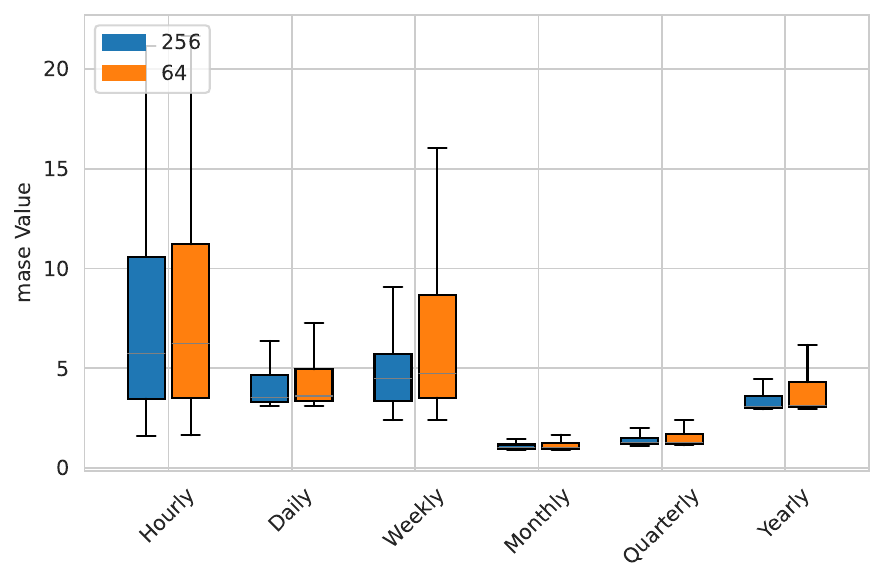}
         \caption{\textit{Hidden Layer Dimensions}}
         \label{fig:exp-appx-dmodel-stf_bp}
     \end{subfigure}
     \hspace{10pt}
    \begin{subfigure}[t]{0.28\textwidth}
         \centering
         \includegraphics[width=\textwidth]{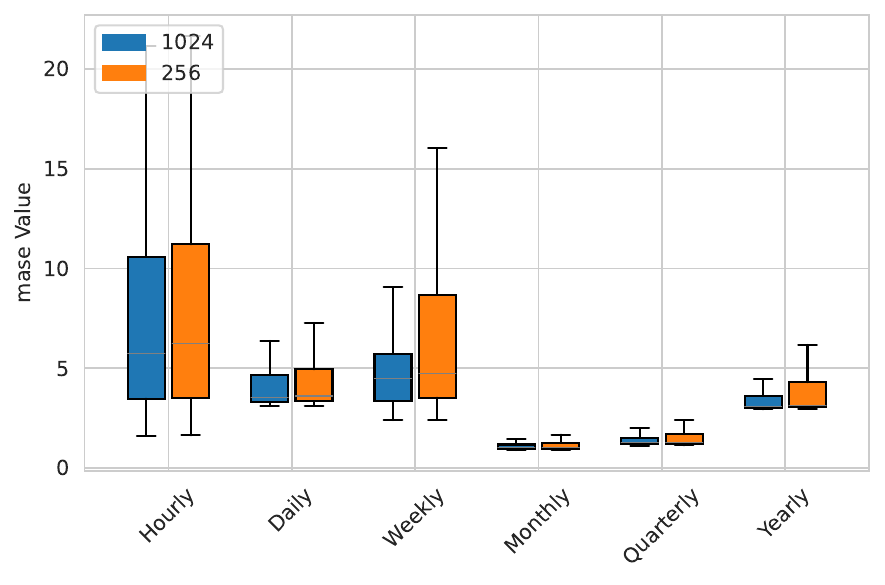}
         \caption{\textit{FCN Layer Dimensions}}
         \label{fig:exp-appx-dff-stf_bp}
     \end{subfigure}
     \hspace{10pt}
    \begin{subfigure}[t]{0.28\textwidth}
         \centering
         \includegraphics[width=\textwidth]{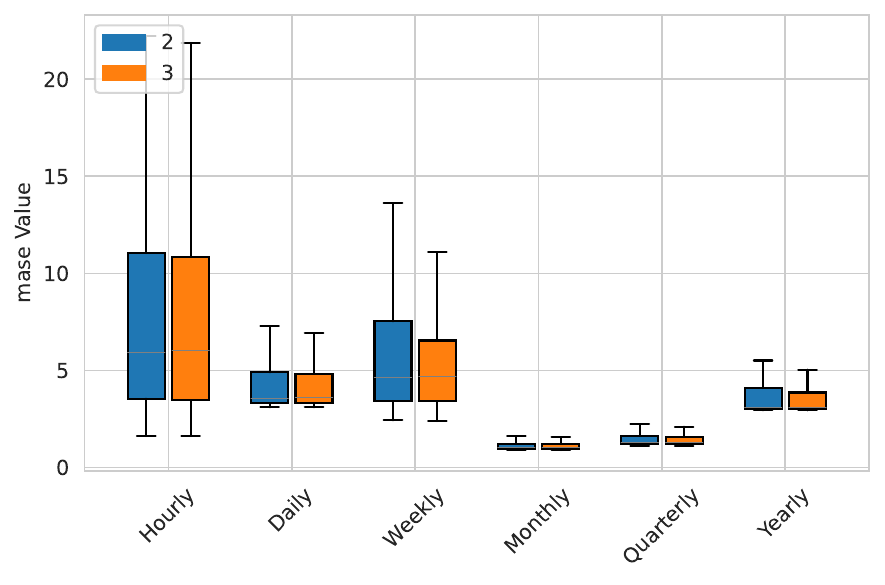}
         \caption{Encoder layers}
         \label{fig:exp-appx-el-stf_bp}
     \end{subfigure}
     \hspace{10pt}
    \begin{subfigure}[t]{0.28\textwidth}
         \centering
         \includegraphics[width=\textwidth]{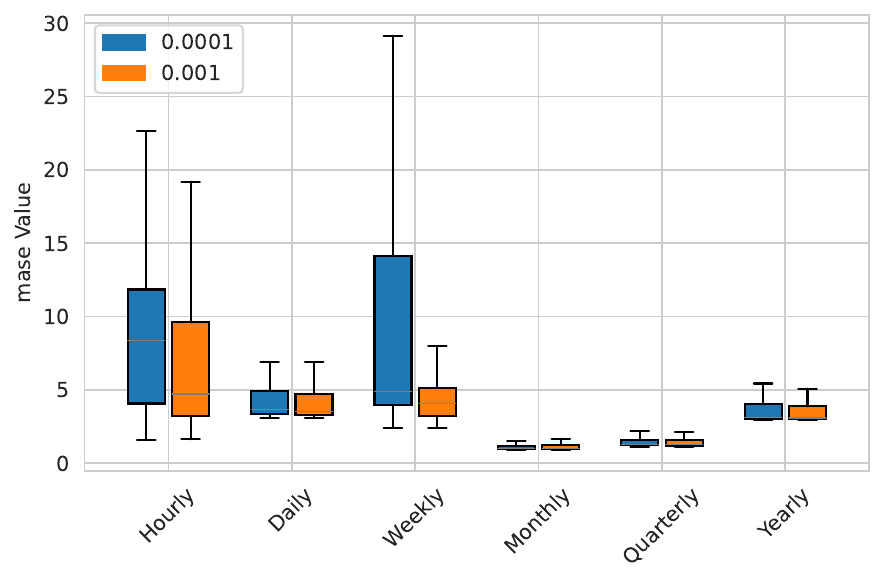}
         \caption{Learning Rate}
         \label{fig:exp-appx-lr-stf_bp}
     \end{subfigure}
     \hspace{10pt}
    \begin{subfigure}[t]{0.28\textwidth}
         \centering
         \includegraphics[width=\textwidth]{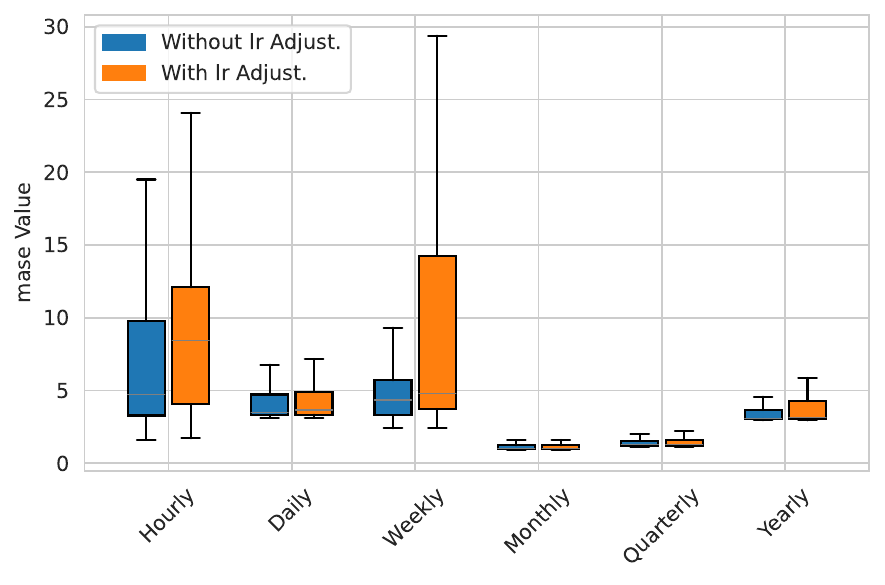}
         \caption{Learning Rate Strategy}
         \label{fig:exp-appx-lrs-stf_bp}
     \end{subfigure}
     \hspace{10pt}
     \caption{Overall performance across all design dimensions in short-term forecasting. The results are based on \textbf{MASE}.}
     \vspace{-0.1in}
     \label{fig:exp-appx-rada-stf_mase_bp}
\end{figure}

\begin{figure}[t!]
     \centering
     \begin{subfigure}[t]{0.28\textwidth}
         \centering
         \includegraphics[width=\textwidth]{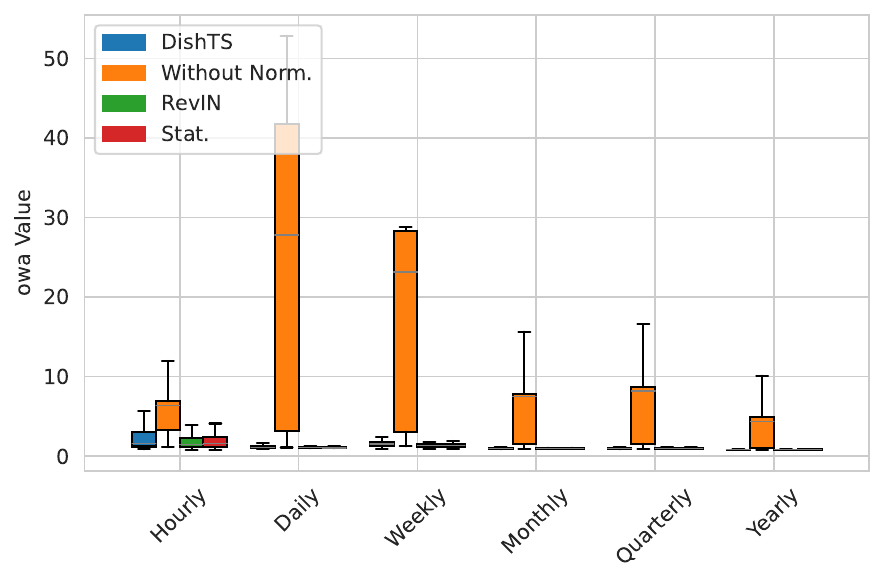}
         \caption{Series Normalization}
         \label{fig:exp-appx-Normalization-stf_owa_bp}
     \end{subfigure}
     \hspace{10pt}
     \begin{subfigure}[t]{0.28\textwidth}
         \centering
         \includegraphics[width=\textwidth]{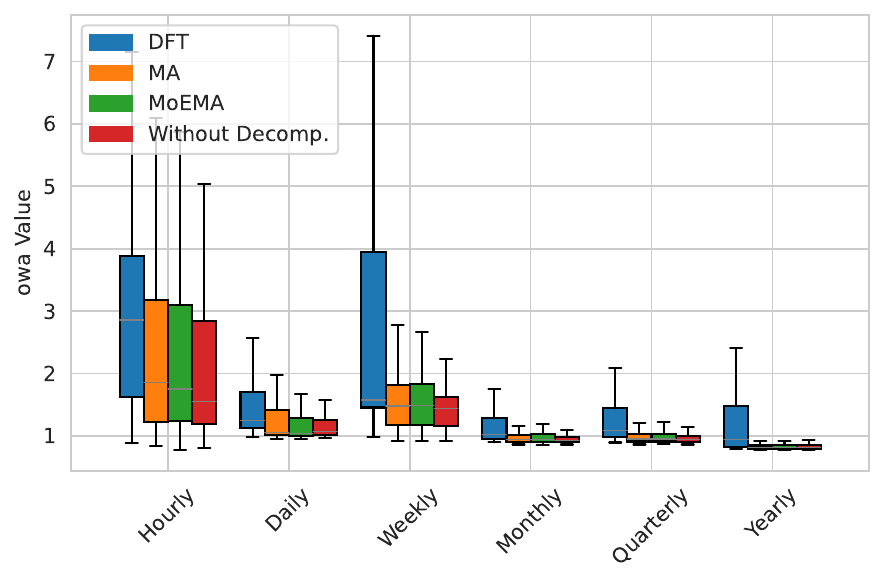}
         \caption{Series Decomposition}
         \label{fig:exp-appx-Decomposition-stf_owa_bp}
     \end{subfigure}
     \hspace{10pt}
    \begin{subfigure}[t]{0.28\textwidth}
         \centering
         \includegraphics[width=\textwidth]{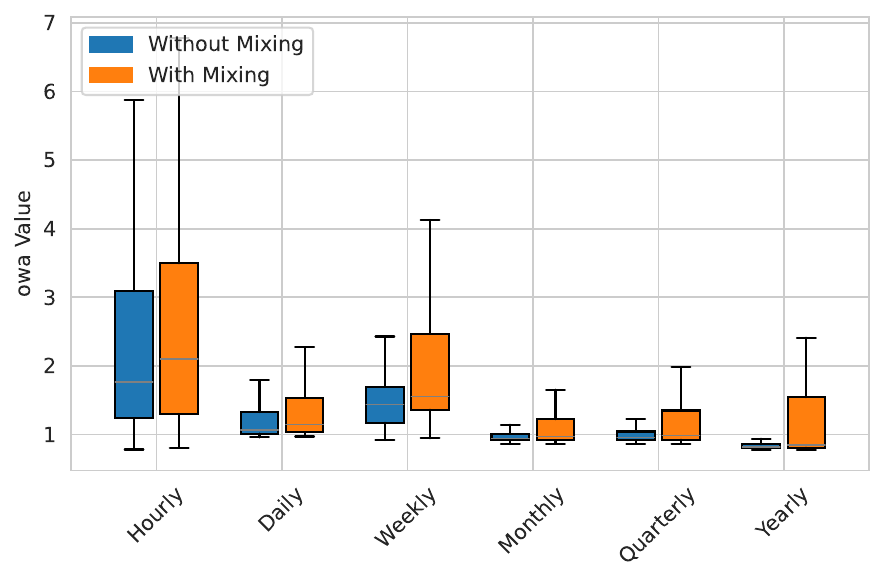}
         \caption{Series Sampling/Mixing}
         \label{fig:exp-appx-mixing-stf_owa_bp}
     \end{subfigure}
     \hspace{10pt}
    \begin{subfigure}[t]{0.28\textwidth}
         \centering
         \includegraphics[width=\textwidth]{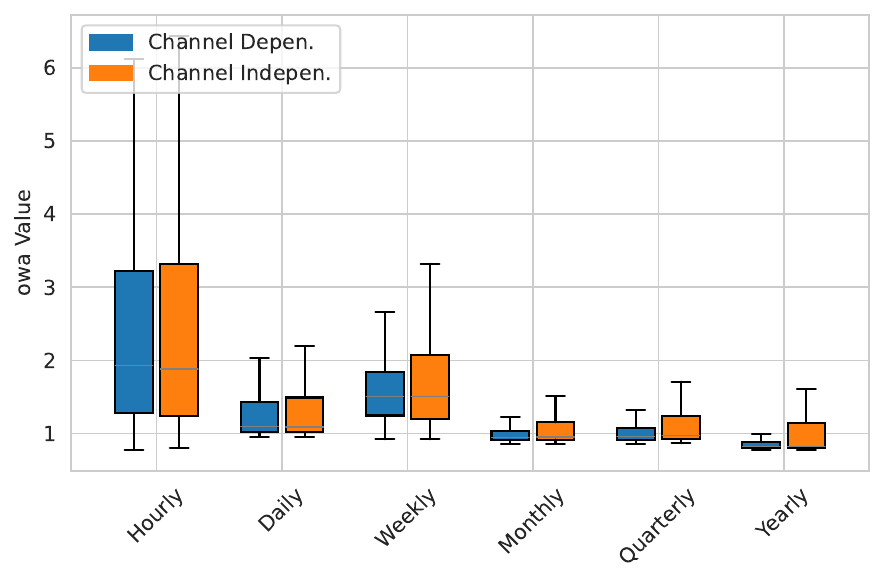}
         \caption{Channel Independent}
         \label{fig:exp-appx-CI-stf_owa_bp}
     \end{subfigure}
     \hspace{10pt}
    \begin{subfigure}[t]{0.28\textwidth}
         \centering
         \includegraphics[width=\textwidth]{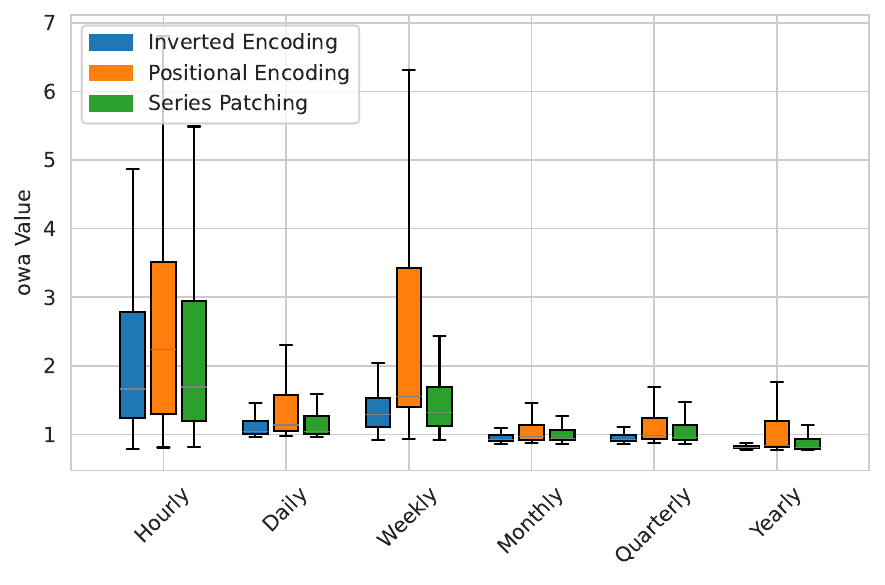}
         \caption{Series Embedding}
         \label{fig:exp-appx-tokenization-stf_owa_bp}
     \end{subfigure}
     \hspace{10pt}
    \begin{subfigure}[t]{0.28\textwidth}
         \centering
         \includegraphics[width=\textwidth]{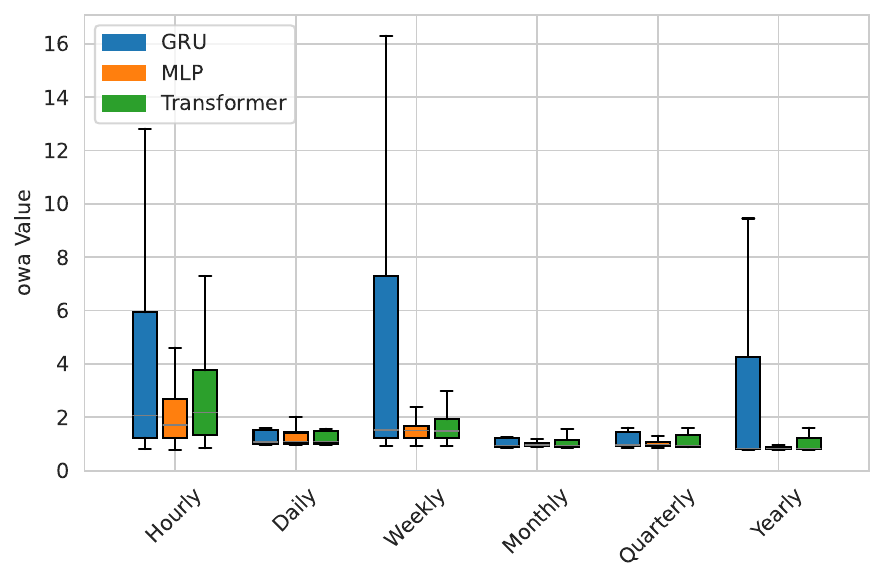}
         \caption{Network Backbone}
         \label{fig:exp-appx-backbone-stf_owa_bp}
     \end{subfigure}
     \hspace{10pt}
    \begin{subfigure}[t]{0.28\textwidth}
         \centering
         \includegraphics[width=\textwidth]{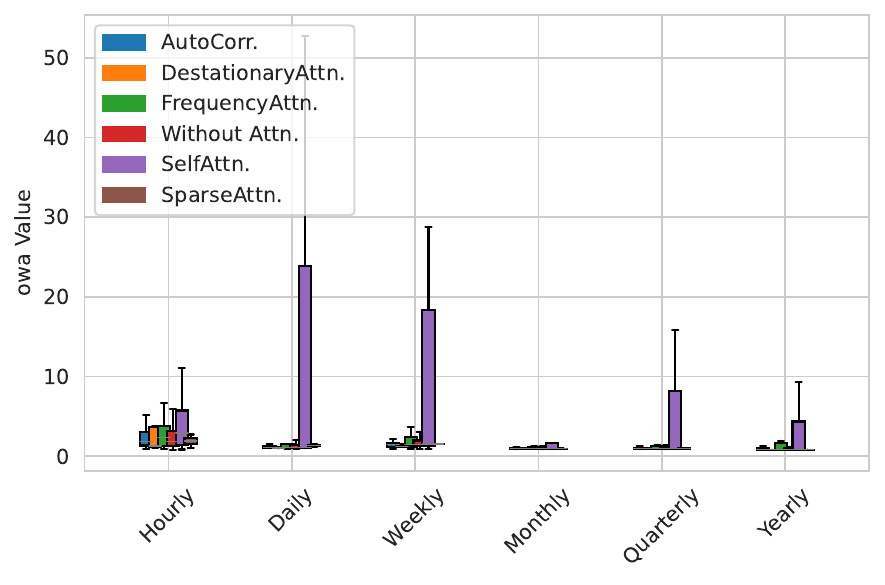}
         \caption{Series Attention}
         \label{fig:exp-appx-attention-stf_owa_bp}
     \end{subfigure}
     \hspace{10pt}
    \begin{subfigure}[t]{0.28\textwidth}
         \centering
         \includegraphics[width=\textwidth]{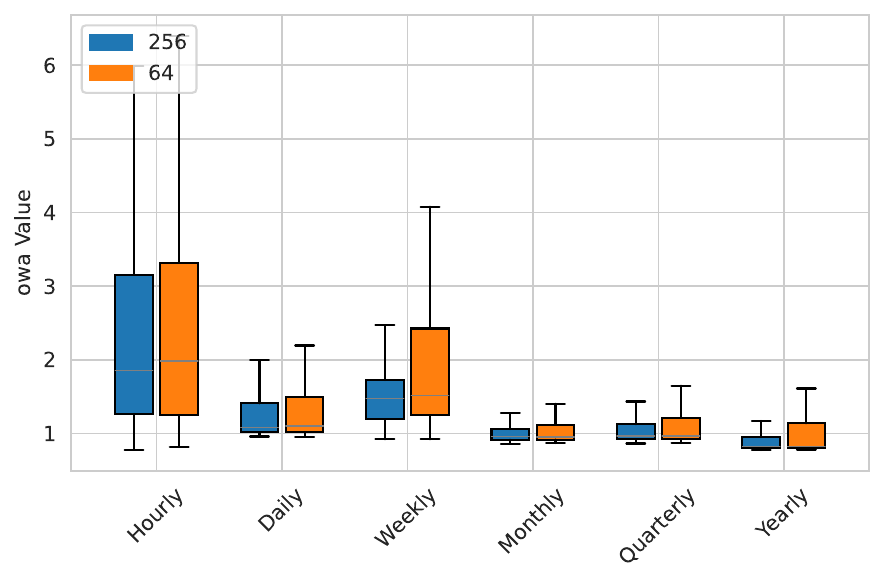}
         \caption{\textit{Hidden Layer Dimensions}}
         \label{fig:exp-appx-dmodel-stf_owa_bp}
     \end{subfigure}
     \hspace{10pt}
    \begin{subfigure}[t]{0.28\textwidth}
         \centering
         \includegraphics[width=\textwidth]{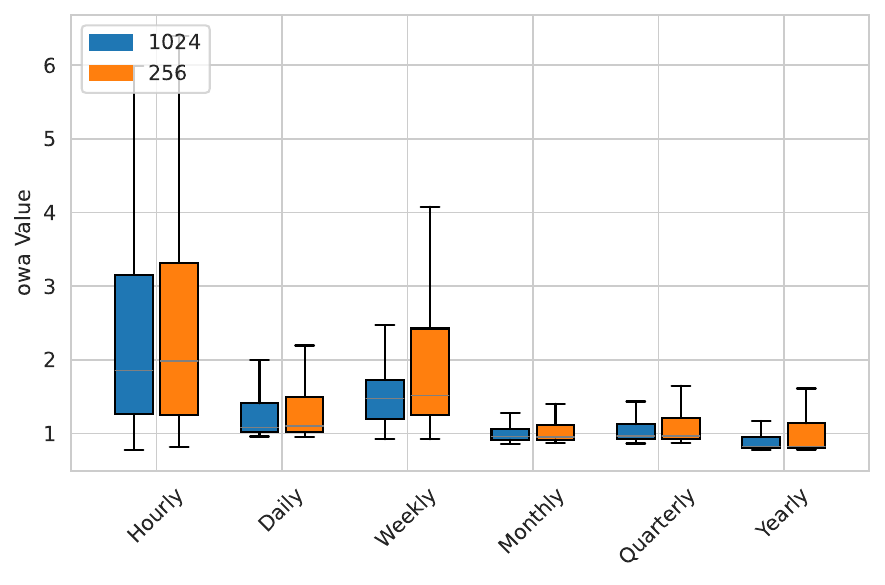}
         \caption{\textit{FCN Layer Dimensions}}
         \label{fig:exp-appx-dff-stf_owa_bp}
     \end{subfigure}
     \hspace{10pt}
    \begin{subfigure}[t]{0.28\textwidth}
         \centering
         \includegraphics[width=\textwidth]{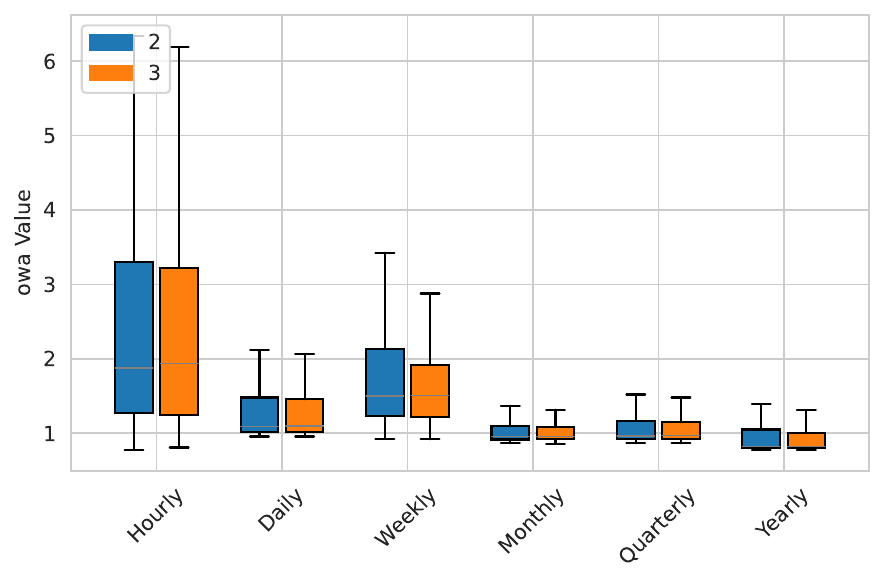}
         \caption{Encoder layers}
         \label{fig:exp-appx-el-stf_owa_bp}
     \end{subfigure}
     \hspace{10pt}
    \begin{subfigure}[t]{0.28\textwidth}
         \centering
         \includegraphics[width=\textwidth]{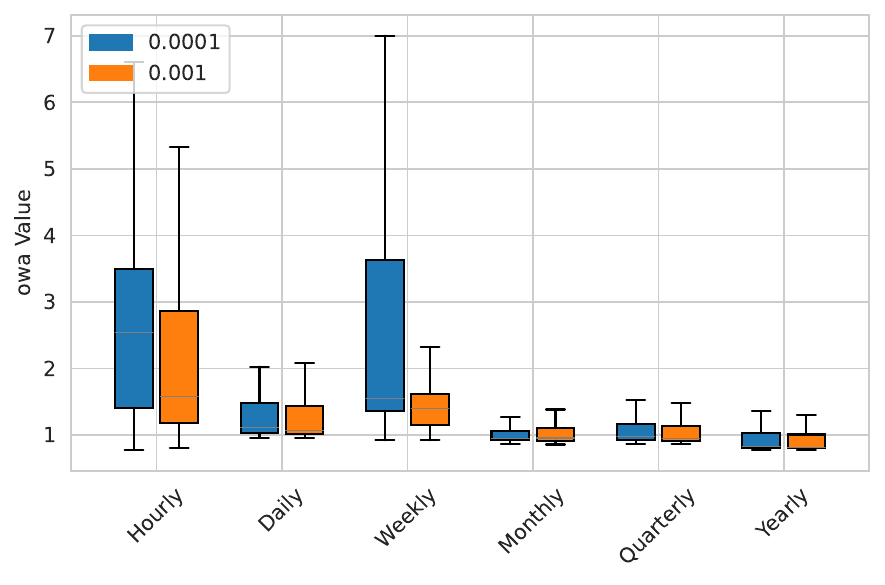}
         \caption{Learning Rate}
         \label{fig:exp-appx-lr-stf_owa_bp}
     \end{subfigure}
     \hspace{10pt}
    \begin{subfigure}[t]{0.28\textwidth}
         \centering
         \includegraphics[width=\textwidth]{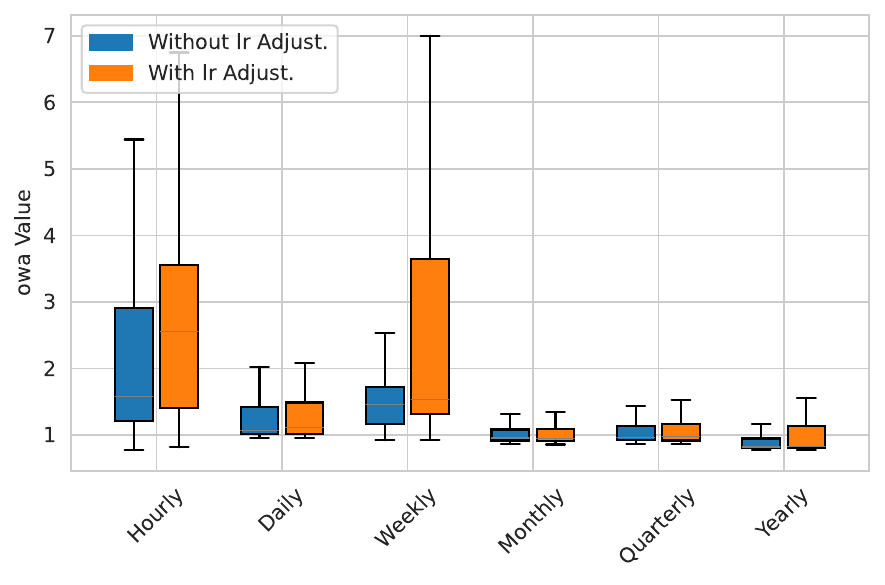}
         \caption{Learning Rate Strategy}
         \label{fig:exp-appx-lrs-stf_owa_bp}
     \end{subfigure}
     \hspace{10pt}
     \caption{Overall performance across all design dimensions in short-term forecasting. The results are based on \textbf{OWA}.}
     \vspace{-0.1in}
     \label{fig:exp-appx-rada-stf_owa_bp}
\end{figure}

\begin{figure}[t!]
     \centering
     \begin{subfigure}[t]{0.28\textwidth}
         \centering
         \includegraphics[width=\textwidth]{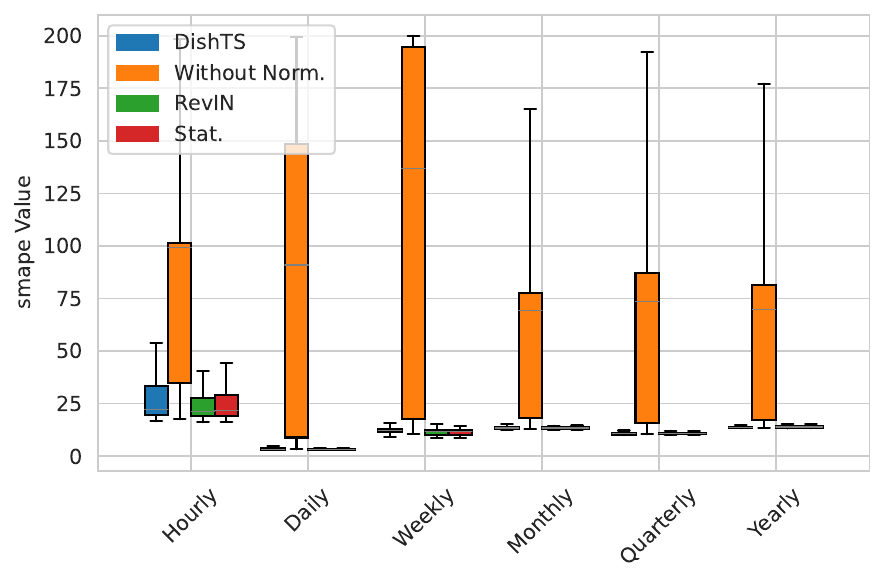}
         \caption{Series Normalization}
         \label{fig:exp-appx-Normalization-stf_smape}
     \end{subfigure}
     \hspace{10pt}
     \begin{subfigure}[t]{0.28\textwidth}
         \centering
         \includegraphics[width=\textwidth]{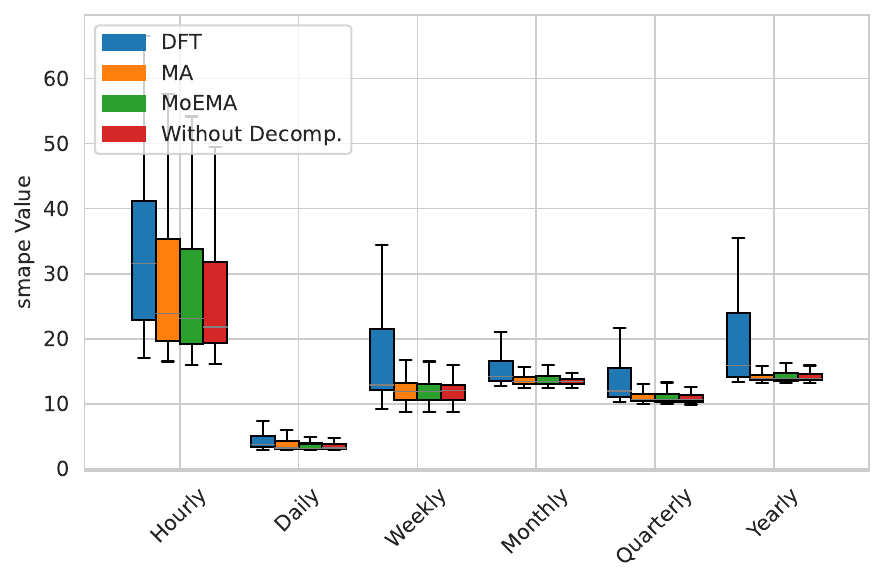}
         \caption{Series Decomposition}
         \label{fig:exp-appx-Decomposition-stf_smape_bp}
     \end{subfigure}
     \hspace{10pt}
    \begin{subfigure}[t]{0.28\textwidth}
         \centering
         \includegraphics[width=\textwidth]{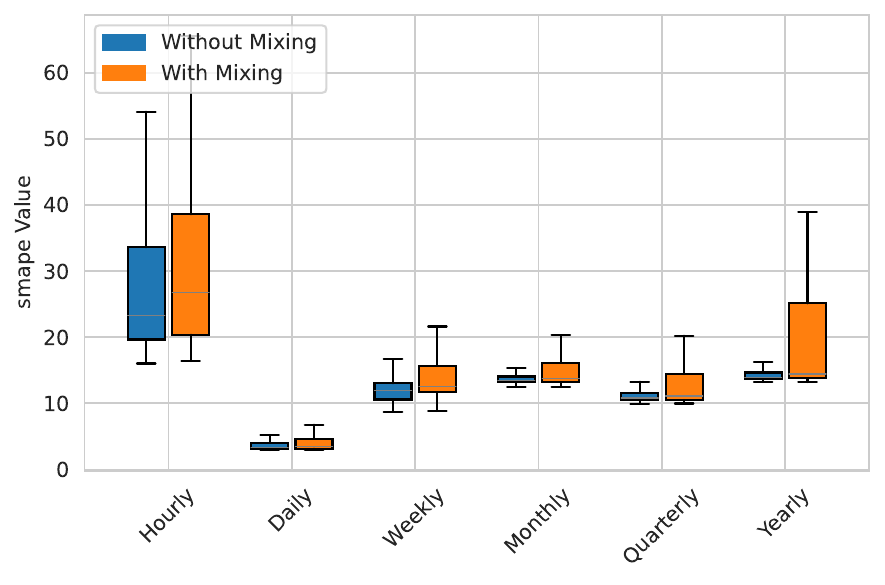}
         \caption{Series Sampling/Mixing}
         \label{fig:exp-appx-mixing-stf_smape_bp}
     \end{subfigure}
     \hspace{10pt}
    \begin{subfigure}[t]{0.28\textwidth}
         \centering
         \includegraphics[width=\textwidth]{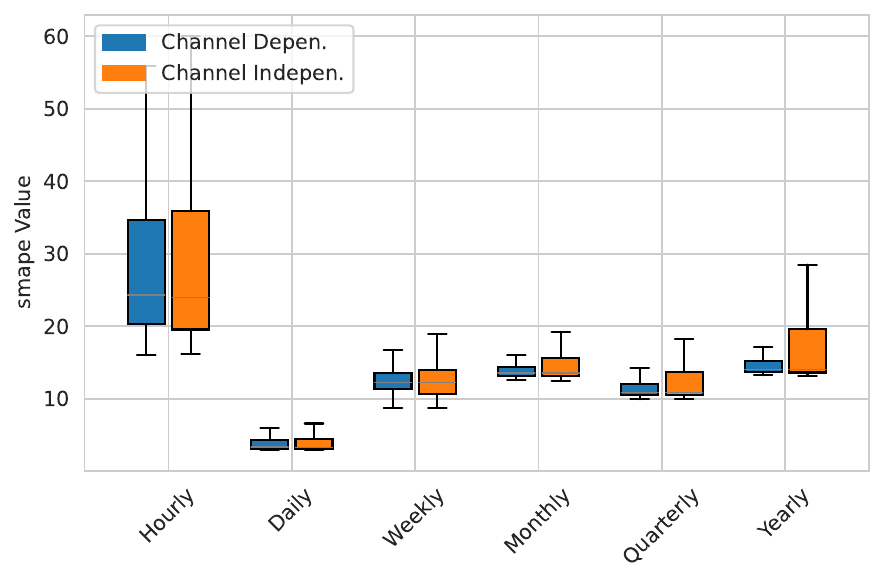}
         \caption{Channel Independent}
         \label{fig:exp-appx-CI-stf_smape_bp}
     \end{subfigure}
     \hspace{10pt}
    \begin{subfigure}[t]{0.28\textwidth}
         \centering
         \includegraphics[width=\textwidth]{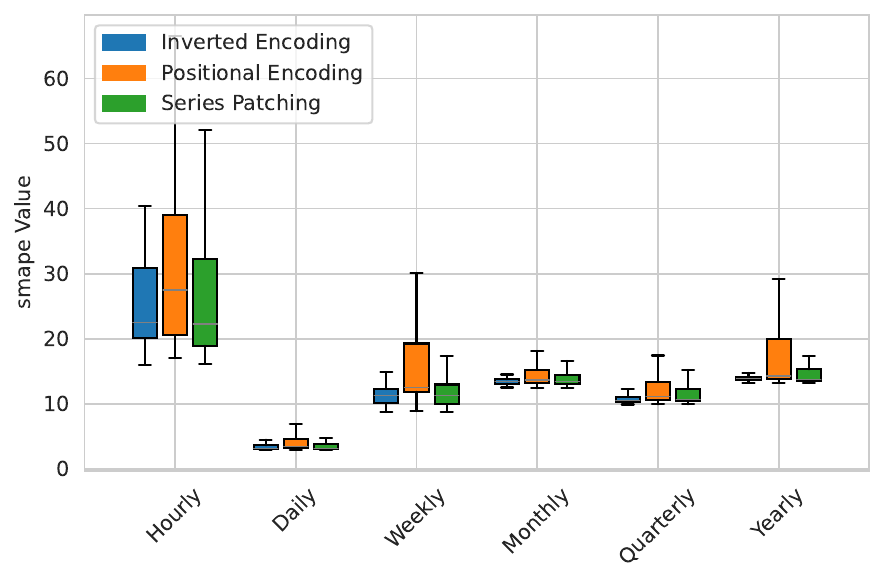}
         \caption{Series Embedding}
         \label{fig:exp-appx-tokenization-stf_smape_bp}
     \end{subfigure}
     \hspace{10pt}
    \begin{subfigure}[t]{0.28\textwidth}
         \centering
         \includegraphics[width=\textwidth]{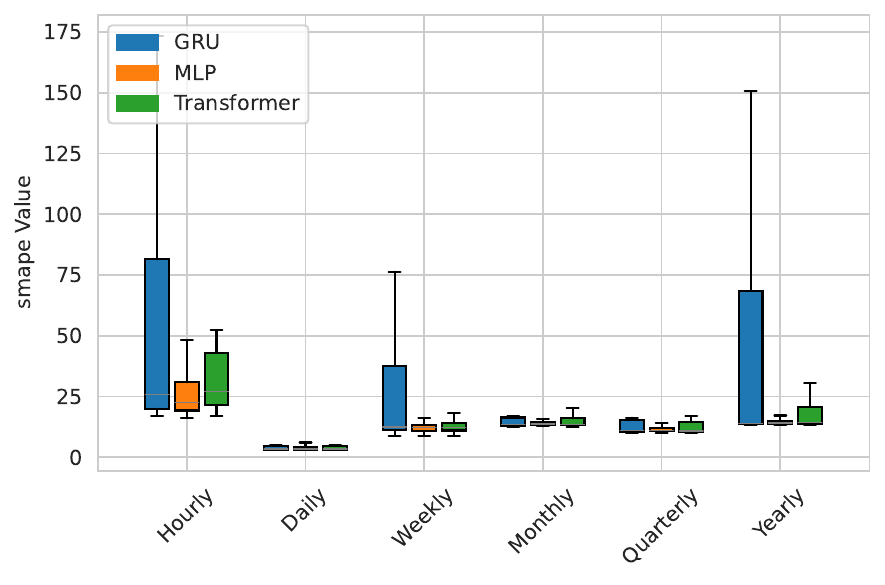}
         \caption{Network Backbone}
         \label{fig:exp-appx-backbone-stf_smape_bp}
     \end{subfigure}
     \hspace{10pt}
    \begin{subfigure}[t]{0.28\textwidth}
         \centering
         \includegraphics[width=\textwidth]{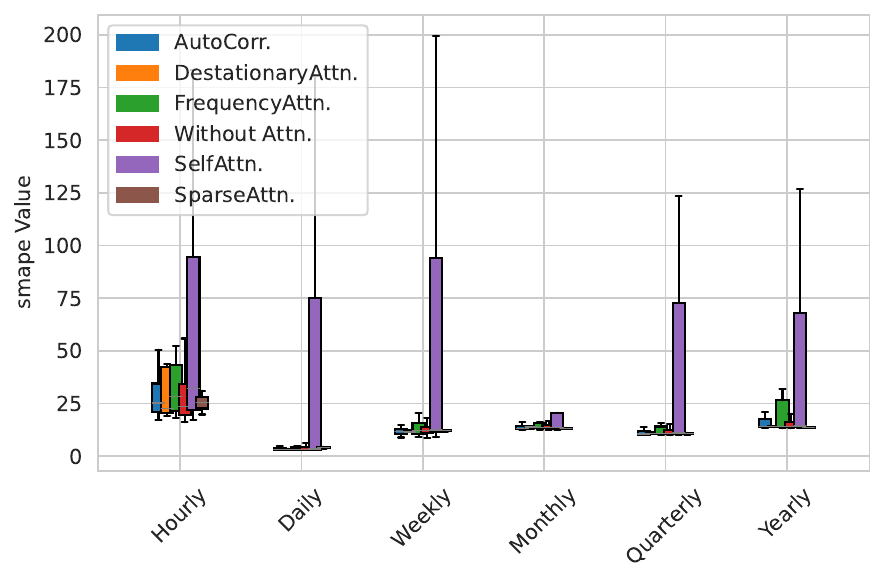}
         \caption{Series Attention}
         \label{fig:exp-appx-attention-stf_smape_bp}
     \end{subfigure}
     \hspace{10pt}
    \begin{subfigure}[t]{0.28\textwidth}
         \centering
         \includegraphics[width=\textwidth]{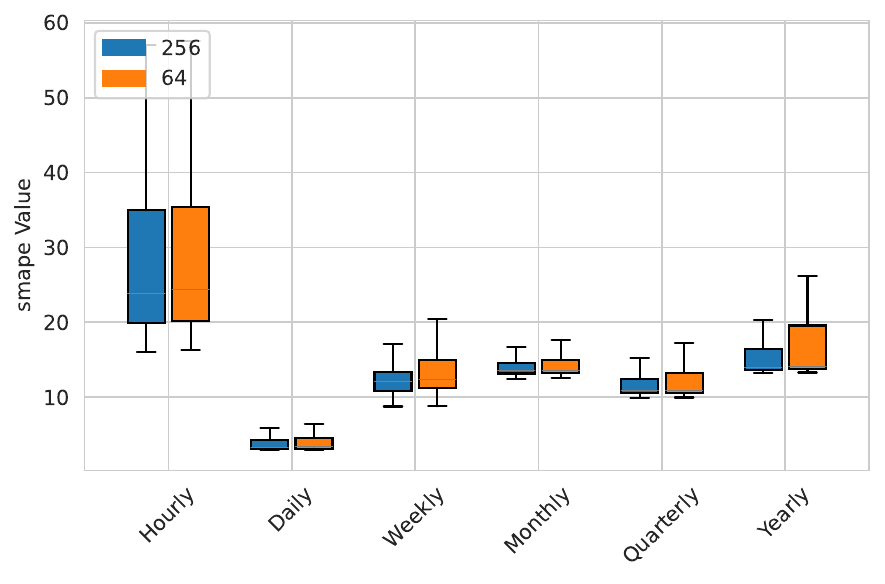}
         \caption{\textit{Hidden Layer Dimensions}}
         \label{fig:exp-appx-dmodel-stf_smape_bp}
     \end{subfigure}
     \hspace{10pt}
    \begin{subfigure}[t]{0.28\textwidth}
         \centering
         \includegraphics[width=\textwidth]{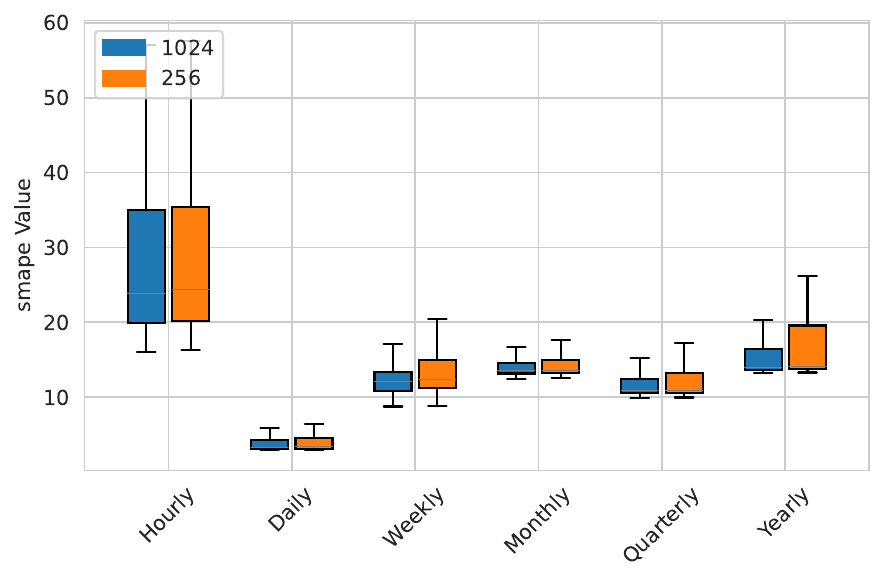}
         \caption{\textit{FCN Layer Dimensions}}
         \label{fig:exp-appx-dff-stf_smape_bp}
     \end{subfigure}
     \hspace{10pt}
    \begin{subfigure}[t]{0.28\textwidth}
         \centering
         \includegraphics[width=\textwidth]{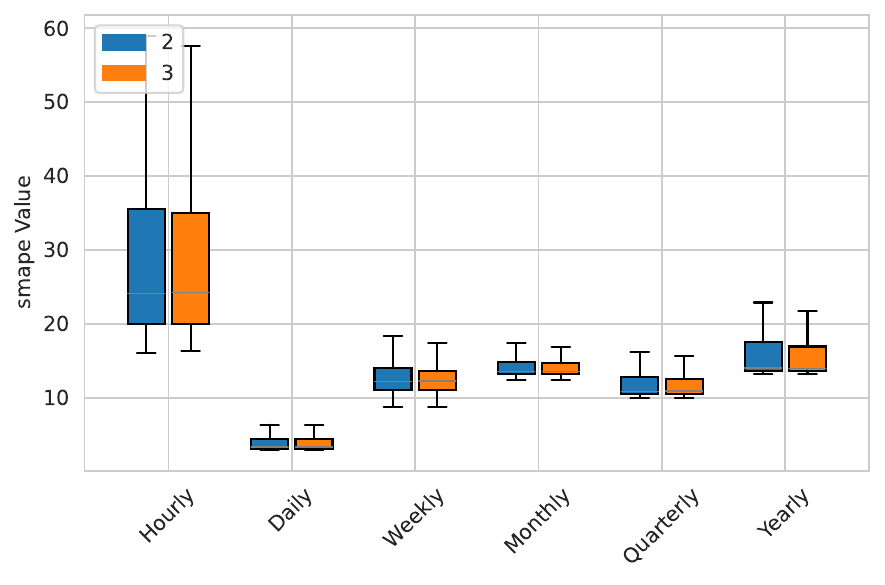}
         \caption{Encoder layers}
         \label{fig:exp-appx-el-stf_smape_bp}
     \end{subfigure}
     \hspace{10pt}
    \begin{subfigure}[t]{0.28\textwidth}
         \centering
         \includegraphics[width=\textwidth]{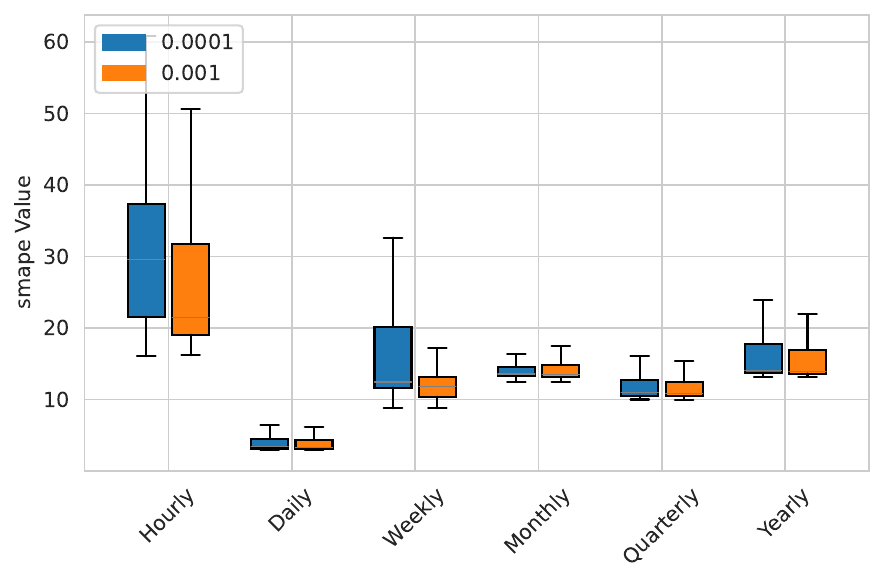}
         \caption{Learning Rate}
         \label{fig:exp-appx-lr-stf_smape_bp}
     \end{subfigure}
     \hspace{10pt}
    \begin{subfigure}[t]{0.28\textwidth}
         \centering
         \includegraphics[width=\textwidth]{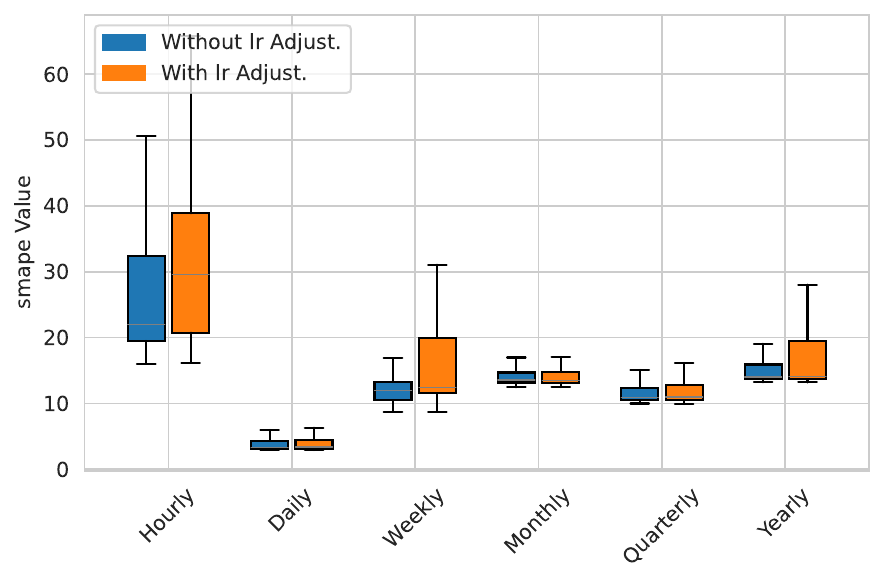}
         \caption{Learning Rate Strategy}
         \label{fig:exp-appx-lrs-stf_smape_bp}
     \end{subfigure}
     \hspace{10pt}
     \caption{Overall performance across all design dimensions in short-term forecasting.  The results are based on \textbf{SMAPE}.}
     \vspace{-0.1in}
     \label{fig:exp-appx-rada-stf_smape_bp}
\end{figure}

Complementary box plots are provided in Figure~\ref{fig:exp-appx-rada-stf_mase_bp}, Figure~\ref{fig:exp-appx-rada-stf_owa_bp}, and Figure~\ref{fig:exp-appx-rada-stf_smape_bp}, offering a statistical perspective on the distribution and robustness of performance across evaluation metrics.

Overall, the relative performance trends observed under MASE, OWA, and sMAPE metrics are consistent with those found in long-term forecasting tasks, reinforcing the generalizability and stability of our architectural choices.

\rv{\subsection{Ablation-based experiment for investigating four phases.}}

\rv{To quantitatively investigate the influence of four phases during MTSF pipeline (shown in Fig. \ref{fig:pipeline})—Series Preprocessing, Series Encoding, Network Architecture, and Network Optimization, we design an ablation-based experiment. We find that, compared to complex model structures and optimization methods, \textbf{the series preprocessing and encoding phases are more crucial}.}

\rv{Firstly, we establish the typical design choices as a baseline for 9 original datasets and 3 additional datasets (AQShunyi, AQWan, and CzeLan), which is defined as follows: \textit{Series Normalization=Null, Series Decomposition=Null, Series Sampling=False, Channel Independent=False, Sequence Length=48, Series Embedding=Positional Encoding, Network Backbone=RNN, Series Attention=Null, Feature Attention=Null, Encoder Layers=2, Epochs=10, Loss Function=MSE, Learning Rate=1e-3, Learning Rate Strategy=Null.} This baseline is applied to 48 experimental settings across 12 datasets and 4 different forecasting lengths.}

\rv{Next, to evaluate the impact of each phase, we keep the other three phases fixed at their simplest design choices and select a range of combinations for the current phase under evaluation. For example, when evaluating the impact of the Network Architecture phase, we fix the other three phases to the combinations \textit{Series Normalization=Null, Series Decomposition=Null, Series Sampling=False, Channel Independent=False, Series Embedding=Positional Encoding, Loss Function=MSE, Learning Rate=1e-3, Learning Rate Strategy=Null} and isolate the results varied during this phase.}

\rv{To account for differences across datasets, we measured the influence of each phase by comparing the relative improvement of the current phase against the corresponding baseline for each dataset and forecasting length. This is formally defined as:}

\rv{$$AvgRelErr_{phase} = \frac{1}{|S_{phase}|} \sum_{s \in S_{phase}} ( -\frac{MSE_{s}^{variant} - MSE_{s}^{baseline}}{MSE_{s}^{baseline}} )$$
where
$S_{phase}$ is the set of all (dataset, horizon) settings that satisfy the ablation criteria for the phase;
$MSE_{s}^{variant}$ is the MSE of a candidate configuration under setting s;
$MSE_{s}^{baseline}$ is the MSE of the minimal baseline under the same setting s.}

\rv{The table \ref{tab:analysis_phrases} reports the maximum, mean, and standard deviation of the relative improvement for each phase compared to the corresponding baseline. These values represent the potential, average improvement level, and stability of the current phase, respectively. In summary, compared to complex model structures and optimization methods, \textbf{the series preprocessing and encoding phases are more crucial}.}

\begin{table}[]
\centering
\caption{\rv{Performance Improvement of Each Phrase over the Baseline.}}\label{tab:analysis_phrases}
\begin{tabular}{c|ccc}
\hline
Phase                & Max Improvement & Mean Improvement & Std Improvement \\ \hline
Series Preprocessing & 90.87\%                              & 19.96\%                               & 0.394                                \\ 
Series Encoding      & 83.90\%                              & 30.98\%                               & 0.248                                \\ 
Network Architecture & 66.75\%                              & 7.12\%                                & 0.290                                \\ 
Network Optimization & 26.54\%                              & 10.01\%                               & 0.109                                 \\ \hline
\end{tabular}
\end{table}

\rv{\subsection{Explaining Design Drivers via Meta-Feature Importance Analysis}} \label{appx:meta_feature_importance}
\rv{To directly investigate the impact of individual meta-features, we conducted an additional analysis using an interpretable XGBoost-based meta-learner. Although this machine learning–based variant slightly underperforms compared to the original deep learning–based meta-learner (average MAE 0.447 vs. 0.426), due to its limited capacity in modeling rich, high-dimensional interactions among features, it remains competitive and provides a clear advantage in interpretability.}

\rv{This analysis allows us to quantify the relative importance of meta-features and structural design dimensions in determining model performance. As summarized in Table~\ref{tab:xgboost_meta_feature_importance}, certain temporal and spectral features—such as MFCC descriptors and Negative Turning Points—consistently appear among the most influential across datasets. In addition, architectural design choices like series normalization emerge as universally important factors, further validating the findings of our component-level ablation study.}
\begin{table}
\centering
\caption{\rv{Top 5 Most Important Meta-Features per Dataset Estimated via XGBoost}}
\resizebox{1\textwidth}{!}{
\begin{tabular}{c|rrrrr}
    \toprule
    \textbf{Dataset} & \multicolumn{5}{c}{\textbf{Top 5 Meta-Features (Importance)}} \\
    \midrule
    \textbf{ETTm1} & mean\_Negativeturningpoints (0.08) & series norm (0.06) & mean\_Centroid (0.05) & mean\_MFCC\_0 (0.05) & min\_Positiveturningpoints (0.05) \\
    \textbf{ETTm2} & mean\_Negativeturningpoints (0.10) & series norm (0.05) & mean\_MFCC (0.05) & mean\_Centroid (0.05) & q25\_Kurtosis (0.04) \\
    \textbf{ETTh1} & mean\_MFCC\_10 (0.08) & series norm (0.06) & mean\_Spectralroll-on (0.05) & mean\_Spectraldistance (0.04) & mean\_MFCC (0.04) \\
    \textbf{ETTh2} & mean\_MFCC\_0 (0.07) & series norm (0.05) & mean\_Medianfrequency (0.05) & mean\_Centroid (0.04) & min\_Meanabsolutediff (0.04) \\
    \textbf{ECL} & mean\_MFCC (0.08) & std\_MFCC (0.08) & series norm (0.07) & mean\_Centroid (0.06) & mean\_Maxpowerspectrum (0.04) \\
    \textbf{Traffic} & q25\_Kurtosis (0.08) & series norm (0.07) & mean\_Centroid (0.06) & mean\_Maxpowerspectrum (0.05) & mean\_MFCC (0.05) \\
    \textbf{Weather} & mean\_MFCC (0.14) & mean\_Negativeturningpoints (0.11) & series norm (0.06) & min\_Negativeturningpoints (0.05) & mean\_Centroid (0.04) \\
    \textbf{Exchange} & std\_MFCC (0.11) & mean\_Negativeturningpoints (0.10) & mean\_MFCC (0.07) & series norm (0.06) & mean\_Medianfrequency (0.03) \\
    \textbf{ILI} & mean\_MFCC (0.09) & mean\_Maximumfrequency (0.06) & channel independent (0.05) & series norm (0.05) & mean\_LPCC (0.05) \\
    \bottomrule
\end{tabular}
}
\label{tab:xgboost_meta_feature_importance}
\end{table}

\rv{\subsection{Meta-Feature Similarity Enables Targeted Knowledge Transfer}} \label{appx:meta_transfer}

\rv{In Fig.~\ref{fig:meta_features_pca}, we visualize the dimension-reduced meta-features across different datasets using PCA. The visualization confirms that datasets tend to cluster based on inherent properties, such as domain (e.g., ETT family) and temporal frequency (e.g., M4-Hourly vs. M4-Yearly). This indicates that meta-feature similarity reflects structural characteristics of datasets, and suggests the potential for targeted knowledge transfer between similar datasets.}

\rv{To further explore this, we conducted a case study focusing on the ILI dataset—a relatively difficult and data-scarce task. We enriched the meta-learner's training pool by adding two datasets (COVID-19 and FRED-MD) that are more similar to ILI in the meta-feature space. As shown in Table~\ref{tab:transfer_ili}, TSGym’s performance on ILI improves significantly, while performance on other datasets remains stable or even improves slightly. This highlights the potential of incorporating similar datasets to enhance performance on low-resource or underperforming tasks.}
\begin{table}[htbp]
\centering
\caption{\rv{Performance Comparison Before and After Adding Similar Datasets (COVID-19 and FRED-MD) to the Meta-Learner Training Pool}}
\begin{tabular}{l|cc|cc}
\toprule
 & \multicolumn{2}{c|}{\textbf{TSGym}} & \multicolumn{2}{c}{\textbf{+COVID-19, FRED-MD}} \\
\textbf{Metric} & \textbf{MSE} & \textbf{MAE} & \textbf{MSE} & \textbf{MAE} \\
\midrule
\textbf{ETTm1} & \textbf{\textcolor{red}{0.357}} & \textbf{\textcolor{red}{0.383}} & 0.360 & 0.386 \\
\textbf{ETTm2} & 0.261 & \textbf{\textcolor{red}{0.319}} & \textbf{\textcolor{red}{0.258}} & 0.320 \\
\textbf{ETTh1} & \textbf{\textcolor{red}{0.426}} & 0.440 & 0.429 & \textbf{\textcolor{red}{0.436}} \\
\textbf{ETTh2} & \textbf{\textcolor{red}{0.358}} & \textbf{\textcolor{red}{0.400}} & 0.368 & 0.406 \\
\textbf{ECL} & 0.170 & \textbf{\textcolor{red}{0.265}} & 0.170 & 0.269 \\
\textbf{Traffic} & 0.435 & 0.313 & \textbf{\textcolor{red}{0.428}} & \textbf{\textcolor{red}{0.306}} \\
\textbf{Weather} & 0.229 & 0.268 & \textbf{\textcolor{red}{0.228}} & \textbf{\textcolor{red}{0.266}} \\
\textbf{Exchange} & 0.410 & 0.431 & \textbf{\textcolor{red}{0.373}} & \textbf{\textcolor{red}{0.408}} \\
\textbf{ILI} & 2.233 & 1.015 & \textbf{\textcolor{red}{2.098}} & \textbf{\textcolor{red}{0.941}} \\
\bottomrule
\end{tabular}
\label{tab:transfer_ili}
\end{table}

\rv{\subsection{Meta-Learner Performance Scaling with Candidate Pool Size}}
\label{appx:meta-Learner-Scaling}

\rv{To investigate how the size of the candidate model pool $M_s$ affects meta-learner performance, we conducted a scaling analysis across all datasets. We trained the meta-learner on progressively larger subsets of $M_s$ (ranging from 5\% to 100\%), and measured the average rank of the model selected by TSGym.}

\rv{As shown in Table~\ref{tab:poolsize_rank}, the performance improves significantly as the pool size increases up to 25\%, after which the gains plateau. Remarkably, even with just 10\% of the full pool, TSGym already outperforms strong baselines such as DUET (which achieves average ranks of 4.11 for MSE and 3.67 for MAE). This highlights the high sample efficiency of TSGym and suggests that a moderately sized pool is sufficient to reach near-optimal performance. These results further motivate the use of smarter sampling strategies, such as Bayesian Optimization, to construct high-quality training pools with minimal cost.}

\begin{table}[htbp]
  \centering
  \caption{\rv{Effect of Candidate Pool Size on Meta-Learner Selection Accuracy}}
  \begin{tabular}{c|c|c}
    \toprule
    \textbf{Subset Size of \( M_s \)} & \textbf{MSE (Avg. Rank)} & \textbf{MAE (Avg. Rank)} \\
    \midrule
    5\%  & 3.67 & 3.44 \\
    10\% & 2.67 & 3.00 \\
    25\% & 1.67 & 1.78 \\
    50\% & 1.89 & 2.67 \\
    75\% & 1.89 & 2.22 \\
    100\% & 1.67 & 2.00 \\
    \bottomrule
  \end{tabular}
  \label{tab:poolsize_rank}
\end{table}

\begin{rve}
\section{Comparative Experiments with AutoML Methods}

\subsection{Experimental Setup}
We investigated a number of well-known AutoML and NAS (Neural Architecture Search) libraries, including TPOT\cite{Olson2016EvoBio}, H2O-3\cite{h2o_package_or_module}, Microsoft NNI\cite{nni2021}, Auto-Keras\cite{JMLR:v24:20-1355}, Auto-Sklearn\cite{feurer-arxiv20a}, and NASLib\cite{naslib-2020}. We noted, however, that \textbf{most of these frameworks are not specifically designed for time-series forecasting}, which would make adapting them for MTSF tasks both burdensome and likely result in an unfair comparison. For this reason, we \textbf{selected two leading AutoML libraries that explicitly support MTSF} for a direct and meaningful comparison: AutoGluon-TimeSeries\cite{agtimeseries} and AutoTS\cite{AutoTS_github}.

The experimental settings for AutoGluon and AutoTS were aligned with those of TSGym. We adopted the same training, validation, and test set splits as used in TSGym, with all base configurations set to their default values. To ensure computational efficiency, AutoGluon was configured with the ``high\_quality'' preset, while AutoTS was tested using the ``superfast'' model setting.

\subsection{Performance Comparison}

We benchmarked TSGym against these two methods on both short-term and long-term forecasting tasks. The results are presented in Table \ref{tab:auto-ml-short_term_comparison} and Table \ref{tab:auto-ml-long_term_comparison}, respectively.

For short-term forecasting (Table \ref{tab:auto-ml-short_term_comparison}), TSGym demonstrates clear superiority, achieving the best scores across all three metrics: Overall Weighted Average (OWA), Symmetric Mean Absolute Percentage Error (SMAPE), and Mean Absolute Scaled Error (MASE). This indicates a robust and consistently better performance in short-horizon predictions compared to the established AutoML baselines.

In the more challenging long-term forecasting tasks (Table \ref{tab:auto-ml-long_term_comparison}), TSGym continues to show a strong competitive advantage. It secures the lowest (best) Mean Squared Error (MSE) and Mean Absolute Error (MAE) on the majority of datasets, including ETTm1, ETTm2, ETTh1, ETTh2, ECL, and Weather. It is worth noting that AutoGluon achieves better performance on the Exchange dataset and a lower MAE on the ILI dataset, while AutoTS shows a competitive MAE on the Traffic dataset. Nevertheless, TSGym's dominant performance across a wide range of datasets underscores its effectiveness and robustness for long-horizon prediction.

\begin{table}[]
\centering
\caption{Short-term Forecasting Comparison}
\label{tab:auto-ml-short_term_comparison}
\begin{tabular}{@{}lccc@{}}
\toprule
Model    & TSGym (ours)     & AutoGluon & AutoTS  \\ \midrule
\textbf{OWA}   & \textbf{0.872}   & 0.95      & 2.002   \\
\textbf{SMAPE} & \textbf{12.013}  & 13.178    & 18.977  \\
\textbf{MASE}  & \textbf{1.575}   & 1.775     & 4.981   \\ \bottomrule
\end{tabular}
\end{table}

\begin{table}[htbp]
\centering
\caption{Long-term Forecasting Comparison}
\label{tab:auto-ml-long_term_comparison}
\begin{tabular}{@{}l|cc|cc|cc@{}}
\toprule
\multirow{2}{*}{Dataset} & \multicolumn{2}{c|}{\textbf{TSGym (ours)}} & \multicolumn{2}{c|}{\textbf{AutoGluon}} & \multicolumn{2}{c}{\textbf{AutoTS}} \\ \cmidrule(l){2-7} 
                       & MSE              & MAE              & MSE         & MAE         & MSE         & MAE         \\ \midrule
ETTm1                  & \textbf{0.357}   & \textbf{0.383}   & 0.482       & 0.408       & 0.744       & 0.546       \\
ETTm2                  & \textbf{0.261}   & \textbf{0.319}   & 0.273       & 0.337       & 0.392       & 0.389       \\
ETTh1                  & \textbf{0.426}   & \textbf{0.440}   & 0.503       & 0.473       & 0.981       & 0.610       \\
ETTh2                  & \textbf{0.358}   & \textbf{0.400}   & 0.419       & 0.430       & 0.589       & 0.488       \\
ECL                    & \textbf{0.170}   & \textbf{0.265}   & 0.265       & 0.328       & 0.327       & 0.355       \\
Traffic                & \textbf{0.435}   & 0.313            & 0.555       & 0.325       & 0.739       & \textbf{0.311} \\
Weather                & \textbf{0.229}   & \textbf{0.268}   & 0.236       & 0.270       & 0.519       & 0.372       \\
Exchange               & 0.410            & 0.431            & \textbf{0.330} & \textbf{0.393} & 0.588       & 0.494       \\
ILI                    & \textbf{2.233}   & 1.015            & 2.271       & \textbf{0.979} & 2.533       & 1.049       \\ \bottomrule
\end{tabular}%
\end{table}

\subsection{Conclusion}
As the results demonstrate, TSGym consistently and significantly outperforms both AutoGluon and AutoTS across the vast majority of datasets for both short-term and long-term forecasting tasks.

In summary, while TSGym shares the goals of AutoML, its distinct methodology—automated model construction via fine-grained component decomposition and meta-learning—offers a clear advantage over existing AutoML methods for MTSF. This specialized design, now validated against both state-of-the-art deep-learning MTSF methods and the newly included AutoML baselines, represents a significant and valuable contribution to the MTSF community. 

\end{rve}

\rv{\section{TSGym Performance Comparison Across Sampling Strategies}\label{appx:optuna} }

\rv{To reduce computational cost and improve search efficiency, we adopted Optuna, which is based on Bayesian optimization, as a more efficient sampling strategy than random search. After an initial 50 random trials, Optuna intelligently sampled an additional 50 configurations. The experimental results confirms that \textbf{Optuna significantly improves search efficiency, discovering superior model combinations with far fewer evaluations}. }

\rv{Table~\ref{tab:optuna_vs_random_distribution} reports the MSE distribution statistics for configurations sampled by Optuna and random search across various datasets. “Trials to Exceed Best Random” indicates how many Optuna trials were needed to outperform the best result from random search. Optuna consistently achieves better results with significantly fewer evaluations (less than 100 trials vs. 1.5-5 times more for random search), validating its sampling efficiency. This confirms that a meta-learner trained on this intelligently curated set is more effective.}
\begin{table}[htbp]
  \centering
  \caption{\rv{Comparison of MSE Distribution Between Optuna and Random Search Across Datasets}}
  \resizebox{1\textwidth}{!}{
    \begin{tabular}{c|c|c|c|c|c|c|c|c}
    \toprule
    \textbf{Dataset} & \textbf{Method} & \textbf{Optuna Best Over Random Experiments} & \textbf{Min mse} & \textbf{Q1 mse} & \textbf{Median mse} & \textbf{Q3 mse} & \textbf{Max mse} & \textbf{Total Experiment Count} \\
    \toprule
    \multirow{2}[3]{*}{ECL} & Optuna & 56    & \textbf{0.131} & \textbf{0.158} & \textbf{0.182} & \textbf{0.213} & \textbf{0.414} & \textbf{400} \\
          & Random &       & 0.134 & 0.179 & 0.212 & 0.247 & 0.862 & 1493 \\
    \midrule
    \multirow{2}[4]{*}{ETTh1} & Optuna & 70    & \textbf{0.355} & \textbf{0.416} & \textbf{0.450} & \textbf{0.519} & \textbf{1.210} & \textbf{400} \\
          & Random &       & 0.376 & 0.442 & 0.494 & 0.581 & 1.860 & 1129 \\
    \midrule
    \multirow{2}[4]{*}{ETTh2} & Optuna & 91.5  & \textbf{0.268} & \textbf{0.342} & \textbf{0.400} & \textbf{0.539} & \textbf{9.060} & \textbf{400} \\
          & Random &       & 0.270 & 0.391 & 0.464 & 1.017 & 30.771 & 1147 \\
    \midrule
    \multirow{2}[4]{*}{ETTm1} & Optuna & 59.5  & \textbf{0.293} & \textbf{0.341} & \textbf{0.405} & \textbf{0.472} & \textbf{4317.839} & \textbf{400} \\
          & Random &       & 0.295 & 0.378 & 0.447 & 0.547 & 304538.500 & 623 \\
    \midrule
    \multirow{2}[4]{*}{ETTm2} & Optuna & 56.5  & \textbf{0.159} & \textbf{0.221} & \textbf{0.279} & \textbf{0.385} & \textbf{9.344} & \textbf{400} \\
          & Random &       & 0.167 & 0.259 & 0.371 & 0.584 & 198.023 & 762 \\
    \midrule
    \multirow{2}[4]{*}{Exchange} & Optuna & 90.25 & 0.081 & \textbf{0.169} & \textbf{0.280} & \textbf{0.681} & 15.054 & \textbf{400} \\
          & Random &       & \textbf{0.080} & 0.189 & 0.391 & 0.959 & \textbf{12.106} & 2033 \\
    \midrule
    \multirow{2}[4]{*}{ili} & Optuna & 54    & 1.506 & \textbf{1.891} & \textbf{2.302} & \textbf{3.080} & \textbf{7.503} & \textbf{400} \\
          & Random &       & \textbf{1.495} & 2.363 & 2.842 & 4.027 & 7.532 & 3976 \\
    \midrule
    \multirow{2}[4]{*}{traffic} & Optuna & 93    & 0.387 & \textbf{0.438} & \textbf{0.491} & \textbf{0.612} & \textbf{1.051} & \textbf{400} \\
          & Random &       & \textbf{0.379} & 0.487 & 0.589 & 0.686 & 1.473 & 1018 \\
    \midrule
    \multirow{2}[4]{*}{weather} & Optuna & 74    & 0.144 & \textbf{0.193} & \textbf{0.245} & \textbf{0.312} & \textbf{29.895} & \textbf{400} \\
          & Random &       & \textbf{0.143} & 0.207 & 0.263 & 0.344 & 109.467 & 656 \\
    \bottomrule
    \end{tabular}}%
  \label{tab:optuna_vs_random_distribution}%
\end{table}%

\rv{We also note that Optuna can provide interpretability. Table~\ref{tab:optuna_importance} shows the importance of each design dimension estimated by Optuna’s built-in fANOVA analysis. Sequence Length and Series Normalization contribute the most to performance variation, suggesting their critical role in architecture design.}
\begin{table}[htbp]
  \centering
  \caption{\rv{Relative Importance of Design Dimensions Estimated by Optuna’s fANOVA Analysis}}
    \begin{tabular}{c|c|c}
    \toprule
    \textbf{Rank} & \textbf{Design Dimensions} & \textbf{Importance} \\
    \midrule
    \textbf{1} & Sequence Length & 0.270 \\
    \textbf{2} & Series Normalization & 0.255 \\
    \textbf{3} & Series Embedding & 0.134 \\
    \textbf{4} & Feature Attention & 0.077 \\
    \textbf{5} & Series Decomposition & 0.053 \\
    \textbf{6} & Channel Independent & 0.050 \\
    \textbf{7} & Series Sampling/Mixing & 0.029 \\
    \textbf{8} & Epochs & 0.025 \\
    \textbf{9} & d\_model d\_ff & 0.020 \\
    \textbf{10} & Learning Rate & 0.020 \\
    \textbf{11} & With/Without Timestamps & 0.018 \\
    \textbf{12} & Network Type & 0.017 \\
    \textbf{13} & Encoder Layers & 0.013 \\
    \textbf{14} & Learning Rate Strategy & 0.012 \\
    \textbf{15} & Loss Function & 0.010 \\
    \textbf{16} & Series Attention & 0.000 \\
    \bottomrule
    \end{tabular}%
  \label{tab:optuna_importance}%
\end{table}%

\rv{\section{Deeper Insights into Design Choices}\label{appx:deep_analysis}}

\rv{To delve deeper into how each component behaves, TSGym formulate 7 most contentious claims in the reaserch community and clarify them with this benchmark. Beyond the 9 datasets already reported in our paper, we added 3 new ones—AQShunyi, AQWan, and CzeLan—selected by jointly considering dataset profiles and computational cost, and every component underwent at least 2,000 experimental runs. Following the dataset-characterization framework proposed in TFB~\cite{qiu2024tfb}, we describe each dataset in terms of dimensionality, sample size, stationarity, distributional drift, and channel correlation, as shown in Table~\ref{tab:dataset_chara_tfb}. }
\begin{table}[]
\centering
\caption{\rv{Dataset Characteristics Derived from TFB\cite{qiu2024tfb}.}}\label{tab:dataset_chara_tfb}
\begin{tabular}{c|llllll}
\hline
dataset  & freq   & dim & length & stationary & shifting & correlation \\ \hline
ETTm1    & mins   & 7   & 57600  & 9.73E-05   & -0.06298 & 0.612       \\
ETTm2    & mins   & 7   & 57600  & 0.003043   & -0.40555 & 0.504       \\
ETTh1    & hourly & 7   & 14400  & 0.0012     & -0.06136 & 0.63        \\
ETTh2    & hourly & 7   & 14400  & 0.021788   & -0.40381 &  0.509       \\
ECL      & hourly & 321 & 26304  & 0.00515    & -0.07494 &  0.802       \\
traffic  & hourly & 862 & 17544  & 3.71E-08   & 0.066992 & 0.814       \\
weather  & mins   & 21  & 52696  & 1.04E-08   & 0.213569 &  0.694       \\
Exchange & daily  & 8   & 7588   & 0.359774   & 0.325341 &  0.565       \\
ili      & weekly & 7   & 966    & 0.169155   & 0.721088 &  0.674       \\
czelan   & mins   & 11  & 19934  & 0.159633   & -0.15823 &  0.683       \\
AQShunyi & hourly & 11  & 35064  & 0.000302   & 0.018716 &  0.613       \\
AQWan    & hourly & 11  & 35064  & 0.000275   & -0.01141 &  0.624      \\ \hline
\end{tabular}
\end{table}

\rv{\textbf{Claim 1: Model scale should match data scale.}}

\rv{\textbf{Conclusion 1: Yes.} We validate this claim from 2 aspects:  1) Hidden-dimension alignment: On the two high-dimensional datasets (ECL and Traffic), the 256 hidden dimensions surpasses 64 across all four prediction horizons, while for the remaining datasets, 64 outperforms 256 in 32 of 40 settings. 2) Model Backbone: Transformer dominates on the three largest datasets (ECL, Traffic, Weather) across 11 of 12 settings, whereas MLP prevails in 29 of 36 smaller-scale settings.}

\rv{\textbf{Claim 2: Transformers are less robust than MLPs.}}

\rv{\textbf{Conclusion 2: Yes.} Using the inter-quartile range (IQR) as the robustness metric, Transformers fare worse than MLPs in 34 out of 48 settings. The average relative IQR gap, $\frac{1}{N} \sum_{i=1}^{N} \frac{IQR_{Transformer,i} - IQR_{MLP,i}}{IQR_{MLP,i}}$, reaches as high as 54.6\%.}

\rv{\textbf{Claim 3: Transformers exhibit a higher upper bound than MLPs.}}

\rv{\textbf{Conclusion 3: No.} Even when incorporating multiple attention variants, Transformers do not guarantee a performance advantage over MLPs. We compared the upper bound defined as the best result achieved across all of their respective settings. Transformers outperform in 27 of the 48 settings, while underperforming in 21, with an average margin of only +0.038\%.}

\rv{\textbf{Claim 4: The superiority of Channel-Independent (CI) vs. Channel-Dependent (CD) correlates with inter-channel correlation.}}

\rv{\textbf{Conclusion 4: No.} In the 13 settings where CD models outperform CI models, the average inter-channel correlation is 0.696, versus 0.624 in the 35 settings where CI is preferred. Yet ECL, with the second-highest correlation 0.802, still favors CI on every horizon. This finding suggests the CI/CD decision must consider the data's semantic context, not just statistical correlation.}

\rv{\textbf{Claim 5: Series normalization mitigates distribution shift.}}

\rv{\textbf{Conclusion 5: Yes.} On the 6datasets with the highest drift (ETTm2, ETTh2, weather, Exchange, ILI, CzeLan), enabling series normalization wins in all 24 settings. Conversely, on the two near-drift-free datasets (AQShunyi and AQWan), disabling it is better in 8/8 settings. Thus, series normalization effectively combats distribution shift but can discard useful information when no shift is present.}

\rv{\textbf{Claim 6: Novel sequence encodings outperform the traditional single-timestep encoding.}}

\rv{\textbf{Conclusion 6: Yes.} Across all 48 settings, classic positional encoding is best only twice, whereas inverted encoding and series patching lead in 17 and 20 settings, respectively.}

\rv{\textbf{Claim 7: Novel attention mechanisms outperform vanilla self-attention.}}

\rv{\textbf{Conclusion 7: Yes.} Across 48 settings, vanilla self-attention and auto-correlation each win only 4 times, whereas de-stationary, frequency-enhanced, and sparse attention dominate with 14, 14, and 12 best results, respectively. Notably, on the two least-stationary datasets (Traffic and Weather), De-stationary attention clearly surpasses all other attention and non-attention variants.}

\end{document}